%% file: main.tex
\documentclass{siamonline220329}

\usepackage{url}
\expandafter\def\csname ver@subfig.sty\endcsname{}
\usepackage{graphicx,subfigure} 
\usepackage{natbib} 
\setcitestyle{numbers,sort&compress} 

\usepackage[utf8]{inputenc} 
\usepackage[T1]{fontenc}    
\usepackage{hyperref}       
\usepackage{url}            
\usepackage{booktabs}       
\usepackage{amsfonts}       
\usepackage{nicefrac}       
\usepackage{microtype}      
\usepackage{xcolor}         
\usepackage{nicematrix} 

\usepackage{etoolbox}
\providetoggle{todos}
\providetoggle{long}
\settoggle{todos}{false}
\settoggle{long}{false}
\providetoggle{arxiv}
\settoggle{arxiv}{true}

\providetoggle{unitcircle}
\settoggle{unitcircle}{false}
\usepackage{tikz}
\usetikzlibrary{topaths, shapes ,arrows ,positioning}
\tikzset{bend angle=15,
}
\usepackage{pgfplots}
\newtoks\mycoords
\usepackage{xparse,fp,xfp}
\usetikzlibrary{decorations.pathreplacing}
\usepackage{microtype}
\usepackage{graphicx}
\usepackage{booktabs} 
\usepackage{amsmath}
\usepackage{amsfonts}
\usepackage{lastpage}
\usepackage{multirow}
\usepackage{tabu}
\usepackage{lipsum}
\usepackage{clrscode}
\usepackage{caption}
\usepackage{subfig}
\usepackage{float}
\usepackage{placeins} 

\usepackage{mathrsfs}
\usepackage{indentfirst}

\newcommand{\bone}{\boldsymbol{1}}

\newcommand{\lasso}{Lasso} 
\newcommand{\params}{\theta} 
\newcommand{\paramspace}{\Theta} 
\newcommand{\regparams}{\theta_w} 
\newcommand{\nn}[2]{f_{#1}\left({#2};\theta \right)} 
\newcommand{\nnbiased}[2]{f^{\mathrm{skip}}_{#1}\left({#2};\theta \right)} 
\newcommand{\NN}[2]{f^{\mathrm{NN}}_{#1}\left({#2};\theta\right)} 
\newcommand{\Wul}[2]{\mathbf{W}^{\left(#1,#2\right)}} 
\newcommand{\bul}[2]{\mathbf{b}^{\left(#1,#2\right)}} 

\newcommand{\Xul}[2]{\hat{\mathbf{X}}^{\left(#1,#2\right)}} 

\newcommand{\sul}[2]{\mathbf{s}^{\left(#1,#2\right)}} 
\newcommand{\ssul}[2]{s^{\left(#1,#2\right)}} 

\newcommand{\wl}[1]{\mathbf{W}^{\left({#1}\right)}}
\newcommand{\xl}[1]{\hat{\mathbf{X}}^{\left({#1}\right)}}
\newcommand{\bl}[1]{\mathbf{b}^{\left({#1}\right)}}
\newcommand{\midpt}[2]{m_{{#1},{#2}}}
\newcommand{\midptname}{normalized midpoint}

\newcommand{\Amat}{\mathbf{A}}
\newcommand{\Apos}{\Amat{}_+} 
\newcommand{\Aneg}{\Amat{}_-} 
\newcommand{\Asign}[1]{\Amat{}^{(s)}_{#1}} 
\newcommand{\Athresh}[1]{\Amat{}^{(t)}_{#1}} 
\newcommand{\zpos}{\mathbf{z}_+} 
\newcommand{\zneg}{\mathbf{z}_-} 
\newcommand{\zposscalar}{\left({z}_+\right)} 
\newcommand{\znegscalar}{\left({z}_-\right)} 

\newcommand{\posslope}{a^+}
\newcommand{\negslope}{a^-}
\newcommand{\posbias}{b^+}
\newcommand{\negbias}{b^-}

\newcommand{\nc}{non-convex}
\newcommand{\nd}[1]{{#1}{-}D}
\newcommand{\effL}{\tilde{L}} 

\newcommand{\deepnarrow}{deep narrow}
\newcommand{\deepReLU}[1]{\deepnarrow{} network}

\newcommand{\ReLU}[2]{\textrm{ReLU}^{#1}_{#2}}
\newcommand{\ramp}[3]{\textrm{Ramp}^{#1}_{{#2},{#3}}}
\newcommand{\ReLUname}{ramp}
\newcommand{\rampname}{capped ramp}
\newcommand{\deepsign}{rectangular} 
\newcommand{\kink}{breakpoint}
\newcommand{\unscal}{unscal}

\newcommand{\ReLUdicindices}{\mathbf{s},\mathbf{j},k}
\newcommand{\ReLUdicfun}[1]{f^{\left({#1},\sigma \right)}_{\ReLUdicindices{}}}

\newcommand{\dicindexset}{\mathcal{M}}

\newcommand{\reflect}[2]{R_{\left({#1},{#2}\right)}}
\newcommand{\reflectsmall}[2]{R_{\big({#1},{#2}\big)}}
\newcommand{\reflecttiny}[2]{R_{({#1},{#2})}}
\newcommand{\switchmatrix}[1]{\mathbf{H}^{({#1})}}
\newcommand{\concat}{\oplus}
\newcommand{\lassoSolSet}{\Phi}
\newcommand{\hkt}{\mathbf{h}^{(T)}}
\newcommand{\hktname}{square wave}
\newcommand{\betafrac}{\beta_T}
\newcommand{\biasparams}{\params_{b}}
\newcommand{\salpha}{\tilde{\alpha}}
\newcommand{\slope}{\mu} 
\newcommand{\adam}{Adam}
\newcommand{\labelvec}{target vector} 
\newcommand{\switchsetname}{switching set}

\newcommand{\regfunc}{r}

\newcommand{\skipslope}{\omega}

\newcommand{\symbeta}{\tilde{\beta}}
\newcommand{\symmeterized}{symmetrized} 

\newcommand{\secnameSignSol}{Solution path for sign activation and binary, periodic labels}
\newcommand{\secnameMain}{Main results}

\newcommand{\secnameSignAct}{Deep neural networks with sign activation}
\newcommand{\secnameSolSet}{The solution sets of \lasso{} and the training problem}
\newcommand{\secnameMinNorm}{Solution sets of \lasso{} under minimal regularization}

\newcommand\scalemath[2]{\scalebox{#1}{\mbox{\ensuremath{\displaystyle #2}}}} 

{
\par\noindent
  \textbf{#1 (Informal)}\begin{itshape}%
}%
{\end{itshape}\par\vspace{\baselineskip}}

\usepackage{amsmath,bm}
\usepackage{mathtools}

\newtheorem{remark}[theorem]{Remark}

\newcommand{\proofsketch}[1]{ \textit{Proof sketch.} #1 \hfill $\square$}

\newcommand{\pulse}[2]{ 
	\draw ( {#1},  0) -- ( {#1}, -{#2}) -- (0,  -{#2})  -- ( 0, -15);
}

\newcommand{\multipulse}[7]{ 
	\draw ( {#1},  0) -- ( {#1}, -{#2}) -- (0,  -{#2})  -- ( 0,-{#3}) -- ({#1}, -{#3}) -- ({#1}, -{#4}) -- (0, -{#4}) -- (0, -{#5}) -- ({#1}, -{#5}) -- ({#1},-{#6}) -- (0, -{#6}) -- (0,-{#7}) -- ({#1},-{#7} ) -- ({#1},-15)  ;
}

\newcommand{\snegshape}{circle} 

\newcommand{\jonecol}{yellow}
\newcommand{\jtwocol}{red}
\newcommand{\jthreecol}{blue}
\newcommand{\jonejtwocol}{orange}
\newcommand{\jtwojthreecol}{purple}

\newcommand{\incline}[2]{ 
    \draw[dashed] (0,0) -- (6,0);
    \draw (0,-3) -- (6,3);
    \fill (3,0)   node [fill={#1}, {#2}, scale=0.4]  {};
}

\newcommand{\decline}[2]{ 
    \draw[dashed] (0,0) -- (6,0);
    \draw (0,3) -- (6,-3);
    \fill (3,0)   node [fill={#1}, {#2}, scale=0.4]  {};
}

\newcommand{\incReLU}[2]{ 
    \draw[dashed] (0,0) -- (6,0);
    \draw (0,0) -- (3,0) -- (6,3);
    \fill (3,0)   node [fill={#1}, {#2}, scale=0.4]  {};
}

\newcommand{\downshiftincReLU}[4]{ 
    \draw[dashed] (0,0) -- (6,0);
    \draw (0,-1) -- (3,-1) -- (6,2);
    \fill (3,0)   node [fill={#1}, {#2}, scale=0.4]  {};
    \fill (4,0)   node [fill={#3}, {#4}, scale=0.4]  {};
}

\newcommand{\decReLU}[2]{
    \draw[dashed] (0,0) -- (6,0);
    \draw (0,3) -- (3,0) -- (6,0);
    \fill (3,0)   node [fill={#1}, {#2}, scale=0.4]  {};
}

\newcommand{\downshiftdecReLU}[4]{ 
    \draw[dashed] (0,0) -- (6,0);
    \draw (0,2) -- (3,-1) -- (6,-1);
    \fill (2,0)   node [fill={#1}, {#2}, scale=0.4]  {};
    \fill (3,0)   node [fill={#3}, {#4}, scale=0.4]  {};
}

\newcommand{\negdecReLU}[4]{ 
    \draw[dashed] (0,0) -- (6,0);
    \draw (0,2) -- (3,2) -- (6,-1);
    \fill (3,0)   node [fill={#1}, {#2}, scale=0.4]  {};
    \fill (5,0)   node [fill={#3}, {#4}, scale=0.4]  {};
}

\newcommand{\negincReLU}[4]{ 
    \draw[dashed] (0,0) -- (6,0);
    \draw (0,-1) -- (3,2) -- (6,2);
    \fill (1,0)   node [fill={#1}, {#2}, scale=0.4]  {};
    \fill (3,0)   node [fill={#3}, {#4}, scale=0.4]  {};
}

\newcommand{\incramp}[6]{ 
    \draw[dashed] (0,0) -- (6,0);
    \draw (0,-1) -- (2,-1) -- (4,1) -- (6,1);
    \fill (2,0)   node [fill={#1}, {#2}, scale=0.4]  {};
    \fill (3,0)   node [fill={#3}, {#4}, scale=0.4]  {};
    \fill (4,0)   node [fill={#5}, {#6}, scale=0.4]  {};
}

\newcommand{\incrampfromzero}[4]{ 
    \draw[dashed] (0,0) -- (6,0);
    \draw (0,0) -- (2,0) -- (4,2) -- (6,2);
    \fill (2,0)   node [fill={#1}, {#2}, scale=0.4]  {};
    \fill (4,0)   node [fill={#3}, {#4}, scale=0.4]  {};
}

\newcommand{\upshiftincramp}[4]{ 
    \draw[dashed] (0,0) -- (6,0);
    \draw (0,1) -- (2,1) -- (4,3) -- (6,3);
    \fill (2,0)   node [fill={#1}, {#2}, scale=0.4]  {};
    \fill (4,0)   node [fill={#3}, {#4}, scale=0.4]  {};
}

\newcommand{\downshiftincramp}[4]{ 
    \draw[dashed] (0,0) -- (6,0);
    \draw (0,-3) -- (2,-3) -- (4,-1) -- (6,-1);
    \fill (2,0)   node [fill={#1}, {#2}, scale=0.4]  {};
    \fill (4,0)   node [fill={#3}, {#4}, scale=0.4]  {};
}

\newcommand{\decramp}[6]{ 
    \draw[dashed] (0,0) -- (6,0);
    \draw (0,1) -- (2,1) -- (4,-1) -- (6,-1);
    \fill (2,0)   node [fill={#1}, {#2}, scale=0.4]  {};
    \fill (3,0)   node [fill={#3}, {#4}, scale=0.4]  {};
    \fill (4,0)   node [fill={#5}, {#6}, scale=0.4]  {};
}

\newcommand{\decramptozero}[4]{ 
    \draw[dashed] (0,0) -- (6,0);
    \draw (0,2) -- (2,2) -- (4,0) -- (6,0);
    \fill (2,0)   node [fill={#1}, {#2}, scale=0.4]  {};
    \fill (4,0)   node [fill={#3}, {#4}, scale=0.4]  {};
}

\newcommand{\upshiftdecramp}[4]{ 
    \draw[dashed] (0,0) -- (6,0);
    \draw (0,3) -- (2,3) -- (4,1) -- (6,1);
    \fill (2,0)   node [fill={#1}, {#2}, scale=0.4]  {};
    \fill (4,0)   node [fill={#3}, {#4}, scale=0.4]  {};
}

\newcommand{\downshiftdecramp}[4]{ 
    \draw[dashed] (0,0) -- (6,0);
    \draw (0,-1) -- (2,-1) -- (4,-3) -- (6,-3);
    \fill (2,0)   node [fill={#1}, {#2}, scale=0.4]  {};
    \fill (4,0)   node [fill={#3}, {#4}, scale=0.4]  {};
}

\newcommand{\posramp}[3]{ 
    \pgfmathsetmacro{\dist}{6}
    \pgfmathsetmacro{\a}{{#1}}
    \pgfmathsetmacro{\b}{{#2}}
    \pgfmathsetmacro{\height}{\b - \a}
    \draw[gray, very thin, dashed] (0,0) -- (\dist,0);
    \draw[{#3}] (0,0) -- ({#1},0) -- ({#2},\height) -- (\dist,\height);
}

\newcommand{\negramp}[2]{ 
    \pgfmathsetmacro{\dist}{6}
    \pgfmathsetmacro{\a}{{#1}}
    \pgfmathsetmacro{\b}{{#2}}
    \pgfmathsetmacro{\height}{\a - \b}
    \draw[gray, very thin, dashed] (0,0) -- (\dist,0);
    \draw (0,0) -- ({#1},0) -- ({#2},\height) -- (\dist,\height);
}

\newcommand{\posReLU}[2]{ 
    \pgfmathsetmacro{\dist}{6}
    \pgfmathsetmacro{\a}{{#1}}
    \pgfmathsetmacro{\height}{\dist - \a}
    \draw[gray, very thin, dashed] (0,0) -- (\dist,0);
    \draw[{#2}] (0,0) -- ({#1},0) --  (\dist,\height);
}

\newcommand{\negReLU}[2]{ 
    \pgfmathsetmacro{\dist}{6}
    \pgfmathsetmacro{\a}{{#1}}
    \pgfmathsetmacro{\height}{\a-\dist}
    \draw[gray, very thin, dashed] (0,0) -- (\dist,0);
    \draw[{#2}] (0,0) -- ({#1},0) --  (\dist,\height);
}

\usepackage{comment}

\usepackage[textsize=small]{todonotes}
\renewcommand{\comment}[1]{\vspace{1mm}\todo[inline,color=orange!20,bordercolor=orange]{#1}}
\newcommand{\mert}[1]{\comment{{\bf {\color{red} Mert:}} #1}}
\newcommand{\emi}[1]{{\color{blue} Emi: #1}}

\pgfplotsset{compat=1.18}
\begin{document}


\title{A Library of Mirrors: Deep Neural Nets in Low Dimensions are Convex Lasso Models with Reflection Features}

\author{
Emi Zeger\thanks{Department of Electrical Engineering}
\and 
Yifei Wang\footnotemark[1]
\and
Aaron Mishkin\thanks{Department of Computer Science}
\and
Tolga Ergen\footnotemark[1]
\\
Emmanuel Cand\`es\thanks{Departments of Statistics and Mathematics}
\and
Mert Pilanci\footnotemark[1]
\\
Stanford University
}

\maketitle



\iftoggle{long}{
\begin{figure}
    \centering
\begin{tikzpicture}
    \pgfmathsetmacro{\hint}{7} 
    \pgfmathsetmacro{\scale}{0.2}
    \begin{scope}[scale=\scale, shift={(0,0)}]
		\incReLU{red};
	\end{scope}
    \begin{scope}[scale=\scale, shift={(\hint,0)}]
        \downshiftincReLU{red}{blue};
	\end{scope}
    \begin{scope}[scale=\scale, shift={(2*\hint,0)}]
		\decReLU{blue};
	\end{scope}
    \begin{scope}[scale=\scale, shift={(3*\hint,0)}]
        \downshiftdecReLU{blue}{red};
	\end{scope}
     \begin{scope}[scale=\scale, shift={(4*\hint,0)}]
        \negdecReLU{blue}{red};
	\end{scope}
    \begin{scope}[scale=\scale, shift={(5*\hint,0)}]
        \negincReLU{blue}{red};
	\end{scope}
   \begin{scope}[scale=\scale, shift={(6*\hint,0)}]
        \incrampfromzero{blue}{red};
	\end{scope}
    \begin{scope}[scale=\scale, shift={(7*\hint,0)}]
        \incramp{blue}{red}{green};
	\end{scope}
    \begin{scope}[scale=\scale, shift={(8*\hint,0)}]
        \upshiftincramp{blue}{red};
	\end{scope}
    \begin{scope}[scale=\scale, shift={(0,-10)}]
        \downshiftincramp{blue}{red};
	\end{scope}
    \begin{scope}[scale=\scale, shift={(\hint,-10)}]
        \decramp{blue}{red}{green};
	\end{scope}
    \begin{scope}[scale=\scale, shift={(2*\hint,-10)}]
        \upshiftdecramp{red}{blue};
	\end{scope}
     \begin{scope}[scale=\scale, shift={(3*\hint,-10)}]
        \downshiftdecramp{red}{blue};
	\end{scope}
    \begin{scope}[scale=\scale, shift={(11*\hint,0)}]
        \decramptozero{blue}{green};
	\end{scope}
    \begin{scope}[scale=\scale, shift={(10*\hint,0)}]
        \decline{blue};
	\end{scope}
    \begin{scope}[scale=\scale, shift={(9*\hint,0)}]
        \incline{blue};
	\end{scope}
    \begin{scope}[scale=\scale, shift={(4*\hint,-10)}]
        \posramp{2}{4};
	\end{scope}
     \begin{scope}[scale=\scale, shift={(5*\hint,-10)}]
        \posReLU{2};
	\end{scope}
    \begin{scope}[scale=\scale, shift={(6*\hint,-10)}]
        \negReLU{2};
	\end{scope}
    \begin{scope}[scale=\scale, shift={(7*\hint,-10)}]
        \negramp{2}{4};
	\end{scope}
\end{tikzpicture}
\caption{}
\end{figure}
}


\begin{abstract}
	We prove that training neural networks 
 on \nd{1} data is equivalent to solving convex \lasso{} problems with discrete, explicitly defined dictionary matrices. 
 We consider neural networks with piecewise linear activations and depths ranging from $2$ to an arbitrary but finite number of layers.
 We first show that two-layer networks with piecewise linear activations are equivalent to Lasso models using a discrete dictionary of ramp functions, with breakpoints corresponding to the training data points.
In certain general architectures with absolute value or ReLU activations, a third layer surprisingly creates features that reflect the training data about themselves. Additional layers progressively generate reflections of these reflections.
    The \lasso{} representation provides valuable insights into the analysis of globally optimal networks, elucidating their solution landscapes and enabling closed-form solutions in certain special cases.
   Numerical results show that reflections also occur when optimizing standard deep networks using standard non-convex optimizers. Additionally, we demonstrate our theory with autoregressive time series models.

\end{abstract}
\section{Introduction}\label{sec:intro}

Training deep neural networks is an important optimization problem. However, the \nc{}ity of neural nets makes their training challenging. In this paper, we show that for low-dimensional data, e.g., \nd{1} or \nd{2}, training a deep neural network can be simplified to solving a convex \lasso{} problem with an easily constructable dictionary matrix.

Neural networks are used as predictive models for low-dimensional data in 
acoustic signal processing 
\citep{bianco2019machine,freitag2017audeep,hsu2009improvement,mavaddati2020novel,purwins2019deep, 
serra2019blow}, physics-informed machine learning problems, uncertainty quantification \citep{chen2020projected,chen2019projected,stuart2014uncertainty,wang2022projected,wang2022optimal,zahm2022certified}, and predicting financial data 
(\Cref{sec:autoreg}). 
In \citep{savarese2019infinite, ergen2021convex, ergen2021revealing}, the problem of learning \nd{1} data is studied for two-layer ReLU networks, and it is proved that an optimal two-layer ReLU neural network precisely interpolates the training data as a piecewise linear function for which the {\kink}s are at the data points. Recent work in \citep{Joshi2023,karhadkar:2023, kornowski:2023} also studied $2$-layer ReLU neural networks and their behavior on \nd{1} data.

However, even for low-dimensional data, the current literature still lacks analysis on the expressive power and learning capabilities of deeper neural networks with generic activations. This motivates us to study the optimization of $2$ and $3$-layer networks with piecewise linear activations and deeper neural networks with sign and ReLU activations. For \nd{1} data, we simplify the training problem by recasting it as a convex \lasso{} problem, which has been extensively studied
\citep{efron2004least,Tibshirani:1996,  tibshirani2013unique}.

Convex analysis of neural networks was developed in several prior works. As an example, infinite-width neural networks enable the convexification of the overall model \citep{bach2017breaking,bengio2005convex,fang2019convex}. However, due to the infinite-width assumption, these results do not reflect finite-width neural networks in practice. Recently, a series of papers \citep{ergen2020aistats, ergen2021convex,pilanci2020neural} developed a convex analytic framework for the training problem of two-layer neural networks with ReLU activation. As a follow-up work, a similar approach is used to formulate the training problem for threshold activations with data in general $d$-dimensions as a \lasso{} problem \citep{ergen2023globally}. However, the dictionary matrix is described implicitly and requires high computational complexity to create \citep{ergen2023globally}. 
By focusing on \nd{1} data, we can provide simple, explicit \lasso{} dictionaries and consider additional activations. For example, we analyze networks with sign activation, which is useful in contexts such as saving memory to meet hardware constraints \citep{Bulat:2019,Kim:2018}. 

Throughout the paper, all scalar functions extend to vector and matrix inputs component-wise. We write vectors as $\mathbf{v}=(v_1, {\cdots}, v_n)$ and denote the set of $n$-dimensional, real-valued column and row vectors by $\mathbb{R}^n$ and $\mathbb{R}^{1\times n}$, respectively. For $L \geq 2$, an $L$-layer \emph{neural network} for $d$-dimensional data is denoted by $\nn{L}{\mathbf{x}}: \mathbb{R}^{1 \times d} \to \mathbb{R}$, where $\mathbf{x} \in \mathbb{R}^{1 \times d}$ is an input row vector and
$\params$ is the \emph{parameter set}. The set $\params$ may contain matrices, vectors, and scalars representing weights and biases. We let $\params\in\paramspace$, where $\paramspace$ is the \emph{parameter space}.
Let $\mathbf{X} \in \mathbb{R}^{N \times d}$ be a \emph{data matrix} consisting of $N$ \emph{training samples} $\mathbf{x}_1, {\cdots} \mathbf{x}_N \in \mathbb{R}^{1 \times d}$. We consider regression tasks and call $\mathbf{y} \in \mathbb{R}^N$ the \emph{\labelvec{}}. The (\nc{}) neural net \textit{training problem} is 
\begin{align} \label{eq:nonconvex_generic_loss_gen_L_intro}
	\min_{\params \in \Theta}  \frac{1}{2} \| \nn{L}{\mathbf{X}}  - \mathbf{y} \|^2_2 + \frac{\beta}{\effL{}}  \|\regparams\|^{\effL{}} 
\end{align}
where $\beta>0$ is a regularization coefficient for a subset of parameters $\regparams \subset \params$ that incur a weight penalty when training. We denote $\|\regparams\|^{\effL{}} = \sum_{q \in \regparams} \|q\|_2^{\effL{}}$, which penalizes the total network weight. The results can be generalized to other $p$-norm regularizations as well. $\tilde{L}$ is the \textit{effective regularized depth}, defined to be the usual depth $L$ for ReLU, leaky ReLU, and absolute value activations, but defined to be just $2$ for threshold and sign activations. This is motivated by the property that neurons with sign or threshold activations are invariant to the magnitude of their neuron weights, so it does not make sense to penalize the inner weights (\Cref{remk:signdet_inweights0}). 
%

A central element of this paper is the \textit{\lasso{} problem} 

\begin{equation}\label{eq:genericlasso}
\min_{\mathbf{z}, \xi}  \frac{1}{2} \| \Amat{} \mathbf{z} + \xi \mathbf{1} - \mathbf{y} \|^2_2 + \symbeta \|\mathbf{z}\|_1
\end{equation}
%
where $\mathbf{z}$ is a vector, $\xi \in \mathbb{R}$, $\mathbf{1}$ is a vector of ones, and $\symbeta > 0$ (whose relation to $\beta$ in \eqref{eq:nonconvex_generic_loss_gen_L_intro} is defined in \Cref{sec:deepnarrownets}). $\Amat{}$ is called the \textit{dictionary matrix}, with columns $\Amat_i \in \mathbb{R}^N.$ 

A neural net is \textit{trained} by searching for $\params$ that optimizes \eqref{eq:nonconvex_generic_loss_gen_L_intro}. A neural net $\nn{L}{\mathbf{x}}$ is called \emph{optimal} if $\params$ is a global minimizer in \eqref{eq:nonconvex_generic_loss_gen_L_intro}. Unfortunately, training is complicated by the \nc{}ity of 
the optimization problem. However, for data of dimension $d{=}1$, we reformulate the training problem \eqref{eq:nonconvex_generic_loss_gen_L_intro} into the equivalent but simpler \lasso{} problem \eqref{eq:genericlasso},
where $\Amat{}$ is a fixed matrix that is constructed based on the training data $\mathbf{X}$ and neural net architecture. 
Moreover, the dictionary matrix columns
$\mathbf{A}_i$ correspond to piecewise linear functions we call \textit{features} that satisfy the following: $1)$ the value of the feature when evaluated on $x_n$ is $\mathbf{A}_{i,n}$, and $2)$ the function consisting of their affine combination, where the coefficients are elements of a \lasso{} solution, is equal to an optimal neural net. The set of features for a given network is a \textit{dictionary}. We call a collection of dictionaries a \textit{library} when referring to the features from multiple depths.

We explicitly provide the elements of $\Amat{}$, making it straightforward to build and solve the convex \lasso{} problem instead of solving the \nc{} training problem. This reformulation allows for exploiting fast \lasso{} solvers based on the proximal gradient method and Least Angle Regression (LARS) \cite{efron2004least}.
\iftoggle{long}{and FISTA \cite{}. }

 Whereas in the training problem \eqref{eq:nonconvex_generic_loss_gen_L_intro}, the quality of the neural net fit to the data is measured by the $l_2$ loss as $\frac{1}{2}|| \nn{L}{\mathbf{X}} -\mathbf{y} ||^2_2$,
our results generalize to a wide class of convex loss functions $\mathcal{L}_{\mathbf{y}}:\mathbb{R}^N \to \mathbb{R}$. With a general loss function, \eqref{eq:nonconvex_generic_loss_gen_L_intro} becomes  $\displaystyle \min_{\params \in \Theta} \mathcal{L}_{\mathbf{y}} (\nn{L}{\mathbf{X}}) + \frac{\beta}{\effL{}}  \|\regparams\|^{\effL{}}$. This is shown to be equivalent to the generalization of \eqref{eq:genericlasso}, namely $\displaystyle \min_{\mathbf{z}, \xi}  \mathcal{L}_{\mathbf{y}}(\Amat{}\mathbf{z} + \xi \mathbf{1} - \mathbf{y} )+ \symbeta \|\mathbf{z}\|_1$. 

The \lasso{} problem selects solutions $\mathbf{z}$ that generalize well by penalizing their total weight in $l_1$ norm \cite{Tibshirani:1996}. The $l_1$ norm typically selects a small number of elements in $\mathbf{z}$ to be nonzero.
The \lasso{} equivalence demonstrates that neural networks can learn a sparse representation of the data by selecting certain features to fit $\mathbf{y}$. 

The \lasso{} representation also elucidates the solution path of neural networks. The \emph{solution path} for the \lasso{} or training problem is the map from $\beta \in (0, \infty)$ to the solution set.
The \lasso{} solution path is well understood \citep{Tibshirani:1996,tibshirani2013unique,efron2004least}, providing insight into the solution path of the ReLU training problem \citep{mishkin2023solution}.

This paper is organized as follows. \Cref{section:nn_arch} defines various neural network architectures. \Cref{section:main_results} describes our main theoretical result: neural networks are solutions to \lasso{} problems. One important consequence is that deep networks with ReLU and absolute value activations learn geometric reflections in the data. The next sections describe applications of the \lasso{} equivalence. \Cref{section_main:square_wave} uses our theory to find explicit optimal neural networks for examples of binary data, and \Cref{sec:autoreg} uses the \lasso{} formulation to improve training of neural networks that predict financial time-series.
\Cref{sec:solsetslassovsnn} examines the relationship between the entire set of optimal neural nets given by the training problem versus the \lasso{} problem, while \Cref{sec:minnorm} applies the \lasso{} model to examine neural net behavior under minimum regularization, finding closed-form solutions. \Cref{sec:num_results} presents experiments that support our theoretical results, and shows examples where neural networks trained with \adam{} naturally exhibit \lasso{} features.

\subsection{Contributions}

\iftoggle{long}{First, note that this paper always assumes the data is \nd{1}.  Given the data is in \nd{1}, without loss of generality we always assume the training samples are ordered.}

Our contributions can be summarized as follows:

\begin{itemize}
    \item Training various neural network architectures on \nd{1} data is equivalent to solving \lasso{} problems with finite, explicit and fixed dictionaries of basis signals that grow richer with depth (Theorems \ref{thrm:lasso_deep_1nueron_ReLU}, \ref{lemma:numfeatures} and \ref{thm:3lyr_sign}). We identify dictionaries for various architectures in closed form (\Cref{def:ReLU dic func}).
    
    \item Features with reflections of training data appear in libraries for simple $3$-layer and deeper architectures with ReLU (\Cref{ReLU2neuroninformal}) or absolute value activation (\Cref{absvalinformal}). Experimentally, training these networks using the \adam{} optimizer leads to the same reflection features that we prove and matches our theoretical results on the global optima (\Cref{sec:num_results}).  In contrast, no reflection features are generated for the sign activation for any depth.


     \item Although the sign activation does not produce reflection features, its features become richer for depth $3$ networks compared to depth $2$. Accordingly, for certain binary classification tasks, we analytically observe that optimal $3$-layer sign activation networks generalize better than their $2$-layer counterparts in the sense that their predictions are more uniform. Moreover, the \lasso{} problem yields closed-form expressions for these neural networks (Corollaries \ref{cor:solfor2lyrwave} and \ref{cor:solfor3lyrwave})
    
    \item After depth $3$, the libraries freeze for networks with ReLU activation that have the same number of biased neurons in the middle and final layers (\Cref{lemma:ReLUsaturates}), and for networks with sign activation that have a constant number of neurons per layer (\Cref{thm:3lyr_sign}). But, the libraries grow when the width expands to twice as many neurons in the middle layer as the final layer for networks with ReLU activation (\Cref{thrm:lasso_deep_1nueron_ReLU}), and a tree structure for networks with sign activation (\Cref{thm:3lyr_sign}).
    
    \item A similar \lasso{} equivalence extends to neural networks trained on \nd{2} data in the upper half plane (\Cref{lemma:2D_upperhalfplane_signact}).

\iftoggle{long}{The results above are stated in section \ref{section:main_results} and proved in the appendix \ref{app_section:2lyr_nn}, \ref{app_section:3lyr_sign}. Using these three results, we also show the following.}
\iftoggle{todos}{\begin{itemize}
        \item We construct the set of all global
            minima of the convex Lasso problem and 
            leverage this to derive an analytical expression for the 
            global minima and Clarke stationary points of the non-convex
            training problem.
	\end{itemize}
}


    \iftoggle{long}{
    , neural nets with sign activation fit the data exactly under minimal regularization . The solution to the \lasso{} problem with sign activation is unique, and as $\beta \to 0$, yields a closed-form expression. 
    }
 
\end{itemize}

\subsection{Notation}\label{section:notation}
Assume \nd{1} training data is ordered as 
$x_1>x_2>{\cdots} > x_N$. 
The indicator function of a logical statement $z$ is $\mathbf{1}\{z\}$. 
Let $[n]=\{1,2,\dots,n\}$.
For a matrix $\mathbf{Z}$, let $\mathbf{Z}_S$ be the submatrix of $\mathbf{Z}$ corresponding to indices in $S$.  
\iftoggle{long}{For a matrix $\mathbf{Z} \in \mathbb{R}^{N_1 \times N_2}$ and a subset of indices $S \subset [N_1] \times [N_2]$, denote $\mathbf{Z}_S$ to be the submatrix of $\mathbf{Z}$ corresponding to the indices in $S$. }
\iftoggle{long}{The concatenation accounts for bias weights. For a vector $z \in \mathbb{R}^d$, denote $\mathrm{Diag}(z) \in \mathbb{R}^{d \times d}$ to be the diagonal matrix with $\mathrm{Diag}(z)_{i,i}=z_i$. 
}
The number of nonzero elements in a vector $\mathbf{z}$ is $||\mathbf{z}||_0$.  
Let $\mathbf{1}, \mathbf{0} \in \mathbb{R}^N$ be the all-ones and all-zeros vectors, respectively.

A network that uses ReLU activation is a "ReLU network" or "ReLU-activated network," and a feature in a \lasso{} problem that is equivalent to a ReLU network is a "ReLU feature." Similar terminology holds for other activations and architectures.
\iftoggle{long}{
, and ReLU and absolute value are special cases. }  
\iftoggle{long}{The activation function may be occasionally denoted as $\sigma_s$ to explicitly denote its dependence on $s$. We next define the following neural net architectures. }
\iftoggle{long}{\subsection{Standard neural nets}} 

\section{Neural net architectures}\label{section:nn_arch}
This section is devoted to defining neural net terminology and notation to be used throughout the rest of the paper.
 Let $L \geq 2$ be the depth of a neural network (which has $L{-}1$ hidden layers). 
\iftoggle{long}{The final, outer layer has $m_{L-1}$ neurons. 
}
We assume the activation $\sigma: \mathbb{R}\to\mathbb{R}$ is \textit{piecewise linear around $0$} , i.e., of the form \\ 
$$	\begin{aligned}
		\sigma(x) & =
		\begin{cases}
			\negslope x + \negbias & \mbox{ if } x < 0     \\
			\posslope x + \posbias & \mbox{ else } 
		\end{cases}
	\end{aligned}
 $$
 for some $\negslope, \posslope, \negbias, \posbias \in \mathbb{R}$. As shorthand, "piecewise linear" will mean "piecewise linear around $0$."
We focus on the piecewise linear activations of ReLU, leaky ReLU, absolute value, sign, and threshold functions. 
The ReLU activation is $\sigma(x)=(x)_+ := \max\{x,0\}$, and absolute value activation is $\sigma(x)=|x|$.
\iftoggle{long}{Appendix  \ref{app_section:act_func} defines these activation functions and generalizes results to activations satisfying properties such as being piecewise linear about $0$, symmetric, or bounded. 
}
 The leaky ReLU generalizes ReLU and absolute value as $\sigma(x)=(\posslope\mathbf{1}\{x>0\}+\negslope\mathbf{1}\{x<0\})x$ where $\posslope \neq \negslope$. ReLU, leaky ReLU and absolute value activations will be referred to as "continuous piecewise linear." The threshold activation is $\sigma(x)=\mathbf{1}\{x\geq0\}$, and the sign activation is $\sigma(x) =   \mbox{sign}(x)$, where $\mbox{sign}(x)$ is $-1$ if $x<0$, and $1$ if $x \geq 0$. Note $\mbox{sign}(0)=1$. 

For $\mathbf{Z} \in \mathbb{R}^{n \times m}, \mathbf{s} \in \mathbb{R}^{m}$, let $\sigma_{\mathbf{s}}(\mathbf{Z}) = \sigma(\mathbf{Z})\text{Diag}(\mathbf{s})$. 
When the columns of
$\sigma(\mathbf{Z})$ are neuron outputs, the $i^{\text{th}}$ column of $\sigma_{\mathbf{s}}(\mathbf{Z}) \in \mathbb{R}^{N \times m}$ represents a neuron scaled by an \textit{amplitude parameter} $s_i \in \mathbb{R}$. Amplitude parameters are (trainable) parameters for only the sign and threshold activations, and are to be ignored by interpreting them as $1$ for ReLU, leaky ReLU, and absolute value activations. 

Next, we define some neural net architectures. The parameter set is partitioned into $\params = \regparams \cup \biasparams \cup \{\xi\}$, where $\biasparams$ is a set of \emph{internal bias} terms, and $\xi$ is an \emph{external} bias term. We define the elements of each parameter set below. We define neural nets by their output on row vectors $\mathbf{x} \in \mathbb{R}^{1 \times d}$. Their outputs then extend to matrix inputs $\mathbf{X} \in \mathbb{R}^{N \times d}$ row-wise.



\subsection{Standard networks}\label{sec:stdnetworks}
The following is a commonly studied neural net architecture. 
Let $L\geq 2$, the number of layers. Let $m_0=d,m_{L-1}=1$ and $m_l \in \mathbb{N}$ for $l \in [L-2]$, which are the number of neurons in each layer. For $l \in [L-1]$, let $\mathbf{W}^{(l)} \in \mathbb{R}^{m_{l-1} \times m_l}, \mathbf{s}^{(l)} \in \mathbb{R}^{m_l}, \mathbf{b}^{(l)} \in \mathbb{R}^{1 \times m_l}, \xi \in \mathbb{R}$, which denote weights, amplitude parameters, internal biases, and external bias, respectively. Let $ \mathbf{X}^{(1)} = \mathbf{x} \in \mathbb{R}^{1\times d}$ be the input to the neural net and $\mathbf{X}^{(l+1)} \in \mathbb{R}^{1 \times m_{l}}$ be viewed as the inputs to layer $l+1$, defined by

\begin{equation}\label{eq:std_nn}
    \mathbf{X}^{(l+1)} = \sigma_{\mathbf{s}^{(l)}} \left( \mathbf{X}^{(l)}\mathbf{W}^{(l)}+ \mathbf{b}^{(l)} \right). 
\end{equation}

 Let $\boldsymbol{\alpha} \in \mathbb{R}^{m_L}$, which is the vector of final layer coefficients. A \textit{standard neural network} is $\nn{L}{\mathbf{x}} = \xi + \mathbf{X}^{(L)}\boldsymbol{\alpha}$. The regularized and bias parameter sets are \\ $\regparams {=} \left\{ \boldsymbol{\alpha}, \mathbf{W}^{(l)}, \mathbf{s}^{(l)}: l \in [L{-}1] \right \}$ and $\biasparams {=} \left \{ \mathbf{b}^{(l)}: l \in [L-1] \right\} $, respectively. 

The ultimate goal is analyzing the training problem for standard networks, but this appears to be challenging. However, by changing the architecture to a parallel structure defined next, we show that the training problem simplifies to the \lasso{} problem. These alternative architectures allow neural nets to be reconstructed more tractably from a \lasso{} solution than with a standard network.
In the parallel architecture, $m_L$ is the number of neurons in the final layer and for $i \in [m_L]$, we define the disjoint unions $\displaystyle \regparams = \bigcup_{i \in [m_L]}\regparams^{(i)}$ and $\displaystyle \biasparams = \bigcup_{i\in[m_L]}\biasparams^{(i)}$. 

\subsection{Parallel networks}\label{sec:parnet}

A parallel network is a linear combination of standard networks in parallel, as we now define. Each standard network is called a \textit{parallel unit}.
Let $L \geq 2, m_0=d, m_{L-1}=1$ and $m_l \in \mathbb{N}$ for $l \in [L]-\{L-1\}$.  For $i \in [m_{L}], l \in [L-1]$, let $\Wul{i}{l} \in \mathbb{R}^{m_{l-1} \times m_{l}}$, $\sul{i}{l} \in \mathbb{R}^{m_l}$, $\bul{i}{l} \in \mathbb{R}^{1 \times m_l}, \xi \in \mathbb{R}$, which are the weights, amplitude parameters, and biases of the $i^{th}$ parallel unit. Let $\Xul{i}{1} = \mathbf{x}\in \mathbb{R}^{1\times d}$ be the input to the neural net and $ \Xul{i}{l+1} \in \mathbb{R}^{1 \times m_{l}}$ be viewed as the input to layer $l+1$ in unit $i$, defined by 
\begin{equation}\label{eq:parallelnn_rec}
    \Xul{i}{l+1}= \sigma_{\sul{i}{l}} \left( \Xul{i}{l}\Wul{i}{l}+\bul{i}{l} \right). 
\end{equation}
 Let $\boldsymbol{\alpha} \in \mathbb{R}^{m_L}$. A \textit{parallel neural network} is $\nn{L}{\mathbf{x}} = \xi + \displaystyle \sum_{i=1}^{m_{L}}  \Xul{i}{L}\alpha_i$. 
The regularized and bias parameter sets are $\regparams^{(i)} = \left \{\alpha_i, \sul{i}{l}, \Wul{i}{l}: l \in [L-1] \right \}, \biasparams^{(i)} = \left \{ \bul{i}{l}: l \in [L-1] \right\}$, for $ i \in [m_{L}]$. 
%
%
%
We view parallel units as functions $\Xul{i}{L}: \mathbb{R}^{1\times d} \to \mathbb{R}$ and with abuse of notation write
$\Xul{i}{L}(\bf{x}) \in \mathbb{R}$ as the output of $\Xul{i}{L}$ evaluated on a sample $\mathbf{x}$. 
For a training dataset $\mathbf{X}\in \mathbb{R}^{N\times d}$, we denote the evaluations of the functions $\Xul{i}{L}$ on the training data as $\Xul{i}{L}(\mathbf{X})$.
This paper primarily focuses on parallel architectures. However, a parallel network can be converted into a standard network (\Cref{parallel2std}). 
%
%
\iftoggle{long}{
A $L=2$ layer neural network (one hidden layer) is defined as
%
%
\begin{equation}\label{eq:2layernn}
\nn{2}{\mathbf{x}}
	=\sigma(\mathbf{x} \mathbf{W_0} + \mathbf{b})\boldsymbol{\alpha}+ \xi .
\end{equation}
\iftoggle{long}{
	Which can also be written as,
	\begin{equation}
		\begin{aligned}
			f_2(x)
			 & =\sigma([x,1] \mathbf{W})\boldsymbol{\alpha} + \xi              \\
			 & = \sum_{i=2}^{m_1} \alpha_i \sigma(x \mathbf{w_i} + b_i) + \xi, \\
		\end{aligned}
	\end{equation}
}

 In \eqref{eq:nonconvex_generic_loss_gen_L_intro}, if the activation $\sigma$ is sign or threshold then the set of $2$-layer trainable parameters is $\params = \{\boldsymbol{\alpha} \in \mathbb{R}^{m_1}, \xi \in \mathbb{R}, \mathbf{W_0} \in \mathbb{R}^{d \times m_1}, \mathbf{b} \in \mathbb{R}^{1 \times m} \}$, and  the subset of parameters with weight penalty is  $\regparams = \{ \boldsymbol{\alpha}, \mathbf{W_0} \}$. If the activation is ReLU, leaky ReLU or absolute value, then $\params = \{\boldsymbol{\alpha} \in \mathbb{R}^{m_1}, \xi \in \mathbb{R}, s \in \mathbb{R}^{m_1}, \mathbf{W_0} \in \mathbb{R}^{d \times m_1}, \mathbf{b} \in \mathbb{R}^{1 \times m} \}$, and  $\regparams = \{\mathbf{W_0}, s, \boldsymbol{\alpha} \}$. The parameters $b_i$ are considered to be bias terms within neurons while $\xi$ is referred to as an external bias term outside the neural net. Both $\mathbf{b}$ and $\xi$ bias terms are excluded from regularization in $\regparams$.

\iftoggle{long}{and $ \mathbf{W} := \begin{pmatrix} \mathbf{W_0} \\ \mathbf{b} \end{pmatrix} \in \mathbb{R}^{(d+1) \times m}$ }

\iftoggle{long}{Note $m_1$ is the number of neurons. In \eqref{eq:2layernn}, $ \mathbf{w_i} \in \mathbb{R}^{d}$ denotes the $i^{th}$ column of $\mathbf{W_0}$. When the data is one-dimensional, $\mathbf{x}^T \mathbf{w_i} = xw_i$, where $w_i \in \mathbb{R}$. }

\iftoggle{long}{For 2-layer networks, we consider activations $\sigma$ to be ReLU, leaky ReLU, absolute value, sign, or threshold functions. For sign and threshold activations, the trainable parameters also include the amplitude parameter $s$. We only consider $3$-layer networks with sign or threshold activation $\sigma$.\\}

A $3$-layer \textit{standard} neural net has the form, $\nn{3,std}{\mathbf{x}}  :=  \sigma \left ( \left[ \sigma \left([\mathbf{x},1] \mathbf{W^{(1)}} \right), \mathbf{1}  \right ]   \mathbf{W^{(2)}}\right) \boldsymbol{ \alpha } + \xi$ \label{eq:3lyr_nn_std}.
\iftoggle{long}{In the standard neural net defined in  \eqref{eq:3lyr_nn_std}, $\mathbf{W^{(1)}}  \in \mathbb{R}^{2 \times m_1}, \mathbf{W^{(2)}}  \in \mathbb{R}^{m_1}, \boldsymbol{\alpha}  \in \mathbb{R}^{m_2}, \xi \in \mathbb{R}$ are trainable parameters in $\Theta$. Note $m_1$ and $m_2$ are the number of first and second hidden layer neurons, respectively. For sign and threshold activations, the trainable parameters also include two amplitude parameters $s$, one for each of the inner and outer activations.}
\iftoggle{long}{\subsection{Parallel neural nets}}
However, for $L \geq  3$ layers, instead of the standard architecture we consider the more general $3$-layer \textit{parallel} neural net, which is  defined as the sum of independent units of $3$-layer standard neural nets:
\begin{equation}\label{eq:3lyr_nn_par} 
 \nn{3,par}{\mathbf{x}}
 := \sum_{i=1}^{m_2} \sigma \left ( \left [ \sigma \left([\mathbf{x},1] \Wul{i}{1}   \right), \mathbf{1} \right ] \mathbf{W^{(i,2)}} \right) \alpha_i + \xi.
\end{equation}

If the activation $\sigma$ is sign or threshold then $\params=\{\boldsymbol{\alpha}  \in \mathbb{R}^{m_2}, \mathbf{W^{(i,l)}}  \in \mathbb{R}^{2 \times m_1}, \mathbf{W^{(i,2)}} \in \mathbb{R}^{m_1+1}, \xi, s^{(i,1)}, s^{(i,2)} \in \mathbb{R}: i \in [m_2] \}$ and 
$\regparams = \{\boldsymbol{\alpha}  , \Wul{i}{1}_{1:d,1:m_1}, \mathbf{W^{(i,2)}}, s^{(i,1)}, s^{(i,2)}: i \in [m_2] \}$. If the activation $\sigma$ is ReLU, leaky ReLU or absolute value then $\params=\{\boldsymbol{\alpha}   \in \mathbb{R}^{m_2}, \mathbf{W^{(i,l)}}   \in \mathbb{R}^{2 \times m_1}, \mathbf{W^{(i,2)}}  \in \mathbb{R}^{m_1}, \xi \in \mathbb{R}: i \in [m_2] \}$ and 
$\regparams = \{\boldsymbol{\alpha}, 
\Wul{i}{l}_{1:d,1:m_1}, \Wul{i}{2}: i \in [m_2] \}$.
\iftoggle{long}{Note and $m_1$ and $m_2$ are the number of first and second hidden layer neurons, respectively. For sign and threshold activations, the trainable parameters also include $2m_2$ amplitude parameters $s$, one for each of the inner and outer activations in each of the $m_2$ parallel units. By increasing the number of neurons in the standard neural  net,}
Deeper $L$-layer neural nets are defined similarly (see equation \eqref{parLlayernn_signdet} in \Cref{app_sec:parallel_nn_def}). 
A parallel neural net can be converted to a standard neural net
(\Cref{section:par2std}). 
\iftoggle{todos}{\textcolor{red}{todo: revise this appendix section}}

\iftoggle{long}{\subsection{Tree structured neural nets}}

For $L>3$ layers, we also consider a fully parallel, or ``tree" neural network, defined recursively. While each branch of the final layer of parallel neural network is a standard network, every branch of every layer of the tree network is a tree network. \Cref{app_subsection:tree_nn_def} contains a formal definition. 
The definitions of standard, parallel, and tree neural nets coincide for $L=2$ layers. The definitions of parallel and tree neural nets coincide for $L=3$ layers. For $L \geq 4$ layers, the definitions all diverge.

\iftoggle{long}{
We consider the squared loss function $\mathcal{L}_{\mathbf{y}}(\mathbf{z})=\frac{1}{2}||\mathbf{z}-\mathbf{y}||^2_2$ for simplicity.  In the appendix proofs, we generalize to an
arbitrary loss function $\mathcal{L}_{\mathbf{y}}:\mathbb{R}^N \to \mathbb{R}$ that is parameterized by the target vector $\mathbf{y}$, and which is convex and lower semicontinuous \cite{Borwein:2000} \label{assump:Loss_func}. Generalizing to arbitrary losses also allows absorption of the external bias $\xi$ into the loss function to simplify derivations (see
\Cref{app_section:loss_absorbs_xi}.)
}
}
%
\section{\secnameMain}\label{section:main_results}

In this section, we show that \nc{} deep neural net training problems are \textit{equivalent} to \lasso{} problems, that is, their optimal values are the same, and given a \lasso{} solution, we can reconstruct a neural net that is optimal in the training problem. We say that a neural network \emph{model} is equivalent to a \lasso{} model to mean that the optimal models are convertible to each other. We note that the optimal solutions may not be unique. Discussion of the relation between the solution sets of the \lasso{} and \nc{} training problems with respect to a specified straightforward reconstruction is given in \Cref{sec:solsetslassovsnn}. Further analysis of all possible reconstruction maps between the models is an area of future work. Analysis of the span or uniqueness and the generalizing abilities of different optimal or stationary solutions to the training problem is also an area for future work.

Unless otherwise stated, for the rest of this paper assume the data is \nd{1}. Define $\symbeta=\frac{\beta}{2}$ in \eqref{eq:nonconvex_generic_loss_gen_L_intro},\eqref{eq:genericlasso} for $3$-layer \symmeterized{} networks (defined below). For all other networks, let $\symbeta=\beta$.  


\iftoggle{long}{As stated in \eqref{eq:nonconvex_generic_loss_gen_L_intro}, the general training problem for a neural net with an arbitrary number of layers $L$, dimension $d$, number of samples $N$, data matrix $X \in \mathbb{R}^{N \times d}$ and loss function $\mathcal{L}_{\mathbf{y}}$ is,
\begin{align} \label{eq:nonconvex_generic_loss_gen_L}
	\min_{\Theta} \mathcal{L}_{\mathbf{y}} (f_L(\mathbf{X})) + \frac{\beta}{2} \|\regparams\|_2^2.
\end{align}}




\iftoggle{long}{

We define a $L$-layer, (ReLU) \textit{\deepReLU{}} as $\nn{L}{x}: \mathbb{R} \to \mathbb{R}$,

\begin{equation}\label{Llyr_seriesnn}
    \nn{L}{x} = \sum_{j=1}^{m_{L-1}} \sigma \left( \left( \cdots \sigma \left(\sigma \left( x w^{(j,1)} + b^{(j,1)} \right) w^{(j,2)} + b^{(j,2)}  \right) \cdots \right) w^{(j,L-1)} + b^{(j,L-1)}\right )\alpha_j + \xi.
\end{equation}

This network has parallel structure, ReLU activations and one neuron in each layer, and takes in \nd{1} data. The parameter set is $\params = \{ \alpha_j, w^{(j,l)}, b^{(j,l)}, \xi: j \in [m_{L-1}], l \in [L-1]\}$.
We define the following \nc{} \textit{training problem} for the \deepReLU{} \eqref{Llyr_seriesnn}:

\begin{equation}\label{eq:noncvxtrain_LlyrLLnorm}
    \min_{\params \in \Theta}  \mathcal{L}_{\mathbf{y}} (\nn{L}{\mathbf{X}})
     + \frac{\beta}{L}  \|\regparams\|_L^L,
\end{equation}

where the regularized parameters are $\regparams^{(j)} = \left \{w^{(j,l)}, \alpha_j:l \in [L-1] \right \}$ for $j \in [m_{L-1}]$ and $\regparams =  \bigcup_{l \in [m_{L-1}]} \regparams^{(l)}$, and they are penalized by their $l_L$ norm as $ \|\regparams\|_L^L = \sum_{q \in \regparams} q^L$. 
}

\subsection{$2$-layer networks with piecewise linear activations and deep networks with continuous piecewise linear activations}\label{sec:deepnarrownets}
 
A \emph{\deepReLU{}} is a parallel network where the number of neurons in each parallel path is equal to 1, i.e., $m_1{=}{\cdots}{=}m_{L-1}{=}1$. In other words, a \deepReLU{} has $m_L$ parallel units, each of which has one neuron in every layer, i.e., $m_L$ neurons across units in each layer, adding up to $Lm_L$ total neurons. For a $2$-layer network, the parallel and \deepnarrow{} networks are the same as the standard network. A \emph{\symmeterized{} 3-layer network} is a parallel network with continuous piecewise linear activation where $m_1{=}2$ (which means each $i^{\text{th}}$ unit out of $m_3$ parallel units has \nd{2} weight vectors $\Wul{i}{1}$ and $\Wul{i}{2}$ for the first and second layer, respectively) and the parameter space $\paramspace$ enforces the constraint $\left|\Wul{i}{l}_{1}\right|{=}\left|\Wul{i}{l}_{2}\right|$ for $l \in [2], i \in [m_3]$. Therefore a \symmeterized{} 3-layer network has $2m_3$ neurons in the "middle" layer. A \symmeterized{} network extends the expressibility of a \deepReLU{} network by expanding its width. For other architectures, see \Cref{sec:deepsign}.

\begin{figure}[t!]
    \centering
    \begin{tikzpicture}
    \pgfmathsetmacro{\hint}{12}
    \pgfmathsetmacro{\vint}{-6}
    \pgfmathsetmacro{\scale}{0.5}
    \begin{scope}[scale=\scale,shift={(0*\hint,0)}]
        \draw[dashed] (-2,0) -- (8,0);
    \draw (-2,4) -- (2,0) -- (3,1) -- (4,0) --(8,4);
    \node[scale=0.5] (a1) at (2,0) [circle,fill=\jtwocol, label=below left:$x_{j_2}$]{};
    \node[scale=0.5] (a2) at (3,0) [circle,fill=\jonecol, label=below:$x_{j_1}$]{};
    \node[scale=0.5] (a3) at (4,0) [circle,fill=\jonecol, draw=\jtwocol, line width=1.5, label=below right:$\reflect{x_{j_2}}{x_{j_1}}$]{}; 
    \end{scope}
    \begin{scope}[scale=\scale,shift={(\hint,0)}]
    \draw[dashed] (0,0) -- (12,0);
    \draw (0,2) -- (2,0) -- (6,4) -- (10,0) --(12,2);
    \node[scale=0.5] (a1) at (2,0) [circle,fill=\jtwocol, draw = \jonejtwocol, line width = 2, label=below left:\small$\reflect{\frac{x_{j_1}+x_{j_2}}{2}}{x_{j_2}}$]{};
    \node[scale=0.5] (a1) at (4,0) [circle,fill=\jtwocol, label=below :\small$x_{j_2}$]{};
    \node[scale=0.5] (a2) at (6,0) [circle,fill=\jonejtwocol, label=below:\small$\frac{x_{j_1}+x_{j_2}}{2}$]{};
    \node[scale=0.5] (a1) at (8,0) [circle,fill=\jonecol, label=below :\small$x_{j_1}$]{};
    \node[scale=0.5] (a3) at (10,0) [circle,draw=\jonejtwocol, line width = 1.5, fill=\jonecol, label=below right:\small$\reflect{\frac{x_{j_1}+x_{j_2}}{2}}{x_{j_1}}$]{}; 
    \end{scope}
   \begin{scope}[scale=2*\scale,shift={(0,0.7*\vint)}]
        \draw[dashed] (0,0) -- (14,0);
        \draw[] (0,3) -- (3,0) -- (4,1) -- (5,0) -- (7,2) -- (9,0) -- (10,1) -- (11,0) -- (14,3);
        \node[scale=0.5] (xj) at (4,0) [circle,fill=\jtwocol, label=below : \footnotesize$x_{j_2}$]{};
        \node[scale=0.5] (xk) at (5,0) [circle,fill=\jthreecol, label=below :\footnotesize$x_{j_3}$]{};
        \node[scale=0.5] (xi) at (7,0) [circle,fill=\jonecol, label=below:\footnotesize$x_{j_1}$]{};
        \node[scale=0.5] (xi) at (3,0) [circle,draw=\jthreecol, fill=\jtwocol, line width=1.5pt, label=below :\footnotesize$\reflect{x_{j_3}}{x_{j_2}}$]{};
        \node[scale=0.5] (xi) at (9,0) [circle,draw=\jthreecol, fill=\jonecol, line width=1.5pt, label=below left:\footnotesize$\reflect{x_{j_3}}{x_{j_1}}$]{};
        \node[scale=0.5] (xi) at (10,0) [circle,draw=\jtwocol, fill=\jonecol, line width=1.5pt, label=below :\footnotesize$\reflect{x_{j_2}}{x_{j_1}}$]{};
        \coordinate (RR) at (11,0);
        \draw[fill=\jthreecol,draw=\jthreecol] (RR) circle (0.15);
        \draw[fill=\jtwocol,draw=\jtwocol] (RR) circle (0.1);
        \draw[fill=\jonecol,draw=\jonecol] (RR) circle (0.05);
        \node[scale=0.5] (xi) at (RR) [label=below right:\footnotesize$\reflect{\reflect{x_{j_3}}{x_{j_2}}}{x_{j_1}}$]{};
   \end{scope}
    \end{tikzpicture}
    \caption{Example features, not including reversed directions, for \deepnarrow{} networks with absolute value activation. Top row: $3$-layer features. The top left feature contains a \kink{} at the reflection of $x_{j_2}$ (\jtwocol{}) across $x_{j_1}$ (\jonecol{}), which is denoted as $\reflect{x_{j_2}}{x_{j_1}}$ (\jtwocol{} encircling \jonecol{}). Other \kink{}s are colored similarly. 
    Bottom row: an example of a $4$-layer feature, which contains a double reflection of $x_{j_3}$ (\jthreecol{}) reflected across $x_{j_2}$ (\jtwocol{}), then reflected across $x_{j_1}$ (\jonecol{}), which is denoted as $\reflectsmall{\reflecttiny{x_{j_3}}{x_{j_2}}}{x_{j_1}}$ (\jthreecol{} encircling \jtwocol{}, encircling \jonecol{}).
    All lines have slopes $\pm1$, and $x_{j_1},x_{j_2}, x_{j_3}$ are training data. 
    }
    \label{fig:absvalfeattikz}
\end{figure}

In this section, we focus on the architectures discussed above to derive explicit, simple features that provide some of the first steps towards intuitively understanding the representation power of neural networks. We reformulate the training problem for $2$-layer networks and for deeper depths with absolute value, ReLU and leaky ReLU activations into a convex \lasso{} problem. In general, our convexification approach and proofs provide a framework to analyze neural networks with more arbitrary widths. Additional results are deferred to \Cref{app:reconstrct} due to space, and proofs are deferred
to \Cref{app:deepReLU}.

We first discuss networks with absolute value activation, as the symmetry of $|x|$ significantly simplifies the features. However, absolute value activation models are equivalent to ReLU activation models, as long as a skip connection is present.
A $2$-layer network with a \textit{skip connection} is $\nnbiased{L}{x}=\nn{L}{x}+\skipslope x$ if $x \in \mathbb{R}$ (or more generally, $\nnbiased{L}{\mathbf{x}}=\nn{L}{\mathbf{x}}+\mathbf{x}\skipslope$) where $\skipslope \in \mathbb{R}^{d}$ is a trainable parameter in $\params$.

\begin{lemma}\label{lemma:absvalReLUskipconnection}
    The training problem for a $2$-layer network with skip connection and ReLU activation remains equivalent if the activation is changed to absolute value, and
there is a map between the solutions for either activation.
\end{lemma} 
By \Cref{lemma:absvalReLUskipconnection}, the absolute value activation is of interest to analyze as it can map to ReLU when skip connections are incorporated. Next we define some terms used to state our results.

%
For a piecewise linear function $f:\mathbb{R} \to \mathbb{R}$, $x$ \emph{is a \kink} of $f$ (alternatively, $f$ \emph{has a \kink{} at} $x$) if $f$ changes slope or is discontinuous at $x$. A \kink{} is a "kink" in the graph of $f$. For $a,b,c \in \mathbb{R}$, the \emph{reflection of $a$ about $b$} is $\reflect{a}{b} = 2b-a$. A \emph{double reflection} is a reflection of a reflection, i.e., is of the form $\reflect{\reflect{a}{c}}{b}$ or $\reflect{b}{\reflect{a}{c}}$. The \emph{generalized reflection of $a$ and $c$ about $b$} is $a+c-b = \reflect{b}{\frac{a+c}{2}}$, the reflection of $b$ about the average of $a$ and $c$. When $a=c$, the generalized reflection of $a$ and $c$ about $b$ is the reflection of $a$ about $b$. A \kink{} that is of the form $\reflect{x_{j_1}}{x_{j_2}}$ for training data $x_{j_1},x_{j_2}$ is called a \textit{reflection \kink{}} and a feature with a reflection \kink{} is called a \textit{reflection feature}; and similarly for double reflections. Networks with absolute value activation can be modeled as \lasso{} problems with reflection features, as stated next. 

\begin{figure}[t!]
    \centering
    \begin{minipage}[t!]{\textwidth}
        \includegraphics[width=0.9\linewidth]{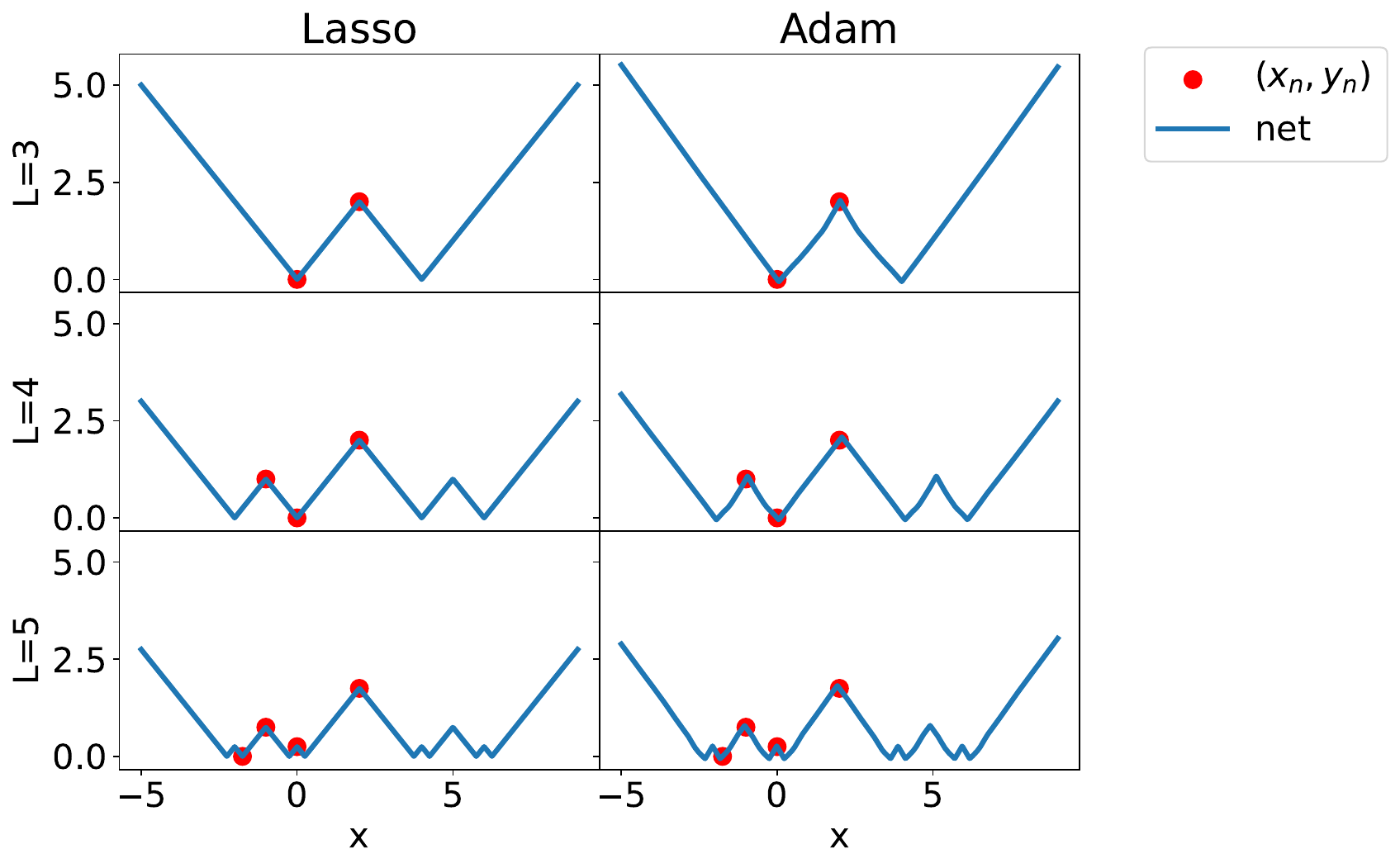}
    \end{minipage}
    \caption{\lasso{} and \adam{}-trained \deepnarrow{} networks with absolute value activation. 
    For $L{=}3$, the \kink{} at $2$ is not a training point; it is the reflection of $x_2{=}0$ across $x_1{=}1$. For $L{=}4$, the \kink{} at $6$ is not a training point; it is $x_2{=}0$ reflected about $x_3{=}{-}1$ to ${-}2$ (which not a training point) and then reflected across $x_1{=}2$. Similarly the $5$-layer network contains more complex reflections.}
    \label{fig:absval3lyrnet}
\end{figure}

\begin{theorem}[\lasso{} equivalent of deep absolute value networks]\label{absvalinformal}
    A \deepnarrow{} network of arbitrary depth with $\sigma(x)=|x|$ is equivalent\footnote{See Section \ref{section:main_results} for the definition of model equivalence.} to a \lasso{} model with a finite set of features. Its dictionary matrix for $2$ layers is $\Amat{}_{i,j}=|x_i-x_j|$. For $3$ and $4$ layers, its library includes features whose $i^{\mathrm{th}}$ element is $\big||x_i-x_{j_1}|-|x_{j_2}-x_{j_1}|\big|$ and $\Big|\big||x_i {-}x_{j_1} | {-} |x_{j_2}{-}x_{j_1}| \big| {-}
   \big||x_{j_3} {-}x_{j_1} | {-} |x_{j_2}{-}x_{j_1}| \big|  \Big|$, respectively, over all training samples $x_i,x_{j_1},x_{j_2}, x_{j_3}$. A similar pattern holds for deeper networks. The $2$-layer features have \kink{}s exactly at training data. The libraries for $3$ and $4$ layers additionally include reflection and double reflection features, respectively.
\end{theorem}

\Cref{absvalinformal} implies that an absolute value network learns to model data with a discrete and fixed dictionary of features. \Cref{fig:absvalfeattikz} plots these features for $L{=}3$ and $L{=}4$. It illustrates that the \kink{}s of the $L{=}3$ features occur at training data reflections, averages, and reflections about averages. The bottom row plots a possible feature for $4$ layers, which shows \kink{}s at reflections and double reflections.
Even with just one neuron per layer (per path for parallel networks), adding a third layer in neural networks with absolute value activation adds features to the dictionary, and adds new locations of \kink{} to those features, namely at reflections $\reflect{x_i}{x_j}$ of data points about themselves. Adding a fourth layer creates double reflections. 
In \Cref{fig:absval3lyrnet}, standard $3,4,$ and $5$-layer networks are trained with \adam{}. We make $\beta{=}10^{-7}$ close to zero, and also solve the \lasso{} problem as $\beta {\to} 0$ (\Cref{sec:minnorm}). The \adam{}-trained networks closely match the \lasso{} solutions. Moreover, the $5$-layer networks suggest that features continue to gain more complex reflections with depth.
Simulations details are given in \Cref{sec:num_results}.

\Cref{absvalinformal} describe a subset of the library for $L \geq 3$ layers. The full library for $3$ layers is defined in \Cref{thrm:lasso_deep_1nueron_ReLU} and the features are explicitly described in \Cref{def:ReLU dic func}. Our approach lays a foundation to enumerate the full library for $L\geq4$ layers as an area of future work.

In contrast, as will be shown in \Cref{thm:3lyr_sign}, the sign activation dictionary has no reflection features at any depth, which may limit its expressibility \citep{Minsky:2017}. A similar argument applies to the threshold activation. Reflection features allow neural networks to fit functions with \kink{}s at locations in between data points. The reflection \kink{}s for absolute value networks suggest that they can learn geometric structures or symmetries from the data.
Moreover, as the depth increases, this dictionary expands, which deepens its representation power.

With intuition from absolute value networks, we next discuss ReLU networks. Since reflection features appear in $3$-layer networks with absolute value activation and $|x| {=} \frac{(x)_+{+}(-x)_+}{2}$, we might expect reflections to also appear in $3$-layer ReLU networks with twice as many neurons in the middle layer as the absolute value network. This is indeed the case, as shown below. First, we define parameterized families of functions from $\mathbb{R}$ to $\mathbb{R}$ that will be used to describe features for ReLU networks.
\begin{definition}\label[definition]{def:ramps}
      Let $a_1 \in [-\infty, \infty), a_2 \in (-\infty, \infty]$. 
  The \emph{\rampname} functions are 
         \begin{align*}
                \ramp{+}{a_1}{a_2}(x) &= 
                \begin{cases}
                    0 & \mbox{ if } x \leq a_1 \\
                    x-a_1 & \mbox{ if } a_1 \leq x \leq a_2 \\
                    a_2-a_1  & \mbox{ if } x \geq a_2
                \end{cases},
&
                \ramp{-}{a_1}{a_2}(x) &= 
                \begin{cases}
                    a_2-a_1 & \mbox{ if } x \leq a_1 \\
                    a_2-x & \mbox{ if } a_1 \leq x \leq a_2 \\
                    0  & \mbox{ if } x \geq a_2 
                \end{cases}
            \end{align*}
 provided that $a_1 \leq a_2$, and otherwise $\ramp{+}{a_1}{a_2}=\ramp{-}{a_1}{a_2}=0$. 
 In particular, the \emph{\ReLUname} functions are $\ReLU{+}{a}(x)=\ramp{+}{a}{\infty} = (x-a)_+$ and $ \ReLU{-}{a}(x)=\ramp{-}{-\infty}{a}=(a-x)_+$. 
 \iftoggle{long}{
 , which will be used in \Cref{def:ReLU dic func}.}
\end{definition}
\hfill \break
In \Cref{def:ramps}, the parameters $a, a_1, a_2$ are the {\kink}s of {\ReLUname} and {\rampname} functions. This is illustrated in \Cref{fig:ReLUramp}. 
 \begin{figure}[t!]
    \centering
    \begin{tikzpicture}
    \pgfmathsetmacro{\hint}{10} 
    \pgfmathsetmacro{\scale}{0.4}
    \begin{scope}[scale=\scale, shift={(0*\hint,0)}]
        \incReLU{green}{circle} 
        \node[] () at (3,-3) {$\ReLU{+}{a}(x)$};
        \node [] at (3,-1)  {\footnotesize $a$};
	\end{scope}
    \begin{scope}[scale=\scale, shift={(1*\hint,0)}]
        \decReLU{green}{\snegshape}
        \node[] () at (3,-3) {$\ReLU{-}{a}(x)$};
        \node [] at (3,-1)  {\footnotesize $a$};
	\end{scope}
    \begin{scope}[scale=\scale, shift={(2*\hint,0)}]
        \incrampfromzero{green}{circle}{red}{circle}
        \node[] () at (3,-3) {$\ramp{+}{a_1}{a_2}(x)$};
        \node [] at (2,-1)  {\footnotesize $a_1$};
        \node [] at (4,-1)  {\footnotesize $a_2$};
	\end{scope}
    \begin{scope}[scale=\scale, shift={(3*\hint,0)}]
        \decramptozero{green}{circle}{red}{circle}
        \node[] () at (3,-3) {$\ramp{-}{a_1}{a_2}(x)$};
        \node [] at (2,-1)  {\footnotesize $a_1$};
        \node [] at (4,-1)  {\footnotesize $a_2$};
	\end{scope}

    \end{tikzpicture}
    \caption{Examples of {\rampname} functions in \Cref{def:ramps}. If $a, a_1, a_2 {\in} \{x_1,{\cdots}, x_N\}$, these functions are \lasso{} features for $3$-layer \deepnarrow{} networks with ReLU activation. 
    }
    \label[figure]{fig:ReLUramp}
\end{figure}
Using these features, we state our results on \lasso{} equivalence for ReLU networks. 

\begin{theorem}[\deepnarrow{} ReLU network representation capability stagnates]\label{ReLUinformal}
A \deepReLU{} of arbitrary depth with ReLU activation 
is equivalent to a \lasso{} model with a finite set of features. Its library contains only \ReLUname{}s and \rampname{}s, and beyond $3$ layers, ReLU features do not change as the network deepens.
\end{theorem}

In contrast to absolute value activation, for \deepnarrow{} networks, the ReLU library never gains reflection features. However, a \symmeterized{} ReLU architecture creates reflections.

\begin{theorem}[wider ReLU networks do not stagnate and generate reflections]  \label{ReLU2neuroninformal}
A three-layer \symmeterized{} network with ReLU activation is equivalent to a \lasso{} model with a finite set of features, including those with \kink{}s at reflections.
\end{theorem}

\begin{figure}[t!]
    \centering
     \begin{minipage}[t!]{\textwidth}
    \includegraphics[width=\linewidth]{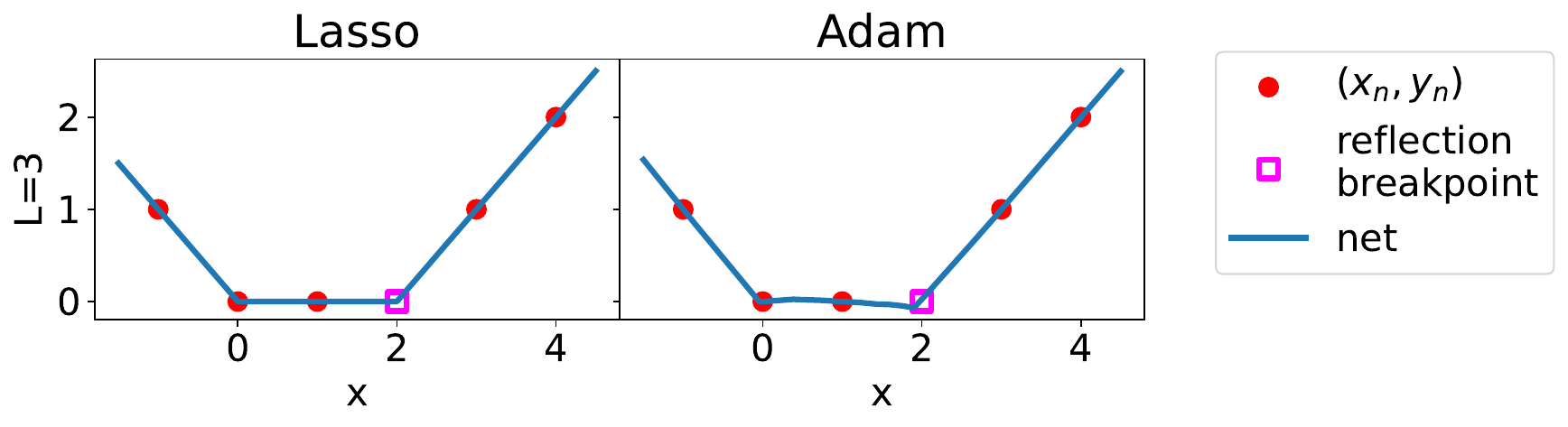}
    \end{minipage}
    \caption{\lasso{} and \adam{}-trained $3$-layer  \symmeterized{} ReLU networks. 
    Most crucially, the \kink{} at $2$ is not a training point; it is the reflection of $x_4=0$ across $x_3=1$.}
    \label{fig:ReLU3lyrnet}
\end{figure}

\Cref{fig:ReLUfeatdiagram} plots ReLU features. A $2$-layer ReLU network learns ramp features. Extending the depth to one more layer without changing the width creates a \deepnarrow{} $3$-layer network, which additionally learns \rampname{} features. Extending both the depth and width enables creating a $3$-layer \symmeterized{} network, which learns an additional host of features, including generalized reflection features shown in the bottom row of the figure. The generalized reflections $x_j{+}x_k{-}x_i$ are reflections when $j{=}k$. Reflection features appear in \adam{}-trained ReLU networks as predicted by the \lasso{} formulation, as shown in \Cref{fig:ReLU3lyrnet}. Details of this simulation are given in \Cref{sec:num_results}. Next, we formally define a feature function and its properties.
%
%

%
\iftoggle{long}{
\begin{definition}\label[definition]{def:rampfromzero}
Let $\ramp{}{a_1}{a_2}$ be $\ramp{+}{a_1}{a_2}$ if $a_1  \leq a_2$ and $\ramp{-}{a_2}{a_1}$ otherwise.
\end{definition} 
\Cref{def:rampfromzero} states that  $\ramp{}{a_1}{a_2}(x)$ is the \rampname{} function with breakpoints at $a_1, a_2$ such that $\ramp{}{a_1}{a_2}(a_1)=0$.
When $L=4$, \deepnarrow{} features can have \kink{}s at reflections. However, not all reflections of data points are \kink{}s, as described in the next result.
For a \ReLUname{}, its slope must be in the direction of the reflection: if  $x_{j_2} > x_{j_1}$ then $\reflect{x_{j_1}}{x_{j_2}} > x_{j_2} $ and a \ReLUname{} must increase for $x > \reflect{x_{j_1}}{x_{j_2}}$. If $ x_{j_2} < x_{j_1}$ then $\reflect{x_{j_1}}{x_{j_2}} < x_{j_2}$ and a \ReLUname{} must decreases for $ x < \reflect{x_{j_1}}{x_{j_2}}$.
For a \rampname{}, at most one \kink{} can be a reflection. If a \rampname{} evaluates to $0$ at a data point \kink{} and the other \kink{} is a reflection $\reflect{x_{j_1}}{x_{j_2}}$, then $x_{j_1}$ has to reflect across $x_{j_2}$ towards the \kink{} that is a data point. For example, an increasing \rampname{} has a reflection \kink{} satisfying $\reflect{x_{j_1}}{x_{j_2}} < x_{j_2} < x_{j_1} $. 
If the capped ramp has a value of $0$ at a reflection \kink{} $\reflect{x_{j_1}}{x_{j_2}}$, then the other \kink{} must be $x_{j_2}$.
\iftoggle{arxiv}{
See 
 \Cref{fig:L4featureswithkinks}. }
\begin{lemma}\label[lemma]{lemma:L4rampshapesreflect}
    All $L=4$ \deepnarrow{} functions with \kink{}s at a reflection are of the form $\ReLU{+}{\reflect{x_{j_1}}{x_{j_2}}}$ if ${j_1} \geq {j_2}$, $\ReLU{-}{\reflect{x_{j_1}}{x_{j_2}}}$ if ${j_1} \leq {j_2}$,  $\ramp{}{x_{j_3}}{\reflect{x_{j_1}}{x_{j_2}}}$ where $j_1 < j_2 \leq j_3$ or $j_1 > j_2 \geq j_3$, and $\ramp{}{\reflect{x_{j_1}}{x_{j_2}}}{x_{j_2}}$ where $j_1 \neq j_2$.
\end{lemma}
}
%
\iftoggle{long}{
\iftoggle{arxiv}{
\begin{figure}[t]
    \centering
    \pgfmathsetmacro{\scale}{1}
    \begin{tikzpicture}
        \begin{scope}[scale=1]
            \draw[red] (-7,0) -- (0,0) -- (2,2);
            \draw[blue] (-2,2) -- (0,0) -- (7,0); 
            \fill (0,0) node[diamond, fill = {black}, {black}, label={[text=black]below:$\reflect{x_{j_1}}{x_{j_2}}$},scale=0.4] {};
            \node [red] at (4,2)  {$\ReLU{+}{\reflect{x_{j_1}}{x_{j_2}}}$};
            \node [blue] at (-4,2)  {$\ReLU{-}{\reflect{x_{j_1}}{x_{j_2}}}$};
            \fill (4,0) node[circle, fill = {blue}, {blue}, label={[text=blue]below:$x_{j_1}$},scale=0.4] {};
            \fill (2,0) node[circle, fill = {blue}, {blue}, label={[text=blue]below:$x_{j_2}$},scale=0.4] {};
            \fill (-4,0) node[circle, fill = {red}, {red}, label={[text=red]below:$x_{j_1}$},scale=0.4] {};
            \fill (-2,0) node[circle, fill = {red}, {red}, label={[text=red]below:$x_{j_2}$},scale=0.4] {};
        \end{scope}
        \begin{scope}[scale=\scale, shift={(0,-4)}]
        \draw[red] (-7,0) -- (0,0) -- (2,2) -- (7,2);
        \draw[blue] (-7,2) -- (-2,2) -- (0,0) -- (7,0);
        \node [red] at (4,2.5)  {$\ramp{+}{x_{j_3}}{\reflect{x_{j_1}}{x_{j_2}}}$};
        \node [blue] at (-4,2.5)  {$\ramp{-}{x_{j_3}}{\reflect{x_{j_1}}{x_{j_2}}}$};
        \filldraw[black] (0,0) circle (2pt) node[anchor=north]{$x_{j_3}$};
        \fill (2,0) node[star, fill = {red}, {red}, label={[text=red]above:$\reflect{x_{j_1}}{x_{j_2}}$},scale=0.4] {};
        \fill (-2,0) node[star, fill = {blue}, {blue}, label={[text=blue]above:$\reflect{x_{j_1}}{x_{j_2}}$},scale=0.4] {};
        \filldraw[blue] (-6.5,0) circle (2pt) node[anchor=north]{$x_{j_1}$};
        \filldraw[blue] (-4.25,0) circle (2pt) node[anchor=north]{$x_{j_2}$};
        \fill (3.5,0) node[diamond, fill = {blue}, {blue}, label={[text=blue]below:$x_{j_1}$},scale=0.4] {};
        \fill (0.75,0) node[diamond, fill = {blue}, {blue}, label={[text=blue]below:$x_{j_2}$},scale=0.4] {};
        \filldraw[red] (6.5,0) circle (2pt) node[anchor=north]{$x_{j_1}$};
        \filldraw[red] (4.25,0) circle (2pt) node[anchor=north]{$x_{j_2}$};
        \fill (-3.5,0) node[diamond, fill = {red}, {red}, label={[text=red]below:$x_{j_1}$},scale=0.4] {};
        \fill (-0.75,0) node[diamond, fill = {red}, {red}, label={[text=red]below:$x_{j_2}$},scale=0.4] {};
	\end{scope}
 
    \begin{scope}[scale=\scale, shift={(0,-8)}]
        \draw[red] (-7,0) -- (-2,0) -- (0,2) -- (7,2);
        \draw[blue] (-7,2) -- (0,2) -- (2,0) -- (7,0);
        \node [red] at (4,2.5)  {$\ramp{+}{\reflect{x_{j_1}}{x_{j_2}}}{x_{j_2}}$};
        \node [blue] at (-4,2.5)  {$\ramp{-}{x_{j_2}}{\reflect{x_{j_1}}{x_{j_2}}}$};
        \fill (0,0) node[circle, fill = {black}, {black}, label={[text=black]below:$x_{j_2}$},scale=0.4] {};
        \fill (2,0) node[star, fill = {black}, {black}, label={[text=blue]above right:$\reflect{x_{j_1}}{x_{j_2}}$},scale=0.4] {};
        \fill (-2,0) node[star, fill = {black}, {black}, label={[text=red]above left:$\reflect{x_{j_1}}{x_{j_2}}$},scale=0.4] {};
        \fill (-2,0) node[star, fill = {black}, {black}, label={[text=blue]below left:$x_{j_1}$},scale=0.4] {};
        \fill (2,0) node[star, fill = {black}, {black}, label={[text=red]below right:$x_{j_1}$},scale=0.4] {};
    \end{scope}

    \end{tikzpicture}
    \caption{$L=4$ \deepnarrow{} features that have a \kink{} at a reflection point. Top: \ReLUname{} features. Middle: $\ramp{}{x_{j_3}}{\reflect{x_{j_1}}{x_{j_2}}}$. Bottom: $\ramp{}{\reflect{x_{j_1}}{x_{j_2}}}{x_{j_2}}$.} 
    \label{fig:L4featureswithkinks}
\end{figure}
}
}
%
%
\begin{remark}\label{rk:featuredef}
    \Cref{thrm:lasso_deep_1nueron_ReLU} and \Cref{def:reconstructfunc} below show that the training problem is equivalent to a \lasso{} problem with solution $(\mathbf{z}^*,\xi^*)$ and a dictionary matrix whose $i^{\text{th}}$ column $\mathbf{A}_i$ maps to a parallel unit function $\Xul{i}{L}:\mathbb{R}\to\mathbb{R}$ such that $\Xul{i}{L}\left(\mathbf{X}\right)=\mathbf{A}_i$ and $\sum_i z^*_i \Xul{i}{L}(x) + \xi^*$ is the same function as an optimal neural net. In this section we define a \emph{feature} as described in \Cref{sec:intro} to be a parallel unit function $\Xul{i}{L}$ corresponding to $\mathbf{A}_i$ such that $z^*_i\neq 0$.
\end{remark}
In addition to ReLU and absolute value networks, leaky ReLU networks are also equivalent to \lasso{} models. Moreover, we can upper bound the size of the libraries. 

\begin{theorem}\label{lemma:numfeatures}
    A \deepnarrow{} network of any depth $L\geq 2$ 
    and piecewise linear activation
    is equivalent to a \lasso{} problem with a finite set of features. The number of features is $O(N^2)$ for ReLU activation and $O(N^{L-1}2^L L!)$ for leaky ReLU and absolute value activations. 
\end{theorem}
In particular, the number of features is finite and at most polynomial in the number of training points and exponential in the depth. However, the number of features is an overestimate and in fact saturates for certain activations and architectures, as seen in the next result.

\begin{lemma}\label{lemma:ReLUsaturates}
    Training a \deepnarrow{} ReLU network with an arbitrary number of layers ($L \geq 2$) is a \lasso{} problem where features are ReLU or \rampname{} functions with \kink{}s at data points. The number of features is $O(N^2)$.
\end{lemma}

Adding only one neuron per layer limits the expressibility of ReLU networks. In contrast to absolute value, ReLU networks rely more heavily on wider layers for expressibility.

\begin{definition}
    For $\alpha, \beta , \posslope, \negslope \in \mathbb{R}$ such that $\posslope\neq \negslope$,
    let the \emph{\midptname} be 
    \begin{equation}\label{def:malphabeta}
 \midpt{\alpha}{\beta} = -\frac{\posslope\max\{\alpha, \beta\}  - \negslope\min\{\alpha,\beta\} }{\posslope- \negslope}.
\end{equation}
\end{definition}
\newcommand{\datafeature}{data feature}
Note $\midpt{\alpha}{\beta}$ is the midpoint $\frac{\alpha{+}\beta}{2}$ between $\alpha$ and $\beta$ if $\sigma(x){=}|x|$. If $\sigma(x){=}(x)_+$, or if $\sigma(x){=}|x|$ and $\alpha{=}\beta$, then $\midpt{\alpha}{\beta}{=}\alpha$. We use the \midptname{} to define bias parameters for features. Recall that $\Xul{1}{L}$ denotes a parallel unit, which is a standard network (\Cref{sec:parnet}). Let $\xl{l}{=}\Xul{1}{l},\wl{l}{=}\Wul{1}{l}$, and $\bl{l}{=}\bul{1}{l}$. The output $\xl{l}$ of each layer can be interpreted as a feature extracted by that layer. This motivates the following definition.
\begin{definition}\label{def:datafeaturebias}
    A bias parameter $\bl{l}$ is a \emph{data feature bias} if $\bl{l}{=}{-}\xl{l}(\mathbf{x}_0 )\wl{l}$ for some column vector $\mathbf{x}_0 {\in} \{x_1,{\cdots},x_N\}^{m_l}$. The bias parameter $\bl{1}$ 
 is a first-layer \emph{midpoint feature bias} if $\bl{1} {=} -\midpt{{x_{j_1}}}{{x_{j_1'}}}\wl{1}$ for some $j_1, j_1' \in [N]$ and 
    it is in a $3$-layer \deepnarrow{} network with a non-monotone, continuous piecewise linear activation (such as $\sigma(x)=|x|$.)
\end{definition}

In the simplest case when $L{=}2$, a \datafeature{} bias is of the form $b=-w x_n $. 
These bias parameters are used to define a deep library.

\begin{definition}[Deep Library]\label{def:FeatDicStdNN}
  Consider a $3$-layer \symmeterized{} or $L$-layer \deepnarrow{} network. 
 The \emph{deep library} is the set of all network outputs $\xl{L}(\mathbf{X})$ defined in \eqref{eq:parallelnn_rec} with data or midpoint feature biases and all elements of $\wl{1}$ being $\pm 1$.
\end{definition}
The deep library contains a finite number of standard networks evaluated on the training data.
The next result shows that a neural network learns features in the deep library.

\begin{theorem}[complete \lasso{} libraries for general activations]\label{thrm:lasso_deep_1nueron_ReLU}
Consider a \deepnarrow{} $L$-layer network where $L {\in} \{2,3\}$ and with activation $\sigma$ which is ReLU, leaky ReLU, absolute value, sign, or threshold if $L{=}2$, and ReLU, leaky ReLU or absolute value if $L{=}3$. 
Consider a \lasso{} problem whose dictionary is the deep library
and where $\xi{=}0$ if $\sigma$ is sign or threshold. Suppose $(\mathbf{z}^*, \xi^*)$ is a solution, and let $m^*{=}\|\mathbf{z}^*\|_0$.
This \lasso{} problem is equivalent to the training problem for the network, provided $m_L \geq m^*$.

\iftoggle{long}{

The training problem for the $L=2, 3, 4$-layer \deepReLU{}, where the activation is any of those mentioned above, and ReLU if $L > 2$, is equivalent to the \lasso{} problem \eqref{eq:genericlasso}.
\iftoggle{long}{
\eqref{eq:genericlasso}
     \begin{equation}\label{LLlasso}
     \min_{\mathbf{z}, \xi}  \frac{1}{2}\left(\Amat{} \mathbf{ z} + \xi \mathbf{1} \right)+ \beta \|\mathbf{z}\|_1, 
 \end{equation}
 }
The columns of $\Amat{}$ are indexed by the tuple $\ReLUdicindices{}$. For a $L$-layer network, the $(\ReLUdicindices{})^{th}$ column of $\Amat$ is $\ReLUdicfun{L}(\mathbf{X}) \in \mathbb{R}^N$.
This holds provided that the number of parallel units is $m_{L} \geq m^* := \| \mathbf{z}^*\|_0$.
 Depending on the activation $\sigma$, the \lasso{} problem is simplified as follows:

	\begin{enumerate}
		\item \label{symm_simp} sign and absolute value: 
		$\mathbf{z}_{(\ReLUdicindices{})} = \mathbf{0}$ for $\mathbf{s} = -1$. 
		\item \label{bounded_simp} sign and threshold: 
  $ \xi = 0$.
	\end{enumerate}

%
%
%
\iftoggle{long}{
  We can also write,

  \begin{equation}
  \begin{aligned}
    &\Amat{}^r_{s,i,j_1,j_2,j_3} = \sigma \left( s^{(3)} \left(\sigma \left( s^{(2)} \left(\sigma \left( s^{(1)} \left( x_i -  \left(2x_{j_2} - x_{j_1}\right) \right) \right) - \sigma \left( s^{(1)} \left( x_{j_1} - x_{j_2}   \right) \right) \right) \right) - \\
     & \hspace*{1.0cm}\sigma \left( s^{(2)} \left(\sigma \left( s^{(1)} \left( x_{j_3} - \left(2x_{j_2} - x_{j_1}\right) \right) \right) - \sigma \left( s^{(1)} \left(  x_{j_1} - x_{j_2}  \right) \right) \right) \right) \right) \right) \Big) \Big ). 
  \end{aligned}
  \end{equation}

  }

}
%
 \end{theorem}
The notion of equivalence between optimization problems is defined in the beginning of \Cref{section:main_results}. \Cref{def:reconstructfunc} defines a map to reconstruct an optimal neural network from a \lasso{} solution (\Cref{lemma:parreconstructionworks}). The map is especially straightforward for $2$-layer networks (\Cref{rk:2lyrRecon}). These results are given in \Cref{app:reconstrct} due to space.  
The next theorem generalizes \Cref{ReLU2neuroninformal} to leaky ReLU activations. 

\begin{theorem}\label{theorem:symmeterizednetworks}
    The training problem for a $3$-layer \symmeterized{} network with monotone activations such as ReLU is equivalent to a \lasso{} problem with solution $(\mathbf{z}^*,\xi^*)$ and whose dictionary contains the deep library, provided $m_L \geq m^*$, where $m^*=\|\mathbf{z}^*\|_0$.
\end{theorem}

\Cref{theorem:symmeterizednetworks} states that the deep library is a sub-dictionary for \symmeterized{} networks. Finding the full dictionary for a \symmeterized{} network is an area of future work.
\Cref{thrm:lasso_deep_1nueron_ReLU} shows that instead of training a neural network with a \nc{} problem and reaching a possibly local optimum, we can simply solve a straightforward \lasso{} problem whose convexity guarantees that gradient descent approaches global optimality. \Cref{fig:1d_func_approx} shows an example where a network trained with \lasso{} achieves a better function fit than training with the \nc{} problem. 
 In previous work \citep{ergen2023globally}, a similar \lasso{} formulation is developed for networks with threshold activation but requires up to $2^N$ features of length $N$ in the dictionary for a $2$-layer network.
 In contrast, with \nd{1} data, \Cref{thrm:lasso_deep_1nueron_ReLU} shows that at most $2N^2$ features are needed for a $2$-layer network.


 \begin{remark}
    Note that when the network is $2$ layers, the equivalent \lasso{}  dictionary only contains features with breakpoints at training data, leading to a prediction with \kink{}s only at data locations. In contrast, when the network has $3$ layers there can be \kink{}s at \textbf{reflections} of data points with respect to other data points due to the reflection features. As a result, for activations such as absolute value and ReLU with \symmeterized{} networks, the sequence of dictionaries as the network gets deeper converges to a richer library that includes reflections.
\end{remark}

 \begin{figure}[t]
    \centering 
    \scalebox{1}{
    \begin{tikzpicture}
    \pgfmathsetmacro{\hint}{11} 
    \pgfmathsetmacro{\vint}{12} 
    \pgfmathsetmacro{\scale}{0.22} 
     \pgfmathsetmacro{\scaletwo}{0.5} 
    \pgfmathsetmacro{\xi}{4}
    \pgfmathsetmacro{\xj}{11}
    \pgfmathsetmacro{\xk}{15}
    \pgfmathsetmacro{\len}{26}
    \pgfmathsetmacro{\reflectkink}{\xj+\xk-\xi}
   \draw[decoration={brace,mirror,raise=5pt},decorate]
  (-0.3,\vint*\scale*0.5+0.2) -- node[yshift=12pt,left=12pt, rotate=90] {\footnotesize$L{=}2$} (-0.3,-0.5);
  \draw[decoration={brace,mirror,raise=5pt},decorate]
  (0,\vint*\scale*0.5) -- node[yshift=12pt,left=12pt, rotate=90] {\footnotesize$L{=}3$} (0,-2.2*\vint*\scale);
  \draw[decoration={brace,raise=5pt},decorate]
  (0,\vint*\scale*0.5+0.3) -- node[above=6pt] {\footnotesize $m_{L-1}{=}1 (\text{ \deepnarrow{}})$} (\hint*\scale,\vint*\scale*0.5+0.3);
  \draw[decoration={brace,raise=5pt},decorate]
  (0,\vint*\scale*0.5) -- node[above=6pt] {\footnotesize $m_{L-1}{=}2$ ($3$-layer \symmeterized{})} (5.5*\hint*\scale,\vint*\scale*0.5);
    \begin{scope}[scale=\scale*1.5, shift={(0*\hint,0)}]	
        \incReLU{\jonecol}{circle} 
        \node [] at (3,-1)  {$x_{j_1}$};  
    \end{scope}
    \begin{scope}[scale=\scale, shift={(0*\hint,-\vint)}]
        \draw[dashed] (0,0) -- (10,0);
        \draw (0,0) -- (5,0) -- (7,2) -- (10,2);
        \node[scale=0.5] (a2) at (5,0) [circle,fill=\jonecol, label=below:\footnotesize$x_{j_1}$]{}; 
        \node[scale=0.5] (a3) at (7,0) [circle,fill=\jtwocol, label=below:\footnotesize$x_{j_2}$]{};
    \end{scope}
    \begin{scope}[scale=\scale, shift={(1.1*\hint,-\vint)}]	
        \draw[dashed] (0,0) -- (10,0);
        \draw (0,0) -- (5,0) -- (7,4) -- (10,7);
        \node[scale=0.5] (a2) at (5,0) [circle,fill=\jonecol, label=below:\footnotesize$x_{j_1}$]{}; 
        \node[scale=0.5] (a3) at (7,0) [circle,fill=\jtwocol, label=below:\footnotesize$x_{j_2}$]{};
    \end{scope}
    \begin{scope}[scale=\scale,shift={(2.2*\hint,-1*\vint)}]
       \draw[dashed] (0,0) -- (9,0);
        \draw (0,0) -- (2,0) -- (5,3) -- (7,3) -- (9,5);
        \node[scale=0.5] (a1) at (2,0) [circle,fill=\jtwocol, label=below left:\footnotesize$x_{j_2}$]{};
        \node[scale=0.5] (a2) at (5,0) [circle,fill=\jonecol, label=below:\footnotesize${x_{j_1}}$]{}; 
        \node[scale=0.5] (a3) at (7,0) [circle,fill=\jthreecol, label=below:\footnotesize$x_{j_3}$]{};
    \end{scope}
    \begin{scope}[scale=\scale,shift={(3.3*\hint,-1*\vint)}]
        \draw[dashed] (0,0) -- (9,0);
        \draw (0,0) -- (2,0) -- (5,3) -- (7,7) -- (9,9);
        \node[scale=0.5] (a1) at (2,0) [circle,fill=\jonecol, label=below left:\footnotesize$x_{j_1}$]{};
        \node[scale=0.5] (a2) at (5,0) [circle,fill=\jtwocol, label=below:\footnotesize$x_{j_2}$]{}; 
        \node[scale=0.5] (a3) at (7,0) [circle,fill=\jthreecol, label=below:\footnotesize$x_{j_3}$]{};
    \end{scope}
    \begin{scope}[scale=\scale,shift={(4.4*\hint,-1*\vint)}]
        \draw[dashed] (0,0) -- (9,0);
        \draw (0,0) -- (2,0) -- (5,6) -- (7,8) -- (9,8);
        \node[scale=0.5] (a1) at (2,0) [circle,fill=\jonecol, label=below left:\footnotesize$x_{j_1}$]{};
        \node[scale=0.5] (a2) at (5,0) [circle,fill=\jtwocol, label=below:\footnotesize$x_{j_2}$]{}; 
        \node[scale=0.5] (a3) at (7,0) [circle,fill=\jthreecol, label=below:\footnotesize$x_{j_3}$]{};
    \end{scope}
    \begin{scope}[scale=\scale,shift={(1*\hint,-2*\vint)}]
        \draw[dashed] (0,0) -- (\len,0);
        \draw (0,\xi) -- (\xi,0) -- (\reflectkink,0) -- (\len,\len-\reflectkink);
        \node[scale=0.5] (xi) at (\xi,0) [circle,fill=\jonecol, label=below:\footnotesize $x_{j_1}$]{}; 
        \node[scale=0.5] (xj) at (\xj,0) [circle,fill=\jtwocol, label=below:\footnotesize $x_{j_2}$]{};
        \node[scale=0.5] (xk) at (\xk,0) [circle,fill=\jthreecol, label=below:\footnotesize $x_{j_3}$]{};
        \node[scale=0.5] (avg) at (0.5*\xk+0.5*\xj,0) [circle,fill=\jtwojthreecol, label=above:\footnotesize $\frac{x_{j_2}{+}x_{j_3}}{2}$]{};
        \node[scale=0.5] (reflect) at (\reflectkink,0) [circle,fill=\jtwojthreecol,draw=\jonecol, line width = 1.5pt, label=below:\footnotesize $x_{j_2}{+}x_{j_3}{-}x_{j_1}$]{};
    \end{scope}
    \begin{scope}[scale=\scale,shift={(3.5*\hint,-2*\vint)}]
        \draw[dashed] (0,0) -- (\len,0);
        \draw (0,0) -- (\xi,0) -- (\xj,\xj-\xi) -- (\xk, \xj-\xi)--(\reflectkink,0)--(\len,0);
        \node[scale=0.5] (xi) at (\xi,0) [circle,fill=\jonecol, label=below:\footnotesize $x_{j_1}$]{}; 
        \node[scale=0.5] (xj) at (\xj,0) [circle,fill=\jtwocol, label=below:\footnotesize $x_{j_2}$]{};
        \node[scale=0.5] (xk) at (\xk,0) [circle,fill=\jthreecol, label=below:\footnotesize $x_{j_3}$]{};
        \node[scale=0.5] (avg) at (0.5*\xk+0.5*\xj,0) [circle,fill=\jtwojthreecol, label=above:\footnotesize $\frac{x_{j_2}{+}x_{j_3}}{2}$]{};
        \node[scale=0.5] (reflect) at (\reflectkink,0) [circle,fill=\jtwojthreecol,draw=\jonecol, line width=1.5pt, label=below:\footnotesize $x_{j_2}{+}x_{j_3}{-}x_{j_1}$]{};
    \end{scope}
\end{tikzpicture}
}
\caption{Example ReLU features, excluding reverse directions. Bottom row: $3$-layer \symmeterized{} ReLU features with \kink{}s at generalized reflections of $x_{j_1}$ (\jonecol{}) across $x_{j_2}$ (\jtwocol{}) and $x_{j_3}$ (\jthreecol{}) $x_{j_2}{+}x_{j_3}{-}x_{j_1}$ (\jonecol{} encircling \jtwojthreecol{}). Lines have slopes $\pm 2, \pm 1$ or $0$.
}
\label{fig:ReLUfeatdiagram}
\end{figure}

Our approach lays the foundation to further analyze the evolution of feature libraries over expanding depth and widths as an area of future work. For $2$-layer networks, the dictionary (\Cref{thrm:lasso_deep_1nueron_ReLU}) is simple, as described next.

\begin{corollary}[$2$-layer libraries]\label{Amat2lyrs}
    Let $\Apos, \Aneg \in \mathbb{R}^{N \times N}$ with $(\Apos{})_{i,n} {=} \sigma(x_i {-} x_n), (\Aneg{})_{i,n} {=} \\ \sigma(x_n {-} x_i)$.
     We can write the dictionary matrix for $2$-layer networks as $\mathbf{A} {=} \Apos$ for absolute value and sign activations, and $\mathbf{A} {=} [\Apos{}, \Aneg{}] \in \mathbb{R}^{N \times 2N}$ for ReLU, leaky ReLU, and threshold activations. 
\end{corollary}

In \Cref{Amat2lyrs}, $\Apos$ and $\Aneg$ contain features $\xl{2}(\mathbf{X})$ where $\wl{1}=1$ and $\wl{1}=-1$, respectively (\Cref{def:FeatDicStdNN}).
\Cref{fig:Amatrix} illustrates $\Apos$ for the ReLU and sign activations. 
Using the notation of \Cref{def:FeatDicStdNN}, the $3$-layer \deepnarrow{} absolute value features are of the form $\xl{3}(x){=}\big||x{-}x_{j_1}|{-}|x_{j_2}{-}x_{j_1}|\big|$, and for $4$ layers, include features of the form $\xl{4}(x){=}\Big|\big||x{-}x_{j_1} | {-} |x_{j_2}{-}x_{j_1}| \big| {-}
\big||x_{j_3} {-}x_{j_1} | {-} |x_{j_2}{-}x_{j_1}| \big|  \Big|$. These features are plotted in \Cref{fig:absvalfeattikz}, which highlights their reflection and double reflection \kink{}s. \Cref{fig:ReLUfeatdiagram} shows a subset of the ReLU library, with generalized reflection features for $3$-layer \symmeterized{} networks. 
%
%
\Cref{fig:ReLUfeatdiagram} illustrates general feature shapes not including mirrored directions. In \Cref{fig:ReLU2neurons}, we choose distinct training samples $x_i,x_j,x_k {\in} \{-1,0,2\}$ and numerically compute all possible deep library features for $3$-layer, \symmeterized{} ReLU networks using \Cref{def:FeatDicStdNN}. 
Since the features $\xl{L}(x)$ are continuous with respect to $x_i, x_j, x_k$, the features for non-distinct $x_i,x_j,x_k$ can be extrapolated by merging adjacent training points. The numerically plotted features are consistent with \Cref{fig:ReLUfeatdiagram}. In addition to graphical representations, \Cref{def:ReLU dic func} in \Cref{app:reconstrct} gives examples of simple, explicit expressions for features defined in \Cref{def:FeatDicStdNN}. 
 
 So far, we have emphasized deep ReLU and absolute value networks as having reflection features. Next, we show that in contrast, deep networks with sign activation do not have reflection features, even if they have many neurons per layer, suggesting the importance of choice of activation. The results for sign activation can be similarly extended to threshold activation \cite{ergen2023globally}. 
%
%
  %
\iftoggle{long}{Theorem \ref{theo:1d_all} shows that in order to globally minimize the training objective in \eqref{eq:nonconvex_generic_loss_2lyr}, there is no need to perform an exhaustive search over all possible subsets of samples to form hyperplane arrangements for the data matrix X as in \eqref{eq:background_ReLU_bidual} and \eqref{eq:background_threshold_bidual}. Instead, one can construct the $N \times N$ dictionary matrices $A_+, A_-$ and solve a standard Lasso using $A=[A_+, A_-]$ as the dictionary matrix. }
\iftoggle{long}{
\Cref{lemma:sign_thresh_a} can be seen in the left plot of \Cref{fig:Amatrix}, where each column of $\Amat{}$ consists of a purple and yellow region representing $-1$ and $1$, respectively.
Similarly the columns of $\mathbf{A_+}, \mathbf{A_-} $ for sign activation are vectors in $\{0,1\}^N$ that switch values at most once.
}
%
%
\iftoggle{long}{This is seen in bottom plot of figure \ref{fig:Amatrix}, which is a matrix that is just $0$ (purple) in the top right and bottom left corners, and $1$ everywhere else, unaffected by the random data matrix $X$. So for sign and threshold activation on $\nd{1}$ data,  the training problem depends only on the labels $y$, not the data $X$.}
 %


%

\iftoggle{long}{In the left plot, the elements of the ReLU dictionary are all nonnegative, while elements of the sign dictionary are $-1$ or $1$. For sign activation, by simplification \ref{symm_simp} of \Cref{thrm:lasso_deep_1nueron_ReLU}, just the right (or left) half of the $A$ matrix can be used. This would make $A$ be a square matrix that is $1$ in the upper right corner and $-1$ in the lower left corner, as can be seen in the right plot of \Cref{fig:Amatrix}.  }



\iftoggle{long}{
\begin{theorem}\label{theo:1d_all}
	Consider the two-layer neural net in \eqref{eq:2layernn} and the non-convex training problem given in \eqref{eq:nonconvex_generic_loss_2lyr}. If $m \geq m^*$, the optimal value of \eqref{eq:nonconvex_generic_loss_2lyr} is equal to the Lasso model
	\begin{equation}\label{eqref:Lasso}
 \iftoggle{long}{\min_{\mathbf{z_+}, \mathbf{z_-} \in \mathbb{R}^N, \xi \in \mathbb{R}}  \mathcal{L}_{\mathbf{y}} (\mathbf{A_+ z_+} + \mathbf{A_- z_-} + \xi \mathbf{1})
		+ \beta (\|\mathbf{z_+}\|_1+\|\mathbf{z_-}\|_1),}
\min_{\mathbf{z} \in \mathbb{R}^{2N}, \xi \in \mathbb{R}} \frac{1}{2 }\| \mathbf{Az}  + \xi \mathbf{1} - \mathbf{y} \|^2_2
		+ \beta \|\mathbf{z}\|_1,
	\end{equation}

	and a globally optimal solution to \eqref{eq:nonconvex_generic_loss_2lyr}  can be reconstructed from the Lasso \eqref{eq:genericlasso} (\Cref{thrm:2layer_reconstruct}, \Cref{section:reconstruction}). Here, $m^* = ||\mathbf{z}^*||$, where $\mathbf{z^*} 
$ is  optimal in \eqref{eq:genericlasso}. 
	\iftoggle{long}{ie, the cardinality of the support of the optimal Lasso solution. }
 The dictionary matrix is defined as $\Amat{} = [\mathbf{A_+} \mathbf{A_-}] \in \mathbb{R}^{N \times 2N}$ where $\mathbf{A_+}, \mathbf{A_-} \in \mathbb{R}^{N \times N}$ are given by $(\mathbf{A_+})_{i,j} = \sigma_{s=1}(x_i - x_j), (\mathbf{A_-})_{i,j} = \sigma_{s=1}(x_j - x_i)$.  Define the subvectors $\mathbf{z_+} = \mathbf{z}_{1:N}$ and $ \mathbf{z_-} = \mathbf{z}_{N+1:2N}$.

 Depending on the activation $\sigma$, problem \eqref{eq:genericlasso} is simplified as follows:

	\begin{enumerate}
		\item \label{symm_simp} sign and absolute value $(\sigma(x)=s \cdot \mathrm{sign}(x), |x|):
			      \mathbf{z_-} = \mathbf{0}$. 
		\item \label{bounded_simp} sign and threshold $(\sigma(x)=s \cdot \mathrm{sign}(x), s\mathbf{1}(x)): \xi = 0$.
	\end{enumerate}
\end{theorem}
}

\iftoggle{long}{
\begin{corollary}\label{cor:dic_elts_Llyr_ReLU}
    The shapes of the non-zero feature vectors in the \deepReLU{}'s dictionary matrix in \Cref{thrm:lasso_deep_1nueron_ReLU} are a ReLU or its reflection, or a flattened ramp 
    or its reflection. For $L=3$, the {\kink}s of the feature vector shapes are at data points $x_n$. For $L=4$, the {\kink}s points can also be at reflections of data points about each other.
\end{corollary}
}

\iftoggle{long}{
\begin{figure}
    \centering
    \begin{tikzpicture}
    \pgfmathsetmacro{\scale}{0.5}
    \pgfmathsetmacro{\hshift}{10}
    \pgfmathsetmacro{\vshift}{-8}
    \begin{scope}[scale=\scale, shift={(0,0)}]
        \posReLU{2}{};
    \end{scope}
    \begin{scope}[scale=\scale, shift={(0,\vshift)}]
        \negReLU{4}{};
    \end{scope}
    
    \begin{scope}[scale=\scale, shift={(\hshift,0)}]
        \posramp{2}{4}{};
    \end{scope}
        \begin{scope}[scale=\scale, shift={(\hshift,\vshift)}]
        \negramp{2}{4}{};
    \end{scope}
    \end{tikzpicture}

    \caption{Dictionary features}
    \label{fig:Llyr_features}
\end{figure}
}

\iftoggle{long}{

\subsection{Two layer neural networks}
\iftoggle{long}{Here, we consider two layer neural networks \eqref{eq:2layernn}.
}

The regularized training problem for $2$-layer nets is given by plugging the neural net \eqref{eq:2layernn} into \eqref{eq:nonconvex_generic_loss_gen_L_intro}:

\iftoggle{todos}{\textcolor{red}{todo: make generic loss theorem in appendix, which we will be proving}}

\begin{align} \label{eq:nonconvex_generic_loss_2lyr}
\iftoggle{long}{\min_{\boldsymbol{\alpha}, \mathbf{W_0}, \mathbf{b}, \xi} \mathcal{L}_{\mathbf{y}} \left(\sigma_s( \mathbf{X} \mathbf{W_0}  + b)\alpha + \xi  \right) + \frac{\beta}{2} \|\regparams\|_2^2} 
 \min_{\boldsymbol{\alpha}, \mathbf{W_0}, \mathbf{b}, \xi} \frac{1}{2} \| \sigma_s( \mathbf{X} \mathbf{W_0}  + b)\alpha + \xi \mathbf{1} - \mathbf{y} \|^2_2   + \frac{\beta}{2} \|\regparams\|_2^2,
\end{align} 

where $s$ is omitted if $\sigma_s$ is not sign or threshold and $\regparams$ denotes all trainable parameters that are regularized.
\iftoggle{long}{In equation \eqref{eq:nonconvex_generic_loss_2lyr}, $\beta>0$ is the regularization coefficient.  (The activation parameter $s \in \mathbb{R}$ is omitted from the activation $\sigma_s$ and the parameter sets $\Theta$ and $\Theta_s$ if the activation is not sign or threshold.)}
}


\iftoggle{long}{
\begin{figure}[t]
\includegraphics[width=\textwidth]{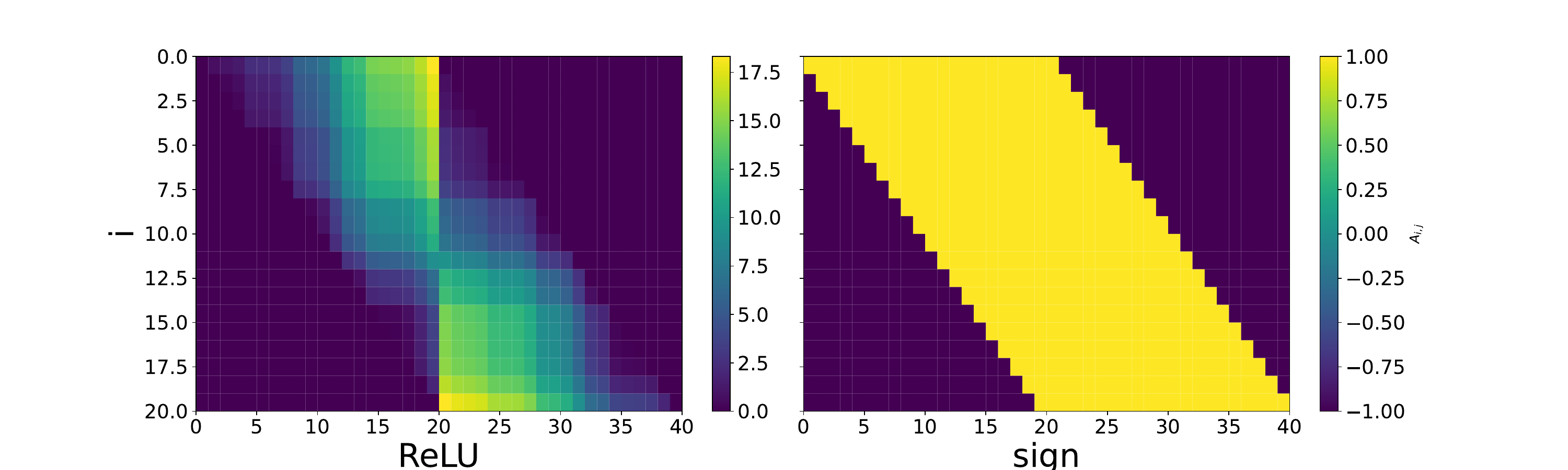}
\caption{Dictionary matrices $\Amat{}$ in \Cref{thrm:lasso_deep_1nueron_ReLU} for ReLU (left) and sign (right) activation. The data $\mathbf{X}$ is $20$ samples uniformly drawn at random from $[-10,10]$. The colors indicate the value of $\Amat{}_{i,j}$. }
\label{fig:Amatrix}
\end{figure}

}


\subsection{\secnameSignAct{}}\label{sec:deepsign}
In this section, we analyze the training problem of an $L$-layer deep network with sign activation, which need not be a \deepReLU{}. The formal statements of a few results are deferred to \Cref{app:treenetsec} due to space. Proofs in this section are deferred to \Cref{appsec:1Ddatasignact}. 

We say the vector $\mathbf{h} {\in} \{-1,1\}^N$  \emph{switches} 
    at 
    $n > 1$
\iftoggle{long}{$i \in \{2,\cdots,N\}$}
if $h_n \neq h_{n-1}$.
%
\iftoggle{long}{The number of switches of $h$ is the number of indices at which $\mathbf{h}$ switches, ie $|\{i \in \{2,{\cdots},N\}: \mbox{sign}(a_i) \neq \mbox{sign}(a_{i-1}) \}|$.}  For $n \in \mathbb{N}$, let the \emph{\switchsetname{}}  $\switchmatrix{n}$ be the set of all vectors in $\{-1,1\}^N$ that start with $1$ and switch at most $n$ times. For a set of vectors $\mathcal{Z}$, let $[\mathcal{Z}]$ be a matrix whose set of columns is $\mathcal{Z}$.  The next result relates the switching set to the \lasso{} dictionary matrix in \Cref{thrm:lasso_deep_1nueron_ReLU}.

\begin{lemma}\label[lemma]{lemma:sign_thresh_a}
	The dictionary matrix for a $2$-layer network with sign activation is $\left[\switchmatrix{1}\right]$.
\end{lemma}
We will show in \Cref{thm:3lyr_sign} that the training problem \eqref{eq:nonconvex_generic_loss_gen_L_intro} for deeper networks with sign activation is also equivalent to a \lasso{} problem \eqref{eq:genericlasso} whose dictionary is a \switchsetname{}. 

\begin{figure}[t!]
    \centering

    
    \begin{tikzpicture}[ 
    scale=0.7,
    declare function={         ReLU(\x) = max(0,\x); 
    incfun(\x)=\x;
    }]
    \pgfplotsset{set layers}
    \begin{axis}[
        domain=1:5,
        xlabel=$i$, ylabel=$j$,
        zmin = 0,
        axis lines=none,
        set layers,
    ]
    \addplot3 [
        samples y=6,
        samples=25,
        mesh,
        patch type=line,
        thick,
    ] {ReLU(incfun(x)-incfun(y))};
    \begin{pgfonlayer}{axis foreground}
        \node[] () at (5.3,5) {$n$};
        \node[] () at (0.7,1.7) {$i$};
    \end{pgfonlayer}
    \end{axis}
    \end{tikzpicture}\hspace{2cm}
    %
    %
    \qquad\qquad
    \begin{tikzpicture}[ 
        scale=0.5,
        declare function={         
        incfun(\x)=\x; 
        signsigma(\x) =  { \x == 0 ? 1 : 1*sign(\x) }; 
        }]
        \pgfplotsset{set layers}
        \begin{axis}[
            domain= 1:5, 
            xlabel=$n$, ylabel=$i$,
            axis lines=none,
            set layers,
        ]
        \addplot3 [
            samples y=5,
            samples=25,
            mesh,
            patch type=line,
            thick,
        ] {signsigma(incfun(x)-incfun(y))}; 
        \begin{pgfonlayer}{axis foreground}
            \node[] () at (5.3,5) {$n$};
            \node[] () at (0.9,1.7) {$i$}; 
        \end{pgfonlayer}
        \end{axis}
    \end{tikzpicture}

    \caption{Generic shape of $\Apos \in \mathbb{R}^{N \times N}$ defined by ${\Apos}_{i,n}=\sigma(x_i-x_n)$, where $\sigma$ is ReLU (left) and sign activation (right). Each $i^{th}$ curve represents a feature. The points $\left(i,n,{\Apos}_{i,n}\right)$ are plotted in \nd{3}, with ${\Apos}_{i,n}$ represented by the curve height and color. Here, $n \in [N]$ but each curve interpolates between integer values of $n$.  
}
    \label[figure]{fig:Amatrix}
\end{figure}

%
We now consider another architecture.
While each unit of the parallel neural network is a standard network, every branch of a \emph{tree network} is a parallel network. A detailed definition is given in \Cref{def:treenet} due to space.
Parallel and tree nets have the same architecture for $L=3$ layers. 
For $L \geq 3$, a \emph{\deepsign ~network} is a parallel network (\Cref{sec:parnet}) with $m_1 = \cdots = m_{L-2}$. A \deepnarrow{} network is a special case of a \deepsign{} network. Now we state the main result of this section: networks with sign activation are equivalent to \lasso{} models with libraries that increase until depth $3$ and then freeze for \deepsign{} networks but continue multiplying geometrically for tree networks.

%
\iftoggle{long}{
For $L=3$ layers, plugging the neural net \eqref{eq:3lyr_nn_par} into problem \eqref{eq:nonconvex_generic_loss_gen_L_intro} gives the \nc{} training problem
\begin{align} \label{eq:nonconvex_generic_lossL_3yr}
 \iftoggle{long}{\min_{\Theta} \mathcal{L}_{\mathbf{y}} \left( \sum_{i=1}^{m_2} \sigma_{s^{(i,2)}} \left ( \sigma_{s^{(i,1)}} \left([X,1] \Wul{i}{1}  \right) \mathbf{W^{(i,2)}}\right) \alpha_i + \xi \right) + \frac{\beta}{2}  \|\regparams\|_2^2,}
 \min_{\params \in \Theta} \frac{1}{2} \left \|   \sum_{i=1}^{m_2} \sigma_{s^{(i,2)}} \left ( \left [ \sigma_{s^{(i,1)}} \left([\mathbf{X},1] \Wul{i}{1}  \right), \mathbf{1}\right ] \mathbf{W^{(i,2)}}\right) \alpha_i + \xi - \mathbf{y} \right \|^2_2 + \frac{\beta}{2}  \|\regparams\|_2^2.
\end{align}
}
\iftoggle{long}{casts the training problem  \eqref{eq:nonconvex_generic_lossL_3yr} as Lasso, which }

\iftoggle{long}{
by showing that the $3$ and $4$-layer training problem \eqref{eq:nonconvex_generic_lossL_3yr} can also be solved with Lasso, just with a larger dictionary.
}

\begin{theorem}\label{thm:3lyr_sign}
Consider a \lasso{} problem whose dictionary is the \switchsetname{} $\switchmatrix{K}$, $\xi=0$, and with solution $\mathbf{z}^*$. Let $m^*= \| \mathbf{z}^* \|_0$. This \lasso{} problem 
is equivalent to the training problem for a neural network with sign activation, $m_L \geq m^* $,  and $m_{L-2} = K$ when it is a \deepsign{} network, and $ \prod_{l=1}^{L-1} m_l=K$ when it is a tree network.
 %
	%

\end{theorem}

Similar to \Cref{rk:featuredef}, we can formally define features for \deepsign{} and tree networks as parallel units and subtrees, respectively.

\begin{corollary}\label{Cor:signfeaturesdef} 
The features defined and described in \Cref{rk:featuredef} also apply to arbitrarily deep parallel networks with sign activation, and analogously for tree networks.
\end{corollary}

\begin{figure}[t]
\centering
\begin{tikzpicture}[scale=0.7]
\begin{axis}[axis x line=middle, axis y line=middle, xticklabels={,,}, axis line style={-}]
\addplot[ycomb, color=black,solid,mark=*] 
plot coordinates{
(1,1) 
(2,1) 
(3,1) 
(4,1)
(5,1)
(6,-1) 
(7,-1) 
(8,-1) 
(9,-1)
(10,-1)

					(11,1)
					(12,1)
					(13,1)
					(14,1)
					(15,1)
					(16,-1)
					(17,-1)
					(18,-1)
					(19,-1)
					(20,-1)

					(21,1)
					(22,1)
					(23,1)
					(24,1)
					(25,1)

					(26,-1)
					(27,-1)
					(28,-1)
					(29,-1)
					(30,-1)

					(31,1)
					(32,1)
					(33,1)
					(34,1)
					(35,1)
					(36,-1)
					(37,-1)
					(38,-1)
					(39,-1)
					(40,-1)
				};

			\addplot[ycomb, color=red,solid,mark=*]
			plot coordinates{
					(1,0.8)
					(2,0.8)
					(3,0.8)
					(4,0.8)
					(5,0.8)
					(6,-0.6)
					(7,-0.6)
					(8,-0.6)
					(9,-0.6)
					(10,-0.6)

					(11,0.6)
					(12,0.6)
					(13,0.6)
					(14,0.6)
					(15,0.6)
					(16,-0.6)
					(17,-0.6)
					(18,-0.6)
					(19,-0.6)
					(20,-0.6)

					(21,0.6)
					(22,0.6)
					(23,0.6)
					(24,0.6)
					(25,0.6)

					(26,-0.6)
					(27,-0.6)
					(28,-0.6)
					(29,-0.6)
					(30,-0.6)

					(31,0.6)
					(32,0.6)
					(33,0.6)
					(34,0.6)
					(35,0.6)
					(36,-0.8)
					(37,-0.8)
					(38,-0.8)
					(39,-0.8)
					(40,-0.8)
				};

			\addplot[ycomb, color=magenta,solid,mark=*]
			plot coordinates{
					(1,0.5)
					(2,0.5)
					(3,0.5)
					(4,0.5)
					(5,0.5)

					(36,-0.5)
					(37,-0.5)
					(38,-0.5)
					(39,-0.5)
					(40,-0.5)
				};

			\addplot[ycomb, color=blue,solid,mark=*]
			plot coordinates{
					(1,0.1)
					(2,0.1)
					(3,0.1)
					(4,0.1)
					(5,0.1)

					(36,-0.1)
					(37,-0.1)
					(38,-0.1)
					(39,-0.1)
					(40,-0.1)
				};
		\end{axis}

        \node[] at (3,6.2) {\scriptsize $L=2$};
		\node[] at (7.3,2.9) {\scriptsize $n$};
  
		\draw [decorate,
			decoration = {brace, mirror}] (0.9,5.2) --  (0.9,5.7);
		\node[] at (1.2,5.4) {\scriptsize $\textcolor{red}{\betafrac{}}$};

		\draw [decorate,
			decoration = {brace, mirror}] (2.7,4.6) --  (2.7,5.7);
		\node[] at (3.1,5.2) {\scriptsize $2\textcolor{red}{\betafrac{}}$};

		\draw [decorate,
			decoration = {brace}] (-0.19,2.9) --  (-0.19,3.2);
		\node[] at (-0.8,3.1) {\scriptsize $\frac{1-\textcolor{blue}{\betafrac{}}}{2}$};

		\draw [decorate,
			decoration = {brace, mirror}] (7,1.4) --  (7,2.9);
        \node[] () at (8.4,2) 
                {\scriptsize   
                        $
                            \begin{aligned}
                                \frac{1-\textcolor{magenta}{\betafrac}}{2} = \textcolor{magenta}{\betafrac}
                            \end{aligned}
                       $
                };

	\end{tikzpicture}
 	\begin{tikzpicture}[scale=0.7]
		\centering
		\begin{axis}[axis x line=middle, axis y line=middle, xticklabels={,,}, axis line style={-}]
			\addplot[ycomb,color=black,solid,mark=*]
			plot coordinates{
					(1,1)
					(2,1)
					(3,1)
					(4,1)
					(5,1)
					(6,-1)
					(7,-1)
					(8,-1)
					(9,-1)
					(10,-1)

					(11,1)
					(12,1)
					(13,1)
					(14,1)
					(15,1)
					(16,-1)
					(17,-1)
					(18,-1)
					(19,-1)
					(20,-1)

					(21,1)
					(22,1)
					(23,1)
					(24,1)
					(25,1)
					(26,-1)
					(27,-1)
					(28,-1)
					(29,-1)
					(30,-1)

					(31,1)
					(32,1)
					(33,1)
					(34,1)
					(35,1)
					(36,-1)
					(37,-1)
					(38,-1)
					(39,-1)
					(40,-1)
				};

			\addplot[ycomb,color=magenta,solid,mark=*]
			plot coordinates{
					(1,1/2)
					(2,1/2)
					(3,1/2)
					(4,1/2)
					(5,1/2)
					(6,-1/2)
					(7,-1/2)
					(8,-1/2)
					(9,-1/2)
					(10,-1/2)

					(11,1/2)
					(12,1/2)
					(13,1/2)
					(14,1/2)
					(15,1/2)
					(16,-1/2)
					(17,-1/2)
					(18,-1/2)
					(19,-1/2)
					(20,-1/2)

					(21,1/2)
					(22,1/2)
					(23,1/2)
					(24,1/2)
					(25,1/2)
					(26,-1/2)
					(27,-1/2)
					(28,-1/2)
					(29,-1/2)
					(30,-1/2)

					(31,1/2)
					(32,1/2)
					(33,1/2)
					(34,1/2)
					(35,1/2)
					(36,-1/2)
					(37,-1/2)
					(38,-1/2)
					(39,-1/2)
					(40,-1/2)
				};
		\end{axis}

        \node[] at (3,6.2) {\scriptsize $L=3$};
        \node[] at (7.1,2.9) {\scriptsize $n$};
		\draw [decorate,
			decoration = {brace, mirror}] (0.9,4.3) --  (0.9,5.7);
		\node[] at (1.2,5) {\scriptsize $\textcolor{magenta}{\betafrac}$};

	\end{tikzpicture}
	\caption{Each of the two figures depicts  $(n,y_n)$ with black dots, where  $\mathbf{y}=\hkt$ with $T=10, N=40$. Sign-activated neural net predictions are depicted as $\left(n, \nn{L}{x_n}\right)$ with blue, magenta, and red dots for $\betafrac = \frac{4}{5} \in \left[\frac{1}{2} , 1 \right], \betafrac = \frac{1}{2} $, and $ \betafrac =\frac{1}{5} \leq \frac{1}{2}$, respectively.
}
\label[figure]{fig:pulse_ex_overlap_2lyr}
\end{figure}

Features for sign activation are step functions that take on the value $\mathbf{A}_{i,n}{=}\pm1$ at $x_n$ (where $\mathbf{A}_i$ is the corresponding column of the dictionary matrix) and remain at that value until the next data point $x_{n+1}$. They only switch value at data points, and do not have \kink{}s at reflections. The green graph in \Cref{fig:3lyr_arch_sign_tikz} illustrates an example of a sign activation feature.

\Cref{thm:3lyr_sign} generalizes \Cref{thrm:lasso_deep_1nueron_ReLU} for sign networks.
By \Cref{lemma:sign_thresh_a}, for $L{=}2$, $1{=}d{=}m_0{=}L{-}2$, so the sign activation dictionary is also $\switchmatrix{m_{L-2}}{=}\switchmatrix{1}$. 
Adding a third layer to a parallel network with sign activation expands the library to encompass features with up to $m_1$ switches. But no new features enter the library if the depth increases beyond $3$. This limited library suggests that the representation power of sign activation networks may stagnate after three layers, unless the architecture is changed. Indeed, for a tree architecture, the library continues to expand, and the features have as many \kink{}s as the product of the number of neurons in each layer.

\iftoggle{long}{
that gives the same outputs as an optimal neural net for \eqref{eq:nonconvex_generic_lossL_3yr} with sign activation, and is used in its reconstruction from \lasso{} \eqref{thm_eq:3lyrsign} . }

\iftoggle{long}{
The next two theorems generalize \Cref{thm:3lyr_sign} to deep neural networks: they too are \lasso{} problems.

\begin{theorem}\label{thm:deeper_sign_par}
	\Cref{thm:3lyr_sign} still holds as is, if additional layers with $m_1$ neurons each are added between the first and final layer of the parallel neural net (which has $m_2$ neurons).
\end{theorem}

\begin{theorem}\label{thm:3lyr_sign_tree}
	For a tree neural net with $m_l$ layers in layer $L$, \Cref{thm:3lyr_sign} holds with the change that the dictionary matrix $\Amat{}$ contains as columns all vectors  $a \in \{-1,1\}^N$ that start with $a_1=1$ and have at most $\min \left \{N-1, \prod_{l=1}^{L-1} m_l \right \}$ switches.
\end{theorem}

}

\iftoggle{long}{The proof of  \Cref{thm:3lyr_sign_tree} is in \Cref{app_section:3lyr_sign}.}

An explicit and efficient reconstruction of an optimal $3$-layer neural net with sign activation is described in \Cref{thm:3lyr_sign_reconstruction} in \Cref{app:treenetsec}.
\iftoggle{arxiv}{It is drawn in \Cref{fig:3lyr_arch_sign_tikz}. }

Reconstructions for other architectures are given in \Cref{appsec:1Ddatasignact}.
%
%
The \lasso{} dictionary for deep neural nets in previous work \citep{ergen2023globally} uses a dictionary that depends on the training data. However, \Cref{lemma:sign_thresh_a} and \Cref{thm:3lyr_sign} show that
networks with sign activation have dictionaries that are invariant to the training data. So to train multiple neural nets on different data, the dictionary matrix $\Amat{}$ only needs to be constructed once. 

%
%
\iftoggle{long}{The proof of \Cref{Cor:beta->0_overfit} is in \Cref{app_section:3lyr_sign}.}
\iftoggle{long}{The rate at which the neural nets converge to exact interpolation can be seen in an example in \Cref{section_main:square_wave}.}

Our theory for \nd{1} data provides a lens to analyze higher dimensional data. 
Using a similar approach, \Cref{lemma:2D_upperhalfplane_signact} in \Cref{app:treenetsec} gives an example of \nd{2} data for which a neural net can be recast as a \lasso{} model. Extending this to more general data and higher dimensions is an area for future work.
 The next remark summarizes the expansion of libraries over depth.  
\begin{remark}\label{rk:dictsgetbigger}
    The dictionary for an architecture discussed in \Cref{thrm:lasso_deep_1nueron_ReLU}, \Cref{thm:3lyr_sign} or \Cref{lemma:2D_upperhalfplane_signact} is a superset of any dictionary with the same architecture but shallower depth.
\end{remark}
 This section analyzed the equivalence of neural networks with sign activation and \lasso{} models. While their features do not contain reflections, they are straightforward binary features that elucidate how depth increases the representation power of $2$ versus $3$-layer networks. For example, \Cref{cor:3lyr_m1x2lyr} in \Cref{app:treenetsec} uses the \lasso{} problem to  bound the training loss for $2$-layer networks with sign activation between those of depth $3$. The next section leverages the sign activation features to show that in a case of binary, periodic data, $3$-layer networks have a solution path indicating better generalization properties than $2$-layer networks.

\iftoggle{long}{
	\section{$L$ layer parallel networks with sign activation }

	The next theorem states that for sign activation, adding more layers cannot make the dictionary wider. Deepening neural nets with more layers of the same size does not improve them if sign activation is used.

	\begin{theorem}\label{thm:deeper_sign_par_long}
		Consider a $L$-layer neural net where all hidden layers have the same size as the first layer, with $m_1$ neurons. Let $m_2$ be the number of neurons in the final outer layer. Then theorem \ref{thm:3lyr_sign} also holds. In particular, the dictionary matrix $A$ still contains only the vectors in $\{-1,1\}^N$ with at most $m_2$ switches.
	\end{theorem}

	The proof is in appendix  \ref{app_section:3lyr_sign}. Theorem \ref{thm:deeper_sign_par} states that in a parallel neural network with $L>3$ layers, the dictionary $A$ is still the same as for $L=2$ layers, so the training objective can't decrease with more layers.

	While more layers beyond three doesn't improve the training loss of a parallel network, it may for a tree architecture, due to the widening of its matrix. This is stated in the next theorem \ref{thm:3lyr_sign_tree}.
}

\iftoggle{long}{\emi{if there is time: can we find a case where the tree network does strictly better than the parallel case? Would need at least 4 layers. We could come up with exact linear combination of vectors that switch many times but only a few neurons in each layer, for the case when $\beta \to 0$. And then the total weight, which is "equivalently" the total weight of the outer layers, should be less than using a parallel or even 2 layer net? Compare against using a wider 3 layer net.}}

\iftoggle{long}{

	\subsection{$L$-layer tree networks with sign activation, 1D}

	\begin{theorem}\label{thm:3lyr_sign_tree_long}
		Consider the $L$-layer tree neural net with sign activation $\sigma_s(x)= s \cdot \mbox{sign}(x)$ and $m_1, m_2, \cdots, m_{L-1}, m_L$ neurons in the layers $1$ to $L$.  Consider the non-convex training problem given in \eqref{eq:nonconvex_generic_loss_gen_L_intro}. The optimal value is equal to

		\begin{equation}
			\min_{z} \mathcal{L}_{\mathbf{y}}(Az) + \beta ||z||_1,
		\end{equation}

		and a  globally optimal solution to \eqref{eq:nonconvex_generic_loss_gen_L_intro} can be found from the solution to Lasso \eqref{thm_eq:3lyrsign}, if the number of final layer hidden neurons $m_L \geq ||z^*||_0$.  $A$ is a matrix containing as columns all vectors  $a \in \{-1,1\}^N$ that start with $a_1=1$ and have at most $\min\{N, \prod_{l=1}^{L-1} m_l\}$ switches between $1$ and $-1$.
	\end{theorem}

	The proof is in appendix  \ref{app_section:3lyr_sign}. The tree dictionary matrix has more columns for a tree than the parallel neural net so the training objective could be decreased with more layers.
}

\iftoggle{long}{\textcolor{red}{todo: if there is time, verify, clean and insert theorem from the appendix about closed form solution for 2 layer threshold activation nn when $\beta \to 0$.}}

\begin{figure}[t]
\centering
\begin{minipage}[t]{0.45\textwidth}
        \includegraphics[width=1\linewidth]{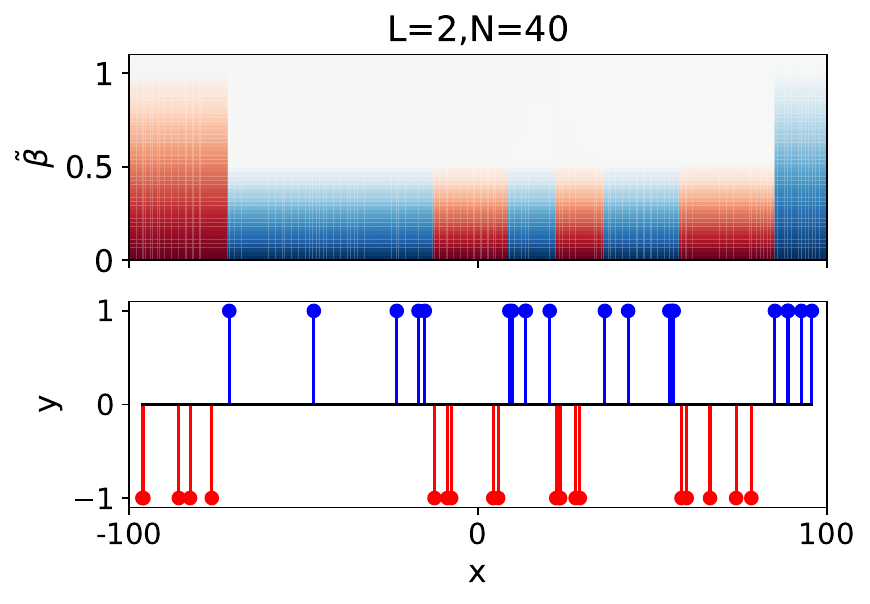}
    \end{minipage}
    \begin{minipage}[t]{0.498\textwidth}
        \includegraphics[width=1\linewidth]{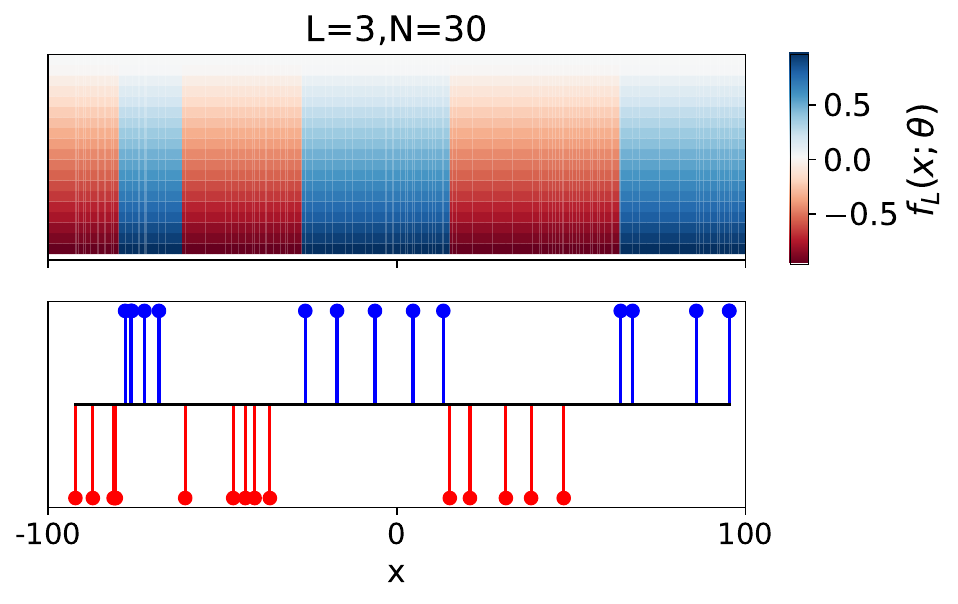}
    \end{minipage}

\caption{The bottom figures plot training data $(x_n,y_n)$. The top figures plot sign-activated neural net predictions by color for each $x$, as parameterized by $\beta$ on the vertical axis. With abuse of notation, in this figure (only), $\tilde{\beta}=\beta/T$. 
}
\label[figure]{fig:decision_regions}
\end{figure}

\section{\secnameSignSol}\label{section_main:square_wave}

\iftoggle{long}{
\nd{1} binary labeled data can be viewed as samples from non-overlapping intervals in $\mathbb{R}$ with alternating labels. Suppose our training data has $k$ samples from each interval. In other words, the first $k$ samples $x_1, \cdots, x_k$ have label $y_1 = \cdots = y_k =1$, the second $k$ samples have label $-1$, and this alternation of label groups repeats. The vector $\textbf{y}$ is a discrete square wave
\iftoggle{long}{when plotted against indices from $1$ to $N$,}
}

This section examines solution paths of \lasso{} problems for $2$ and $3$-layer neural nets
with sign activation and \nd{1} data where the target vector $\mathbf{y}$ is binary. Such data appears in temporal sequences such as binary encodings of messages communicated digitally \citep{kim:2018a}, neuron firings in the brain \citep{fang:2010}, and other applications, where $x_n$ represents time. These real-world sequences are in general aperiodic. 
However, in the special case that the \labelvec{} is periodic and binary, the \lasso{} problem gives tractable solutions for optimal neural networks. This offers a step towards analyzing neural network behavior for more general, aperiodic data, which is an area for future work. We call the binary, periodic sequence a square wave, defined as follows.
%
For a positive even integer $T$ that divides $N$, define a \emph{\hktname{}} to be $\hkt \in \{-1,1\}^N$ that starts with $1$ and is periodic with period $T$. If the real line is split into a finite number of regions by binary labels,
the \hktname{} represents the labels of a monotone sequence of points, with the same number of samples in each region.
The black dots in \Cref{fig:pulse_ex_overlap_2lyr} plot an example of a \hktname{}.
\iftoggle{long}{
For integers $k, T$ such that $N=kT$, we define the square wave vector $ \in \{-1,1\}^N$ to be a discrete-sample square wave of $k$ cycles, each of length $T$ over its indices. Each cycle consists of a half cycle of length $T/2$ indices over which $\hkT$ is $-1$, and a half cycle over which $\hkT$ is $1$. Without loss of generality since $\hkT$ contains the same number of $1$'s as $-1$'s, assume the first element ${\hkT}_1=-1$. \label{def:informal_hKT}
A formal definition of $\hkT$ is given in \Cref{def:h_{k,T}}, \Cref{app_sec:squarewave-2lyr}.
\iftoggle{long}{\subsection{2 layer neural net}}
}

	\iftoggle{long}{Note $w_{bdry} < 0, w_{cycle}>0$.}

	\iftoggle{long}{\textcolor{red}{todo: can add noisy versions too, but should verify with sims more. }}

\iftoggle{todos}{\textcolor{red}{The proof is by lemma \ref{lemma:2lyr_sign_square_wave_big_beta} in the appendix for the case when $\beta \geq \frac{T}{2}$, and lemma \ref{lemma:lasso_sol_square_wave_noise} in the appendix for $\beta \leq \frac{T}{2}$.}}

Consider training a $2$-layer neural net with sign activation when $\mathbf{y}=\mathbf{h}^{T}$. When $\beta > T$, an optimal neural net is the constant zero function. 
When $\beta \leq \frac{T}{2}$, we can find an optimal neural net $\nn{2}{\mathbf{X}}$ which is
 is periodic over $\left[\frac{T}{2}, N - \frac{T}{2}\right]$ with period $T$, and has amplitude $2\frac{\beta}{T}$ less than that of $\mathbf{y}$. \Cref{thm:no_noise_square_wave_2lyr} 
 gives the entire \lasso{} solution path and \Cref{cor:solfor2lyrwave} gives a closed form expression for the resulting optimal neural net. 
%
%
\iftoggle{long}{\subsection{3 layer neural net}}
\Cref{thm:3layer_no_noise} and \Cref{cor:solfor3lyrwave} give the solution path and an optimal neural net when $L=3$.
%
Only one parallel unit is active in this network, and so it is also a standard neural net. 
The neural net has output $\nn{3}{x}(\mathbf{X})=(1-\betafrac)_+\mathbf{y}$. If $\beta > N$, then the optimal neural net is the constant zero function. These results are in \Cref{sec:squarewaveapp} due to space.

\iftoggle{long}{
	\iftoggle{todos}{\subsection{Noisy labels}
	}

	3 layers generalizes better.

	\iftoggle{todos}{\begin{itemize}
			\item 2 layer solution for noncritical $\beta$, and noisy label at one point
			\item 2 layer neural net solution for label being a square wave without noise and big $\beta$
			\item 3 layer neural net with noise, at critical beta
			\item test/train errors when there is no noise. and when there is noise at critical beta, compare 2 and 3 layers (\textcolor{red}{todo: finish})
		\end{itemize}}
}

\iftoggle{long}{

\begin{figure}[h]
	\centering
	\includegraphics[width=9cm]{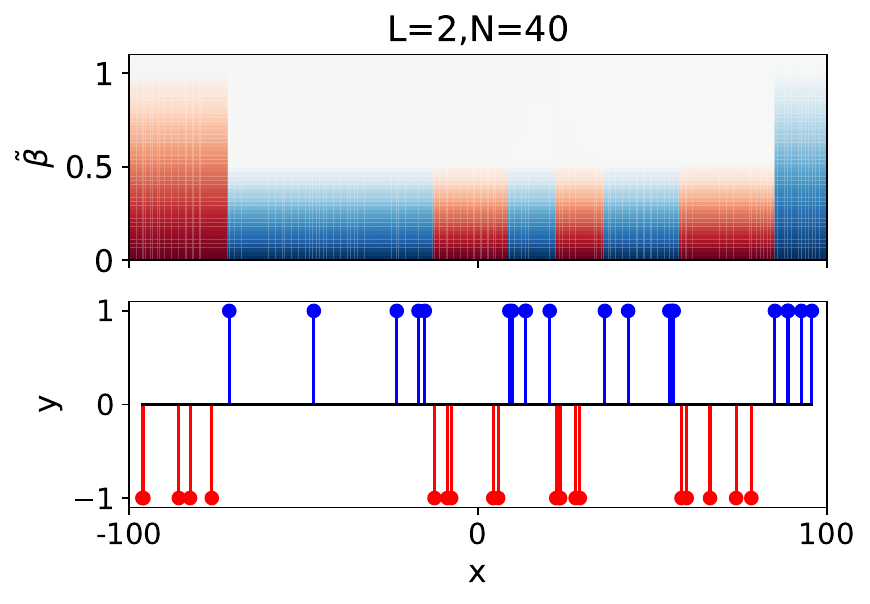}
	\caption{Decision regions of the optimal 2-layer neural network solution path. The activation is the sign function. The label is a square wave: alternating regions of opposite $1$ and $-1$ classes. The training data $X$ were $40$ randomly chosen points uniformly between $-100$ and $100$, and then sorted. The labels consist of a square wave of $k=4$ periods of length $T=10$. So every $T/2 = 5$ samples, the training data $X$ changes class. Decision regions are plotted over the horizontal axis as $\frac{\beta}{T}$ varies over the vertical axis. The colors are the neural net $\nn{L}{\mathbf{X}}$ predictions. $\frac{\beta}{T}=1$ is the critical $\beta$ (magenta dotted line), beyond which all decision regions are 0 (white). When $\frac{1}{2} < \frac{\beta}{T} < 1$ (between magenta and yellow dotted lines), only the outer 2 neurons are active (blue and red columns). Overall, as $\beta$ increases, the predicted neural net gets smaller. The Lasso formulation was used to train. This is consistent with \Cref{thm:no_noise_square_wave_2lyr}.}
	\label{fig:2lyr_square_wave}
\end{figure}

\begin{figure}[h]
	\centering
	\includegraphics[width=9cm]{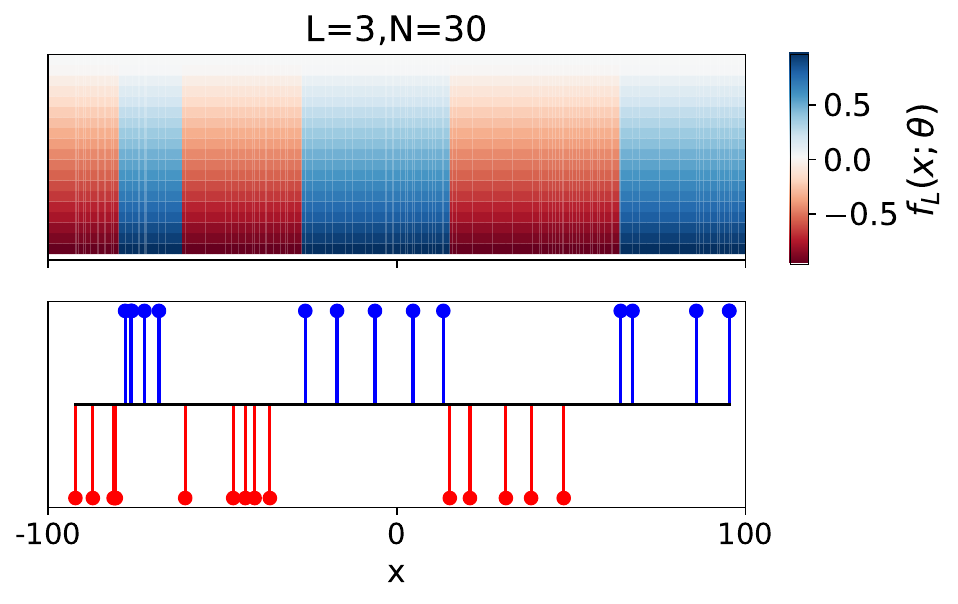}
	\caption{Decision regions of a reconstructed $3$-layer neural network with sign activation as a function of $\beta$. The \labelvec{} is a square wave of period $T=10$. The training data consists of $N=40$ points chosen uniformly at random between $-100$ and $100$. 
 So every $T/2 = 5$ samples, the training data $X$ changes class. Decision regions are plotted over the horizontal axis as $\frac{\beta}{N}$ varies over the vertical axis, where $N$ is the number of samples. The colors are the neural net $\nn{L}{\mathbf{X}}$ predictions. $\frac{\beta}{N}=1$ is the critical $\beta$ (magenta dotted line), beyond which all decision regions are 0 (white). Overall, as $\beta$ increases, the predicted nueral net gets smaller with uniform amplitude over the sample points. The Lasso formulation was used to train. This is consistent with
 \Cref{thm:3layer_no_noise}. }
	\label{fig:3lyr_square_wave}
\end{figure}

}
\Cref{fig:pulse_ex_overlap_2lyr} 
illustrates the solution path of $ \nn{L}{\mathbf{X}} $ when $\mathbf{y}=\hkt$. It suggests that as the regularization increases, the $2$-layer network focuses on preserving the boundary points of the data (first and last $T$ points) to closely match the \labelvec{}. Therefore the network will generalize well if noise occurs in the middle of the data. In contrast, if noise occurs uniformly over the data, the $3$-layer network will generalize well.

We verify our theoretical predictions for optimal neural nets in \Cref{thm:no_noise_square_wave_2lyr} and 
\Cref{thm:3layer_no_noise} 
by solving the \lasso{} problem on sample training data with \labelvec{} $\hkt$.  \Cref{fig:decision_regions} illustrates the training data in the bottom plot and neural net predictions for each $\beta$ in the rows of the top plot. The $2$-layer network is biased towards predicting strongly in the first and last intervals, suggesting worse generalizability than the $3$-layer network. In addition, the $3$-layer net changes more uniformly with $\beta$ than the $2$-layer net, making it easier to tune $\beta$.



\section{Application: Time-series modeling}\label{sec:autoreg}

\begin{figure}[t]
\centering
\includegraphics[width=\linewidth]{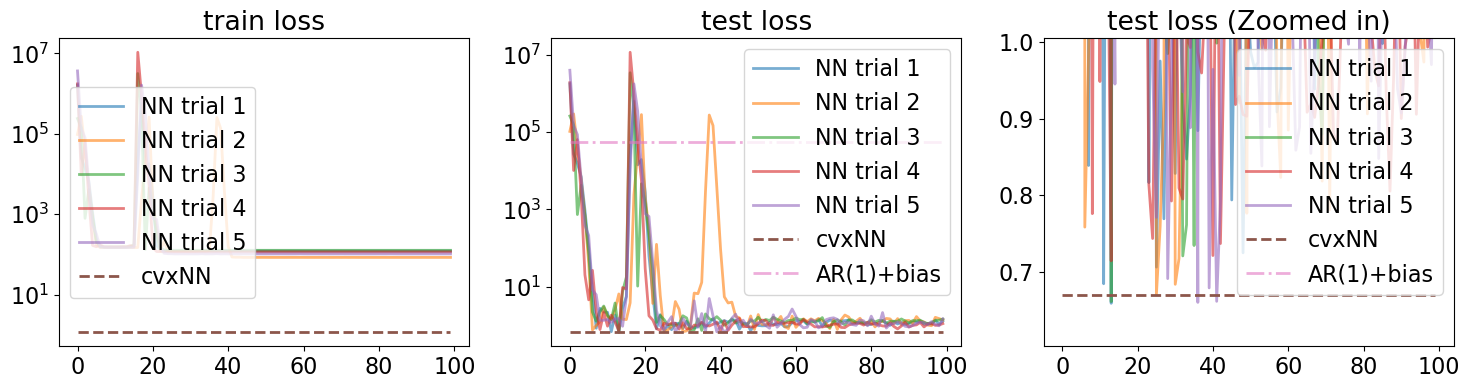}
\caption{Comparison of neural autoregressive models of the form $x_{t}=f(x_{t-1};\params)+\epsilon_t$ using convex and non-convex optimizers and the classical linear model AR(1) for time series forecasting. The horizontal axis is the training epoch. The dataset is BTC-2017min from Kaggle, which contains all $1$-minute Bitcoin prices in 2017 \citep{kaggle}. The non-linear models outperform the linear AR(1) model. Moreover, SGD underperforms in training and test loss compared to the convex model which is guaranteed to find a global optimum of the NN objective.}\label[figure]{fig:BTC_main_paper}
\end{figure}

In this section, we apply the \lasso{} problem for neural networks to an autoregression problem. Suppose at times $1, \cdots, T+1$ we observe data points $x_1, \cdots, x_{T+1} \in \mathbb{R}$ that follow the time-series model
\begin{equation}\label{equ:ar}
	x_{t} =  \nn{}{x_{t-1}} +\epsilon_t,
\end{equation}
where $f:\mathbb{R}\to\mathbb{R}$ is parameterized by some parameter $\theta$ and $\epsilon_t\sim \mathcal{N}(0,\sigma^2)$ represents observation noise. The parameter $\theta$ is unknown, and the goal is find $\theta$ that best fits the model \eqref{equ:ar} to the data $x_1, \cdots, x_{T+1}$. For example, the \textit{auto-regressive model with lag $1$} (AR(1)) is a linear model $f(x;\theta)=ax$ where $\theta=\{a\}$ is chosen as a solution to
$\displaystyle
 	\min_{\params \in \paramspace}\sum_{t=1}^{T} \left( \nn{}{x_t}-x_{t+1}\right)^2.
$
For a more expressive model, instead of $f(x;\params)=ax$ suppose we use a $2$-layer neural network 
\begin{equation}\label{eq:nn_autoreg}
	\NN{2}{x}  = \sum_{i=1}^m \left|xw_i+b_i \right |\alpha_i,
\end{equation}
which has $m$ neurons and absolute value activation. The parameter set is $\theta=\{w_i,b_i,\alpha_i\}_{i=1}^m$. 

 Here, we show how to find a neural network model $ \NN{2}{x_t}$ \eqref{eq:nn_autoreg} that represents the $\tau$-quantile of the distribution of $x_{t+1}$ given the observation $x_t$, where $\tau\in[0,1]$, by using the \textit{quantile regression loss} $L_\tau(z)=0.5|z|+(\tau-0.5)z$ and choosing $\theta$ that solves the \textit{neural net (NN) quantile regression (QR) training problem}
\begin{equation}\label{eq:quantile_reg}
	\min_{\params \in \paramspace}\frac{1}{T}\sum_{t=1}^{T}
 L_\tau \left(\NN{2}{x_t}-x_{t+1} \right)+\frac{\beta}{2}\|\regparams\|_2^2. 
\end{equation}
Problem \eqref{eq:quantile_reg} can be solved by converting it to an equivalent \lasso{} problem. \Cref{fig:BTC_main_paper} shows that using the \lasso{} model to predict Bitcoin price reaches a lower loss than training with the \nc{} model. More details are found in \Cref{app:yifeictd}.

\section{Conclusion}
Our results show that deep neural networks with a variety of activation functions trained on \nd{1} data with weight regularization can be recast as convex Lasso models with simple dictionary matrices. This provides critical insight into their solution path as the weight regularization changes. The \lasso{} problem also provides a fast way to train neural networks for \nd{1} data. Moreover, the understanding of the neural networks through Lasso models could also be used to explore designing better neural network architectures.

We proved that reflection features can emerge in the \lasso{} dictionary when the depth is $3$ or deeper. This leads to predictions that have breakpoints at reflections of data points about other data points. In contrast, for networks of depth $2$, the breakpoints are only located at a subset of training data. We believe that this mechanism enables deep neural networks to generalize to the unseen by encoding a geometric regularity prior.

\iftoggle{long} {For general higher dimensions, the training problem with sign and threshold activations can also be shown to be a Lasso problem, and the proof in the appendix holds for general dimensions (also see \citep{ergen2023globally}). However in general for dimensions greater than $1$ the Lasso dictionaries are implicit.}
Our analysis of various architectures provides a foundation for studying more complex network topologies.
The \nd{1} results can extend to sufficiently structured or low rank data in higher dimensions. 
Generalizing to higher dimensions is also an area of future work. Building on a similar theme, \citep{pilanci2023complexity} showed that  the structure of hidden neurons can be expressed through convex optimization and Clifford's Geometric Algebra. The techniques developed in this paper can be combined with the Clifford Algebra to develop higher-dimensional analogues of the results.


\newpage
\bibliographystyle{siamplain}
\bibliography{references}

\iftoggle{todos}{\section{Introduction}
Background on DNNs and non-convex models.
\subsection{Prior work}

\begin{itemize}
    \item Background on convex neural networks.
    \item Neural networks were recast as convex LASSO problems in \citep{}. There could be multiple solutions \cite{}.
    \item For one-dimensional problem settings, neural networks were found to perform dataset interpolation in \cite{}.
    \item However the parameters of the LASSO problem (the dictionary matrices) were implicit \cite{} or especially for high dimensional data required potentially large exhaustion to find, or subsampling \cite{}.
    \item For $1$-D data, we find explicit dictionary matrices which provide insight into the neural network.
\end{itemize}

\subsection{Motivation}
(Why study low dimensional (e.g., 1D, 2D, 3D data)?)
\begin{itemize}
    \item Gives a complete and rigorous theoretical description of a deep neural network model with arbitrary depth (black box to white box)
    \item There are implicit image/object models (fitting a NN to a single image or 3D object) that work with low dimensional data (d=2,3). Audio data such as speech and music are one dimensional as well. Low dimensional NN models are actually useful.
    \item There are applications of neural networks in solving differential equations, for which $d=1$ is a common setting.
    \item Pedagogical value of an explanation that is transparent and exact is high. We can interpret DNNs better as convex models.
\end{itemize}
\section{Problem formulation}
Consider an $L$-layer DNN with a parallel architecture and biases at each layer. We will also consider tree NN.

Assumption: without loss of generality by reordering, if the data samples $x_1, \dots, x_N$ are one-dimensional, assume that $x_1 > \dots > x_N$. This will give a convenient structure to the dictionary matrices in the resulting LASSO problems.

\subsection{Contributions}
\begin{itemize}
\item 2-layer NNs with ReLU/abs/threshold/sign activation are identical to convex Lasso models with dictionary matrix $A_{i,j}=\sigma(x_i-x_j)$.
\item L-layer NNs with sign activation are identical to convex Lasso models. Tolga's paper on threshold activations characterized the dictionary by $\mathcal{H}$. Here, for $1D$ data on sign activation, the dictionary has columns with a specific number of switches between $-1,1$.
\end{itemize}

The reconstruction is described in the appendix.

\section{Main results (1D)}

\subsection{2-layer NN with ReLU/abs/thresh/sign ($1D$)}
\begin{itemize}
    \item dictionary matrix (\textcolor{red}{todo} +picture)
    \item recover nn from Lasso
    \item sign activation: unique solution
    \item sign activation: interpolate exactly as $\beta \to 0$: $z=A^{-1}y$.
    \item threshold activation: min solution as $\beta \to 0$ (\textcolor{red}{todo: check, clean})
\end{itemize}

\subsection{L-layer NN with sign activation}
\begin{itemize}
    \item for L=2, $D=1$, dictionary has one-switch columns
    \item for L=3, $D=1$, dictionary has m-switch columns
    \item for $L>3$ and $D>1$, refer to appendix 
    \item for parallel neural net, if $D=1$, $>3$ layers doesn't change the dictionary matrix (\textcolor{red}{todo: + plot})
\end{itemize}

\section{sign activation and more layers}

\begin{itemize}
    \item For $D=1$, $p_{3,\beta} \leq p_{2,\beta} \leq p_{3,m_1\beta}$ 
\end{itemize}

\section{Toy applications for intuition}

\subsection{sign activation and square wave label: 2 vs 3 layers}

-Characterization of solution path : 2 and deeper models. 3 layers generalizes better. 

-X is a disjoint union of points, of alternating labels. WLOG assume same number of points in each.

\begin{itemize}
    \item solution is 0 below critical $\beta$
    \item 2 layer solution for noncritical $\beta$, and noisy label at one point 
    \item 2 layer solution for critical $\beta$, no noise
    \item 3 layer solution, no noise and appropriate number of neurons 
    \item test/train errors when there is no noise (\textcolor{red}{todo: finish})
    \item \textcolor{red}{todo: picture of 2 and 3 layer at big, small, medium $\beta$ with heights labeled}
\end{itemize}

\subsection{special cases of 2D data}

\begin{itemize}
    \item data on a unit circle
    \item data on the upper half plane
    \item rank 1 data (see Mert comment) (\textcolor{red}{todo})
    \item \textcolor{red}{todo: how to recover nn from Lasso}
\end{itemize}

\section{Why use parallel NN}

-Mention but mostly refer to appendix?

\begin{itemize}
    \item standard NN can achieve strong duality if $m=N$
    \item example where standard NN can fail to achieve strong duality as $\beta\to0$ if $m<N$ (\textcolor{red}{todo: read through, fix reference})
    \item parallel to standard conversion is in appendix
\end{itemize}

-sign vs sgn def

\section{Experiments}+

\subsection{Discussion}
(Why are these results interesting).
\begin{itemize}
    \item Hidden convexity in non-convex optimization
    \item Computational cost to find the global minimum. All of our models can be globally optimized in polynomial time with respect to all parameters. 
     \item size of LASSO matrix shows how neural networks may admit unique solutions depending on the data and the activation function.
\end{itemize}
}

\newpage

\appendix


\section{Detailed results for \Cref{section:nn_arch}}

\begin{remark}[Parallel to standard architecture conversion]\label{parallel2std}
    Let $\mathbf{W}^{(1)} = \left[ \Wul{1}{1} \cdots \Wul{m_1}{1} \right]$. For $l \geq 1$, let $\mathbf{b}^{(l)} = \left( \bul{1}{l} \cdots \bul{m_l}{l} \right)$. For $l > 1$, let
$\mathbf{W}^{(l)} = \mathrm{blockdiag}\left( \Wul{1}{l} \cdots \Wul{m_l}{l} \right)$. 
And let $\boldsymbol{\alpha}, \xi$ be the same in the standard network as the parallel one.
\end{remark} 



\section{Detailed results for \Cref{sec:deepnarrownets}} \label{app:reconstrct} Proofs are defered to \Cref{app:deepReLU}.

\begin{figure}
    \centering
    \begin{minipage}[t]{0.9\textwidth}
    \includegraphics[width=\linewidth]{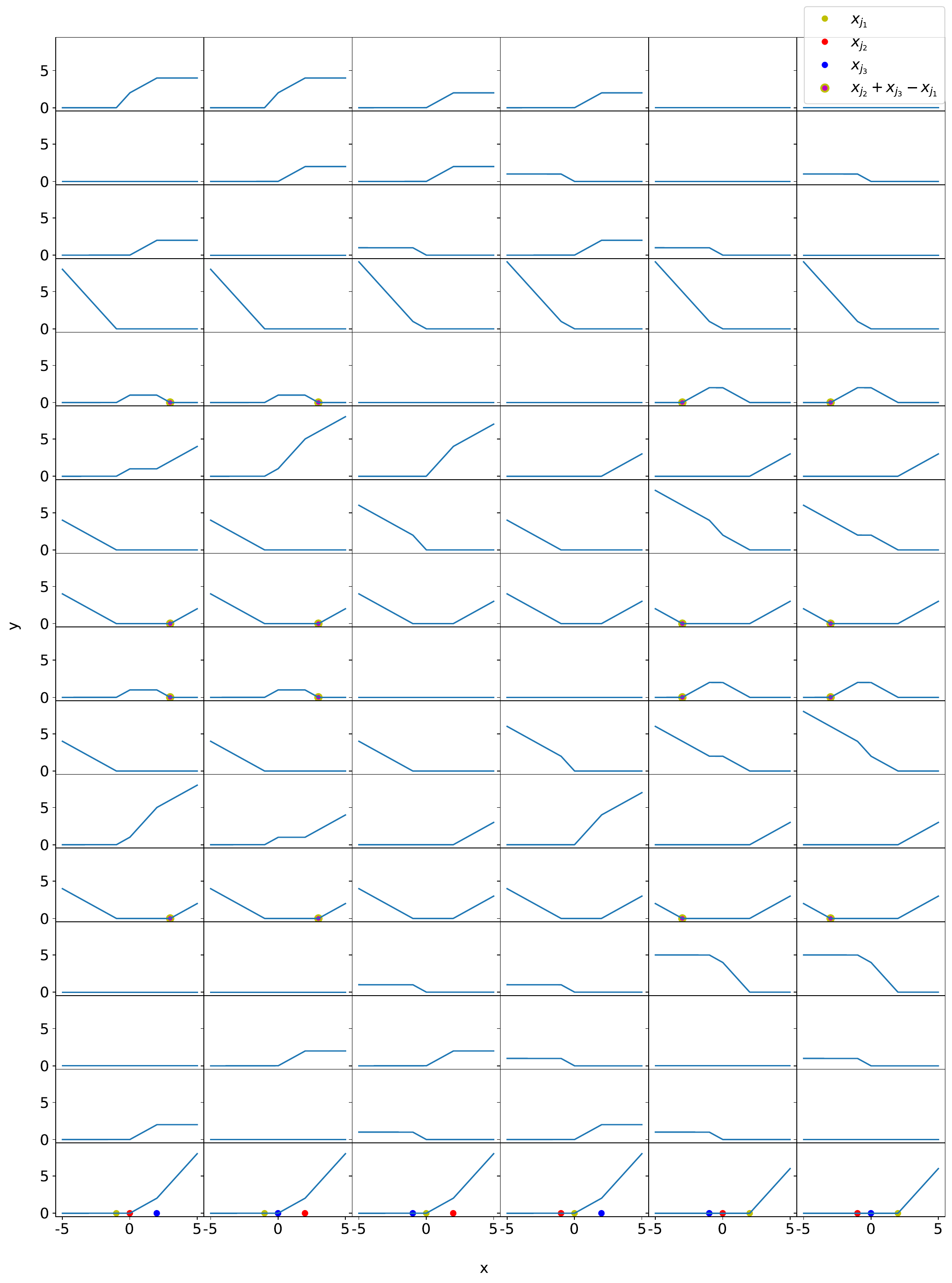}
    \end{minipage}
    \caption{Figure for \Cref{app:reconstrct}. 
 Deep library features for a $3$-layer \symmeterized{} ReLU network. Each row corresponds to a different set of weights $\wl{1} {\in} \{{-}1,1\}^{1\times 2},\wl{2} {\in} \{{-}1,1\}^{2 \times 1}$. Each column corresponds to a different ordering of $x_i, x_j, x_k$. The generalized reflections of $x_{j_1}$ (\jonecol{}) across $x_{j_2}$ (\jtwocol{}) and $x_{j_3}$ (\jthreecol{}) are depicted by \jonecol{} encircling \jtwojthreecol{}.}
    \label{fig:ReLU2neurons}
\end{figure}

%

\begin{definition}\label{def:wfeat}
    Define the function $\mathcal{W}_{\alpha,\beta}:\mathbb{R}\to\mathbb{R}$ parameterized by $\alpha, \beta \in \mathbb{R}$ as
    \begin{equation*}
    \begin{aligned}
       \mathcal{W}_{\alpha,\beta}(x)  
        &= \begin{cases}
            \alpha - x &\mbox{ if } x \leq \alpha\\
            x - \alpha &\mbox{ if } \alpha \leq x \leq \beta \\
            \reflect{\alpha}{\beta} - x &\mbox{ if } \beta \leq x \leq  \reflect{\alpha}{\beta}\\
            x - \reflect{\alpha}{\beta}&\mbox{ if } x \geq \reflect{\alpha}{\beta}.
        \end{cases}
    \end{aligned}
\end{equation*}
\end{definition}

\begin{lemma}[Examples of the Deep Library]\label{def:ReLU dic func}
ReLU features $\xl{L}(x)$ are those plotted in \Cref{fig:ReLUfeatdiagram}. As shown in the figure, \symmeterized{} ReLU network features contain generalized reflections of the form $x_i{+}x_j{-}x_k$. We now describe features for other activations. Let $L \in \{2, 3\}$. \\
%
%
\textbf{If } $\mathbf{L=2}$: for $\bl{1}=-x_j\wl{1}$,
\begin{equation*}
    \begin{aligned}
    \xl{L}(x)=
        \begin{cases}
            \sigma(x_{j_1}-x) & \mbox{ if } \wl{1} = -1 \\
            \sigma(x-x_{j_1}) & \mbox{ if } \wl{1} = 1 \\
        \end{cases}
    \end{aligned}
\end{equation*}

\textbf{If $\mathbf{L=3}$:} 

\quad if $\sigma(x)=\text{ReLU}(x)$ and $m_1=1$: for $\bl{1} = -x_{j_1}\wl{1}$,

\quad \quad if $\wl{2}=1$:

\begin{equation*}
    \begin{aligned}
    \xl{2}(x)=
        \begin{cases}
            \ReLU{-}{\min \left\{x_{j_1},x_{j_2}\right\}}(x) & \mbox{ if } \wl{1} = -1 \\
            \ReLU{+}{\max \left\{x_{j_1},x_{j_2} \right\}}(x) & \mbox{ if } \wl{1} = 1 \\
        \end{cases}
    \end{aligned}
\end{equation*}

\quad \quad if $\wl{2} =-1$:
\begin{equation*}
    \begin{aligned}
    \xl{L}(x)=
        \begin{cases}
            \ramp{+}{x_{j_2}}{x_{j_1}}(x) & \mbox{ if } \wl{1}  = -1 \\
            \ramp{-}{x_{j_1}}{x_{j_2}}(x) & \mbox{ if } \wl{1}  = 1.
        \end{cases}
    \end{aligned}
\end{equation*}

\quad if $\sigma(x)=|x|$ and $m_1=1$: for $a \in \left\{x_{j_1}, \frac{x_{j_1}+x_{j_2}}{2}\right\}$,
\begin{equation*}
    \begin{aligned}
        \xl{L}(x)=
        \begin{cases}
            \mathcal{W}_{\min \left\{x_{j_2}, \reflect{x_{j_2}}{a} \right\},a}(x) &\mbox{ if } \bl{1} = -a\wl{1} \\
            \mathcal{W}_{\min \left\{ \reflect{a}{x_{j_1}},\reflect{a}{x_{j_2}} \right\},a}(x) &\mbox{ if }  \bl{1} = -a\wl{1} .
        \end{cases}
    \end{aligned}
\end{equation*}


\iftoggle{long}{
\textbf{If $\mathbf{L=4}$: }

\quad  if $s_{2}=s_{3}=1$:

\begin{equation*}
    \begin{aligned}
    \ReLUdicfun{L}(x)=
        \begin{cases}
            \ReLU{-}{\min \left\{a_1,x_{j_2},x_{j_3} \right\}}(x) & \mbox{ if } s_{1} = -1 \\
            \ReLU{+}{\max \left\{a_1,x_{j_2},x_{j_3} \right\}}(x) & \mbox{ if } s_{1} = 1 \\
        \end{cases}
    \end{aligned}
\end{equation*}

\quad if $s_{2}=1, s_{3}=-1$:

\begin{equation*}
    \begin{aligned}
     \ReLUdicfun{L}(x)=
        \begin{cases}
            \ramp{+}{x_{j_3}}{ \min \left\{a_1,x_{j_2} \right\}}(x) & \mbox{ if } s_{1} = -1 \\
            \ramp{-}{\max \left \{a_1,x_{j_2} \right\}}{x_{j_3}}(x) & \mbox{ if } s_{1} = 1 \\
        \end{cases}
    \end{aligned}
\end{equation*}

\quad if $s_{2}=-1, s_{3}=1$:

\begin{equation*}
    \begin{aligned}
     \ReLUdicfun{L}(x)=
        \begin{cases}
            \ramp{+}{\max\{x_{j_2},x_{j_3}\}}{a_1}(x) & \mbox{ if } s_{1} = -1 \\
            \ramp{-}{a_1}{\min\{x_{j_2},x_{j_3}\}}(x) & \mbox{ if } s_{1} = 1 \\
        \end{cases}
    \end{aligned}
\end{equation*}

\quad if $s_{2}=-1, s_{3}=-1$:

\begin{equation*}
    \begin{aligned}
     \ReLUdicfun{L}(x)=
        \begin{cases}
            \ramp{-}{x_{j_2}}{\min\{a_1,x_{j_3}\}}(x)
            & \mbox{ if }  
            s_{1} = -1\\
            \ramp{+}{\max\{a_1, x_{j_3}\}}{x_{j_2}}(x) &   \mbox{ if } 
            s_{1} = 1 .
        \end{cases}
    \end{aligned}
\end{equation*}
}
%


\end{lemma}

\subsection{Reconstruction results}
In this section, let $(\mathbf{z}^*, \xi^*)$ be a solution to the \lasso{} problem. 
We give a map to efficiently and explicitly reconstruct an optimal neural net from $(\mathbf{z}^*, \xi^*)$ by leveraging the structure of the deep library (\Cref{def:FeatDicStdNN}). Proofs are deferred
to \Cref{app:deepReLU}.
%
%

 \begin{definition}[Reconstructed parameters]\label{def:reconstructfunc}
     The \emph{reconstructed parameters} for a parallel network are constructed as follows. For each $i^{\text{th}}$ column $\mathbf{A}_i$ of the dictionary matrix such that $z^*_i\neq0$, let $\Xul{i}{L}$ be the parallel unit $\xl{L}$ corresponding to that column in the deep library. 
%
%
  %
  %
  Let  $ \boldsymbol{\alpha} {=} \mathbf{z}^*$ and $\xi{=}\xi^*.$ For sign and threshold activation, let all amplitude parameters be $1$.
  %
  %
   Finally, \unscal{}e parameters (\Cref{parameterunscale}).
  \iftoggle{long}{
  For each $(\ReLUdicindices{}) \in \dicindexset{}$, let $\gamma =  \left | \alpha_{i} \right | ^{1/\effL{}}$. 
  Then for all regularized parameters $q \in \regparams^{(\mathbf{j})}$, rescale $q \to 
  \mbox{sign}(q) \gamma $ and for all neuron biases $q=\bul{i}{l}$, rescale $q \to q \gamma $. $\gamma_i = \sqrt{|z_i|}$ and transforming $s^{(i,L-1)} \to \gamma_i$ and $\alpha_i \to \alpha_i/\sqrt{|z_i|}$.
  }
\iftoggle{long}{ 
  , where $\regparams^{(\mathbf{j})}$ are rescaled parameters \eqref{eq:parallelrescaled} for sign-determined activation and original parameters for homogeneous activation. }
  \iftoggle{long}{
  $\alpha^{(\mathbf{j})} \to \mbox{sign} \left(\alpha^{(\mathbf{j})}\right) \gamma, w^{(\mathbf{j},l)} \to \mbox{sign}(w^{(\mathbf{j},l)}) \gamma, b^{(\mathbf{j},l)} \to \mbox{sign}(b^{(\mathbf{j},l)}) \gamma$.}
  \iftoggle{long}{ 
  Finally, if $\mathbf{z}_{i}=0$ then set all parameters indexed by $\ReLUdicindices$ to zero.
  }
\end{definition}
%
 A \textit{reconstructed network} is a network with reconstructed parameters. 
\begin{lemma}\label{lemma:parreconstructionworks}
   A reconstructed parallel network is optimal in the training problem.  
\end{lemma}
%
%
%
\iftoggle{long}{
\begin{remark}
Suppose $L=2$. Then the dictionary indices $\ReLUdicindices{}$ are all scalars. 
Consider a parallel network. All weighs and biases have layer index $l=1$.
 We denote $i = (\ReLUdicindices)  \in \dicindexset{}$ and write $w_i = \Wul{i}{l=1}, b_i  =\bul{i}{l=1}$. 
 Also denote $\mathbf{w}=(w_1, \cdots, w_{m_2}), \mathbf{b}=(b_1, \cdots, b_{m_2})$.
\end{remark}
}
%
Recall that each $2$-layer feature corresponds to $\wl{1} {\in} \{{-}1,1\}, \bl{1} {=} {-}\wl{1}x_n$ for some $n {\in} [N]$. 

\begin{definition}\label{rk:2lyrRecon}
Consider a $2$-layer ReLU, absolute value, or leaky ReLU network.
Let $R^{z\to\alpha}(z){=}\mathrm{sign}(z)\sqrt{|z|}$. 
For $w {\in} \{-1,1\}, b {\in} \{-x_n: n {\in} [N]\}$, let $ 
R^{\alpha,w,b \to \params}(\alpha) {=}\left(\alpha, \alpha w, - \alpha wb \right) $. 
Let $R^{w,b}(z){=}R^{\alpha,w,b \to \params} \left(R^{z\to\alpha}(z) \right)$. Define the $2$-layer \emph{reconstruction function} $R(\mathbf{z})$ which outputs a vector whose $i^{th}$ element is $R^{w,b}(z_i)$, where $(w,b)=\left(\wl{1},\bl{1}\right)$ corresponds to the $i^{\mathrm{th}}$ feature. 
\end{definition}

For a $2$-layer network, let $w_i {=} \Wul{i}{1}, b_i {=} \bul{i}{1} {\in} \mathbb{R}$ and $\mathbf{w}, \mathbf{b}, \boldsymbol{\alpha}$ be vectors stacking together all $w_i,  b_i, \alpha_i$ respectively.
The reconstructed parameters are $(\boldsymbol{\alpha}, \mathbf{w}, \mathbf{b})=R(\mathbf{z}^*)$ and $\xi=\xi^*$.

\iftoggle{arxiv}{
\begin{figure}[tb]
	\centering
    \scalebox{0.85}{
	\begin{tikzpicture}

		\node[] at (6.5, 3) {\large $ \nn{3}{\mathbf{X}}=\sum_{i=1}^{m_2}$ \textcolor{blue}{$\alpha_i $} $  $ \textcolor{green}{$\sigma_{\sul{i}{3}}$} $( $ \textcolor{red}{$\sigma$} $ ( \mathbf{X} $ \textcolor{purple}{$\Wul{i}{1}$} + $\mathbf{1} \cdot$ \textcolor{purple}{$\bul{i}{1}$}$)  $  \textcolor{orange}{$\mathbf{W^{(i,2)}}$} +   $ \mathbf{1 } \cdot $ \textcolor{orange}{$\mathbf{b^{(i,2)}}$} $)$  };

		\begin{scope}[scale=0.2, shift={(-13,20)}, color = purple]
			\draw ( 0,  0) -- (0, -1) -- (1,  -1)  -- (1 ,-2) -- (2,-2) -- (2,-3) -- (3, -3) -- (3, -4) -- (4, -4) -- (4, -5) -- (5, -5) -- (5,-6) -- (6, -6) -- (6,-7) -- (7,-7) -- (7,-8 ) -- (8,-8) -- (8,-9) -- (9,-9) --(9,-10) -- (10,-10) -- (10,-11) -- (11,-11) -- (11,-12) -- (12,-12) -- (12,-13) -- (13,-13) -- (13, -14) -- (14,-14) --(14,-15) ;
			\node[] (p0) at (5,-7) {};
			\node [color = black] at (5,1) {\small $\left(\mathbf{X} \Wul{i}{1} + \mathbf{1} \cdot \bul{i}{1}\right)_{j}=\mathbf{X}-x_{{I^{(i)}_j}} \mathbf{1}$}; 
			\node [] at (0, -5) {\large $\cdots$};
		\end{scope}

		\begin{scope}[scale=0.2, shift={(-16,1)}, color = red]
			\pulse{1}{1}  
			\node[] (p1) at (0.5,-15) {};
			\node[color=black] at (2,-8) {$-$};
			\node[color=black] at (-0.6, 1) {\small $-1$};
			\node[color=black] at (1.4, 1) {\small $1$};
			\node[color=black] at (-2, -15) {\small $j=1$};
		\end{scope}
		\begin{scope}[scale=0.2, shift={(-12,1)}, color = red]
			\pulse{1}{2}
			\node[] (p2) at (0.5,-15) {};
			\node[color=black] at (2,-8) {$+$};
		\end{scope}
		\begin{scope}[scale=0.2, shift={(-8,1)}, color = red]
			\pulse{1}{4}
			\node[] (p3) at (0.5,-15) {};
			\node[color=black] at (2,-8) {$-$};
			\node[] (p3top) at (0.5,0) {};
			\node[color=black, circle, draw] (oneswitchlabel) at (0.5,4) { $\sigma$};
			\node[color=black] at (3,-4) {\small $I^{(i)}_j$};
		\end{scope}
		\begin{scope}[scale=0.2, shift={(-4,1)}, color = red]
			\pulse{1}{7}
			\node[] (p4) at (0.5,-15) {};
			\node[color=black] at (2,-8) {$+$};
		\end{scope}
		\begin{scope}[scale=0.2, shift={(0,1)}, color = red]
			\pulse{1}{12}
			\node[] (p5) at (0.5,-15) {};
			\node[color=black] at (3,-8) {$-$};
		\end{scope}
		\begin{scope}[scale=0.2, shift={(4,1)}, color = red]
			\pulse{1}{14}
			\node[] (p6) at (0.5,-15) {};
			\node[color=black] at (4, -15) {\small $j=m_1$};
		\end{scope}

		\begin{scope}[scale=0.2,  shift={(10,-18)}, color = orange]
			\multipulse{2}{1}{2}{4}{7}{12}{14} 
			\node[] (multip_sum) at (0,-7) {};
			\node[] (multip_sumr) at (2,-7) {};
			\node[circle, text=black]  at (-1,1) {\small $-2$};
			\node[circle, text=black]  at (2,1) {\small $0$};
		\end{scope}
		\begin{scope}[scale=0.2,  shift={(20,-18)}, color = green]
			\multipulse{1}{1}{2}{4}{7}{12}{14} 
			\node[] (multip_sigma) at (0,-7) {};
			\node[] (multip_sigmar) at (1,-7) {};
			\node[circle, text=black]  at (-1,1) {\small $-1$};
			\node[circle, text=black]  at (1,1) {\small $1$};
		\end{scope}
		\begin{scope}[scale=0.2,  shift={(35,-18)}, color = blue]
			\multipulse{3}{1}{2}{4}{7}{12}{14} 
			\node[] (multip_alpha) at (0,-7) {};
			\node[] (multip_alphar) at (3,-7) {};
			\node[circle, text=black]  at (-1,1) {\small $-\alpha_i$};
			\node[circle, text=black]  at (3,1) {\small $\alpha_i$};
		\end{scope}

		\node[circle, draw, minimum size=1.2cm] (insum) at (1,-5) {\large $\sum$};

		\node[circle, draw] (outsigma) at (3.2,-5) {$\sigma$};

		\node[rectangle, draw, align = left] (alphai) at (5.5,-5) {\small $\alpha_i = \frac{z_{i}}{\sqrt{|z_{i}|}}$ \\ \small $\sul{i}{3} = \sqrt{|z_{i}|}$ }; 
		\node[circle, draw] (outsum) at (9,-5) {$\sum$};

		\node[rectangle, draw] (alpha1) at (8,-2) {${z}^*_{1}$};

		\node[rectangle, draw] (alpham) at (8,-8) {${z}^*_{{m_3}}$};

		\node[circle] (f) at (11,-5) {$\nn{3}{\mathbf{X}} $};

		\path[->, draw = black] (p0) edge (oneswitchlabel);
		\path[->, draw = black] (oneswitchlabel) edge (p3top);
		\path[->, draw = black] (p1) edge 
		(insum) ;
		\path[->, draw = black] (p2) edge (insum);
		\path[->, draw = black] (p3) edge (insum);
		\path[->, draw = black] (p4) edge (insum);
		\path[->, draw = black] (p5) edge (insum);
		\path[->, draw = black] (p6) edge (insum);
		\path[->, draw = black] (insum) edge (multip_sum);
		\path[->, draw = black] (multip_sumr) edge (outsigma);
		\path[->, draw = black] (outsigma) edge (multip_sigma);
		\path[->, draw = black] (multip_sigmar) edge (alphai);
		\path[->, draw = black] (alphai) edge (multip_alpha);
		\path[->, draw = black] (multip_alphar) edge (outsum);
		\path[->, draw = black] (outsum) edge (f);

		\path[->, draw = blue, dashed] (alpha1) edge (outsum);

		\path[->, draw = blue, dashed] (alpham) edge (outsum);


	\end{tikzpicture}
 }
	\caption{Figure for \Cref{app:reconstrct}. Output of an optimal $3$-layer neural net with sign activation reconstructed from a \lasso{} solution $\mathbf{z}^*$ using \Cref{thm:3lyr_sign_reconstruction}
 \iftoggle{long}{
 and is optimal in the \nc{} training problem. 
 }.
 \iftoggle{long}{
 $\Amat{}_{I_i}$ }
The pulse colors correspond to network operations.
 \iftoggle{long}{
 The output of the $j^{\mathrm{th}}$ neuron of the first layer is depicted before and after activation by the purple and red lines, respectively. 
 }
 \iftoggle{long}{
 The inner weight is $1$, corresponding to the first row of $\Wul{i}{1}$, and the bias is one of the data samples, corresponding to the second row of $\Wul{i}{1}$. }
\iftoggle{long}{The vertical red pulses in figure \ref{fig:3lyr_arch_sign_tikz} represent the output of the first hidden layer of a $2$-layer neural net unit.}
\iftoggle{long}{
Each red pulse represents 
\iftoggle{long}{
a vector in $\{-1,1\}^N$ with one switch, which is }
a column of the dictionary matrix $\Amat{}$ for the $2$-layer \lasso{} (\Cref{lemma:sign_thresh_a}). 
} The alternating $+,-$ represent $\Wul{i}{2} = (1,-1,1,-1,\cdots)$. 
\iftoggle{long}{ linearly combining them with  $W^{(i,2)}$ which is a vector alternating between $+1$ and $-1$.}
\iftoggle{long}{The location of the kinks (sign switches) in the red pulse of figure \ref{fig:3lyr_arch_sign_tikz} are locations of kinks (horizontal edges) in the orange pulse.}
The red and green pulses illustrate $2$ and $3$-layer dictionary features, respectively (\Cref{thrm:lasso_deep_1nueron_ReLU}, \Cref{thm:3lyr_sign}), while the other colors represent multiplication by weights and amplitudes. 
\iftoggle{long}{
with multiple switches between $-1$ and $1$ }
\iftoggle{long}{is the output of the second hidden layer of a parallel unit, which }
}
	\label{fig:3lyr_arch_sign_tikz}
\end{figure}
}
%
%

\section{Detailed results for \Cref{sec:deepsign}}\label{app:treenetsec}
Proofs 
are deferred to \Cref{appsec:1Ddatasignact}. 
\subsection{Tree network definition}\label{def:treenet}
Let $L \geq 3, m_2, \cdots, m_L \in \mathbb{N}$. 
Given $l \in \{0, \cdots, L-2\}$, let $\mathbf{u}$ be an $l$-tuple where if $l=0$, we denote $\mathbf{u} = \emptyset$ and otherwise, $\mathbf{u}=(u_1, \cdots, u_l)$ such that $ u_i \in [m_{L-i}]$ for $i \in [l]$. 
For an integer $a$, denote $\mathbf{u}\concat a$ as the concatenation $(u_1, \cdots, u_l, a)$. 
For $l \in [L-1]$, and $\mathbf{u}$ of length $l$, let $\alpha^{(\mathbf{u})}, s^{(\mathbf{u})}, {b}^{(\mathbf{u})}, \mathbf{w}^{(\mathbf{u})} \in \mathbb{R} $, except let $\mathbf{w}^{(u_1, \cdots, u_{L-1})} \in \mathbb{R}^d$. For all $\mathbf{u}$ of  length $L-1$, let $\mathbf{X}^{(u_1, \cdots, u_{L-1})} = \mathbf{x}\in \mathbb{R}^{1\times d}$. For $\mathbf{u}$ of length $l \in \{0, \cdots, L-2\}$, let $\mathbf{X}^{(\mathbf{u})} \in \mathbb{R}$ be defined by
\begin{equation}\label{eq:tree_recursion}
    \begin{aligned}
        \mathbf{X}^{(\mathbf{u})} &= \sum_{i=1}^{m_{L-l}}  \sigma_{s^{(\mathbf{u} \concat i)}} 
        \left( \mathbf{X}^{(\mathbf{u} \concat i)} \mathbf{w}^{(\mathbf{u} \concat i)} +  {b}^{(\mathbf{u} \concat i)} \right) \alpha^{(\mathbf{u}\concat i)}.
    \end{aligned}
\end{equation}
A \textit{tree neural network} is $\nn{L}{\mathbf{x}}=\xi + \mathbf{X}^{(\emptyset)} 
$. Visualizing the neural network as a tree, $\mathbf{X}^{(\emptyset)}$ is the ``root," $\mathbf{u}=(u_1, \cdots u_l)$ specifies the path from the root at level $0$ to the ${u_l}^{\text{th}}$ node (or neuron) at level $l$, $\mathbf{X}^{(\mathbf{u})}$ represents a subtree at this node, and \eqref{eq:tree_recursion} specifies how this subtree is built from its child nodes $\mathbf{X}^{(\mathbf{u}\concat i)}$. The leaves of the tree are all copies of $\mathbf{X}^{(u_1, \cdots, u_{L-1})} = \mathbf{X}$. 
Let $\mathcal{U}= \prod_{l=0}^{L-2}[m_{L-l}]$. The regularized and bias parameter sets (\Cref{sec:stdnetworks}) are $\regparams^{(i)} = \left \{ \alpha^{(\mathbf{u})}, s^{(\mathbf{u})}, \mathbf{w}^{\mathbf{(u)}}: \mathbf{u} \in \mathcal{U}, u_1 = i \right \}, \biasparams^{(i)} = \left \{ b^{(\mathbf{u})}: \mathbf{u} \in \mathcal{U}, u_1 = i \right \}$. For tree networks, let $\boldsymbol{\alpha} = \left(\alpha^{(1)}, \cdots, \alpha^{(m_L)}\right) \in \mathbb{R}^{m_L}$.  

The rest of this section includes additional results. The reconstruction defined below uses the \unscal{}ing operation (\Cref{parameterunscale}).
\begin{lemma} 
\label{thm:3lyr_sign_reconstruction}
	Consider a $3$-layer sign-activated network. Suppose $\mathbf{z}^*$ is optimal in the \lasso{} problem, and $m_L \geq ||\mathbf{z}^* ||_0$.
	\iftoggle{long}{   Let $m^* = ||\mathbf{z}^*||_0$. Suppose  the number of final layer neurons is  $m_2 \geq m^*$. }
	Let $\xi{=}0$. 
	\iftoggle{long}{for all $i,l$}
	\iftoggle{long}{ $I$ is the set of indices where $z$ is nonzero.  } 	
	\iftoggle{long}{ active unit }
 Let $\boldsymbol{\alpha}{=}\mathbf{z}^*$.
 Suppose $\Amat{}_i$ switches at $I^{(i)}_1 < I^{(i)}_2 < \cdots < I^{(i)}_{m^{(i)}} $.
Let $\Wul{i}{1}_n{=}1, \bul{i}{1}_n {=} -x_{I^{(i)}_n-1}, \Wul{i}{2}_n {=} (-1)^{n+1}$ and $\bul{i}{2} {=} -\mathbf{1} \left\{{m^{(i)}} \mathrm{\text{ }odd} \right\}$. Let all amplitude parameters be $1$. Let $\mathcal{I} {=} \{i: z_i {\neq} 0\}$. If $i \notin \mathcal{I}$, set $s^{(i,l)}, \alpha_i, \Wul{i}{l}, \bul{i}{l}$ to zero. These parameters are optimal when \unscal{}ed (\Cref{parameterunscale}).
 \iftoggle{long}{
 $\Wul{i}{1} \in \mathbb{R}^{2 \times |S_i|}, \Wul{i}{2} \in \mathbb{R}^{|S_i|+1}$ be defined as

\begin{equation*}
\begin{aligned}
\Wul{i}{1} =
   \begin{pmatrix}
				1             & \cdots & 1                 \\
				-x_{S_i(1)-1} & \cdots & -x_{S_i(|S_i|)-1}
			\end{pmatrix}, 
\Wul{i}{2} = 
\begin{cases}
    (1,-1,\cdots,1,-1,0)^T &\mbox{ if $|S_i|$ is even}\\
    (1,-1,\cdots,1,-1,-1)^T &\mbox{ if $|S_i|$ is odd}
\end{cases}.
\end{aligned}
\end{equation*}
i.e., $\Wul{i}{2}_j = (-1)^{j+1}$ for $j \in [|S_i|]$ and $\Wul{i}{2}_{|S_i+1|} = -\mathbf{1}\{|S_i| \mbox{ odd}\}$.
}

%
%
\iftoggle{long}{
\begin{equation*}
\Wul{i}{1} =
   \begin{pmatrix}
				1             & \cdots & 1                  & 1\\
				-x_{S_i(1)-1} & \cdots & -x_{S_i(|S_i|)-1}  & 0
			\end{pmatrix} \in \mathbb{R}^{2 \times |S_i|+1}, 
\Wul{i}{2} = (1,-1,\cdots,1,-1,-1)^T \in \mathbb{R}^{|S_i|+1}
\end{equation*}
}
\iftoggle{long}{
 Here, $\Wul{i}{1} \in \mathbb{R}^{2 \times m}, \Wul{i}{2} \in \mathbb{R}^{m}$ where $m=|S_i|$ if $|S_i|$ is even and $m=|S_i|+1$ if $|S_i|$ is odd.
 For  $j \in [|S_i|]$, let $S_i(j)$ be the $j^{th}$ smallest index in $S_i$, let column $j$ of $\Wul{i}{1}$ be 
$[1, -x_{S_i(j)-1}]^T$, and let $\Wul{i}{2}_{j}=(-1)^{j+1}$. Additionally, if $|S_i|$ is odd, let column $|S_i|+1$ of $\Wul{i}{1}$ be $[1,0]^T$, and let $\Wul{i}{2}_{|S_i|+1} =-1$. 
}
 \iftoggle{long}{

	Let $\alpha_i = z_i$. Let $\mathbf{a}_i \in \{-1,1\}^N$ be the $i^{th}$ column of $\Amat{}$, and $S_i = \{i \in \{2,\cdots,N-1\}: \mbox{sign}(a_i) \neq \mbox{sign}(a_{i-1}) \}$ be its switching indices in ascending order.
	\iftoggle{long}{ $S_i$ is the set of indices in increasing order at which $a_i$ switches sign. }
	Let $\mathbf{W^{(i,2)}} \in \mathbb{R}^{m_1}$ 
 \iftoggle{long}{ be a vector }
alternate between $1$ and $-1$, starting at $\mathbf{W^{(i,2)}}_1=1$ until and including index $|S_i|$, and $\mathbf{W^{(i,2)}}_j=0$ for $j > |S_i|$. Let the $j^{th}$ column of $\Wul{i}{1} \in \mathbb{R}^{2 \times m_1}$ be 
$[1, -x_{S_i(j)-1}]^T$
\iftoggle{long}{
$\Wul{i}{1}_{1,j} =1$ in the first element and let the second element be $\Wul{i}{1}_{2,j} = -x_{S_i(j)-1}$},
where $S_i(j)$ is the $j^{th}$ (smallest) index in $S_i$. Let all other columns of  $\Wul{i}{1}$ be $0$. In addition, if $|S_i|$ is odd, let column $|S_i|+1$ of $\Wul{i}{1}$ be $1$ in the first element and $0$ in the second element, and let $\mathbf{W^{(i,2)}}_{|S_i|+1}=-1$. 
}
\end{lemma}

In the next result, we only consider networks with sign activation. The number of neurons in each layer $l$ of a $2$ and $3$-layer network is denoted by $m_l'$ and $m_l$, respectively. 
\begin{corollary}\label{cor:3lyr_m1x2lyr}
	Consider a $3$-layer sign-activated network.
	\iftoggle{long}{ and dimension $D=1$ with $L=3$ layers, and $m_1$ neurons in the first hidden layer, $m_2$ neurons in the second hidden layer. }
	There exists an equivalent $2$-layer network with $m_{2}'=m_1m_3$ neurons.
	\iftoggle{long}{ and whose $l_1$ norm of the outer weights that is $m_1$ times greater}
	Let $p^*_{L,\beta}$ be the optimal value of the training problem \eqref{eq:nonconvex_generic_loss_gen_L_intro} for $L$ layers, regularization  $\beta$ and sign activation. Then, for $2$-layer nets trained with 
	$m_{2}'\geq m_1m_2$ neurons, $p^*_{L=3,\beta} \leq p^*_{L=2,\beta} \leq p^*_{L=3, m_1\beta}$.

\end{corollary}

\Cref{cor:3lyr_m1x2lyr} states that a $3$-layer net can achieve lower training loss than a $2$-layer net, but only while its regularization $\beta$ is at most $m_1$ times stronger. \iftoggle{long}{If the regularization is $m_1$ times as  $L=2$-layer regularization coefficient $\beta$ is $m_1$ times more than for $L=3$ layers, the optimal $L=3$ layer net cannot get lower training loss
	(achieve a lower $L=2$ primal objective  \eqref{equ:nn_train_signdetinput_rescale}) than the optimal $L=2$ layer network.  }
%

\subsection{Example of \nd{2} data}\label{sec:2Ddata}

The next result extends \Cref{thm:3lyr_sign} to \nd{2} data on the upper half plane. 
We consider parallel neural nets without internal bias parameters, that is, $\bul{i}{l} {\notin} \params$ (they can be thought of as set to $0$).
Proofs are located in \Cref{app_section_2d_data}.

\begin{theorem}\label{lemma:2D_upperhalfplane_signact}
\iftoggle{long}{
	Let $X \in \mathbb{R}^{N \times d}$ consist of $d$-dimensional data. Define an $L\geq 2$-layer parallel NN with sign activation, i.e., 
 $\sigma(x)=s \cdot \mathrm{sign}(x)$ and }

 Consider a \lasso{} problem whose dictionary is the \switchsetname{} $\switchmatrix{K}$, $\xi=0$, and with solution $\mathbf{z}^*$. Let $m^*= \| \mathbf{z}^* \|_0$. This \lasso{} problem 
is equivalent to the training problem for a sign-activated network without internal biases that is $2$-layer or rectangular, satisfies $m_L \geq m^*$, $m_{L-2} = K$, and is trained on \nd{2} data with unique angles in $(0,\pi)$.
%
%

	
\end{theorem}

The next result reconstructs an optimal neural net from the \lasso{} problem in \Cref{lemma:2D_upperhalfplane_signact}. The unscaling operation is used (\Cref{parameterunscale}).

\begin{lemma}\label{lemma:2Doptnet}
    Let $\mathbf{R}_{\frac{\pi}{2}} = 
\begin{pmatrix}
    0 & -1\\
    1 & 0
\end{pmatrix}$ be the counterclockwise rotation matrix by $\frac{\pi}{2}$. An optimal parameter set for the training problem in \Cref{lemma:2D_upperhalfplane_signact} when $L=2$ is the \unscal{}ed version of $\params {=} \left \{ \alpha_i{=}z_i^*, \sul{i}{1}{=}1, \Wul{i}{1} {=} 
    \mathbf{R}_{\frac{\pi}{2}}  {\left(\mathbf{x}^{(i)}\right)}^T, \xi{=}0: z^*_i {\neq} 0 \right\}$, where $\mathbf{z}^*$ is optimal in the \lasso{} problem. 
\end{lemma}

\section{Detailed results for \Cref{section_main:square_wave}}\label{sec:squarewaveapp}
Proofs in this section are deferred to \Cref{app_sec:squarewave-2lyr}.
Given a square wave of period $T$, let $k{=}\frac{N}{T}$ be the number of cycles it has. There is a \emph{critical value} $\beta_c{=}\max_{n \in [N]} |\Amat{}_n^T \mathbf{y}|$ such that when $\beta {>} \beta_c$,  $\mathbf{z}^*=\mathbf{0}$ is optimal in the \lasso{} problem \citep{efron2004least}. Let $\betafrac {=} \frac{\beta}{\beta_c}$.
\Cref{thrm:lasso_deep_1nueron_ReLU} specifies the \lasso{} problem for a $2$-layer network with sign activation. We will use the $N \times N$ dictionary matrix $\Amat$ with $\Amat_{i,n}=\sigma(x_i-x_n)$, as defined in \Cref{Amat2lyrs}. The \lasso{} solution $\mathbf{z}^* \in \mathbb{R}^N$ is unique, by \Cref{cor:sign_full_rk_unique}.

\begin{theorem}
\label{thm:no_noise_square_wave_2lyr}
Consider the \lasso{} problem for a $2$-layer net with sign activation and square wave \labelvec{} of period $T$. The critical value is $\beta_c = T$. And the solution is 
 %

	\begin{equation}\label{eq:signdet2overindices}
		\begin{aligned}
			{z}_{\frac{T}{2}i}^* & =
			\begin{cases}
                \begin{cases}
                    \frac{1}{2}\left(1-\betafrac{} \right)_+ & \mbox{ if }  i \in \{1, 2k-1\} \\
                    0 & \mbox{ else } \\
                \end{cases}
            &\mbox{ if }  \betafrac{} \geq  \frac{1}{2} \\
                \begin{cases}
                     1 -\frac{3}{2}\betafrac{}  & \mbox{ if } i \in \{1, 2k-1\} \\
                    (-1)^{i+1}\left(1-2\betafrac{}\right) & \mbox{ else }
                \end{cases}
            &\mbox { if } \betafrac{} \leq \frac{1}{2}.
		  \end{cases},
		\end{aligned}
	\end{equation}

for $i \in [2k-1]$ and $z^*_n = 0$ at all other $n \in [N]$.
\end{theorem}
\begin{corollary}\label{cor:solfor2lyrwave}
 For a square wave \labelvec{} with period $T$, there is an optimal $2$-layer neural network with sign activation specified by
   \begin{equation*}
       \begin{aligned}
            \nn{2}{x} &= 0, &\mbox { if } \betafrac{} \geq 1 \\
            \nn{2}{x} &=  
                \begin{cases}
                    -\left( 1-\betafrac\right) &\mbox{ if } x < x_{N-\frac{T}{2}} \\
                    0 &\mbox{ if } x_{N-\frac{T}{2}} \leq x < x_{\frac{T}{2}} \\
                    1-\betafrac &\mbox{ if } x \geq x_{\frac{T}{2}} 
                \end{cases}
            \quad &\mbox{ if } \frac{1}{2} \leq \betafrac{} \leq 1 \\
            \nn{2}{x} &= 
            \begin{cases}
               - \left( 1-\betafrac{}\right) &\mbox{ if } x < x_{N-\frac{T}{2}} \\
                (-1)^{i} \left(1-2\betafrac{} \right)&\mbox{ if } x_{\frac{T}{2}(i+1)} \leq x < x_{\frac{T}{2}i}, \quad i \in [2k-2] \\
                 1-\betafrac{} &\mbox{ if } x \geq x_{\frac{T}{2}} 
            \end{cases}
            \quad &\mbox{ if } \betafrac{}\leq \frac{1}{2}
    \end{aligned}
\end{equation*}
\end{corollary}

\begin{theorem}\label{thm:3layer_no_noise}
  Consider the \lasso{} problem for a $3$-layer network with sign activation and \labelvec{} a square wave of period $T$ and
   $m_3 \geq 2\frac{T}{N}-1$. Then $\beta_c = N$ and $\Amat{}_i =-\hkt$ for some $i$.
    The solution to the \lasso{} problem is $z_i^* = -\left( 1 - \betafrac\right)_+$ and $z^*_n=0$ at all other $n$. 
\end{theorem}

\begin{corollary}\label{cor:solfor3lyrwave}
 Let $x_0 = \infty$. For a square wave \labelvec{} with period $T$, there is an optimal $3$-layer neural net with sign activation specified by $\nn{3}{x} = (1-\betafrac)_+(-1)^{(i-1)}$ if $x_{\frac{T}{2}i} \leq x < x_{\frac{T}{2}(i-1)}$ for $i \in [2k-1]$, and $\nn{3}{x} = -(1-\betafrac)_+$ if $x<x_{N-\frac{T}{2}}$.
 \end{corollary}

Consider the $2$-layer network in the left plot of \Cref{fig:pulse_ex_overlap_2lyr}. When $\beta \to 0$, the network interpolates the \labelvec{} perfectly. As $\betafrac$ increases to $\frac{1}{2}$ (red dots) from $0$, the magnitude of the middle segments decrease at a faster rate than the outer segments until at $\betafrac{}=\frac{1}{2}$, the net consists of just the outer segments (magenta dots). As $\betafrac$ increases to $1$ from $\frac{1}{2}$ (blue dots), these outer segments decrease until the neural net is the zero function. 

In \Cref{fig:decision_regions}, the training data is chosen randomly from a uniform distribution on $[-100,100]$. 
Suppose we use the neural net as a binary classifier whose output is the sign of $\nn{L}{x}$, where the network is "undecided" if $\nn{L}{x}=0$. 
The red, blue and white indicate classifications of $-1$, $1$, and "undecided," respectively. For $\beta < \beta_c$, the $3$-layer net always classifies the training data accurately, but the $2$-layer net is undecided on all but the first and last interval if $\beta > \beta_c/2$. When used as a regressor, for each $\beta$, the magnitude of the $3$-layer net's prediction is the same over all samples, while the $2$-layer net is biased toward a stronger prediction on the first and last intervals. In this sense, the $3$-layer network generalizes better.

\section{Detailed results of \Cref{sec:autoreg}}\label{app:yifeictd}

 
The \textit{neural net (NN) autoregression training problem} is
\begin{equation}\label{equ:nn}
 \min_{\params \in \paramspace} \frac{1}{T}\sum_{t=1}^{T} 
 \left(  \NN{2}{x_t}-x_{t+1}\right)^2+\frac{\beta}{2}\|\regparams\|_2^2.
\end{equation} 

This model represents predictors of $x_{t+1}$ from $x_t$.
By \Cref{thrm:lasso_deep_1nueron_ReLU}, this \nc{} problem is equivalent to the convex \lasso{} problem \eqref{eq:genericlasso}
where $\Amat{}_{i,j}=|x_i-x_j|$ and $y_i=x_{i+1}$. 

We now compare solving the autoregression \eqref{equ:nn} and quantile regression \eqref{eq:quantile_reg} problems directly with $5$ trials of stochastic gradient descent (SGD) initializations versus using the \lasso{} problem \eqref{eq:genericlasso}. We also compare against the baseline linear method $\nn{}{x}=ax+b$ (AR1+bias), where we include an additional bias term $b$.
%
We test upon real financial data for bitcoin price, including minutely bitcoin (BTC) price (BTC-2017min) and hourly BTC price (BTC-hourly). We consider the training problem on $\tau$-quantile regression (\Cref{app:yifeictd}) with $\tau=0.3$ and $\tau=0.7$. For each dataset, we first choose $T$ data points as a training set and the consecutive $T$ data points as a test set. The numerical results are presented in \Cref{fig:BTC_main_paper}. 
We observe that cvxNN provides a consistent lower bound on the training loss and demonstrates strong generalization properties, compared to large fluctuation in the loss curves of NN. More results can be found in \Cref{sec:App_autoreg}.

We first build a network $\NN{2}{x}$ \eqref{eq:nn_autoreg} with $m$ known, or \textit{planted} neurons. We use this network to generate training samples $x_1,\dots,x_{T+1}$ based on \eqref{equ:ar} with $f(x;\params)= \NN{2}{x} $ 
where $x_1\sim\mathcal{N}(0,\sigma^2)$. 
Using the same model $\NN{2}{x}$, we also generate test samples $x_1^\text{test},\dots,x_{T+1}^\text{test}$ in an analogous way. We use $T=1000$ time samples. Then, we try to recover the planted neurons based on only the training samples by solving the NN AR/QR training problems.

\begin{figure}[!htb] 
\centering
\begin{minipage}[t]{0.9\textwidth}
\centering
\includegraphics[width=\linewidth]{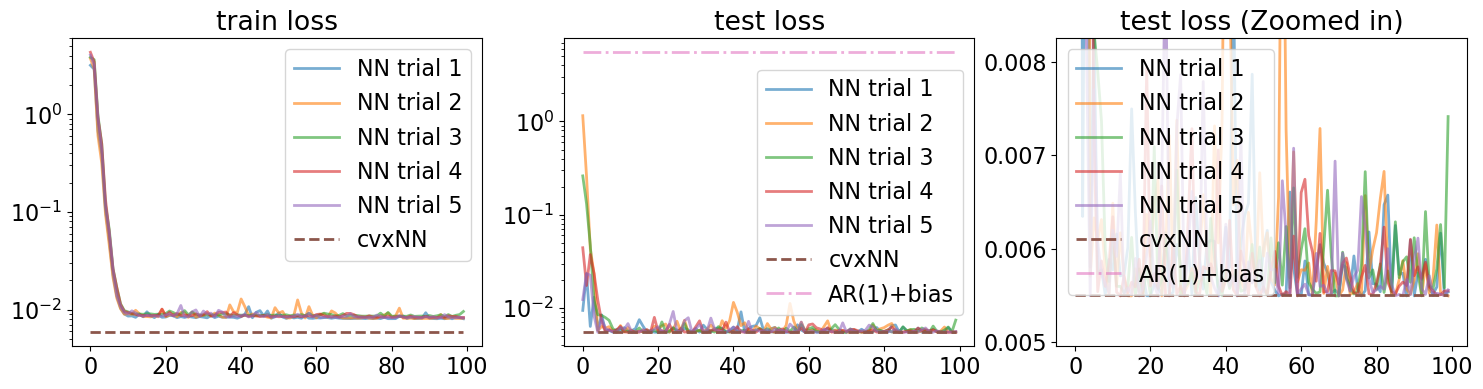}
\caption*{$m=2$}
\end{minipage}
\begin{minipage}[t]{0.9\textwidth}
\centering
\includegraphics[width=\linewidth]{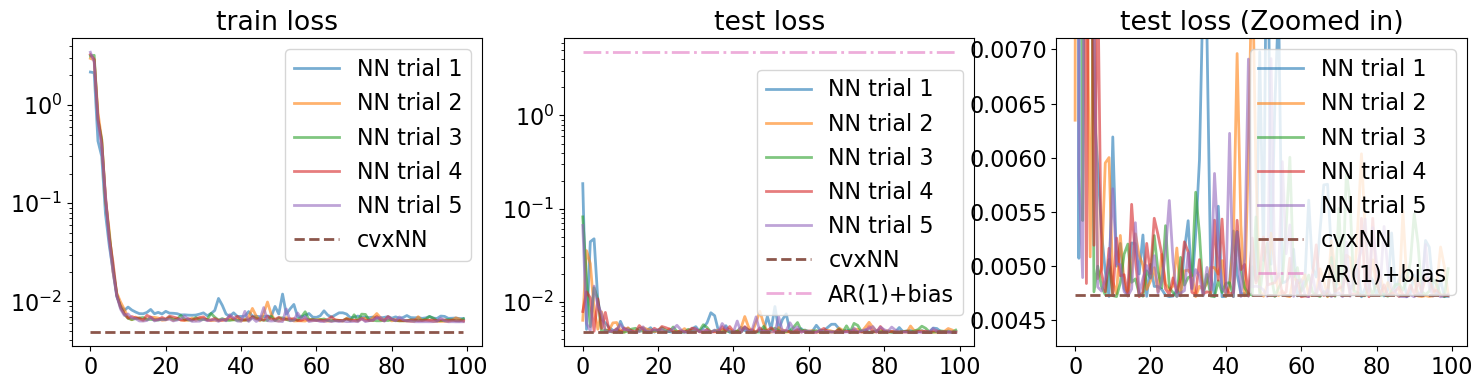}
\caption*{$m=5$.}
\end{minipage}
\begin{minipage}[t]{0.9\textwidth}
\centering
\includegraphics[width=\linewidth]{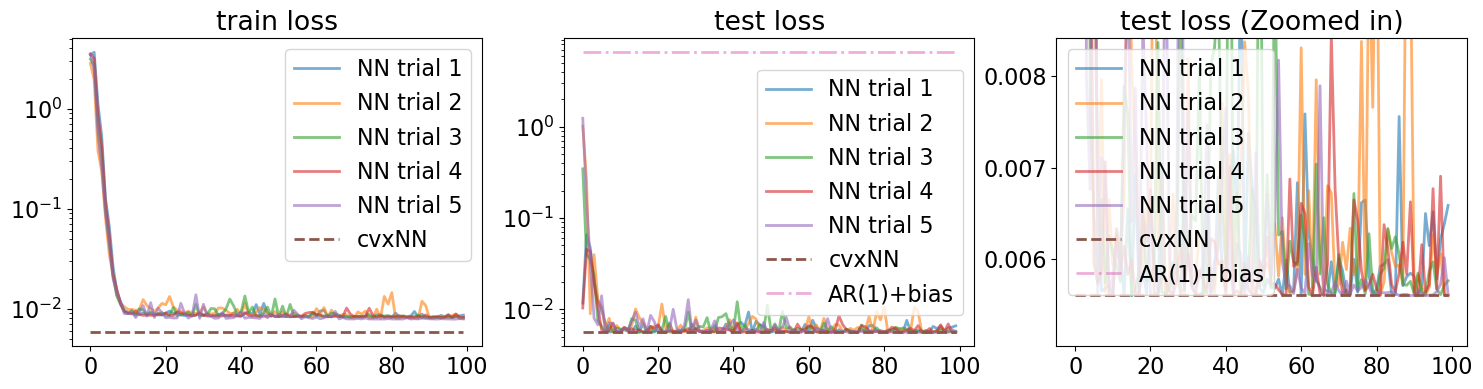}
\caption*{$m=10$.}
\end{minipage}
\caption{Planted data. $\sigma^2=1$. }\label[figure]{fig:planted}
\end{figure}

\begin{figure}[!htb] 
\centering
\begin{minipage}[t]{0.8\textwidth}
\centering
\includegraphics[width=\linewidth]{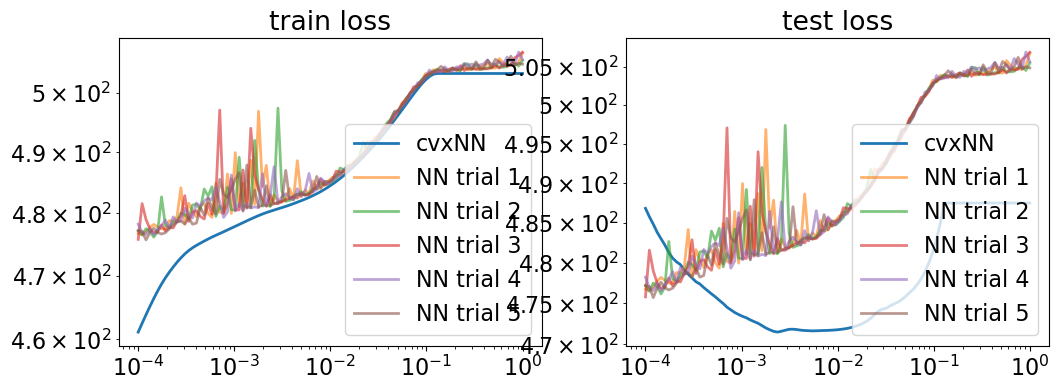}
\end{minipage}
\caption{The regularization path. Here, $\sigma^2=1$, $m=5$.}
\label[figure]{fig:AR_random_reg_path}
\end{figure}

In \Cref{fig:planted}, we present experiments based on the selection of $m$ planted neurons and noise level $\sigma^2$. More results can be found in \Cref{sec:App_autoreg}. The neural net trained with \lasso{} is labeled cvxNN, which we observe has lower training loss. This appears to occur because different trials of NN (neural net trained directly without \lasso{}) get stuck into local minima. The global optimum that cvxNN reaches also enjoys effective generalization properties, as seen by the test loss. The \textit{regularization path} is the optimal neural net's performance loss as a function of the regularization coefficient $\beta$.  \Cref{fig:AR_random_reg_path} plots the regularization path for $\sigma^2=1$ and $m=5$. The regularization path taken by cvxNN is smoother than NN, and can therefore be found more precisely and robustly by using the \lasso{} problem.


\iftoggle{long}{ 
\begin{figure}[t]
\centering
\begin{minipage}[t]{0.9\textwidth}
\centering
\captionsetup{type=subfigure}
    \includegraphics[width=\linewidth]{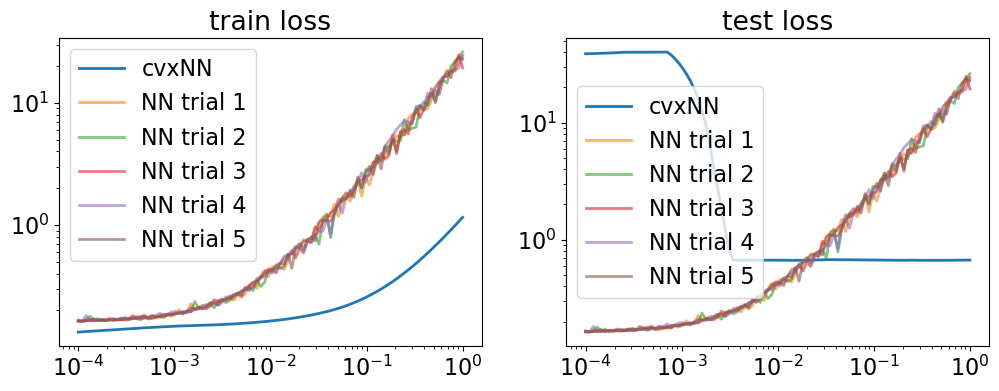}
\caption*{Without skip connection.}
\end{minipage}
\begin{minipage}[t]{0.9\textwidth}
\centering
\captionsetup{type=subfigure}
    \includegraphics[width=\linewidth]{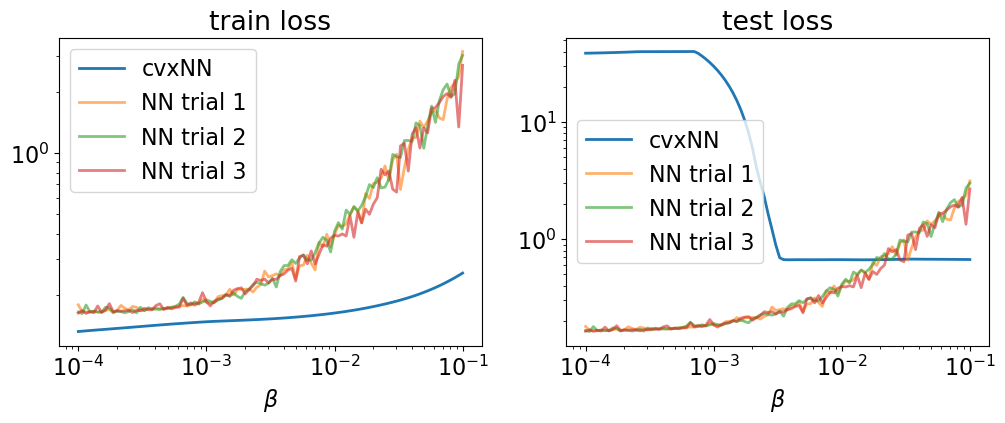}
\caption*{With skip connection.}
\end{minipage}
\caption{BTC-2017min. Regularization path.}
\end{figure}
}

\section{\secnameSolSet}\label{sec:solsetslassovsnn}

\input{sections/solution_set}

\section{\secnameMinNorm{}}\label{sec:minnorm}
One of the insights that the \lasso{} formulation provides is that under minimal regularization, certain neural nets perfectly interpolate the data.  

\begin{corollary}\label[corollary]{Cor:beta->0_overfit}
	For the ReLU, absolute value, sign, and threshold networks with $L=2$ layers, and sign-activated deeper networks,
 \iftoggle{long}{and $\mathcal{L}_y(\mathbf{z})=||\mathbf{z}-\mathbf{y}||^2_2$ }
 if $m_{L} \geq m^*$, then  $\nn{L}{\mathbf{X}} \to \mathbf{y}$ as $\beta \to 0$.
\end{corollary} 

Proofs in this section are deferred to \Cref{app:minNorm}. In \Cref{Cor:beta->0_overfit}, $m^*$ depends on $L$ and the activation and is defined in \Cref{thrm:lasso_deep_1nueron_ReLU} and \Cref{thm:3lyr_sign}.
The \lasso{} equivalence and reconstruction also shed light on optimal neural network structure as regularization decreases.
The \textit{minimum ($l_1$) norm subject to interpolation} version of the \lasso{} problem is 

\begin{equation}\label{rm_gen_lasso_nobeta}
		\min_\mathbf{z, \xi} \|\mathbf{z}\|_1 \mbox { s.t. } \Amat{} \mathbf{z}+\xi \mathbf{1} = \mathbf{y}.
	\end{equation}
 
Loosely speaking, as $\beta\to0$, if $\Amat{}$ has full column rank, the \lasso{} problem \eqref{eq:genericlasso} "approaches" the minimum norm problem \eqref{rm_gen_lasso_nobeta}, where $\xi = 0$ for sign and threshold activations.
%
The rest of this section describes the solution sets of \eqref{rm_gen_lasso_nobeta} for certain networks. 
\iftoggle{long}{
Under the assumption that the number of neurons is sufficiently large ($m \geq m^*$), \eqref{rm_gen_lasso_nobeta} is equivalent to 

Consider the limiting case with $\beta\to0$ for \eqref{eq:nonconvex_generic_loss_2lyr}. Then, the problem becomes the following minimal norm problem:
\begin{align} \label{eq:nonconvex_generic_loss_2lyr_min_norm}
 \min_{\boldsymbol{\alpha}, \mathbf{W_0}, \mathbf{b}, \xi} \frac{1}{2} \| \regparams \|_2^2,\text{ s.t. } \sigma_s( \mathbf{X} \mathbf{W_0}  + b)\boldsymbol{\alpha} + \xi = \mathbf{y},
\end{align} 
under the assumption that the number of neurons is sufficiently large.
The convex program for \eqref{eq:nonconvex_generic_loss_2lyr_min_norm} becomes
\begin{equation}\label{lasso:min}
    \min_{\mfz \in \mathbb{R}^N} \|\mfz\|_1, \text{ s.t. } \Amat{}\mfz+\xi\bone=\mathbf{y}.
\end{equation}
} 
%


\iftoggle{long}{
For each optimal $\mathbf{z}^*$ in  \Cref{rm_gen_lasso_nobeta},  $\xi^* = \left(\mathbf{y}-\mathbf{1} \mathbf{1}^T \mathbf{y}\right) - \left( \Amat{} -\mathbf{1}\mathbf{1}^T \Amat{} \right) \mathbf{z}^*$.
}


\begin{proposition}\label[proposition]{prop:lasso_abs_solset}
    Let $L=2$. Suppose $\sigma$ is the absolute value activation. Let $\mathbf{z}^*$ be a solution to 
    \eqref{rm_gen_lasso_nobeta}. 
    Then, we have $z^*_1z^*_n\leq0$. Moreover, the entire solution set of 
    \eqref{rm_gen_lasso_nobeta}
    for $\mathbf{z}^*$ is 
    \begin{equation}\label{lasso:min_sol}
       \left \{\mathbf{z}^*+t\textup{ sign}(z^*_1)(1,0,\cdots,0,1)^T
       \Big |-|z^*_1|\leq t\leq |z^*_N| \right\}.
    \end{equation}
\end{proposition}

\begin{proposition}
    \label[proposition]{cor:sign_full_rk_unique}
	 For a $2$-layer network with sign activation and
 \iftoggle{long}{ $\mathcal{L}_\mathbf{y}(\mathbf{z}) = ||\mathbf{z}-\mathbf{y}||^2_2$ and}
$\beta \geq 0$, the \lasso{} problem \eqref{eq:genericlasso}
	\iftoggle{long}{(with simplifications \ref{symm_simp}, \ref{bounded_simp})}
	has a unique solution. Furthermore, the minimum norm solution in \eqref{rm_gen_lasso_nobeta}
 is $\mathbf{z}^* = \Amat{}^{-1}\mathbf{y}$  .
	\iftoggle{long}{ where $\Amat{} = \mathbf{A_+}$. 
	}
\end{proposition}

Given an optimal bias term $\xi^*$, if $\mathbf{A}$ is invertible, then $\mathbf{z}^* = \Amat{}^{-1}(\mathbf{y} - \xi^* \mathbf{1})$ is optimal in \eqref{rm_gen_lasso_nobeta}. \Cref{section:dic_matrices} explicitly finds $\Amat{}^{-1}$ for some activation functions. The structure of $\Amat{}^{-1}$ suggests the behavior of neural networks under minimal regularization: sign-activated neural networks act as difference detectors,   
while neural networks with absolute value activation, whose subgradient is the sign activation, act as a second-order difference detectors (see \Cref{rk:diff_detector}). The next result shows that threshold-activated neural networks are also difference detectors, but for the special case of positive, nonincreasing $y_n$. An example of such data is cumulative revenue, e.g. $y_n = \sum_{i=1}^n r_i$ where $r_i$ is the revenue in dollars earned on day $i$.

\begin{proposition}
\label[proposition]{lemma:thresh_informal_sol}
     Let $L=2$. Suppose $\sigma$ is threshold activation and $y_1 \geq \cdots \geq y_N \geq 0$. Then 
     \begin{equation*}
        z^*_n= 
         \begin{cases}
          y_n - y_{n-1} &\mbox{ if } n \leq N-1 \\
          y_N &\mbox{ if } n = N \\
          0  &\mbox{ else }
         \end{cases}
     \end{equation*}
     is the unique solution to the minimum norm problem \eqref{rm_gen_lasso_nobeta}.
         
\end{proposition}

The next result gives a lower bound on the optimal value of the minimum weight problem for ReLU networks. If we can find $\mathbf{z}$ with a $l_1$-norm that meets the lower bound and a $\xi$ such that $\Amat\mathbf{z}+\xi\mathbf{1}=\mathbf{y}$, then we know $\mathbf{z}, \xi$ is optimal. 
In this section, for $n \in [N-1]$, let $\slope_n = \frac{y_{n}-y_{n+1}}{x_{n}-x_{n+1}}$ be the slope between the $n^{th}$ and ${n+1}^{th}$ data points. Let $\slope_{N}=0$.

\begin{lemma}\label{lemma:minNormCert}
     The optimal value $\|\mathbf{z}^*\|_1 $ of the minimum norm problem \eqref{rm_gen_lasso_nobeta} for \deepnarrow{} networks with ReLU or absolute value activation is at least $\max_{n \in [N-1]} |\slope_n|$.
\end{lemma}

In the special case of $2$-layer networks, the next result gives a solution to the minimum weight problem. For $i \in [N]$, let ${\zposscalar}_i$ and $ {\znegscalar}_i$ be the \lasso{} variable corresponding to the features $\ReLU{+}{x_i}$ and $\ReLU{-}{x_i}$, respectively. In other words, $\zpos$ corresponds to $\Apos$, and $\zneg$ corresponds to $\Aneg$ as defined in \Cref{Amat2lyrs}.

\begin{lemma}\label[lemma]{lemma:minnormL2ReLUsol}
The optimal value of the minimum norm problem \eqref{rm_gen_lasso_nobeta} for $L{=}2$ and ReLU activation is $\|\mathbf{z}^*\|_1{=} \sum_{n=1}^{N-1}|\slope_n{-}\slope_{n+1}|$. An optimal solution is $\zposscalar_{n+1} {=}  \slope_{n}{-}\slope_{n+1} $ for $n {\in} [N-1]$, $\zneg{=}\mathbf{0}$, and $\xi{=}y_N$. 
\end{lemma}


\begin{lemma}\label{lemma:minNormCert2neurons}
    For a $3$-layer \symmeterized{}  ReLU network and training data as shown in \Cref{fig:absval3lyrnet}, the optimal value $\|\mathbf{z}^*\|_1$ of the minimum norm problem \eqref{rm_gen_lasso_nobeta} is at least $1$.
\end{lemma}

\Cref{lemma:minNormCert2neurons} can be generalized to more complex sets of training data.

\section{Numerical results}\label{sec:num_results}
Simulations support our theoretical results. 

In \Cref{fig:absval3lyrnet}, a \deepnarrow{}, absolute value network is trained. In \Cref{fig:ReLU3lyrnet}, a $3$-layer \symmeterized{} ReLU network is trained. Both neural nets have standard architecture, $m_L{=}100$, and are trained with \adam{} with the \nc{} problem. The learning rate is $5(10^{-3})$, weight decay is $10^{-4}$, and $\beta {=} 10^{-7} {\approx} 0$. When $\beta{\to}0$, the \lasso{} problem approaches the  minimum norm problem \eqref{rm_gen_lasso_nobeta}. For each $L {\in} {3,4,5}$, the neural net reconstructed from the single feature in the left plot with corresponding \lasso{} parameter $z^*_i=1$ and $\xi^*=0$ is optimal in the minimum norm problem \eqref{rm_gen_lasso_nobeta} by \Cref{lemma:minNormCert} and \Cref{lemma:minNormCert2neurons}. This network is used to initialize a subset of the neurons in the \nc{} training. All other weights are initialized randomly according to Pytorch defaults. 
The figures show that the networks trained with the \nc{} problem closely match the \lasso{} solutions. The networks exhibit \kink{}s at data points and their reflections not in the training data. The standard architecture in the \nc{} model shows the applicability of the \lasso{} formulation. 
SGD gives similar results as \adam{}. 

\begin{figure}
\centering
\begin{minipage}[t]{0.45\textwidth}
\centering
    \includegraphics[width=\linewidth]{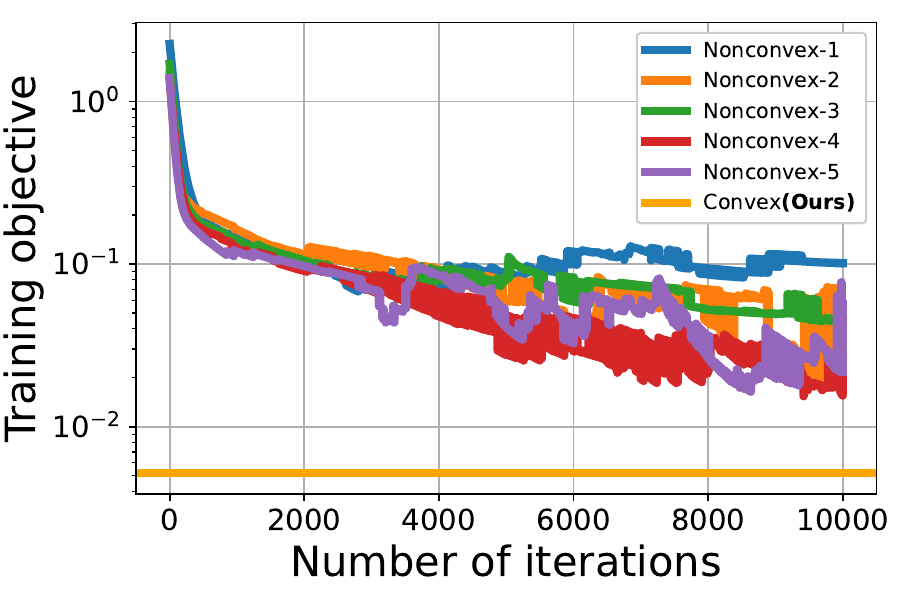}
\caption{Training objective 
}
\end{minipage}
\begin{minipage}[t]{0.45\textwidth}
\centering
    \includegraphics[width=\linewidth]{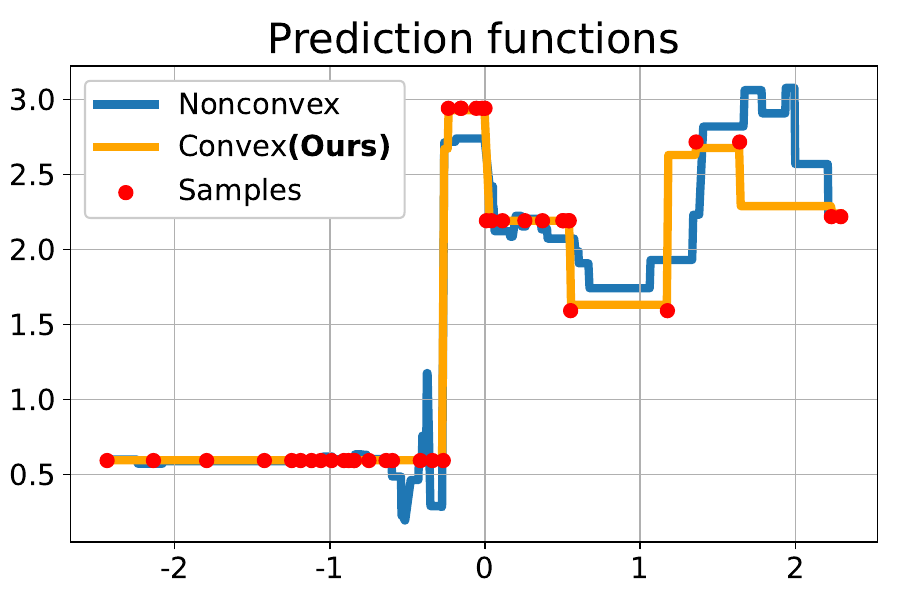}
\caption{Function fit\centering}
\end{minipage}
\caption{Figure for \Cref{sec:num_results}. Training objective (left) and function fit (right) for a neural net using the convex \lasso{} problem versus the \nc{} training problem using STE.} \label{fig:1d_func_approx}
\end{figure}

In addition to training ReLU networks under minimal $\beta$, we  train neural networks with threshold activations and larger $\beta$. 
 We label $N=40$ \nd{1} data samples from an i.i.d. distribution $x \in \mathcal{N}(0,1)$ with a Bernoulli random variable. We use $\beta {=} 10^{-3}$ to train a $2$-layer neural network with threshold activation using the \lasso{} problem \eqref{eq:genericlasso} and a \nc{} training approach based on the Straight Through Estimator (STE) \citep{Bengio2013EstimatingOP}. As illustrated in \Cref{fig:1d_func_approx}, the convex training approach achieves significantly lower objective value than all of the \nc{} trials with different seeds for initialization. \Cref{fig:1d_func_approx} also plots the predictions of the models. We observe that the \nc{} training approach fits the data samples exactly on certain intervals but provides a poor overall function fitting, whereas our convex models yields a more reasonable piecewise linear fit. In particular, the neural net trained with the \nc{} problem fits the data in \Cref{fig:1d_func_approx} poorly compared to  \Cref{fig:ReLU3lyrnet} and \Cref{fig:absval3lyrnet}. This may occur because in \Cref{fig:1d_func_approx}, the data set is larger and more complex, and STE training is used because of the threshold activation, and $\beta$ is larger instead of being close to zero.  


\iftoggle{todos}{\textcolor{red}{todo:place figures in different sections?}}

\clearpage
\newpage

\section*{Supplementary Material}

\section*{Definitions and preliminaries}

\subsection{Activation function} \label{app_section:act_func}

\iftoggle{long}{We also consider the threshold activation $\sigma_s(x) := s \mathbf{1}\{x \geq 0\}$ and the sign activation, $ \sigma_s(x):= s \cdot \mathrm{sign}(x)$
and the amplitude $s\in \mathbb{R}$ is a trainable parameter for each neuron $\sigma_{s^{(i,l)}}(x^{(i,l-1)}W^{(i,l)})$.}

A function $f$ is  \textit{bounded} if there is $M \geq 0$ with $|f(x)| \leq M$ for all $x$.
If $\sigma(x)$ is piecewise linear around zero, $\sigma(x)$ is bounded if and only if $\negslope=\posslope=0$, e.g. $\sigma(x)$ is a threshold or sign activation.
We call $f$ \textit{symmetric} if it is an even or odd function, for example absolute value.
\iftoggle{long}{
$f(x)=f(-x)$ for all $x$ or if $f(x)=-f(-x)$ for all $x$.
In other words,  $f$ is symmetric if it is an even or odd function. 
For any constant $a \in \mathbb{R}$, the scaled absolute value activation, $\sigma(x)=a|x|$, and the linear activation $\sigma(x)=ax$ are symmetric. }
The activation $\sigma(x)$ is defined to be \textit{homogeneous} if for any $a \geq 0$, $\sigma(ax)=a\sigma(x)$.
Homogeneous activations include ReLU, leaky ReLU, and absolute value, and don't have amplitude parameters.
\iftoggle{long}{For homogeneous activations, all the activations are the same throughout the network, so $\sigma$ is written without a subscript.}
%
%
We say $\sigma(x)$ is \textit{sign-determined }if 
its value depends only on the sign of its input and not its magnitude. Threshold and sign activations are sign-determined. 
\iftoggle{long}{Unless $s$ is otherwise understood from the context, we make the amplitude $s$ explicit by denoting the activation as $\sigma_s(z)$. For sign-determined activations, the amplitude parameter $s \Theta$ is trained and included in the regularization penalty.
}


\iftoggle{long}{
\section{Definitions}
The Fenchel conjugate $f^*$ of a real-valued function $f$ is defined as

\begin{equation}\label{def:conjug}
	f^*(x)=\max_z \{ z^T x - f(x) \}.
\end{equation}

For the sign activation $\sigma(x) = s \cdot \mbox{sign}(x)$, we define
\begin{equation}
	\textrm{sign}(x):=
	\begin{cases}
		1 \mbox{ if } x \geq 0 \\
		-1 \mbox{ else}.
	\end{cases}
\end{equation}
Note that we have $\textrm{sign}(0)=1$.
}

%

%

%
\iftoggle{long}{In other words, whenever $\mathcal{L}_{\mathbf{y}}(f_L(\mathbf{X};\theta))$ appears, the change of variables \eqref{eq:change_var_ext_bias} will have occurred, but $f_L(\mathbf{X};\theta)$ by itself will include $\xi$ and not have undergone the change of variables.}

\subsection{Effective depth}\label{appsec:eff_depth}

\begin{remark}\label{remk:signdet_inweights0} 
Plugging \eqref{eq:parallelnn_rec} into itself shows that in parallel network, for $l {\in} [L{-}2]$, 
     \begin{equation*}
        \Xul{i}{l+2}{=}  \sigma_{\sul{i}{l+1}} \left(  \sigma \left( \Xul{i}{l}\Wul{i}{l+1}{+}\bul{i}{l} \right) \sul{i}{l}\Wul{i}{l+1}{+} \bul{i}{l+1} \right).
     \end{equation*}
Similarly, plugging in \eqref{eq:tree_recursion} into itself shows that in a tree network, for $0 \leq l \leq L-3$,
     \begin{equation*}
     \scalemath{0.94}{
\displaystyle \mathbf{X}^{(\mathbf{u})} {=}\sum_{i=1}^{m_{L-l}} \alpha^{(\mathbf{u} \concat i)} \sigma_{s^{(\mathbf{u}\concat i)}} \left( {b}^{(\mathbf{u}\concat i)} {+} \sum_{j=1}^{m_{L-l-1}} \alpha^{(\mathbf{u}\concat i \concat j)} \sigma \left( \mathbf{X}^{(\mathbf{u}\concat i\concat j)} \mathbf{w}^{(\mathbf{u}\concat i\concat j)} {+} {b}^{(\mathbf{u}\concat i\concat j)} \right) s^{(\mathbf{u}\concat i\concat j)} \mathbf{w}^{(\mathbf{u}\concat i)}  \right).
}
     \end{equation*}
The \emph{inner parameters} of a parallel network are $\sul{i}{l}$ for $l {\leq} L-2$ and $\Wul{i}{l}$ for $l {\leq} L-1$. In a tree network, they are $\left(\mathbf{s}^{\mathbf{u}\concat 1}, \cdots \mathbf{s}^{\mathbf{u}\concat m_{L-l}} \right), \left(\alpha^{\mathbf{u}\concat 1}, {\cdots} \alpha^{\mathbf{u}\concat m_{L-l}} \right) $ for $\mathbf{u}$ of positive length, and $\left(\mathbf{w}^{\mathbf{u}\concat 1}, {\cdots} \mathbf{w}^{\mathbf{u}\concat m_{L-l}} \right)$ for $\mathbf{u}$ of any length. 
     Suppose $\sigma$ is sign-determined. 
    Then $\nn{L}{\mathbf{X}}$ is invariant to the value of the inner parameters, so they would be driven to $0$ by weight regularization. We define the minimum value in \eqref{eq:nonconvex_generic_loss_gen_L_intro} as an infimum which is approached as the 
    norms of the inner parameters approach $0$. Therefore the inner parameters are not regularized (the effective depth is $2$), and we optimize for their directions rather than their magnitudes.
\end{remark}

\iftoggle{long}{
\begin{remark}\label[remark]{remk:signdet_inweights0}\iftoggle{long}{\textit{Training problem for parallel networks with sign determined activations:}}
     For a parallel network, by plugging \eqref{eq:parallelnn_rec} into itself, for $1 \leq l \leq L-2$, $\Xul{i}{l+2}= \\ \sigma_{\sul{i}{l+1}} \left(  \sigma \left( \Xul{i}{l}\Wul{i}{l+1}+\bul{i}{l} \right) \sul{i}{l}\Wul{i}{l+1}+ \bul{i}{l+1} \right) $.
    Let the \emph{inner parameters} be the weights $\sul{i}{l}$ for $l \leq L-2$ and  $\Wul{i}{l}$ for $l \leq L-1$. Suppose $\sigma$ is sign-determined. 
    Since the inner parameters do not affect $\nn{L}{\mathbf{X}}$, they will be driven to zero by regularization in \eqref{eq:nonconvex_generic_loss_gen_L_intro}. We define the minimum value in \eqref{eq:nonconvex_generic_loss_gen_L_intro} as an infimum which is approached as $\|\sul{i}{l}\|_{\effL{}} \to 0$ and  $\|\Wul{i}{l}\|_{\effL{}} \to 0$. Therefore it suffices to not regularize the inner parameters, and instead fix their $l_{\effL}$-norms and optimize for $\sul{i}{l}/\|\sul{i}{l}\|_{\effL{}}, \Wul{i}{l}/\|\Wul{i}{l}\|_{\effL{}} $.
\end{remark}

\begin{remark}\label[remark]{remk:signdet_inweights0tree}
For a tree network, by plugging \eqref{eq:tree_recursion} into itself for $0 \leq l \leq L-3$,  $ \mathbf{X}^{(\mathbf{u})} = \\
\displaystyle \sum_{i=1}^{m_{L-l}} \alpha^{(\mathbf{u} \concat i)} \sigma_{s^{(\mathbf{u}\concat i)}} \left( {b}^{(\mathbf{u}\concat i)} + \sum_{j=1}^{m_{L-l-1}} \alpha^{(\mathbf{u}\concat i \concat j)} \sigma \left( \mathbf{X}^{(\mathbf{u}\concat i\concat j)} \mathbf{w}^{(\mathbf{u}\concat i\concat j)} + {b}^{(\mathbf{u}\concat i\concat j)} \right) s^{(\mathbf{u}\concat i\concat j)} \mathbf{w}^{(\mathbf{u}\concat i)}  \right)$. Suppose $\sigma$ is sign-determined. The weight penalty will drive $\left \|\left(\mathbf{s}^{\mathbf{u}\concat 1}, \cdots \mathbf{s}^{\mathbf{u}\concat m_{L-l}} \right) \right \|_{\effL{}}, \left\|\left(\alpha^{\mathbf{u}\concat 1}, \cdots \alpha^{\mathbf{u}\concat m_{L-l}} \right) \right \|_{\effL{}}  \to 0$ where $\mathbf{u}$ has positive length, and $\left\|\left(\mathbf{w}^{\mathbf{u}\concat 1}, \cdots \mathbf{w}^{\mathbf{u}\concat m_{L-l}} \right) \right \|_{\effL{}}  \to 0$. So it suffices to not regularize these parameters.

     \iftoggle{long}{ 
     For each layer $l \in [L]$, and each unit $i \in [m_l]$, the subtree network $f^{(l-1)}(\mathbf{X}; \theta^{(l-1,i)})=\mathbf{1}w^{(l)}_{m_l + 1}+ \sum_{j=1}^{m_{l-2}} \sigma_{s^{(l-2,i,j)}} \left(f^{(l-2)}(\mathbf{X}; \theta^{(l-2,i,j)}) \right) w^{(l-1)}_{i,j}
	$ is invariant to the total magnitude of the weight terms $\sum_{j=1}^{m_{l-2}} ||\mathbf{w^{(l-1,i,j)}}||^2_2$ and to the magnitude of the sign parameters $|s^{(l-2,i,j)}|, j \in [m_{l-2}]$ in $\theta^{(l-2,i,j)}$. 
	So the regularization penalty will send the magnitudes of the non-bias weights and sign parameters to zero, making the regularization also zero.}
    
\end{remark}

}



\section*{Parallel and tree networks with data dimension $d \geq 1$}\\

For convenience, we will omit $\xi$ when writing $\nn{L}{\mathbf{X}}$. This does not change the training problem because we can write \eqref{eq:nonconvex_generic_loss_gen_L_intro} as 
$\displaystyle
	\min_{\params{-}\{\xi\}} \frac{\tilde{\beta}}{\effL} \|\regparams\|_2^2 {+}  \min_{\xi} \left \{ \mathcal{L}_{\mathbf{y}} \left(f_L(X){-}\xi \mathbf{1}{+}\xi \mathbf{1} \right) \right \}
$
and apply the change of variables/functions $       
        \theta' {=} \theta {-} \{\xi\},                   
		f_L(\mathbf{X};\theta)' {=}  f_L(\mathbf{X};\theta){-} \xi \mathbf{1}$ and $       
		\mathcal{L}_{\mathbf{y}}(\mathbf{z})'{=}  \min_{\xi} L_y(\mathbf{z}{+}\xi \mathbf{1} ). 
 $ The loss function absorbs $\xi$ while preserving its convexity.
%
We begin by assuming the data is $d$-dimensional and consider the general \textit{training problem} 

\begin{equation*}
    \min_{\params \in \paramspace} \mathcal{L}_{\mathbf{y}}\left(\nn{L}{\mathbf{X}}\right)+ \frac{\beta}{\effL}\regfunc(\params) 
\end{equation*}
%
%
with a general regularization of the form 
$\regfunc(\params) {=}  \sum_{i=1}^{m_L}  \sum_{\params^{(i,l)} \in \params^{(i)}_w}\left( \regfunc^{(i,l)}\left(\params^{(i,l)}\right) \right)^{\effL}$,  where $\params^{(i,l)}$ is a parameter such as
$\Wul{i}{l}$ and $r^{(i,l)}$ is a regularization function such as $\regfunc^{(i,l)}\left( \Wul{i}{l} \right) {=}  \left \|\Wul{i}{l} \right\|_{p}$. Assume $r^{(i,l)}$ is nonnegative and positively homogeneous, i.e., for any $\alpha\geq0$, $\regfunc^{(i,l)}(\alpha \mathbf{W}) {=}\\ \alpha \regfunc^{(i,l)}(\mathbf{W}) {\geq} 0$.
The training problem \eqref{eq:nonconvex_generic_loss_gen_L_intro} is a special case of the general training problem.

\begin{definition}\label{def:simp_signdefNN}
 A \emph{simplified} neural network with sign-determined activation is 

\begin{equation}\label{eq:parallelrescaled}
    \begin{aligned}
        \nn{L}{\mathbf{x}} &= \xi + \sum_{i=1}^{m_{L}} \Xul{i}{L} \alpha_i s^{(i,L-1)}  \\
        \Xul{i}{l+1} &= \sigma \left( \Xul{i}{l}\Wul{i}{l}+ \bul{i}{l} \right), l \in [L-1] \\
        \regparams^{(i)} &= \left\{\alpha_i, s^{(i,L-1)} \right\}, \biasparams^{(i)} = \left \{\bul{i}{l}, \Wul{i}{l}: l \in [L-1] \right \}  \text{ for } i \in [m_{L}] \\ 
    \end{aligned}
\end{equation}

for a parallel network, and

\begin{equation}
     \begin{aligned}
     \nn{L}{\mathbf{x}} &= \xi + \sum_{i=1}^{m_L}\sigma \left(\mathbf{X}^{(i)}+ b^{(i)}\mathbf{1} \right)  \alpha_i s^{(i)} \\
        \mathbf{X}^{(\mathbf{u})} &=
        \begin{cases}
             \sum_{i=1}^{m_{L-l}} \alpha^{(\mathbf{u}\concat i)} \sigma \left( \mathbf{X}^{(\mathbf{u}\concat i)} +  {b}^{(\mathbf{u}\concat i)}\right) \mbox { if } 1 \leq l \leq L-3 \\
             \sum_{i=1}^{m_{L-l}} \alpha^{(\mathbf{u}\concat i)} \sigma \left( \mathbf{X}\mathbf{w}^{(\mathbf{u}\concat i)} +  {b}^{(\mathbf{u}\concat i)} \right) \mbox { if } l = L-2 \\
        \end{cases} \\
        \regparams^{(i)} &= \left\{\alpha^{(i)}, \mathbf{s}^{(i)}\right\} 
        \biasparams^{(i)} = \left\{\alpha^{(\mathbf{u})}, b^{(\mathbf{u})}: \mathbf{u} \in \mathcal{U}, u_1 = i \right\} \mbox{ for } i \in [m_L]
     \end{aligned}
 \end{equation}
where $\mathbf{u}$ has positive length for $\alpha^{(\mathbf{u})} \in \biasparams^{(i)}$, for a tree network.
\end{definition}

In other words, a simplified network has amplitude parameters only in the last layer.

\begin{lemma}\label[lemma]{lemma:rescale_equiv}
    The simplified neural network is equivalent to the original neural network.
\end{lemma}

\begin{proof}
    By \Cref{remk:signdet_inweights0}, 
    it suffices to only regularize the outermost $\effL{=}2$ layers. For parallel networks, apply a change of variables ${\Wul{i}{l}}'= \sul{i}{l-1}\Wul{i}{l}$ for $2 {\leq} l {\leq} L{-}1$. This removes $\sul{i}{l}$ from $\params$ for $l {\leq} L{-}2$. Similarly for tree networks, for $\mathbf{u}$ of length $1 \leq l \leq L{-}2$ and $i \in [m_{L{-}l{-}1}]$, let ${\alpha^{(\mathbf{u}\concat i)}}'{=}\alpha^{(\mathbf{u}\concat i)} 
 s^{(\mathbf{u}\concat i)} \mathbf{w}^{(\mathbf{u})}$ (note $\mathbf{w}^{(\mathbf{u})} {\in} \mathbb{R} $). This removes $\mathbf{w}^{(\mathbf{u})}$ and $  s^{(\mathbf{u}\concat i)}$ from $\params$. 
\end{proof}

 Henceforth, tree networks are assumed to have sign-determined activation and sign-determined networks are assumed to be simplified.
\newcommand{\efflayers}{\mathcal{D}}
Let $\efflayers=\{L{-}\effL{+}1,{\cdots},L\}$.
By \Cref{lemma:rescale_equiv}, the  training problem's regularization is 
\begin{equation}\label{eq:reg_as_sum}
    \regfunc(\params) {=}  \sum_{i{=}1}^{m_L}  \sum_{l{\in}\efflayers}\left( \regfunc^{(i,l)}\left(\params_w^{(i,l)}\right) \right)^{\effL}
\end{equation}
where $\params^{(i,L)} {=} \alpha_i$ and $ \params^{(i,l)} {=} \Wul{i}{l}$ for $l {<} L$ in parallel networks with homogeneous $\sigma$, $\params^{(i,L)} {=} \alpha_i$ and $\params^{(i,L-1)} {=} \ssul{i}{ L-1 } $ in parallel networks with sign-determined $\sigma$, and $\params^{(i,L)} {=}\alpha^{(i)} $ and $ \params^{(i,L-1)} {=}\mathbf{s}^{(i)}$ in tree networks. 
%
Similar to the parallel network in \Cref{sec:parnet}, extend the standard \eqref{eq:std_nn} and tree \eqref{eq:tree_recursion} network definitions row-wise to the cases where the input is $\mathbf{X} \in \mathbb{R}^{N \times d}$. 
 \begin{lemma}\label{lemma:rescale}
 Let $\mathbf{\tilde{X}}^{(i)} {\in} \mathbb{R}^N$ be $\Xul{i}{L}$ for a parallel network  or $\sigma \left( \mathbf{X}^{(i)}+ b^{(i)}\mathbf{1}\right)$ for a tree network, where the input is $\mathbf{X} \in \mathbb{R}^{N \times d}$. 
     The training problem is equivalent to 
\begin{equation}\label{eq:noncvxtrain_genLlyrLLnorm_rescale}
    \min_{\params \in \paramspace:\regfunc_{(i,l)}\left(\params^{(i,l)}\right)=1, l {\in} \efflayers-\{L\} }  \mathcal{L}_{\mathbf{y}}  \left( \sum_{i=1}^{m_L} \alpha_i \mathbf{\tilde{X}}^{(i)}  \right)
     + \beta \sum_{i=1}^{m_L} \regfunc_{(i,L)}\left(\alpha_i\right)
\end{equation}
%
 \end{lemma}

\begin{proof}
By the AM-GM inequality on \eqref{eq:reg_as_sum}, a lower bound on the training problem is

\begin{equation}\label{eq:rescaledlb}
    \min_{\params \in \paramspace} \mathcal{L}_y \left(\nn{L}{\mathbf{X}} \right) + \beta \sum_{i=1}^{m_L}  \prod_{l\in\efflayers}\regfunc_{(i,l)}\left(\params^{(i,l)}\right).
\end{equation}

Consider the minimization problem

\begin{equation}\label{eq:rescaledlboneconstr}
    \min_{\params \in \paramspace:\regfunc_{(i,l)}\left(\params^{(i,l)}\right)=1, l\in\efflayers{-}\{L\}} \mathcal{L}_y \left(\nn{L}{\mathbf{X}} \right) + \beta \sum_{i=1}^{m_L}  \prod_{l\in\efflayers}\regfunc_{(i,l)}\left(\params^{(i,l)}\right)
\end{equation}

Problem \eqref{eq:rescaledlboneconstr} is an upper bound on \eqref{eq:rescaledlb}. Given optimal $\left\{\params^{(i,l)}\right\}$ in \eqref{eq:rescaledlb}, the rescaled parameters ${\params^{(i,l)}}' {=}  \params^{(i,l)}/\regfunc_{(i,l)}\left(\params^{(i,l)}\right)$ for $l {\in}\efflayers{-}\{L\}$ and ${\params^{(i,L)}}' {=} \params^{(i,L)}\prod_{l\in\efflayers}\regfunc^{(i,l)}\left(\params^{(i,l)}\right)$ (and rescaled bias parameters) achieve the same objective in \eqref{eq:rescaledlboneconstr}. Hence \eqref{eq:rescaledlboneconstr} and \eqref{eq:rescaledlb} are equivalent. Given optimal $\left\{\params^{(i,l)}\right\}$ in \eqref{eq:rescaledlboneconstr}, the rescaled parameters $ {\params^{(i,l)}}' {=} |\params^{(i,L)}|^{\frac{1}{\effL}} \params^{(i,l)} $ (and rescaled bias parameters) achieve the same objective in the training problem, which is therefore equivalent to \eqref{eq:rescaledlboneconstr}. Simplifying \eqref{eq:rescaledlboneconstr} gives \eqref{eq:noncvxtrain_genLlyrLLnorm_rescale}.
\end{proof}


\Cref{lemma:rescale} applies to networks without any weight constraints, i.e., it excludes $3$-layer \symmeterized{} networks.
A $L$-layer \emph{\symmeterized{} network} is a parallel network with homogeneous activation such that $m_{L-3}{=}1$ and the elements of $\Wul{i}{l}$ have the same magnitude for $l {\in} \{L{-}2,L{-}1\}$.
In a \symmeterized{} network, the last two layer's weights are vectors: $\Wul{i}{L-2} {\in} \mathbb{R}^{1 \times m_{L-2}}$ and $ \Wul{i}{L-1} {\in} \mathbb{R}^{m_{L-2}}$. The constraint on the weight magnitudes is encoded in $\paramspace$.
A $3$-layer \symmeterized{} network and a \deepnarrow{} network are special cases of a \symmeterized{} network. 

\begin{lemma}\label{lemma:rescaledsymm}
Let $r^{(i,l)}\left(\Wul{i}{l}\right)=\left\|\Wul{i}{l}\right\|_2$ for $l \in \{L{-}2,L{-}1\}$. The rescaled problem for a \symmeterized{} network is equivalent to  

\begin{equation}\label{eq:gen_rescaled_samemag}
    \min_{\params \in \paramspace: \regfunc^{(i,l)}\left( \Wul{i}{l}\right) = 1 \text{ }\mathrm{for}\text{ }  l \in [L-3]; \left| \Wul{i}{l}_j\right| = 1 \text{ }\mathrm{for}\text{ } l \in \{L-2,L-1\}, j \in [m_{L-3}]} \mathcal{L}_{\mathbf{y}}\left(\nn{L}{\mathbf{X}}  \right) + \symbeta\sum_{i=1}^{m_L}\regfunc_{(i,L)}\left(\alpha_i\right)
    \end{equation}
    where $\symbeta=\frac{\beta}{m_{L-2}} $.
\end{lemma}

\begin{proof}
 Since $m_{L-3}=1$, we have $\Wul{i}{L-2} {\in} \mathbb{R}^{1 \times m_{L-2}}$ and $\Wul{i}{L-1} {\in} \mathbb{R}^{m_{L-2}} $. The constraint states that for $l {\in} \{L{-}2,L{-}1\}$, $\left|\Wul{i}{l}_1\right|{=}{\cdots}{=}\left|\Wul{i}{l}_{m_{L-2}}\right|$. For $l {\in} \{L{-}2,L{-}1\}$, given $\Wul{i}{l},\bul{i}{l}$ and $\alpha_i$ in apply a change of variables ${\Wul{i}{l}}' {=} \sqrt{m_{L-2}} \Wul{i}{l}$ and $ {\bul{i}{l}}'{=}\sqrt{m_{L-2}}^{l+3-L}\bul{i}{l}$ and $\alpha_i' = \frac{1}{m_{L-2}}\alpha_i$ to the parameters in \eqref{eq:noncvxtrain_genLlyrLLnorm_rescale} to arrive at \eqref{eq:gen_rescaled_samemag}. 
\end{proof}

Henceforth, assume $\regfunc_{(i,L)}\left(\alpha_i\right)=|\alpha_i|$.

\begin{definition}
Define the \emph{rescaled training problem} as

\begin{equation}\label{eq:gen_rescale}
    \min_{\params \in \paramspace}  \mathcal{L}_{\mathbf{y}}  \left( \sum_{i=1}^{m_L} \alpha_i \mathbf{\tilde{X}}^{(i)}  \right)
     + \beta \sum_{i=1}^{m_L} |\alpha_i|.
\end{equation}
If the network is \symmeterized{}, $\symbeta {=} \frac{\beta}{m_{L-2}}$ and $\paramspace$ includes the constraints that
$\regfunc^{(i,l)}\left( \Wul{i}{l}\right)\\ {=} 1$ for $l \in [L{-}3]$ and $\left| \Wul{i}{l}_j\right| {=} 1$ for $l \in \{L{-}2,L{-}1\}$. Otherwise, $\symbeta{=}\beta$ and $\regfunc_{(i,l)}\left(\params^{(i,l)}\right){=}1$ for $l \in \efflayers{-}\{L\}$. $\mathbf{\tilde{X}}^{(i)}$ is defined as in \Cref{lemma:rescale}. 
\end{definition}

For $3$-layer
networks with $l_2$ regularization ($r^{(i,l)}\left(\Wul{i}{l}\right)=\|\Wul{i}{l}\|_2$), 
$\paramspace$ constrains the absolute value of all elements of all inner layer weights to be $1$ in the rescaled training problem. The rescaled training problem is equivalent to both \symmeterized{} and non-\symmeterized{} networks.

\begin{lemma}
    The rescaled training problem \eqref{eq:gen_rescale} is equivalent to the training problem for all the architectures and activations discussed above.
\end{lemma}

\begin{proof}
    Follows from \Cref{lemma:rescale} and \Cref{lemma:rescaledsymm}.
\end{proof}

\begin{lemma}
    A lower bound on the rescaled training problem 
    is the dual problem
    \begin{equation}\label{deepgen_dual}
		\begin{aligned}
			\max_{\lambda \in \mathbb{R}^N} & -\mathcal{L}_{\mathbf{y}}^*( \lambda ) \quad 
			\text{ s.t. }                   &
			\max_{\params \in \paramspace}
			\left|\lambda^T  \mathbf{\tilde{X}}  \right| \leq \symbeta,
		\end{aligned}
	\end{equation}
  where $\mathbf{\tilde{X}}=\mathbf{\tilde{X}}^{(1)}$ and $f^*(\mathbf{x}):=\max_{\mathbf{x}}  \left \{ \mathbf{z}^T \mathbf{x} - f(\mathbf{x}) \right \}$ is the convex conjugate of $f$.
\end{lemma}
\begin{proof}
  Find the dual of \eqref{eq:noncvxtrain_genLlyrLLnorm_rescale}, by rewriting \eqref{eq:noncvxtrain_genLlyrLLnorm_rescale} as
	\begin{equation}\label{deepgen_prep_to_dual}
		\begin{aligned}
			\min_{\params \in \paramspace}\,
			              & \mathcal{L}_{\mathbf{y}}( \mathbf{z} ) + \symbeta ||\boldsymbol{\alpha}||_1, \quad
			\text{ s.t. } & \mathbf{z} =  \sum_{i=1}^{m_L} \alpha_i \mathbf{\tilde{X}}^{(i)}  .
		\end{aligned}
	\end{equation}
	The Lagrangian of problem \eqref{deepgen_prep_to_dual} is
%
   $\displaystyle 
			L\left(\lambda, \params \right)
			  = \mathcal{L}_{\mathbf{y}}( \mathbf{z} ) + \symbeta ||\boldsymbol{\alpha}||_1  - \lambda^T \mathbf{z} + \sum_{i=1}^{m_L} \lambda^T  \mathbf{\tilde{X}}^{(i)}     \alpha_i.  
    $
%
	Minimize the Lagrangian over $\mathbf{z}$ and $\boldsymbol{\alpha}$ and use Fenchel duality \citep{boyd:2004}. The dual of \eqref{deepgen_prep_to_dual} is
	\begin{equation}\label{deepgen_dual_int}
		\begin{aligned}
			\max_{\lambda \in \mathbb{R}^N} & -\mathcal{L}_{\mathbf{y}}^*( \lambda ) \quad 
			\text{ s.t. }                   &
			\max_{\params \in \paramspace}
			\left|\lambda^T   \mathbf{\tilde{X}}^{(i)}\right| \leq \symbeta, i \in [m_L].
		\end{aligned}
	\end{equation}
	In the tree and parallel nets, $ {\tilde{\mathbf{X}}}^{(i)}$ is of the same form for all $i \in [m_L]$. So the $m_L$ constraints in \eqref{deepgen_dual_int} collapse to a single constraint. Then we can write \eqref{deepgen_dual_int} as  \eqref{deepgen_dual}.  
\end{proof}
%

%
For a parallel network, 
 the dual problem \eqref{deepgen_dual} 
 is
\begin{equation}\label{gen_parallel_dual}
		\begin{aligned}
			\max_{\lambda \in \mathbb{R}^N} & -\mathcal{L}_{\mathbf{y}}^*( \lambda ) \quad
			\text{ s.t. }                   &
			\max_{\params \in \paramspace}
			\left|\lambda^T \xl{L} \right| \leq \symbeta,
		\end{aligned}
	\end{equation}
where $\xl{1}{=}\mathbf{X}$.
Henceforth, all regularizations are $l_2$-norm: e.g., for a parallel network, $r^{(i,l)}\left(\Wul{i}{l}\right){=}\left\|\Wul{i}{l}\right\|_2$, denoting the square root of sum of squares of $\Wul{i}{l}$'s elements.

\section*{Deep narrow and \symmeterized{} networks with $d{=}1$}
In this section, we assume the data is \nd{1} and find the maximizers of $\left|\lambda^T \xl{L} \right|$ in the dual constraint \eqref{gen_parallel_dual}. To this end, assume the elements of $\wl{l}$ are $\pm 1$.
Note that $\bl{L-1}$ is a scalar and $\mathbf{X}^{(l+1)}{=} \sigma \left( \mathbf{X}^{(l)}\wl{l}{+}\bl{l}\mathbf{1}\right) {\in} \mathbb{R}^N$. 
The next remark refers to the leaky ReLU slopes $\posslope$ and $\negslope$ 
defined in \Cref{section:nn_arch}. Let $\mathcal{K}(f)$ and $\mathcal{Z}(f)$ be the sets of \kink{}s and zeros of a function $f$, respectively.
%
\iftoggle{long}{
\begin{figure}
    \centering
\begin{tikzpicture}
    \draw[dashed] (0,0) -- (12,0);
    \draw (0,0) -- (4,0);
    \draw (4,0) -- (8,4);
    \draw (8,4) -- (10,4);
    \draw[dashed] (8,4) -- (8,0);
    
    \draw[left] (0,0) node {$0$};
    \draw[right] (12,0) node {$b^{(l-1)}$};
    
    \draw[] (4,-0.5) node {$d_{i,j}^{(l-1)} = - \mathbf{X}_i^{(l-1)}$};
    \draw[] (8,-0.5) node {$f_{i,j}^{(l-1)} = - \mathbf{X}_j^{(l-1)}$};
    \draw[right] (10,4) node {$\left|c_{i,j}^{(l-1)}\right| = \mathbf{X}_i^{(l-1)} - \mathbf{X}_j^{(l-1)}$};
\end{tikzpicture}
\caption{Example of $\sigma\left(c^{(l)}_{i,j}\right)$ as a function of $b^{(l-1)}$ when $w^{(l)}=w^{(l-1)}=1, \mathbf{X}^{(l-1)}_i \geq \mathbf{X}^{(l-1)}_j $. }
    \label{fig:sigma(c^l_ij)}
\end{figure}
\begin{lemma}\label{diff_ReLU}
     Suppose $\sigma(x)=(x)_+$. Let $c_{i,j}^{(l)}=w^{(l)}\left(\mathbf{X}^{(l-1)}_i - \mathbf{X}^{(l-1)}_j \right) $. Then 

     \begin{equation*}
     \begin{aligned}
     \sigma \left( c_{i,j}^{(l)} \right)
     =  
        \begin{cases}
           \mathbf{X}^{(l-1)}_i -\mathbf{X}^{(l-1)}_j  \mbox { if } c_{i,j}^{(l-1)} \geq 0, w^{(l)} \geq 0 \\
            \mathbf{X}^{(l-1)}_j -\mathbf{X}^{(l-1)}_i \mbox { if } c_{i,j}^{(l-1)} \leq 0, w^{(l)} \leq 0 \\
            0 \mbox{ else }
     \end{cases},
     \end{aligned}
     \end{equation*}

     which (when nonzero) has {\kink}s at $b^{(l-1)} \in \left \{ \mathbf{X}^{(l-2)}_i w^{(l-1)}, -\mathbf{X}^{(l-2)}_j w^{(l-1)} \right \}$.

     If $ c_{i,j}^{(l-1)} w^{(l)} \leq 0$, then $\sigma \left( c_{i,j}^{(l)} \right)=0$. Otherwise, let $d^{(l-1)}_{i,j} = \min \left \{ -\mathbf{X}^{(l-2)}_i w^{(l-1)},  -\mathbf{X}^{(l-2)}_j w^{(l-1)} \right \}, f^{(l-1)}_{i,j} = \max \left\{ -\mathbf{X}^{(l-2)}_i w^{(l-1)},  -\mathbf{X}^{(l-2)}_j w^{(l-1)} \right\}$. Then

     \begin{equation}\label{eq:sigma_c}
         \begin{aligned}
             \sigma \left( c_{i,j}^{(l)} \right) = 
             \begin{cases}
                  0 & \mbox{ if }  b^{(l-1)}  \leq d^{(l-1)}_{i,j} \\
                 -d^{(l-1)}_{i,j}+ b^{(l-1)} & \mbox{ if }   d^{(l-1)}_{i,j} \leq d^{(l-1)}_{i,j}\leq  f^{(l-1)}_{i,j}\\ 
                  \left | c_{i,j}^{(l-1)} \right | &\mbox { if } b^{(l-1)} \geq  f^{(l-1)}_{i,j}.
             \end{cases}
         \end{aligned}
     \end{equation}
     This is drawn in \Cref{fig:sigma(c^l_ij)}.
\end{lemma}

\begin{proof}

  First suppose $c^{(l-1)}\geq 0$. As a function of $b^{(l-1)}$, we have 

\begin{equation}
  
\begin{aligned}
  \mathbf{X}^{(l-1)}_i - \mathbf{X}^{(l-1)}_j &= \left( \mathbf{X}^{(l-2)}_i w^{(l-1)}+b^{(l-1)} \right)_+ - \left( \mathbf{X}^{(l-2)}_j w^{(l-1)}+b^{(l-1)} \right)_+ \\
&= \begin{cases}
    0 & \mbox{ if }  b^{(l-1)}  \leq -\mathbf{X}^{(l-2)}_i w^{(l-1)} \\
    \mathbf{X}^{(l-2)}_i w^{(l-1)} + b^{(l-1)} & \mbox{ if } -\mathbf{X}^{(l-2)}_i w^{(l-1)} \leq b^{(l-1)} \leq -\mathbf{X}^{(l-2)}_j w^{(l-1)}\\ 
     c_{i,j}^{(l-1)} &\mbox { if } b^{(l-1)} \geq -\mathbf{X}^{(l-2)}_j w^{(l-1)}, 
\end{cases} 
\end{aligned} 
\end{equation} 

which is always nonnegative.
Similarly if $c^{(l-1)}\leq 0$,

\begin{equation}
  
\begin{aligned}
  \mathbf{X}^{(l-1)}_i - \mathbf{X}^{(l-1)}_j &= \left( \mathbf{X}^{(l-2)}_i w^{(l-1)}+b^{(l-1)} \right)_+ - \left( \mathbf{X}^{(l-2)}_j w^{(l-1)}+b^{(l-1)} \right)_+ \\
&= \begin{cases}
 0 &\mbox { if } b^{(l-1)} \leq -\mathbf{X}^{(l-2)}_j w^{(l-1)} \\
    -\left(\mathbf{X}^{(l-2)}_j w^{(l-1)} + b^{(l-1)} \right) & \mbox{ if } -\mathbf{X}^{(l-2)}_j w^{(l-1)} \leq b^{(l-1)} \leq -\mathbf{X}^{(l-2)}_i w^{(l-1)}\\ 
    c_{i,j}^{(l-1)} & \mbox{ if }  b^{(l-1)}  \geq -\mathbf{X}^{(l-2)}_j w^{(l-1)},
\end{cases} 
\end{aligned} 
\end{equation} 

which is always nonpositive. The result follows.
\end{proof}

\begin{figure}
    \centering
\begin{tikzpicture}
    \draw[dashed] (0,0) -- (12,0);
    \draw (0,0) -- (2,0);
    \draw (2,0) -- (6,4);
    \draw (6,4) -- (8,2);
    \draw (8,2) -- (12,2);
    \draw[dotted] (6,4) -- (10,0);
    \draw[dotted] (10,0) -- (12,0);
    \draw[dashed] (6,4) -- (6,0);
    \draw[dashed] (8,2) -- (8,0);
    
    \draw[left] (0,0) node {$0$};
    \draw[right] (12,0) node {$b^{(l-1)}$};
    
    \draw[] (2,-0.5) node {$d_{i,j}^{(l-1)} $};
    \draw[] (6,-0.5) node {$f_{i,j}^{(l-1)} $};
    \draw[top] (6,4.5) node {$c_{i,j}^{(l-1)} = \mathbf{X}_i^{(l-1)} - \mathbf{X}_j^{(l-1)}$};
    \draw[bottom] (8,-0.5) node {$f^{(l-1)}_{k,j}$};
    \draw[bottom] (10,-0.5) node {$r^{(l-1)}_{i,j}$};
\end{tikzpicture}
\caption{Example of $ \sigma \left( w^{(l+1)} \left( \sigma \left( c_{i,j}^{(l)}\right ) - \sigma \left( c_{k,j}^{(l)}\right ) \right) \right)$ as a function of $b^{(l-1)}$. 
}
    \label{fig:diff_sigma(c^l_ij)}
\end{figure}

\begin{lemma}\label{lemma:diff_of_flatramps}
If nonzero, $\sigma \left( w^{(l+1)} \left( \sigma \left( c_{i,j}^{(l)}\right ) - \sigma \left( c_{k,j}^{(l)}\right ) \right) \right) =  \sigma \left( c^{(l)}_{i,k} \right)$ if $d^{(l-1)}_{i,j} = d^{(l-1)}_{k,j}$ or $ f^{(l-1)}_{i,j} = f^{(l-1)}_{k,j}$. This has {\kink}s at $f^{(l-1)}_{i,j}, f^{(l-1)}_{k,j}$ or $d^{(l-1)}_{i,j}, d^{(l-1)}_{k,j}$, respectively.

Let $r^{(l-1)}_{i,j} = 2f^{(l-1)}_{i,j} -d^{(l-1)}_{i,j}$ be the reflection of $d^{(l-1)}_{i,j}$ about $f^{(l-1)}_{i,j}$. If $f^{(l-1)}_{i,j} = d^{(l-1)}_{k,j}$ then 
    \begin{equation}
    \begin{aligned}
     \sigma \left( w^{(l+1)} \left( \sigma \left( c_{i,j}^{(l)}\right ) - \sigma \left( c_{k,j}^{(l)}\right ) \right) \right) = 
       \begin{cases}
            0 & \mbox { if } b^{(l-1)} \leq  d^{(l-1)}_{i,j}\\
            -d^{(l-1)}_{i,j} + b^{(l-1)} & \mbox { if }   d^{(l-1)}_{i,j} \leq b^{(l-1)} \leq f^{(l-1)}_{i,j} \\
           r^{(l-1)}_{i,j}-  b^{(l-1)} & \mbox { if } f^{(l-1)}_{i,j} \leq b^{(l-1)}  \leq \min \left \{ r^{(l-1)}_{i,j}, f^{(l-1)}_{k,j} \right \} \\
           \max \left \{c^{(l)}_{i,k} , 0 \right\} & \mbox { if } b^{(l-1)}  \geq \min \left \{ r^{(l-1)}_{i,j}, f^{(l-1)}_{k,j} \right \} 
       \end{cases}.   
    \end{aligned}
    \end{equation}
   This has {\kink}s at $d^{(l-1)}_{i,j}, f^{(l-1)}_{i,j}$ and $\min \left \{ r^{(l-1)}_{i,j}, f^{(l-1)}_{k,j} \right \} $.
\end{lemma}
\begin{proof}
    Follows from \eqref{eq:sigma_c}.
\end{proof}
}
\iftoggle{long}{
\begin{remark}\label{rk:finiteconstraint}
 Let $g: \mathbb{R} \to \mathbb{R}$ be a piecewise linear function which is not constantly zero.
 Then $g$ is bounded above (below) if and only if $\displaystyle \lim_{|b| \to \infty} g(b) < \infty$ ($-\infty <  \displaystyle \lim_{|b| \to \infty} g(b) $). And $g$ is bounded if it is bounded above and below. A maximizer (minimizer) of $g$ is a finite $b^*$ such that $g(b^*) \geq g(b) $ $(g(b^*)\leq g(b))$ for all $b$. Let $\mathcal{K}$ be the set of {\kink}s of $g$.
If $g$ is bounded, then $\mathcal{K} \neq \emptyset$ and contains a maximizer and minimizer of $g$. Otherwise, a maximizer (minimizer) does not exist if $g$ is bounded below (above). Now let $g(b) =  \displaystyle \sum_{n=1}^N \lambda_n g_n(b)$, where $g_n: \mathbb{R} \to \mathbb{R}$ are piecewise linear functions. Let $\mathcal{K}_n$ be the set of {\kink}s of $g_n$. Then $\mathcal{K} = \displaystyle \bigcup_{n=1}^N \mathcal{K}_n$ will contain a maximizer and minimizer of $g$, if they exist. If $g_n$ is bounded for all $n \in [N]$, then $g$ has a maximizer and a minimizer. Now suppose $g_n$ are all of the form $g_n(b)=c^+ b + a_n^+$ for $b$ large enough and $g_n(b)=c^- b + a_n^-$ for $b$ small enough, where $c^+, c^-, a_n^+, a_n^- \in \mathbb{R}$ are constants. Observe $\left| 
 \displaystyle \lim_{|b| \to \infty} g(b) \right|$ is of the form $ \left| \displaystyle \lim_{|b| \to \infty} \sum_{n=1}^N \lambda_n(c b + a_n) \right| = \left| \displaystyle \lim_{|b| \to \infty} cb \sum_{n=1}^N \lambda_n + \sum_{n=1}^N \lambda_n a_n \right| $. So $g$ has a maximizer and a minimizer if and only if $c^+ = c^- = 0$ or $ \sum_{n=1}^N \lambda_n =0$.
\end{remark}
}

\begin{remark}\label{rk:finiteconstraint}
Let $b {=} \bl{L-1}, a_n {=} \xl{L-1}_n \wl{L-1}, g_n(b) {=} \sigma \left( a_n {+}b\right)$, and $g(b)  {=} \sum_{n{=}1}^N \lambda_n g_n(b) {=}\\ \lambda^T \mathbf{X}^{(L)}$. Let $\mathcal{I} {=} \bigcup_{n{=}1}^N \mathcal{K}\left(g_n\right) \supset \mathcal{K}(g)$.
Then 
$g(b) {=}  \sum_{n{=}1}^N \lambda_n a ( a_n {+} b) {=}  a b \sum_{n{=}1}^N \lambda_n {+} \\ \sum_{n{=}1}^N \lambda_n  a_n a $ for $b$ large enough (with $a {=} \posslope$) and for $b$ small enough (with $a{=}\negslope$). So $\negslope{=}\posslope{=}0$ or $ \sum_{n=1}^N \lambda_n {=}0$ if and only if $g$ is bounded, if and only if $g$ has a (finite) maximizer and minimizer. In this case, assuming $g$ is not a constant function, $\mathcal{I}$ contains a maximizer and minimizer of $g$.
\end{remark}

Henceforth, assume $\lambda^T \mathbf{1} {=} 0$ if $\negslope {\neq} 0$ or $\posslope {\neq} 0$. Suppose $\bl{l}, {\cdots}, \bl{L-1}$ are scalar-valued. We call $\bl{l}{=}{\bl{l}}^*$ \emph{optimal} if ${\bl{l}}^* {\in} \arg\max_{\bl{l}}\max_{\bl{l+1}} {\cdots} \max_{\bl{L-1}}\left| \sum_{n{=}1}^N \lambda_n \xl{L}_n \right|$. 
%
  %
%
%
%
The next result refers to the normalized midpoint defined in \eqref{def:malphabeta}.

\begin{lemma}\label{theorem:kinks_general}
    Let $\sigma$ be leaky ReLU.
%
Let $\alpha,\beta \in \mathbb{R}$ with $\alpha {\neq} \beta$. Let $f(x) {=} \sigma \left( x {+} \alpha \right){-}\sigma \left( x {+} \beta\right)$. For $s \in \{{-},{+}\}$, let $g_s(x) {=} \sigma(sx)$. 
Then, if $\sigma$ is monotone, 
\begin{equation}\label{eq:sigma_mon_kinks}
    \mathcal{K}\left(g_s(f)\right) \subset \{-\alpha,-\beta\}.   
\end{equation}
Otherwise if $\sigma$ is not monotone,
\begin{equation}\label{eq:sigma_notmon_kinks}
  \mathcal{K}(g_s(f))\subset  \{-\alpha, -\beta, m_{\alpha,\beta} \}. 
\end{equation}
%
%
\end{lemma}

%

\begin{proof}
%
Since $f_{\alpha,\beta}$ is piecewise linear,
\begin{equation}\label{eq:kinkszeros}
 \mathcal{K}\left(g_s(f)\right) \subset \mathcal{K}(f)\cup \mathcal{Z}(f).  
\end{equation}
%
Moreover if $f(x)$ has the same sign for all $x$, then observe that
\begin{equation}\label{eq:kinkszerosmon}
    \mathcal{K}\left(g_s(f)\right) {=} \mathcal{K}(f).
\end{equation}
Let $\mathcal{G}(f) = \{(x,f(x)): x \in \mathcal{K}\left(f\right)\}$. We find that 
\begin{equation}\label{eq:falphabeta}
    \mathcal{G}(f) =\left \{ \left({-}\max\{\alpha,\beta\},a^-(\alpha{-}\beta)\right),\left({-}\min\{\alpha,\beta\},a^+(\alpha{-}\beta)\right)\right\}.    
\end{equation}
%
%
In particular, $\mathcal{K}(f){=}\{{-}\alpha, {-}\beta\}$. Observe that $f$ has constant value (an in particular, sign) beyond its \kink{}s. If $\sigma$ is monotone, then $\text{sign}(\posslope){=}\text{sign}(\negslope)$, so \eqref{eq:falphabeta} implies \\$\text{sign}\left(f(x)\right){=}\text{sign}\left(a^+(\alpha{-}\beta)\right)$ for all $x$, so \eqref{eq:kinkszerosmon} implies \eqref{eq:sigma_mon_kinks}. 
%
%
 %
%
If $\sigma$ is not monotone, linearly interpolating between the points in $\mathcal{G}\left(f\right)$ \eqref{eq:falphabeta} shows that $f$ changes sign exactly when $x {=} m_{\alpha,\beta}$, so \eqref{eq:kinkszeros} implies \eqref{eq:sigma_notmon_kinks}. 
\end{proof}


\begin{lemma}\label{lemma:max_constr_Llyrl}
    Let $\sigma$ be piecewise linear if $L{=}2$ and leaky ReLU otherwise. Let $l {\in} \{L{-}1, L{-}2\}$. Consider the bias $\bl{l}$. Let $m_{L-2}{=}1$.
   If $\sigma$ is monotone there is either a data feature bias that is optimal.
  If $\sigma$ is not monotone there is either a data feature bias or a midpoint feature bias $b^{(L-2)} {=} m_{\mathbf{X}^{(L-2)}_{n^{(L-2)}},\mathbf{X}^{(L-2)}_{n^{(L-1)}}} \wl{L-2}$ (if $l{=}L{-}2$) that is optimal. 
%


\end{lemma}

\begin{proof}
 \Cref{rk:finiteconstraint} implies that
   $\bigcup_{n=1}^N \mathcal{K}\left(\xl{L}_n\right)$ contains an optimal $\bl{L-1}$.
The {\kink} of $\xl{L}_n{=} \sigma \left( \mathbf{X}^{(L{-}1)}_n \wl{L-1} {+}\bl{L{-}1}\right)$ as a function of $\bl{L{-}1}$ is $\bl{L{-}1}{=} {-}\mathbf{X}^{(L-1)}_n \wl{L{-}1}$.
Therefore for some $n^{(L{-}1)} {\in} [N]$, the data feature bias $\bl{L{-}1}{=} {-}\mathbf{X}^{(L{-}1)}_{n^{(L{-}1)}} \wl{L{-}1} $ is optimal.
Plugging in this optimal $\bl{L{-}1}$ back into $\xl{L}_n$ gives 
\begin{equation*}
    \lambda^T \mathbf{X}^{(L)}    {=}\sum_{n=1}^N \lambda_n \sigma \left( \left( \xl{L-1}_n{-}\xl{L-1}_{n^{(L{-}1)}}  \right)\wl{L{-}1} \right).
\end{equation*}
For $L>2$,
expanding $\xl{L-1}$ gives
\begin{equation}\label{eq:lambda^TX_L-1}
\scalemath{0.95}{
    \begin{aligned}
         \lambda^T \mathbf{X}^{(L)}    
                                        &{=} \sum_{n=1}^N \lambda_n \sigma \left(  \left( \sigma \left( \xl{L-2}_n \wl{L{-}2}{+} \bl{L{-}2}\right)-\sigma \left( {\xl{L-2}_{n^{(L{-}1)}}} \wl{L{-}2}  {+} \bl{L{-}2}\right) \right) \wl{L{-}1}\right).
    \end{aligned}
    }
\end{equation}
%
Since $m_{L{-}2}{=}1$, we have $\bl{L{-}2} {\in} \mathbb{R}$.
Applying \Cref{theorem:kinks_general} with $x{=}\bl{L-2},\alpha {=} \xl{L-2}_n \wl{L-2}, \\ \beta {=} \xl{L-1}_n \wl{L-2}, s {=} \wl{L-1}$ and a similar argument as \Cref{rk:finiteconstraint} gives the result.
%
\end{proof}


\begin{lemma}\label{lemma:ReLUoptb}
   Let $L \geq 2$. Consider a \deepnarrow{} ReLU network with $L$ layers. For every $l \in \{2,{\cdots},L\}$, a \datafeature{} bias is optimal for layer $L{-}l{+}1$, and with this optimal bias, 
   if $l{<}L$, there exists $N_1, N_2 {\in} [N]$ such that for all $n {\in} [N]$, either $\xl{L}_n{=}0$ or

   \begin{equation}\label{eq:bL-lramp}
       \begin{aligned}
           \xl{L}_n = \pm\left(\sigma\left(\xl{L{-}l}_{N_1} \wl{L{-}l} {+} \bl{L{-}l} \right){-}\sigma\left(\xl{L{-}l}_{N_2} {+} \bl{L{-}l} \right) \right).
       \end{aligned}
   \end{equation}

\end{lemma}

\begin{proof}
    We prove by induction on $l$.
    
    Base case: suppose $l=2$. Then the claim holds by  \eqref{eq:lambda^TX_L-1} in the proof of \Cref{lemma:max_constr_Llyrl}.
    
   Now, suppose the claim holds for $l{=}k{\in}\{2,{\cdots},L{-}1\}$. 
Applying argument similar to the proof of \Cref{lemma:max_constr_Llyrl} to \eqref{eq:bL-lramp} shows that
%
    a \datafeature{} bias $\bl{L{-}k} {=} {-}\xl{L{-}k}_{n^{(L{-}k)}}\wl{L{-}k}$ is optimal for 
$\bl{L-l+1}$ when $l{=}k{+}1$. Plugging in this optimal $\bl{L{-}k}$ into $\xl{L}$ in \eqref{eq:bL-lramp} for $l{=}k$ and expanding $\xl{L{-}k}{=}\sigma\left(\xl{L{-}k{-}1}\wl{L{-}k{-}1}{+}\bl{L{-}k{-}1}\right)$ shows that for some $s^{(1)},s^{(2)} {\in} \{-1,1\}$, either $\xl{L}_n{=}0$ or 
    \begin{equation*}
        \scalemath{0.9}{
        \begin{aligned}
        \xl{L}_n=&\sigma \left(  s^{(1)}\left( \sigma \left(  s^{(2)} \left( \sigma \left(\xl{L{-}k{-}1}_{n'} \wl{L{-}k{-}1}  {+} \bl{L{-}k{-}1} \right)  {-} \sigma \left( \xl{L{-}k{-}1}_{n^{(L{-}2)}}\wl{L-3}  {+} \bl{L{-}k{-}1}  \right) \right) \right) \right. \right. \\
        & \left. \left. \qquad  {-}\sigma \left(  s^{(2)} \left( \sigma \left( \xl{L{-}k{-}1}_{n^{(L{-}1)}}\wl{L{-}k{-}1} {+} \bl{L{-}k{-}1} \right)  {-} \sigma \left( \xl{L{-}k{-}1}_{n^{(L{-}2)}} \wl{L{-}k{-}1} {+} \bl{L{-}k{-}1} \right)\right) \right) \right)\right) \Big) \Big) \\
        & \in \Big \{ s^{(2)}\left(\sigma\left(\xl{L{-}k{-}1}_{n'} \wl{L{-}k{-}1} + \bl{L{-}k{-}1} \right)-\sigma\left(\xl{L-{k}-{1}}_{n^{(L-2)}} + \bl{L{-}k{-}1} \right) \right) , \\
        & \qquad -s^{(2)}\left(\sigma\left(\xl{L{-}k{-}1}_{n^{(L-1)}} \wl{L{-}k{-}1} + \bl{L{-}k{-}1} \right)-\sigma\left(\xl{L-{k}-{1}}_{n^{(L-2)}} + \bl{L{-}k{-}1} \right) \right) , \\
        & \qquad \qquad s^{(2)}\left(\sigma\left(\xl{L{-}k{-}1}_{n'} \wl{L{-}k{-}1} + \bl{L{-}k{-}1} \right)-\sigma\left(\xl{L-{k}-{1}}_{n^{(L-1)}} + \bl{L{-}k{-}1} \right) \right),0 
        \Big\}.
        \end{aligned}
                }
    \end{equation*}

    So \eqref{eq:bL-lramp} holds for $l{=}k{+}1$. Finally if \eqref{eq:bL-lramp} holds for $l{=}L{-}1$ then by a similar argument as the proof of \Cref{lemma:max_constr_Llyrl}, a \datafeature{} bias for $\bl{L-l+1}$ is optimal when $l{=}k{+}1{=}L$. By induction, the result holds.
\end{proof}

\begin{lemma}\label{lemma:ReLUsatperlayer}
    Let $L \geq 2$. Consider a \deepnarrow{} ReLU network with $L$ layers. For $l \in [L-1]$, suppose $\bl{l}$ is a \datafeature{}.
    For every $l {\in} \{2, {\cdots}, L{-}1\}$, there exist $N_1, N_2 {\in} [N]$ such that for all $x$,

    \begin{equation}
        \begin{aligned}
            \xl{l}(x)\in \left\{ \ReLU{\pm}{x_{N_1}}(x), \ramp{\pm}{x_{N_1}}{x_{N_2}}(x)\right\}.
        \end{aligned}
    \end{equation}
\end{lemma}

\begin{proof}
    We prove by induction on $l$. Base case: for $l=2$, $\xl{2}(x){=}\sigma \left(x\wl{1}{+}\bl{1}\right){=}\\\sigma \left(\left(x{-}x_{n^{(1)}}\right)\wl{1}\right){=}\ReLU{\wl{1}}{x_{n^{(1)}}}(x)$. 
    Now, suppose the claim holds for $l=1,\cdots,k{<}L{-}1$.
    Then $\xl{k{+}1}(x){=}\sigma\left(\xl{k}\wl{k}{+}\bl{k}\right){=}\sigma\left(\left(\xl{k}_n{-}\xl{k}_{n^{(k)}}\right)\wl{k}\right)$ so either \\ there exist $N_1', N_2'{\in}\left\{N_1,n^{(k)}\right\}, s {\in} \{+,-\} $ such that for all $x$, \\ $\xl{k{+}1}(x){=}\sigma\left(\left(\ReLU{\pm}{x_{N_1}}(x){-}\ReLU{\pm}{x_{N_1}}\left(x_{n^{(k)}}\right)\right)\wl{k}\right){\in}\left\{\ReLU{s}{x_{N_1'}}(x), \ramp{s}{x_{N_1'}}{x_{N_1'}}(x)\right\}$, or \\there exist $N_1', N_2'\in\left\{N_1,N_2,n^{(k)}\right\}, s \in \{+,-\} $ such that for all $x$, \\ $\xl{k+1}(x){=}\sigma\left(\left(\ramp{\pm}{x_{N_1}}{x_{N_2}}(x){-}\ramp{\pm}{x_{N_1}}{x_{N_2}}\left(x_{n^{(k)}}\right)\right)\wl{k}\right) {=}\ramp{s}{x_{N_1'}}{x_{N_2'}}(x)$. By induction, the result holds.
    %
%
%
%
\end{proof}

\begin{lemma}\label{lemma:numKinks}
Consider a \deepnarrow{} network. Let $L \geq 2$. For all $l \in [L-1]$, there exist  $n^{(L-l)} \in [N]$ and a $l$-tuple $(a_0, a_1, \cdots, a_l)$ such that for all $l \in [L-1]$,

\begin{equation}\label{eq:optb}
    {\bl{L-l}}^*=
    -\sum_{i=1}^{l} a_i \xl{L-l}_{n^{(L-i)}} \wl{L-l} 
\end{equation}

is optimal. Note that optimal $\bl{1},{\cdots},\bl{L-1}$ can be found sequentially, by computing ${\bl{l}}^*$ as a function of ${\bl{l-1}}^*$ and so on. Moreover,
there are $O\left(2^L L!\right)$ options for 
${\bl{1}}^*$. Additionally, for all $l {\in} [L-1]$, under such optimal $\bl{1}, {\cdots}, \bl{l-1}$, for all $n \in [N]$, 

\begin{equation}\label{eq:xl}
\begin{aligned}
    \xl{L-l+1}_n = \sigma \left( \hat{\mathbf{X}}^{(L-l)'}_n  \wl{L-l}\right)
\end{aligned}
\end{equation}
where
\begin{equation}\label{eq:xltilde}
    \begin{aligned}
        \hat{\mathbf{X}}^{(L-l)'}_n  = a_0 \xl{L-l}_n + \sum_{i=1}^l a_i \xl{L-l}_{n^{(L-i)}}.
    \end{aligned}
\end{equation}
\end{lemma}

\begin{proof}
    We prove \eqref{eq:optb}, \eqref{eq:xl} and \eqref{eq:xltilde} by strong induction on $l$. 
    
    Base case: suppose $l=1$. Then \eqref{eq:optb}, \eqref{eq:xl}, \eqref{eq:xltilde} hold by \Cref{lemma:max_constr_Llyrl} and its proof.

    Now, suppose \eqref{eq:optb}, \eqref{eq:xl} and \eqref{eq:xltilde} hold for $l{=}1,{\cdots},k{<}L{-}1$. 
    By \Cref{eq:xl} for $l{=}k$, $\xl{L{-}k{+}1}_n(x)$ is locally a linear combination of the $k{+}1$ terms $\xl{L{-}k{+}1}_{n^{(L{-}k{-}1)}}, \xl{L{-}k{+}1}_{n^{(L{-}1)}},{\cdots} \\ ,\xl{L{-}k{+}1}_{n^{(L{-}k)}}$, and hence so is $\xl{L}$. So the \kink{}s of $\xl{L}$ as a function of $\bl{L{-}(k{+}1)}$ are linear combinations of the $k{+}1$ terms, which proves \eqref{eq:optb} for $l{=}k{+}1$. Plugging in this \kink{} $\bl{L-k-1}$   
    into $\xl{L-k}_n$ proves \eqref{eq:xl} and \eqref{eq:xltilde} for $l{=}k{+}1$. Therefore \eqref{eq:optb}, \eqref{eq:xl} and \eqref{eq:xltilde} hold for all $l {\in} [L{-}1]$.

    Now we prove that the number of options for $\bl{1}$ is $O\left(2^L L!\right)$. 
    By \Cref{eq:xltilde} for $l{=}L{-}1$, as a function of $\bl{1}$, $\hat{\mathbf{X}}^{(L{-}1)'}_j $ has \kink{}s at $-\xl{L{-}l{-}1}_i\wl{L{-}l{-}1}$ for $i {\in} \{n, n^{(L-1)},\\ {\cdots}, n^{(L-l)}\}$, which totals to $l{+}1{=}L$ \kink{}s. A piecewise linear function $f$ has up to plus or minus one as many zeros as \kink{}s, and the zeros become \kink{}s in $\sigma(f)$. So by \Cref{eq:xl}, $\xl{L-l+1}_n$ as a function of $\bl{1}$ has \kink{}s at $O(2L)$ places. Then by \Cref{eq:xltilde} for $l{=}L{-}2$, $ \hat{\mathbf{X}}^{(L{-}l{+}1)'}_n$ as a function of $\bl{1}$ has \kink{}s at $O( 2(L{-}1)(L))$ places. And by \Cref{eq:xl} for $l{=}L{-}2$, $\xl{L{-}k{+}2}_n$ as a function of $\bl{1}$ has \kink{}s at $O\left( 2^2 L (L-1)\right)$ places. By repeating this argument for $l{=}L{-}1,{\cdots},1$, $\xl{L}_n$ as a function of $\bl{1}$ has \kink{}s at $O\left(2^{L}L!\right)$ places. 
    %
    %
\end{proof}

Recall the deep library defined in \Cref{def:FeatDicStdNN}.
Note that any $\mathbf{a} $ in the deep library is of the form $\mathbf{a}=\xl{L}(\mathbf{X})$, with $\mathbf{a}_n = \xl{L}(x_n)$.

\begin{lemma}\label[lemma]{lemma:L234constr_recursive}
%

For $L \in \{2,3\}$ the maximization constraint in \eqref{gen_parallel_dual} is equivalent to: for all vectors $\mathbf{a}$ in the deep library,
\begin{equation}\label{eq:L=234lyr_max_constr}
  \begin{aligned} 
    \left | 
    \lambda^T \mathbf{a}
    \right| 
    \leq \symbeta \\
    & \mathbf{1}^T \lambda = 0 \mbox { if } \posslope \neq 0 \mbox{ or } \negslope \neq 0.
  \end{aligned}
  \end{equation}
%
 
\end{lemma}

\begin{proof}
\Cref{lemma:max_constr_Llyrl} gives \eqref{eq:L=234lyr_max_constr}. 
Now, if the activation is symmetric, 
then $\xl{L}$ is invariant under the sign of the components of $\wl{1}$. Next, recall $x_1 > \cdots > x_N$ and $\text{sign}(0)=1$. If the activation is sign, then for all $n \in [N-1]$, with $\wl{1}=1$ and $\bl{1}=-x_n$ we have
$\xl{L=2}(\mathbf{X})=\sigma(\mathbf{X}-x_{n})=\left(\mathbf{1}_{1:n}^T, -\mathbf{1}^T_{n+1:N} \right) = - \sigma(x_{n+1}-\mathbf{X})$; while for $\wl{1}=-1$ and $\bl{1}=x_{n+1}$ we have $\xl{L=2} (\mathbf{X})=\sigma(x_{n+1}-\mathbf{X})$.
And for $\wl{1}=1$ and $ \bl{1}=-x_N$, we have $ \xl{2}(\mathbf{X})= \mathbf{1}$ while for $\wl{1}=-1$ and $ \bl{1}=x_1$ we have $\xl{2}(\mathbf{X})=\mathbf{1}$. 
So for $L=2$ with symmetric or sign activation, \eqref{eq:L=234lyr_max_constr} is unchanged if $\wl{1} \in \{-1,1\}$ is restricted to be $1$.
\end{proof}

\iftoggle{long}{
\begin{remark}\label{rk:f_sjkwrittenout}
  The function in \Cref{lemma:L234constr_recursive} can be written out as follows.
   For $L=2$ layers: 
\begin{equation}
\begin{aligned} 
    f^{(1)}_{\mathbf{s},\mathbf{j}}(x_n) = \lambda_n \sigma \left(s^{(1)}\left( x_n - x_{j_1} \right)    \right)  .
\end{aligned}
\end{equation}
  
  For $L=3$ layers:
  \begin{equation}
  \begin{aligned} 
 f^{(2)}_{\mathbf{s},\mathbf{j},k}(x_n) = \sigma \left( s^{(2)} \left( \sigma \left( s^{(1)}\left( x_n - x_{j_1} \right)  \right)   -\sigma \left(s^{(1)}\left( x_{j_2}   - x_{j_1} \right)  \right) \right) \right).
  \end{aligned}
  \end{equation}

  For $L=4$ layers: 
\begin{equation}
  \begin{aligned}
       f^{(3)}_{\mathbf{s},\mathbf{j},k}(x_n)=\sigma \left( s^{(3)} \left(\sigma \left( s^{(2)} \left(\sigma \left( s^{(1)} \left( x_n - b \right) \right) - \sigma \left( s^{(1)} \left( x_{j_2} - b \right) \right) \right) \right) - \\
      \sigma \left( s^{(2)} \left(\sigma \left( s^{(1)} \left( x_{j_3} - b \right) \right) - \sigma \left( s^{(1)} \left( x_{j_2} - b \right) \right) \right) \right) \right) \right) \Big ) \Big ) .
  \end{aligned}
  \end{equation}
  where if $k=0$ then $b = x_{j_1}$ and otherwise if $k=1$ then $b = 2x_{j_2} - x_{j_1}$.
\end{remark}
}
\iftoggle{long}{
\begin{lemma}
For $L=2$ layers, the maximization constraint in \eqref{gen_dual_gen_arch_outer_lb_LL} is equivalent to
\begin{equation}\label{eq:L=2lyr_max_constr}
  \begin{aligned} 
\left | \sum_{n=1}^N \lambda_n \sigma \left(w^{(1)}\left( x_n - x_{j_1} \right)    \right) \right| &\leq \beta: w^{(1)} \in \{-1,1\}, j_1 \in [N] \\
      \mathbf{1}^T \lambda &= 0 
  \end{aligned}
  \end{equation}
 
\end{lemma}

\begin{proof}
Apply \Cref{lemma:max_constr_Llyrl} and its proof to $L=2$ layers.
\end{proof}

\begin{lemma}
For $L=3$ layers, the maximization constraint in \eqref{gen_dual_gen_arch_outer_lb_LL} is equivalent to
\begin{equation}\label{eq:L=3lyr_max_constr}
  \begin{aligned} 
\left | \sum_{n=1}^N \lambda_n \sigma \left( w^{(2)} \left( \sigma \left( w^{(1)}\left( x_n - x_{j_1} \right)  \right)   -\sigma \left(w^{(1)}\left( x_{j_2}   - x_{j_1} \right)  \right) \right) \right) \right| &\leq \beta: w^{(1)},w^{(2)} \in \{-1,1\}, j_1, j_2 \in [N] \\
      \mathbf{1}^T \lambda &= 0 
  \end{aligned}
  \end{equation}
 
\end{lemma}

\begin{proof}
Apply \Cref{lemma:max_constr_Llyrl} and its proof to $L=3$ layers.
\end{proof}

\begin{lemma}
For $L=4$ layers, the maximization constraint in \eqref{gen_dual_gen_arch_outer_lb_LL} is equivalent to
\begin{equation}\label{eq:Llyr_max_constr}
  \begin{aligned}
      \left |\sum_{n=1}^N \lambda_n \sigma \left( w^{(3)} \left(\sigma \left( w^{(2)} \left(\sigma \left( w^{(1)} \left( x_n - b \right) \right) - \sigma \left( w^{(1)} \left( x_{j_2} - b \right) \right) \right) \right) - \\
      \sigma \left( w^{(2)} \left(\sigma \left( w^{(1)} \left( x_{j_3} - b \right) \right) - \sigma \left( w^{(1)} \left( x_{j_2} - b \right) \right) \right) \right) \right) \right) \right ) \Big ) \Big | \leq \beta \\
      \mbox { for all } w^{(l)} \in \{-1,1\} \mbox { and  } j_l \in [N] \mbox{ for } l \in [3], \mbox { and } b \in \{x_{j_1}, 2x_{j_2} - x_{j_1} \} \\
      \mathbf{1}^T \lambda = 0 \\
  \end{aligned}
  \end{equation}
 
\end{lemma}

\begin{proof}
Apply \Cref{lemma:max_constr_Llyrl} and its proof to $L=4$ layers.
\end{proof}
}

\begin{lemma}\label[lemma]{lemma:deeplasso_without_simp}
	Let $\Amat{}$ be a matrix whose set of columns is the deep library.
 \iftoggle{long}{Let $ \mathbf{A_+}, \mathbf{A_-} \in \mathbb{R}^{N \times N}$ be defined by $\mathbf{{A_+}}_{i,j}=\sigma(x_i-x_j), \mathbf{{A_-}}_{i,j}=\sigma(x_j-x_i)$.}
	Replace the maximization constraint in  \eqref{gen_parallel_dual} with \eqref{eq:L=234lyr_max_constr}.
 The dual of 
\eqref{gen_parallel_dual} then
  is	\begin{equation}\label{deepgen_lasso_sol}
		\begin{aligned}
  \iftoggle{long}{\min_{\mathbf{z_+}, \mathbf{z_-} \in \mathbb{R}^N, \xi \in \mathbb{R}} & \mathcal{L}_{\mathbf{y}}(\mathbf{ A_+ z_+} +  \mathbf{A_- z_- } + \xi \mathbf{1} )+ \beta (||\mathbf{z_+}||_1 + ||\mathbf{z_-}||_1), 
			\quad \text{ where }                                      & \xi=0 \text{ if }  c_1=c_2=0,}
			\min_{\mathbf{z}, \xi \in \mathbb{R}} & \mathcal{L}_{\mathbf{y}}(\mathbf{ A z} + \xi \mathbf{1} )+ \symbeta \|\mathbf{z}\|_1, 
			\quad \text{ where }                                      & \xi=0 \text{ if }  c_1=c_2=0.
		\end{aligned}
	\end{equation}
\end{lemma}
\iftoggle{long}{\proofsketch{
To take the bidual of \eqref{equ:nn_train_signdetinput_rescale} we take the dual of the dual problem \eqref{L=2_d=1_dual} by minimizing its Lagrangian. Intuitively, the primal and bidual resemble forms: the training loss $\mathcal{L}_{\mathbf{y}}$ that appeared in the primal \eqref{equ:nn_train_signdetinput_rescale} returns as is in the bidual, and like the primal the bidual contains $l_1$ penalties.
}
}

\begin{proof}
\iftoggle{long}{By the definition of $A_+, A_-$ and lemma \ref{lemma:dual_finite}, }
Problem \eqref{gen_parallel_dual}  can be written as
	\begin{equation} \label{L=2_d=1_dual_as_min}
		\begin{aligned}
			-\min_{\lambda \in \mathbb{R}^N}  \mathcal{L}_{\mathbf{y}}^*( \lambda )                        
			\quad \text{ s.t. } \quad
                                     \lambda^T \mathbf{1} = 0  \text{ if } \negslope \neq 0 \text{ or } \posslope \neq 0, \mbox{ and } | \lambda^T \Amat{}| \leq \symbeta \mathbf{1}^T
		\end{aligned}.
	\end{equation}
	 The Lagrangian of the negation of \eqref{L=2_d=1_dual_as_min} with bidual variables $\mathbf{z}, \xi$ is

	\begin{equation} \label{bidual_Lag}
		\begin{aligned}
			L(\lambda, \mathbf{z},\xi)   
         & = \mathcal{L}_{\mathbf{y}}^*(\lambda) - \lambda^T( \mathbf{A z} + \xi \mathbf{1}) - \symbeta \|\mathbf{z}\|_1,
         \text{ where } \xi=0 \text{ if } \negslope=\posslope=0.
		\end{aligned}
	\end{equation}

 \iftoggle{long}{ , e.g. if the activation $\sigma(x)$ is a sign or threshold.}
Equation \eqref{bidual_Lag} holds because the constraint  $|\lambda^T \Amat{}| \leq \symbeta\mathbf{1}^T$ 
\iftoggle{long}{in \eqref{L=2_d=1_dual_as_min}}
i.e.,
\iftoggle{long}{$2N$ scalar constraints}
 $\lambda^T \Amat{} - \symbeta\mathbf{1}^T \leq \mathbf{0}^T, -\lambda^T \Amat{} - \symbeta\mathbf{1}^T \leq \mathbf{0}^T$, appears in the Lagrangian as $\lambda^T \Amat{} \left(\mathbf{z^{(1)}}-\mathbf{z^{(2)}} \right) -  \symbeta\mathbf{1}^T \left(\mathbf{z^{(1)}}+\mathbf{z^{(2)}} \right)$ with bidual variables $\mathbf{z^{(1)}}, \mathbf{z^{(2)}}$, which are combined into one bidual variable $\mathbf{z} = \mathbf{z^{(1)}}-\mathbf{z^{(2)}}$. This makes $\mathbf{z^{(1)}}+\mathbf{z^{(2)}} = \|\mathbf{z}\|_1$. Changing variables $\mathbf{z}'=-\mathbf{z}, \xi' = -\xi$ gives \eqref{bidual_Lag}. Since $\mathcal{L}^{**}=\mathcal{L}$ \cite{Borwein:2000}, $\inf_{\lambda} L(\lambda, \mathbf{z},\xi) = -\mathcal{L}_y - \symbeta\|\mathbf{z}\|_1$ and negating its maximization gives \eqref{deepgen_lasso_sol}.
\iftoggle{long}{
This is equivalent to using one bidual variable $\mathbf{z} \in \mathbb{R}^N$, which can be split into a difference of nonnegative vectors $\mathbf{z^{(1)}},\mathbf{z^{(2)}}$: $\mathbf{z} = \mathbf{z^{(1)}}-\mathbf{z^{(2)}}$, with $||\mathbf{z}||_1 = \mathbf{z^{(1)}}+\mathbf{z^{(2)}}$. }
%
\iftoggle{long}{Since $\mathcal{L}^*$ is convex and lower semicontinuous (\Cref{assump:Loss_func}), $\mathcal{L}^{**}=\mathcal{L}$ \cite{Borwein:2000}. 
By Fenchel duality \cite{boyd:2004} and  $\mathcal{L}^{**}=\mathcal{L}$ (\Cref{assump:Loss_func}, \cite{Borwein:2000}), the dual of \eqref{L=2_d=1_dual} is \eqref{gen_lasso_sol}.
}
\end{proof}

\begin{definition}[Parameter unscaling]\label{parameterunscale}
     Let $\gamma_i =  \left|\alpha_i\right|^{\frac{1}{\effL{}}}$.
    For ReLU, absolute value, or leaky ReLU, parameter {\unscal{}ing} is the following transformation. For \symmeterized{} networks, first change variables as ${\Wul{i}{l}}' = \frac{1}{\sqrt{2}} \Wul{i}{l}$ and $ {\bul{i}{l}}'=\frac{1}{\sqrt{2}^{l}}\bul{i}{l}$  for $l \in [2]$, $\alpha_i' = 2\alpha_i$. Then, for all architectures, change variables as $q' = \mathrm{sign}(q)\gamma_i$ for $q \in \regparams^{(i)}$. For parallel networks, change variables as ${\bul{i}{l}}'= \bul{i}{l}  \left(\gamma_i\right)^l $. For tree networks, change variables as ${\mathbf{b}^{\mathbf{u}}}' = \mathbf{b}^{\mathbf{u}} \left(\gamma_i\right)^l $ for $\mathbf{u} \in \mathcal{U}$ with length $l$ such that $u_1 =i$.  
     For sign and threshold activations, parameter unscaling is the transformation $ \alpha_i' = \mathrm{sign}(\alpha_i )\sqrt{|\alpha_i|}$ and ${s^{(i,L-1)}}' = \sqrt{|\alpha_i|}$ for parallel nets, and ${s^{(i)}}' =  \sqrt{|\alpha_i|}$ for tree nets. 
\end{definition}

\subsection{Proofs of results in \Cref{sec:deepnarrownets}
 }\label{app:deepReLU}

\begin{proof}[Proof of \Cref{lemma:ReLUsaturates}]
    The \lasso{} equivalence follows from a similar argument as the proof of \Cref{thrm:lasso_deep_1nueron_ReLU}. The \lasso{} features follow from \Cref{lemma:ReLUoptb} and \Cref{lemma:ReLUsatperlayer} with $l=L$.
\end{proof}

\begin{proof}[Proof of \Cref{lemma:numfeatures}]
    By the proof of \Cref{lemma:numKinks} and a similar argument as \Cref{thrm:lasso_deep_1nueron_ReLU}, the training problem is equivalent to a \lasso{} problem. Moreover \Cref{lemma:numKinks} and choice of $n^{(1)},\cdots, n^{(L-1)} \in [N]$ give the number of possible $\bl{1}$ for given $\wl{1},\cdots,\wl{L-1}$ as $O\left(N^{L-1}2^{L} {L}!\right)$. Note that by \eqref{eq:xl} and \eqref{eq:xltilde}, $\bl{1}$ determines all other $\bl{l}$. Finally, there are $2^L$ possibilities of $\wl{l} \in \{-1,1\}$.
    The ReLU result follows from \Cref{lemma:ReLUsaturates}.
\end{proof}
 \begin{proof}[Proof of \Cref{thrm:lasso_deep_1nueron_ReLU}]
By \Cref{lemma:deeplasso_without_simp}, problem \eqref{deepgen_lasso_sol} is a lower bound on the training problem \eqref{eq:nonconvex_generic_loss_gen_L_intro}, where the \lasso{} features are all vectors in the deep library. Let $(\mathbf{z}^*,\xi^*)$ be a \lasso{} solution.
From \Cref{def:FeatDicStdNN}, it can be seen that $\Amat{}\mathbf{z}^*{+}\xi^*{=}\sum_{i}z^*_i \mathbf{A}_i+\xi^*{=}\sum_{i}\alpha_i \Xul{i}{L}(\mathbf{X}){+}\xi^*\\{=}\nn{L}{\mathbf{X}}$ and $\|\mathbf{z}^*\|_1{=}\|\boldsymbol{\alpha}\|_1$. Therefore a reconstructed neural net
achieves the same objective in the rescaled training problem, as $(\mathbf{z}^*,\xi^*)$ does in the \lasso{} objective. Parameter \unscal{}ing (\Cref{parameterunscale}) makes them achieve the same objective in the training problem (see \Cref{rk:unscalingOuter}).
Therefore the \lasso{} problem in \Cref{thrm:lasso_deep_1nueron_ReLU} is equivalent to the training problem.
%
%
%
\iftoggle{long}{
  For $L=3$, 
\begin{equation}\label{eq:A_L3ReLU}
     \Amat{}_{\mathbf{s}, i, \mathbf{j},k=0} = \sigma \left( s^{(2)} \left( \sigma \left( s^{(1)} \left( x_i - x_{j_1} \right)  \right)   -\sigma \left(s_1 \left( x_{j_2}   - x_{j_1} \right)  \right)  \right) \right)
\end{equation}
For $L=4$, $\Amat{}= \left [\Amat{}^0, \Amat{}^1 \right]$ where
\begin{equation}\label{eq:A_L4ReLU}
  \begin{aligned}
    \Amat{}_{\mathbf{s},i, \mathbf{j},k} = \sigma \left( s^{(3)} \left(\sigma \left( s^{(2)} \left(\sigma \left( s^{(1)} \left( x_i -  b \right) \right) - \sigma \left( s^{(1)} \left( x_{j_2} -  b \right) \right) \right) \right) - \\
      \sigma \left( s^{(2)} \left(\sigma \left( s^{(1)} \left( x_{j_3} - b\right) \right) - \sigma \left( s^{(1)} \left( x_{j_2} -  b \right) \right) \right) \right) \right) \right) \Big) \Big ),
  \end{aligned}
  \end{equation}
  where $b = x_{j_1}$ if $k=0$ and $b = 2x_{j_2}-x_{j_1}$ if $k=1$. 
Observe that \Cref{fig:ReLU_evol_simp} illustrates the third, sixth, and ninth rows of \Cref{fig:ReLU_evol}, with {\kink} labels as found in the proof of \Cref{lemma:ReLU_3lyr_kinkpts_from_fig} and its proof, respectively. With $a,b,c$ as defined in \Cref{lemma:ReLU_3lyr_kinkpts_from_fig},  \eqref{eq:A_L3ReLU} and \eqref{eq:A_L4ReLU} show that the plots in the rows of \Cref{fig:ReLU_evol} correspond to the dictionary columns for $L=2,3$, and $4$ layers.  
}
%
%
 \end{proof}

 \begin{proof}[Proof of \Cref{lemma:parreconstructionworks}]
By \Cref{thrm:lasso_deep_1nueron_ReLU} and its proof, the reconstructed neural net achieves the same objective as the \lasso{} problem, which is equivalent to the training problem. So the reconstructed neural net is optimal.
\end{proof}

\begin{proof}[Proof of \Cref{theorem:symmeterizednetworks}]
    By considering each component of $\bl{l}$ separately in the proof of \Cref{lemma:max_constr_Llyrl}, we see that optimal bias parameters for a \symmeterized{} network can include \datafeature{}s. The \lasso{} equivalence follows from a similar argument as the proof of \Cref{thrm:lasso_deep_1nueron_ReLU}. 
\end{proof}

\begin{proof}[Proof of \Cref{ReLU2neuroninformal}]
    Follows from \Cref{theorem:symmeterizednetworks} and evaluating the features. See proof of \Cref{fig:ReLUfeatdiagram} below.
\end{proof}

 \begin{proof}[Proof of \Cref{absvalinformal}]
  For $L {\in} \{2,3\}$, apply \Cref{lemma:numfeatures}, \Cref{thrm:lasso_deep_1nueron_ReLU} where $\sigma(x)=|x|$ 
  and \Cref{def:ReLU dic func}.
  For $L{=}4$,
    continue the same line of argument as \Cref{lemma:max_constr_Llyrl} by plugging in  $\bl{L-2}{=}{-}\xl{L-2}_{n^{(L{-}2)}}\wl{L{-}2}$ into \eqref{eq:lambda^TX_L-1}, expanding $\xl{L{-}2}{=}\sigma\left(\xl{L{-}3}\wl{L{-}3}{+}\bl{L{-}3}\right)$, and observing that the \kink{}s of $\lambda^T\xl{L}$ as a function of $\bl{L-3}$ include \datafeature{}s, and hence these can be optimal bias parameters.
    Plugging in these \datafeature{} biases in to $\xl{4}$ gives the features and reflections. Continue repeating a similar argument for $L>4$. The reconstruction and \lasso{} equivalence follows from a similar argument as the proof of \Cref{thrm:lasso_deep_1nueron_ReLU}. 
\end{proof}

\begin{remark}
    In the proof of \Cref{absvalinformal} for $L\geq4$, we only used one of the possible forms of an optimal $\bl{L-2}$ specified by \Cref{lemma:max_constr_Llyrl} for absolute value activation, and we only specified one of the possible forms of a \kink{} in $\lambda^T\xl{L}$ as a function of $\bl{L-3}$. So a $4$-layer network may have additional features not specified.
\end{remark}

\begin{proof}[Proof of \Cref{ReLUinformal}]
    Follows from \Cref{lemma:ReLUsaturates} and \Cref{thrm:lasso_deep_1nueron_ReLU} where $\sigma(x)=(x)_+$ is monotone.
\end{proof}

\begin{proof}[Proof of \Cref{Amat2lyrs}]
    Follows from \Cref{thrm:lasso_deep_1nueron_ReLU} for $L=2$.
\end{proof}

\begin{remark}\label{absvalfeat}

For $b_1, b_2 \geq 0$, 
\begin{equation*}
    \begin{aligned}
        \left| \left| x  -  b_1 \right| - b_2\right| 
        &= \begin{cases}
             b_1 - b_2 - x &\mbox{ if } x \leq  b_1  - b_2 \\
            x - (b_1 - b_2) &\mbox{ if }  b_1 - b_2 \leq x \leq  b_1 \\
            b_1 + b_2 - x &\mbox{ if } b_1 \leq x \leq  b_1 + b_2 \\
            x - (b_1 + b_2) &\mbox{ if } x \geq s_1 b_1 + b_2 
        \end{cases}\\
        &= \mathcal{W}_{b_1-b_2,b_1}.
    \end{aligned}
\end{equation*}

\end{remark}

\begin{proof}[Proof of \Cref{def:ReLU dic func}, \Cref{fig:absvalfeattikz}, \Cref{fig:ReLUfeatdiagram}]
    Follows from direct computation of \Cref{eq:std_nn} and application of \Cref{absvalfeat}. In particular, for ReLU \symmeterized{} networks, it can be verified that for $\wl{1}=(-1,1),\wl{2}=(-1,-1)$; $\wl{1}=(-1,1),\wl{2}=(1,1)$; $\wl{1}=(1,-1),\wl{2}=(-1,-1)$; and $\wl{1}=(1,-1),\wl{2}=(1,1)$, the deep library features can contain reflections. Note ReLU \symmeterized{} features are consistent with \Cref{fig:ReLU2neurons}.
\end{proof}

 \begin{remark}\label[remark]{rk:unscalingOuter}
    For sign-determined activations, by \Cref{remk:signdet_inweights0}, the inner weights can be unregularized. So, reconstructed parameters (as defined in \Cref{def:reconstructfunc}) that are unscaled according to \Cref{parameterunscale} achieve the same objective in the training problem as the optimal value of the rescaled problem. 
 \end{remark}

\begin{proof}[Proof of \Cref{lemma:absvalReLUskipconnection}]
 Since $|x| {=} 2(x)_+ {+} x$, we have $\sum_{i{=}1}^{m_2}  \left|\mathbf{X}\Wul{i}{1}{+}\bul{i}{1}\right| \alpha_i {+} \mathbf{X}\skipslope {+} \xi \\ {=}\sum_{i=1}^{m_2}  2\left(\mathbf{X}\Wul{i}{1}{+}\bul{i}{1}\right)_+ \alpha_i{+}\mathbf{X}\left( \skipslope {-} \sum_{i{=}1}^{m_2}  \Wul{i}{1}\alpha_i \right) {-}\sum_{i{=}1}^{m_2}\bul{i}{1}\alpha_i {+} \xi $. Given a solution ${\Wul{i}{l}}^*, {\bul{i}{l}}^*, {\alpha_i}^*, {\xi}^*, {\skipslope}^*$ to the training problem for absolute value activation with skip connection, the parameters ${\Wul{i}{l}}{=}{\Wul{i}{l}}^*, {\bul{i}{l}}^*{=}{\bul{i}{l}}^*, {\alpha_i}=2{\alpha_i}^*, {\xi}{=}{\xi}^*{-}\sum_{i=1}^{m_2}{\bul{i}{l}}^*{\alpha_i}^*,  {\skipslope}{=}{\skipslope}^*\\{-}\sum_{i=1}^{m_2}{\Wul{i}{1}}^* {\alpha_i}^*$ achieve the same objective in the training problem for ReLU activation with skip connection. Conversely, since $(x)_+ {=} \frac{|x|+x}{2}$, we have $ \sum_{i{=}1}^{m_2}  \left(\mathbf{X}\Wul{i}{1}{+}\bul{i}{1}\right)_+ \alpha_i {+} \mathbf{X}\skipslope {+} \xi \\ {=} \frac{1}{2}\sum_{i{=}1}^{m_2} \left|\mathbf{X}\Wul{i}{1}{+}\bul{i}{1}\right|\alpha_i {+} \mathbf{X}\left( \skipslope {+} \frac{1}{2} \sum_{i=1}^{m_2} \Wul{i}{1}\alpha_i\right) {+} \frac{1}{2}\sum_{i=1}^{m_2} \bul{i}{1} \alpha_i {+} \xi $. Given a solution \\ ${\Wul{i}{l}}^*, {\bul{i}{l}}^*, {\alpha_i}^*, {\xi}^*, {\skipslope}^*$ to the training problem for ReLU activation with skip connection, the parameters ${\Wul{i}{l}}{=}{\Wul{i}{l}}^*, {\bul{i}{l}}^*{=}{\bul{i}{l}}^*, {\alpha_i}=\frac{1}{2}{\alpha_i}^*, {\xi}{=}{\xi}^*{+}\frac{1}{2}\sum_{i=1}^{m_2}{\bul{i}{l}}^*{\alpha_i}^*, \\{\skipslope}{=}{\skipslope}^*{+}\frac{1}{2}\sum_{i=1}^{m_2}{\Wul{i}{1}}^* {\alpha_i}^*$ achieve the same objective in the training problem for absolute value activation with skip connection. Therefore the problems achieve the same optimal value and we have given a map between the solutions.     
\end{proof}

\section*{\secnameSignAct}\label{app:deepSign}

In this section, we assume $\sigma(x)=\text{sign}(x)$. First we discuss parallel architectures. 


\begin{definition}\label[definition]{def:deepH(Z)}
Define the \emph{hyperplane arrangement set} for a matrix $\mathbf{Z} \in \mathbb{R}^{N\times m }$ as
%
\begin{align}\label{eq:deepHdef}
 \mathcal{H}(\mathbf{\mathbf{Z}}) :=  \left \{ \sigma \left(\mathbf{Z} \mathbf{w} + b\mathbf{1} \right): \mathbf{w} \in \mathbb{R}^{m},  b \in \mathbb{R} \right \} \subset \{-1,1\}^N.
\end{align} 
%
%

Let $S_0$ be the set of columns of $\mathbf{X}$. Let $\{S_l\}_{l=1}^{L-1}$ be a tuple of sets satisfying $S_l \subset  \mathcal{H}(\left[S_{l-1}\right])$ and $|S_l |= m_l$. Let ${A_L}_{par}(\mathbf{X})$ be the union of all possible sets $S_{L-1}$. 
\end{definition}
In \Cref{def:deepH(Z)}, since $m_{L-1}=1$ in a parallel network, $S_{L-1}$ contains one vector.
The set $\mathcal{H}(\mathbf{Z})$ denotes all possible $\{1,-1\}$ labelings of the samples $\{\mathbf{z}_i\}_{i=1}^N$ by a linear classifier. Its size is upper bounded by
$
	|\mathcal{H}(\mathbf{Z})| \le 2 \sum_{k=0}^{r-1} {N-1 \choose k}\le 2r\left(\frac{e(n-1)}{r}\right)^r \leq 2^N,
$
where $r:=\mbox{rank}(\mathbf{Z})\leq \min(N,m)$ \cite{cover1965geometrical, stanley2004introduction}. 

\iftoggle{long}{

Next, we define the $L$-layer hyperplane arrangement matrix recursively. Let $\mathcal{H}^{(0)}_{S_0}(\mathbf{X}) = \mathbf{X}$.
For $l \in [L]$, given a matrix $\mathcal{H}^{(l)}_{\{S_j\}_{j=1}^{l}}(\mathbf{X)} \in \mathbb{R}^{N \times |S_l|}$ and a subset of its hyperplane arrangements $S_{l+1} \subset \mathcal{H}\left(  \mathcal{H}^{(l)}_{\{S_j\}_{j=1}^{l}}\left(\mathbf{X}\right) \right)$, let  $\mathcal{H}^{(l+1)}_{\{S_j\}_{j=1}^{l+1}}(\mathbf{X)} = [S_{l+1}] \in \mathbb{R}^{N \times |S_{l+1}|}$ be the concatenation of those arrangements. 
Taking all possible such sequences of sets $\{S_j\}_{j=1}^L$, 
define the $L$-layer dictionary as 

	\begin{equation}\label{def:A_L(X)}
		A_L(X) := \bigcup_{\{S_j:j\in[L-3],|S_j|\leq m_j \}} S_L.
	\end{equation}

the concatenation of their corresponding $L$-layer hyperplane arrangement matrices. \\

Alternatively:

For a vector set $H=\{h_1, \dots, h_{|H|} \} \subset \mathbb{R}^d $ and an ordered subset $S \subset H$, let $[H]_S \in \mathbb{R}^{d \times |S|}$ be the horizontal concatenation of the vectors in $S$ in order. 
For a matrix $\mathbf{Z}$, let $\tilde{H}_S(\mathbf{Z}) =[[\mathcal{H}(\mathbf{Z})]_{S},  \mathbf{1}] $ 
\iftoggle{long}{denote the matrix formed by concatenating $[\mathcal{H}(\mathbf{Z})]_{S}$ with a column of ones}.

Now, denote $\tilde{H}^{(L-1)}_{S_j:j\in[L-1]}(\mathbf{X}) = \tilde{H}^{(L-1)}_{S_{L-1}} \circ \tilde{H}^{(L-1)}_{S_{l-2}} \circ \cdots \circ \tilde{H}^{(1)}_{S_1}(\mathbf{X})$ as the composition of $\tilde{H}_S$ with itself $L-1$ times, where the $j^{\text{th}}$ time uses a subset $S_{j-1} \subset \mathcal{H}\left([(\tilde{H}^{(j-1)}_{S_j} \circ \cdots \circ \tilde{H}^{(1)}_{S_1})(\mathbf{X}),\mathbf{1}]\right)$.

	We define the $L$-layer dictionary set,

	\begin{equation}\label{def:A_L(X)}
		A_L(X) := \bigcup_{\{S_j:j\in[L-3],|S_j|\leq m_j \}} \mathcal{H}(\tilde{H}^{(l)}_{S_j:j\in[l]}(X)).
	\end{equation}

\begin{remark}
	The union in equation \ref{def:A_L(X)} is taken over all possible sequences of $S_1, \dots, S_{L-3}$, where $S_j$ is constrained by the previous $S_i$ for $i < j$, which in turn depends on $X$.
	Propositions \ref{prop:m_1_more_layers_same_size}, \ref{prop:m_1_more_layers} will show that for a fixed number of training samples $N$, regardless of their actual values $x_1,\dots,x_N$, $A_L(X)$ is the same set and with easily enumerable elements.
\end{remark}
}

\begin{lemma}\label[lemma]{lemma:deepsign_rewritedual}
	The lower bound problem \eqref{gen_parallel_dual} 
 is equivalent to

	\begin{equation}\label{deepdsignpar_gen}
		\begin{aligned}
			\max_{\lambda} & -  \mathcal{L}^*_{\mathbf{y}}( \lambda ),                                  
			               \quad  \mathrm{ s.t. } \max_{\mathbf{h} \in {A_L}_{par}(\mathbf{X})} |\lambda^T \mathbf{h} | \leq \beta.
		\end{aligned}
	\end{equation}

\end{lemma}

\iftoggle{long}{
\proofsketch{
Consider the maximization constraint of the dual problem \eqref{gen_dual}. We replace the layer weight variables $W^{(i,l)}$ by all vectors that $\sigma(x^{(L-2)}W^{(L-1)})$ can possibly be over all feasible $W^{(i,l)}$, and collect them into the set $A_L(X)$ as their feasible set.
}
}

\begin{proof}
For $l \in [L]$, there is $S_{l-1} \subset \mathcal{H}\left(\mathbf{X}^{(l-1)}\right)$ with $\mathbf{X}^{(l)} = [S_{l-1}]$. Recursing over $l \in [L]$ 
gives $ \left \{ \mathbf{X}^{(L)} : \params \in \paramspace \right \} = {A_L}_{par}(\mathbf{X})$. So, the constraints of \eqref{gen_parallel_dual} and \eqref{deepdsignpar_gen} are the same.
\iftoggle{long}{
$\{\sigma\left(\mathbf{X^{(L-2)} W^{(1,L-1)}}\right): \theta \in \Theta_s\} = \bigcup_{\theta \in \Theta_s} \mathcal{H}(\mathbf{X}^{(L-2)}) = \bigcup S_{L-1} $ and $\mathbf{X}^{(L-2)} = [\sigma(\mathbf{{X}^{(L-3)}W^{(L-2)}}),\mathbf{1}]$

	For $i \in [m_{L-2}]$, denote $W^{(L-2)}_{i}$ as the $i^{\textrm{th}}$ column of  $W^{(L-2)}$, corresponding to the weights of the $i^{\textrm{th}}$ neuron in layer $L-2$. If the amplitude parameters of the activation functions are all one, then for some sets $S_1, \dots, S_{[L-3]}$, the input to the second to last layer is $X^{(L-2)} = [\sigma_{s=1}(X^{(L-3)}W^{(L-2)}), \mathbf{1}] = [ \sigma_{s=1}(X^{(L-3)}W^{(L-2)}_1), \dots,  \sigma_{s=1}(X^{(L-3)}W^{(L-2)}_{m_{L-2}}), \mathbf{1}] = [[\mathcal{H}(X^{(L-3)})]_{S_{L-3}},  \mathbf{1}] = \tilde{H}_{S_{L-3}}(X^{(L-3)})=\tilde{H}_{S_{L-3}}(\tilde{H}_{S_{L-4}}(X^{(L-4)}))=\cdots = \tilde{H}^{(L-3)}_{S_j:j\in[L-3]}(X^{(0)})=\tilde{H}^{(L-3)}_{S_j:j\in[L-3]}(X)$ can be written in terms of $L-3$ function compositions on the input matrix $X$. Given a fixed feasible $X^{(L-2)}$ and its corresponding sets $S_j:j \in [L-3]$, the set of all possible vectors $\sigma(X^{(L-2)}w)$ generated over all possible vectors $w \in \mathbb{R}^{1+m_{L-2}}$ is, by definition of $\mathcal{H}$, equal to $\mathcal{H}(\tilde{H}^{(l)}_{S_j:j\in[l]}(X))$. So the set of all possible $\sigma(X^{(L-2)}w)$ vectors over all vectors $w$ and all possible, feasible $X^{(L-2)}$ (generated by all possible  $W^{(i,l)}$) is their union $A_L(X)$, as defined in equation \ref{def:A_L(X)}. Therefore we can replace $\sigma(X^{(L-2)W^{(L-1)}})$ in the maximization constraint of the dual problem \eqref{gen_dual} by vectors $h \in A_L(X)$ to get the equivalent problem \ref{dsignpar_gen}.
}
\end{proof}
\begin{remark}\label[remark]{deepHstartswith1}
   Without loss of generality (by scaling by $-1$), assume that the vectors in $\mathcal{H}(\mathbf{Z})$ start with $1$. Under this assumption, \Cref{lemma:deepsign_rewritedual} still holds. 
\end{remark}
\begin{lemma}\label[lemma]{lemma:deeplasso_sign_L}
	Let $\Amat{}=[{A_L}_{par}(\mathbf{X})]$. The lower bound problem \eqref{deepdsignpar_gen} is equivalent to 
	\begin{equation}\label{deepsignLLasso_gen}
		\begin{aligned}
			\min_\mathbf{z} & \quad \mathcal{L}_{\mathbf{y}}( \mathbf{Az}) + \beta||\mathbf{z}||_1.
		\end{aligned}
	\end{equation}
\end{lemma}
\begin{proof}
	Problem \eqref{deepdsignpar_gen} is the dual of \eqref{deepsignLLasso_gen}, and since the problems are convex with feasible regions that have nonempty interior, by Slater's condition, strong duality holds \cite{boyd:2004}. 
\end{proof}
\begin{remark}
 The set $A_{L,par} $ consists of all possible sign patterns at the final layer of a parallel neural net, up to multiplying by $-1$.
\end{remark}

\begin{lemma}\label[lemma]{prop:deepreconstruct_primal_L}
	Let $\Amat{}$ be defined as in \Cref{lemma:deeplasso_sign_L}. Let $\mathbf{z}$ be a solution to \eqref{deepsignLLasso_gen}. Suppose $m_L \geq \| \mathbf{z} \|_0$. There is a parallel neural network 
 satisfying $\nn{L}{\mathbf{X}} =\mathbf{Az}$ which achieves the same objective in the rescaled training problem 
 as $\mathbf{z}$ does in \eqref{deepsignLLasso_gen}.
\end{lemma}
\begin{proof}
By definition of $A_{L,par}$ (\Cref{def:deepH(Z)}), for every $\Amat{}_i \in A_L(\mathbf{X})$, there are tuples 
$\left \{{{\Wul{i}{l}}} \right\}_{l=1}^{L-1}, \left\{{\bul{i}{l}}\right\}_{l=1}^{L-1}$ of parameters such that $\Amat{}_i = {\Xul{i}{L}} $.
Let $\mathcal{I} =\{i: z_i \neq 0\}$. For $i \in \mathcal{I}$, set $\alpha_i = z_i$.
This gives a neural net
 $\nn{L}{\mathbf{X}} = \sum_{i\in\mathcal{I}} \alpha_i \Xul{i}{L} = \mathbf{Az}$ with $|\mathcal{I}| \leq m_L$. 
%
%
%
 %
%
 \iftoggle{long}{
	Denote $W^{(l)}_i$ as the $i^{\textrm{th}}$ column of $W^{(l)}$, which contains the weights for the $i^{\textrm{th}}$ neuron of layer $l$.

 Let $z' \in \mathbb{R}^{m^L}$ be $z$ with $m_L - m^*\geq 0$ of its zero elements removed. Let $A' \in \mathbb{R}^{n \times m_L}$ be the matrix consisting of columns of $A$ that have the same index as the indices of $z$ that were used to make $z'$. Since each $i^{\textrm{th}}$ column $A'_i$ of $A'$ is in $A_L(X)$, there exists  a sequence of sets $S_j: j \in [L-1]$ and a sequence of matrices $ \tilde W^{(i,l-1)}: l \in [L-1]$ such that $\tilde X^{(i,l)} = \sigma([\tilde X^{(i,l-1)}, \mathbf{1}] \tilde W^{(l-2)}_i ) $ is a submatrix of $ \tilde{H}^{(l)}_{S_j:j\in[l]}(X)$ for $ l \in [L-2]$; in particular there exists a vector $\tilde W^{(i,L-1)} \in \mathbb{R}^{m_{L-2} }$ such that $A'_i = \sigma(\tilde X^{(i,L-2)} W^{(i,L-1)})$.
	So the parallel neural network,

	\begin{equation}
		f(X) =  
		\sum_{i=1}^{m_L}z'_i \sigma({{\tilde X^{(i,L-2)}}}  \tilde W^{(i,L-1)}) = \sum_{i=1}^{m_L}z'_i (A')_i=\sum_{i=1}^{|A_L|}z_i (A)_i=Az,
	\end{equation}

	achieves the same objective in the primal \eqref{equ:nn_train_signdetinput_rescale} as the Lasso problem \eqref{signLLasso_gen} with $\alpha=z',  W^{(i,l)}= \tilde{W}^{(i,l)} , X^{(L-2)}=\tilde X^{(L-2)}$. Note we can normalize all the weights $W^{(i,L-1)}_i$ to achieve  $||W^{(i,L-1)}_i||_F=1$ to satisfy the primal \eqref{equ:nn_train_signdetinput_rescale} constraints.
}
\end{proof}
\begin{remark}\label[remark]{rk:tree_reconstruct}
   \Cref{prop:deepreconstruct_primal_L} analogously holds for the tree network by a similar argument: by construction of ${A_{L}}_{tree}$, there is a neural net satisfying $f_L(\mathbf{X};\theta)=\mathbf{Az}$. 
\end{remark} 

%
\begin{proposition}\label[proposition]{prop:deeplasso_equiv_primal}
	For $L$-layer parallel networks with sign activation, the \lasso{} problem \eqref{deepsignLLasso_gen} problem and the original training problem are equivalent.
\end{proposition}
\begin{proof}
    By \Cref{lemma:deeplasso_sign_L}, the \lasso{} problem is a lower bound for the training problem. By the reconstruction in \Cref{prop:deepreconstruct_primal_L}
 (see \Cref{rk:unscalingOuter}), the lower bound is met with equality.
\end{proof}


\begin{definition}\label[definition]{def:ALtree}
%
 Recall the set $\mathcal{H}$ defined in \eqref{eq:deepHdef}. Define a matrix-to-matrix operator
	\begin{equation} 
		J^{(m)}(\mathbf{Z}) := \left[ \bigcup_{|S| = m} \mathcal{H}(\mathbf{Z}_S) \right].
	\end{equation}
For $L=2$, let ${A_L}_{\textrm{tree}}(\mathbf{X})  = \mathcal{H}(\mathbf{X})$ and for $L \geq 2$, let 
${A_L}_{\textrm{tree}}(\mathbf{X}) $ be the set of columns in $J^{(m_{L-1})}\circ \cdots \circ J^{(m_{2})}(\mathcal{H}(\mathbf{X})) $. \\

\end{definition} 

The columns of $J^{(m)}(\mathbf{Z})$ are all hyperplane arrangement patterns of $m$ columns of $\mathbf{Z}$.

\begin{lemma}
	For $L \geq 3$, the lower bound problem \eqref{deepgen_dual} for tree networks is equivalent to

	\begin{equation}\label{gen_dual_tree_A}
		\begin{aligned}
			\max_{z \in \mathbb{R}^N} & -\mathcal{L}_{\mathbf{y}}^*( \lambda ),          
			\quad \mathrm{ s.t. }             & \max_{\mathbf{h} \in {A_L}_{\textrm{tree}}(\mathbf{X})} |\lambda^T \mathbf{h}| \leq \beta.
		\end{aligned}
	\end{equation}
\end{lemma}

\iftoggle{long}{
\proofsketch{
	The proof follows from the recursive definition of the tree structure neural network. Note the standard and parallel networks lack this recursion.
}
}

\begin{proof}
Let $\mathbf{u}$ be a tuple such that $u_1=1$.
First suppose  $\mathbf{u}$ has length $L-2$. For all nodes $i$, $\left \{ \sigma \left( \mathbf{X}\mathbf{w}^{(\mathbf{u}+i)}+ b^{(\mathbf{u}+i)} \mathbf{1} \right): \mathbf{w}^{(\mathbf{u}+i)} \in \mathbb{R}^{d}, b^{(\mathbf{u}+i)} \in \mathbb{R} \right \}=\mathcal{H}(\mathbf{X})$ independently of any other sibling nodes $j \neq i$. So every $\mathbf{X}^{(\mathbf{u})}$ is the linear combination of $m_{2}$ columns in $\mathcal{H}(\mathbf{X})$, with the choice of columns independent of other $\mathbf{u}$ of the same length. Next, for all $\mathbf{u}$ of length $L-3$, 
 the set of all possible $\sigma \left(\mathbf{X}^{(\mathbf{u}+i)} + b^{(\mathbf{u}+i)} \mathbf{1} \right) $ is $J^{(m_2)}\left( \mathcal{H}(\mathbf{X}) \right)$.
 Repeating this for decreasing lengths of $\mathbf{u}$ until $\mathbf{u}$ has length $1$ gives $\Tilde{\mathbf{X}} = \sigma \left(\mathbf{X}^{(i)} + b^{(i)} \mathbf{1} \right)= {A_L}_{\textrm{tree}}(\mathbf{X})$.
 \iftoggle{long}{
 $f^{(l), \Theta^l}(X) = w^{(l)}_{m_l + 1}\mathbf{1} + \sum_{i=1}^{m_{l}} \sigma_{s^{l-1,i}}(f^{(l-1), \Theta^{l-1,i}})w^{(l)}_i \in \mathbb{R}^N $ and at the final layer, $f^{(L-1)}(X) = f^{(L-1), \Theta^{(L-1)}}(X) \in \mathbb{R}^N$. So given layer $l \in [L-2]$ and the set $Z$ of all possible vectors of the form $f^{(l, \Theta^{l-1})}(X)$, i.e., possible outputs at layer $l-1$, any possible $f^{(l)}(X)$ at the next layer $l$ is a linear combination of $m_l$ vectors from $Z$, which is a vector in $J(Z)$. And any linear combination of $m_l$ vectors from $Z$ is also a feasible output $f^{(l)}(X)$  at layer $l$. When $l=2$, by a similar argument as for the parallel networks, the set of all possible outputs at the final layer is $\mathcal{H}(X)$. By recursing, the set of all possible $f^{(L-1)}(X)$ becomes $J$ applied to $\mathcal{H}(X)$ for $L-2$ times.
 }
\end{proof}


\subsection{\nd{1} data}\label{appsec:1Ddatasignact}

In this section, we assume the data is \nd{1}. We will refer to a \switchsetname{} and a
\deepsign{} network defined in \Cref{sec:deepsign} and \Cref{sec:parnet}.

\begin{lemma}\label[lemma]{lemma:hvec_switch_prod}
    Let $m_1, m_2 \in \mathbb{N}, k \in [m_1 m_2]$. A sequence $\{h_i\}$ in $\{-1,1\}$ that starts with $1$ and switches $k$ times is the sum of at most $m_2$ sequences in $\{-1,1\}$ that switch at most $m_1$ times, and the all-ones sequence.
\end{lemma}

\begin{proof}
    Suppose  ${h_i}$ switches at $i_1 < \cdots < i_k$. Let $Q=\left \lceil \frac{k}{m_1} \right \rceil \leq m_2$. For $q \in \left[ Q \right]$, let $\left\{h^{(q)}_i\right\}$ be a sequence in $\{-1,1\}$ that starts with $(-1)^{q+1}$ and switches precisely at $i \in\{I_1 ,\cdots, I_k\}$ satisfying $i = j \mod{m_2}$, which occurs at most $m_1$ times. Let $s_i = \sum_{j} h^{(q)}_i$. Then $s_1 = \mathbf{1}\{ Q \mbox{ odd} \} \in \{h_1, h_1-1\}$. For $i > 1$,
    \begin{align*}
        s_{i} =
        \begin{cases}
            s_{i-1} + 2 &\mbox{ if } h^{(q)}_{i-1} = -1,  h^{(q)}_i = 1 \mbox{ for some } q \\
            s_{i-1} - 2 &\mbox{ if } h^{(q)}_{i-1} = 1,  h^{(q)}_i = -1 \mbox{ for some } q \\
            s_i &\mbox { else. }
        \end{cases}
    \end{align*}
    So $\{s_i\}$ is a sequence in $\{0,-2\}$ or $\{-1,1\}$ that changes value precisely at $i_1, \dots, i_k$. Therefore $\{s_i\}$ is either $\{h_i\}$ or $\{h_i-1\}$.
\end{proof}

\iftoggle{long}{
\begin{lemma}\label{lemma:hvec_switch_prod}
	Let $m_1,m_2 \in \mathbf{N} $. Let $k \in [m_1 m_2]$. 
 Let $h \in \switchmatrix{k}$.
There exists a set $S \subset \{-1,1\}^N$ of at most $m_2$ vectors each with at most $m_1$ switches
\iftoggle{long}{
of at most $ m_2$ $\switchmatrix{m_1}$ and $-\switchmatrix{m_1}$, 
}
such that $h= \sigma \left( \sum_{h' \in S} h' \right)$. 

\end{lemma}

\begin{proof}
Let $I_1, \cdots, I_k$ be the locations at which $h$ switches, such that $I_1 < \cdots < I_k$.
	There 
 are nonnegative integers $Q,R$ 
 such that $0 \leq R<2m_1$ and $k = (2m_1)Q+R$. 
	For $q \in [Q]$, let $h^{(q)} \in \switchmatrix{2m_1}$ with switches precisely at $ I_{1+2m_1(q-1)},\cdots,I_{(2m_1)q} $, and let $u^{(q)}= h^{(q)}-\mathbf{1}$. By \Cref{lemma:hdecomp}, for $q \in [Q]$, there exist ${h^{q}}^{(+)}, {h^{q}}^{(-)} \in \switchmatrix{m_1}$ such that $u^{(q)}={h^{q}}^{(+)}-{h^{q}}^{(-)} $. Let $h^{(Q+1)} \in \switchmatrix{2m_1}$ with switches precisely at $I_{1+(2m_1)q},\cdots, I_k $ and similarly define $ u^{(Q+1)}, {u^{(Q+1)}}^+, {u^{(Q+1)}}^-$. Let $S_{Q+1} = \left\{{h^{q}}^{(+)}, -{h^{q}}^{(-)}: q \in [Q] \right\}$.  
 \iftoggle{long}{
 Let $\mathcal{I}_r = \left\{I_{1+(2m_1)q},I_{2+(2m_1)q},\cdots, I_k \right\}$ be the remaining indices in $[k]$.
 Let $\mathcal{I}^+, \mathcal{I}^-$ be defined as in \Cref{lemma:hdecomp}, and let $h^+ = h(\mathcal{I}^+), h^-= h(\mathcal{I}^-)$.
 Let $S_q = \left \{h^+\left(\mathcal{I}^{(q)}\right): q \in [Q] \} \cup \{-h^-\left(\mathcal{I}^{(q)}\right): q \in [Q] \right\}$.}
 Let
 $
		\begin{aligned}
			S_r & =  
            \begin{cases}
                \emptyset &\mbox { if }  R = 0 \\
                \left \{ h^{(Q+1)} \right \} &\mbox { if } 1 \leq R \leq m_1 \\
                \left\{{h^{(Q+1)}}^{+}, - {h^{(Q+1)}}^{-} \right \} &\mbox { if } m_1 < R <2m_1 
            \end{cases}.
		\end{aligned}
$
 Let $S = S_q \cup S_r \subset \{-1,1\}^N$. Note $S$ only contains vectors with at most $m_1$ switches. Since $m_1 m_2 \geq k = 2m_1 Q + R$, and $m_2$ is an integer, $m_2 \geq 2Q + \lceil \frac{R}{m_1} \rceil = |S| $.
 %


Let $h^{(S_q)} \in \switchmatrix{2m_1 Q}$ switch precisely at $I_1, \cdots, I_{2m_1Q}$ and let $u^{(S_q)} = h^{(S_q)} - \mathbf{1}$.
By \Cref{remk:hu},  $\sum_{h' \in S_q} h'  =  \sum_{q=1}^Q {h^{q}}^+ - {h^{q}}^-  =  \sum_{q=1}^Q u^{(q)}  = u^{(s_q)}$. 
By \Cref{remk:hu}, if $m_1 <  R < 2m_1$ then $\sum_{h' \in S_r} h' = u^{(Q+1)}$. So if $R \neq 0$ then $\sum_{h' \in S_r} h' \in \left\{ u^{(Q+1)}, h^{(Q+1)} \right\} $. Let $u = h - \mathbf{1}$. By \Cref{remk:hu}, $\sum_{h' \in S} h' = \sum_{h' \in S_q} h' + \sum_{h' \in S_r} h' \in \left \{u, h \right \}$. Therefore $\sigma \left( \sum_{h' \in S} h' \right) = h$.
\iftoggle{long}{
By \Cref{lemma:h+h-}, $\mathbf{q} \in \{-2,0\}^N$ and $\sigma(\mathbf{q}) = h(I_1,\cdots,I_{2m_1Q})$. and $\left( \sum_{\mathbf{h} \in S_h} \mathbf{h} \right)_{[2m_1(Q+1):]} = 0$. And $\left( \sum_{\mathbf{h} \in S_r} \mathbf{h} \right)_{[2m_1Q]} \in \{-1,-2\}$.

If $m_1 < R < 2m_1$, $h^+(\mathcal{I}_{r})- h^-(\mathcal{I}_{r}) \in \{-2,0\}^N$

By \Cref{lemma:hremainder}, $h^+(\mathcal{I}_{r})- h^-(\mathcal{I}_{r}) \in \{-2,0\}^N \cup \{-1,1\}^N$. 
 For $q \in [Q]$, by the proof of \Cref{lemma:h+h-}, $h^+ (\mathcal{I}^{(q)}) -h^-(\mathcal{I}^{(q)}) \in \{-2,0\}^N$ and is only nonzero between indices $1+2m_1(q-1)$ and $(2m_1)q$. Therefore $ \sum_{\mathbf{h} \in S_h} \mathbf{h} =  \sum_{q=1}^Q h^+\left(\mathcal{I}^{(q)}\right) - h^- \left(\mathcal{I}^{(q)}\right)\in \{-2,0\}$ is only nonzero between indices $1+2m_1(q-1)$ and $2m_1q$ for $q \in [Q]$. Adding the  vectors in $S_r$ makes $\sigma \left( \sum_{\mathbf{h} \in S} \mathbf{h} \right) = h(\mathcal{I})$.
 }
%
	%
\end{proof}
}

\begin{lemma}\label[lemma]{lemma:decomp_english}
	Let $p, m \in \mathbf{N}$. Let $\mathbf{z} \in \{-1,1\}^N$ with at most $pm$ switches. There is an integer $n \leq m$, $\mathbf{w} \in \{-1,1\}^{n}$, and a $N \times n$ matrix $\mathbf{H} $ with columns in  $\switchmatrix{p}$ such that $\mathbf{z} = \sigma(\mathbf{Hw})$.
\end{lemma}

\begin{proof}
For $x \in \{-1,1\}, \sigma(x)=\sigma(x-1)$. Apply \Cref{lemma:hvec_switch_prod} with $m_2 = p, m_1 = m$.  
%
%
 \iftoggle{long}{
 $\mathbf{z} = \sum_{s =1}^{|S|} w_s \mathbf{\tilde{h}_s}$
 
 $h(\mathcal{I})=\sigma(\sum_{i=1}^p w_i h(\mathcal{I}^{(i)} ) $ for some vectors $\mathcal{I}^{(i)} \in \{-1,1\}^N$ with at most $m_1=m$ switches. Let $H \in \{-1,1\}^{N \times m_2}$ be the matrix whose $i^{\textrm{th}}$ column is  $h(\mathcal{I}^{(i)})$. By lemma \ref{lemma:hvec_switch_prod}, $Hw \in \{-2,0\}^{N}$. And $z = h(\mathcal{I}) = \sigma(Hw)$.
}
\end{proof}

\begin{lemma}\label[lemma]{prop:H_oneswitch}
 $\mathcal{H}(\mathbf{X})$ 
 consists of all columns in $\switchmatrix{1}$.
\end{lemma}

\begin{proof}
	First, $\mathbf{1}=\sigma(\mathbf{0})=\sigma(X \cdot \mathbf{0}) \in \mathcal{H}(\mathbf{X})$.
	Next, let $\mathbf{h} \in \mathcal{H}(\mathbf{X}){-}\{ \mathbf{1}\} \in \{-1,1\}^N$. By definition of $\mathcal{H}(\mathbf{X})$, there exists $w,b\in \mathbb{R}$ such that $\mathbf{h}=\mathbf{X}w+b\mathbf{1}$. 
 Note ${h}_1=1$ (\Cref{deepHstartswith1}).
	Let $i$ be the first index at which  $\mathbf{h}$ switches. 
 So $x_{i}w+b < 0 \leq x_{i-1}w+b  $, which implies $x_{i} w < x_{i-1}w$. 
 For all 
 $j > i$, $x_j < x_i$ so ${h}_j = 
 \iftoggle{long}{
 \sigma((\mathbf{Xw})_j) = \sigma \left((x_j,1)(w,b)^T \right) = 
 }
\sigma(x_j w + b) \leq \sigma(x_i w + b) = \sigma(h_i)=-1$, so $h_j =-1$. So $\mathbf{h}$ switches at most once.

	Now, let $\mathbf{h} \in \{-1,1\}^N$ with $h_1=1$. Suppose $\mathbf{h}$ switches once, at index $i \in \{2, \cdots, N\}$. In particular, $ {h}_i =-1$. Let $w=1,b=-x_{i-1}$. Then at $j<i$, $x_j w+b = x_j - x_{i-1} \geq 
 0$ so $h_j = \sigma \left(x_j w +b\right)=1$. And for $j\geq i$, $x_j w+b = x_j - x_{i-1} 
 < 0$ so ${h}_j = -1$. So $\mathbf{h} \in \mathcal{H}(\mathbf{X})$.
\end{proof}

\begin{proposition}\label[proposition]{prop:mswitches}
	The set 
 $A_{L=3,par}$
 contains all columns of  $\switchmatrix{m_{1}}$.
\end{proposition}

\begin{proof}
	Note 
 $A_{L=3,par}= \bigcup_{\mathbf{X^{(1)}}}\mathcal{H}\left(\mathbf{X^{(1)}}\right)$. From \Cref{prop:H_oneswitch}, any possible column in $\mathbf{X^{(1)}}$ switches at most once.
  Apply \Cref{lemma:decomp_english} with $p=1, m=m_1$ (so that $mp = m_1$ switches), to any column $\mathbf{z}$ of $\mathbf{X^{(1)}}$. 
\end{proof}

\iftoggle{long}{
If we add more layers however, the number of switches does not increase.
}

\begin{proposition}\label[proposition]{prop:m_1_more_layers}
For a \deepsign{} network,	
 \iftoggle{long}{
 activation function $\sigma(z)=\text{sign}(z), X \in \mathbb{R}^{N \times 1}$, 
 }
 $A_{L, par}(\mathbf{X})$ is contained in the columns of  $\switchmatrix{m_{1}}$. 
\end{proposition}

\begin{proof}
 Let $\mathbf{W}^{(1)} \in \mathbb{R}^{1 \times m_1}$. For any $w, b \in \mathbb{R}$, since the data is ordered, $\sigma\left(\mathbf{X}w + b \mathbf{1}\right) \in \{-1,1\}^N$ has at most 1 switch. Then
 $\mathbf{X}^{(1)} = \sigma \left(\mathbf{X}\mathbf{W}^{(1)} + \mathbf{1} \cdot \mathbf{b}^{(1)} \right)$ has $m_1$ columns each with at most one switch. Therefore $\mathbf{X}^{(1)}$ has at most $m_1+1$ unique rows. Let $R$ be the set of the smallest index of each unique row.
 \iftoggle{long}{
 $r_n$ be the smallest index of the $n^{th}$ unique row of $\mathbf{X}^{(1)}$, and let $R= \left\{r_1, \cdots, r_{\tilde{m}} \right\}$, where $\tilde{m} \leq m+1$. In other words, the rows of $\mathbf{X}^{(1)}$ only change at indices in $R$.
 }
 We claim that for all layers $l \in [L]$, the rows of $\mathbf{X}^{(l)}$ are constant at indices in $[N]-R$, that is, for all $i \in [N]-R$, the $i^{\text{th}}$ and $(i-1)^{\text{th}}$ rows of $\mathbf{X^{(l)}}$ are the same. 
 We prove our claim by induction.
 The base case for $l=1$ already holds.
 
 Suppose our claim holds for $l \in \{1, \cdots, L-1\}$. Let $\mathbf{W}^{(l+1)} \in \mathbb{R}^{\left(m_{l} \times m_{l+1}\right)}$. The rows of $\mathbf{X}^{(l)}$ are constant at indices in $[N]-R$, so 
 for any $\mathbf{w} \in \mathbb{R}^{m_{l}}, b \in \mathbb{R}$, the elements of the vector $\mathbf{X}^{(l)}\mathbf{w} + b \mathbf{1}$ are constant at indices in $[N]-R$. This held for any $\mathbf{w} \in \mathbb{R}^{m_l}$, so the rows of $\mathbf{X}^{(l+1)}=\mathbf{X}^{(l)} \mathbf{W}^{(l+1)}+\mathbf{1} \cdot \mathbf{b}^{(l+1)}$ are again constant at indices in $[N]-R$. By induction, our claim holds for all $l \in [L]$.
\iftoggle{long}{
	Now let $W^{(2)} \in \mathbb{R}^{(m_1 +1 \times m_2)}$. $X^{(1)}$ has changing rows only at $r_1, \dots, r_{n}$, so for any vector $w \in R^{m_1 + 1}$, the elements of the vector $X^{(1)}w $ can only switch at indices  $r_1, \dots, r_{n}$. This held for any $w \in \mathbb{R}^{m_1}$, so $X^{(1)} W^{(2)}$ only has at most $m_1$ unique rows, in particular again at rows $r_1, \dots, r_{n}$. So $X^{(2)} = \sigma([X^{(1)}W^{(2)}, \mathbf{1}])$ also only has changing rows at $r_1, \dots, r_{n}$.

	Continuing the same way, $X^{(3)} = \sigma([X^{(2)}W^{(3)}, \mathbf{1}])$ can only have changing rows at $r_1, \dots , r_{n}$, etc and so does $X^{(L-1)} = \sigma([X^{(L-2)}W^{(L-1)}, \mathbf{1}])$.
}
	So $\mathbf{X}^{(L)}$ has at most $|R| \leq m_1+1$ unique rows and hence has columns that each switch at most $m_1$ times. 
 \iftoggle{long}{
 and any feasible $X^{(L-1)}$ (taken over all possible sequences of $W^{(1)}, \dots, W^{(L-1)}$), }
\end{proof}
\iftoggle{long}{
The following proposition provides a converse if the layers all have the same size.
}
\begin{proposition}\label[proposition]{prop:m_1_more_layers_same_size}
For a $\deepsign$ network,
 \iftoggle{long}{
 activation function $\sigma(z)=\text{sign}(z), X \in \mathbb{R}^{N \times 1}$, }
 $A_{L,par}(\mathbf{X})$ contains all columns of $\switchmatrix{m_1}$. 
\end{proposition}
\begin{proof}
	Let $\mathbf{z} \in \{-1,1\}^{N} $ with at most $\min \{N-1, m_1\}$ switches. By \Cref{prop:mswitches}, there exists a feasible $\mathbf{X^{(1)}} = \sigma \left(\mathbf{X}\mathbf{W}^{(1)} + \mathbf{1} \cdot \mathbf{b}^{(1)} \right) \in \mathbb{R}^{N \times m}$ and $\mathbf{W^{(2)}} \in \mathbb{R}^{m \times m}$ such that $\mathbf{z}$ is a column of $\mathbf{X^{(2)}}=\sigma \left(\mathbf{X^{(1)} W^{(2)}} + \mathbf{1} \cdot \mathbf{b}^{(2)} \right)$. Now for every $l \in \{3, \dots, L-1\}$ we can set $\mathbf{W^{(l)}} = \mathbf{I}_m$ to be the $m \times m$ identity matrix and $\mathbf{b}^{(l)}=\mathbf{0}$ so that $\mathbf{X^{(l)}}=  \sigma \left(\mathbf{X^{(l-1)} W^{(l)}} \right) =\mathbf{X^{(l-1)}}$. Then $\mathbf{X}^{(L)}=\mathbf{X}^{(2)}$ contains $\mathbf{z}$ as a column. Therefore $\mathbf{z} \in  A_{L,par}(\mathbf{X})$.
 %
%
\end{proof}

\begin{lemma}\label[lemma]{prop:deepprod_m_tree}
	Let $K=\prod_{l=1}^{L-1} m_l$. The set ${A_L}_{\textrm{tree}}(\mathbf{X})$ consists of all columns in 
 $\switchmatrix{K}$.
\end{lemma}

\begin{proof}
For $l \in [L]$, let $A_l = {A_l}_{tree}(\mathbf{X})$. Let $p_k = \prod_{l=1}^{k-1} m_l$. We claim for all $l \in \{2, \cdots, L\}$, $A_l$ consists of all columns in $\switchmatrix{p_l}$.
	We prove our claim by induction on $l$. The base case when $l=2$ holds by \Cref{prop:H_oneswitch}.
	Now suppose \Cref{prop:deepprod_m_tree} holds when $l=k \in \{2, \cdots, L-1\}$. Observe $A_{k} \subset A_{k+1}$, so if $p_k \geq N-1$, then \Cref{prop:deepprod_m_tree} holds for $l=k+1$. So suppose $p_k < N-1$. Then $A_k$ contains all vectors in $\{-1,1\}^N$ with at most  $p_k$ switches. 

	Let $\mathbf{h} \in {A_{k+1}}$.  
 The set ${A_{k+1}}$ contains all columns in $J^{(m_{k})}([A_{k}])$. So, there exist $\mathbf{w} \in \mathbb{R}^{m_{k}}, b \in \mathbb{R}$ and a submatrix $\mathbf{Z}=[A_k]_S$ where $|S| = m_{k}$
and $\mathbf{h} = \sigma(\mathbf{Zw}+\mathbf{b})$. Each of the at most $ m_{k}$ columns of $\mathbf{Z}$ has at most $p_k$ switches, so the $N$ rows of $\mathbf{Z}$ change at most $m_{k}p_k=p_{k+1}$ times. So  $\mathbf{Zw}+\mathbf{b}$ changes value, and hence
$ \mathbf{h}=\sigma(\mathbf{Zw}+\mathbf{b})$ switches, at most $p_{k+1}$ times.
Conversely, 
by \Cref{lemma:decomp_english}
	with 
 $p=p_k, m = m_{k}$, the set $A_{k+1}$ of all columns in $J^{(m_{k})}([A_k])$ contains all vectors with at most $p_k m_{k}=p_{k+1}$ switches. So our claim holds for $l=k+1$. By induction, it holds for $l=L$ layers.
\end{proof}
\subsection{Proofs of results in \Cref{sec:deepsign}}

\begin{proof}[Proof of \Cref{lemma:sign_thresh_a}]
   The result follows from the definition of $\Amat{}$, the assumption that the data is ordered and that the output of sign activation is $\sigma$ is $\pm 1$ for sign activation.
   \iftoggle{long}{  Since the data is ordered, $\mathbf{X}-x_j \mathbf{1}$ is a vector that changes sign at most once, at index $j+1$, and the output of $\sigma$ is $\pm 1$ for sign activation.}
\end{proof}

\begin{proof}[Proof of \Cref{thm:3lyr_sign}]
For the parallel network, apply
 \Cref{prop:deeplasso_equiv_primal}
 and then apply \Cref{prop:mswitches} for $L=3$, and \Cref{prop:m_1_more_layers} and \Cref{prop:m_1_more_layers_same_size} for $L \geq 3$. For tree networks, by \Cref{rk:tree_reconstruct} the training problem is equivalent to \lasso{} lower bound with dictionary $A_{L,tree}$ given by \Cref{prop:deepprod_m_tree}.
\end{proof}

\begin{proof}[Proof of \Cref{Cor:signfeaturesdef}]
    Follows from \Cref{thm:3lyr_sign}, \Cref{prop:deepreconstruct_primal_L} and \Cref{rk:tree_reconstruct}.
\end{proof}

\iftoggle{long}{

\begin{remark}\label{rk:sign_explicit_recon}
    \Cref{thm:3lyr_sign_reconstruction} (and more generally, \Cref{prop:deepreconstruct_primal_L}) gives parameters that are optimal in the rescaled problem. Rescaling as described in \Cref{def:reconstructfunc} gives an optimal network in the original training problem, by letting $\gamma_i = \sqrt{|z_i|}$ and transforming $s^{(i,L-1)} \to \gamma_i$ and $\alpha_i \to \alpha_i/\sqrt{|z_i|}$. More layers can be added using \Cref{rk:morelayersreconstruct}.    
\end{remark}
}

\iftoggle{long}{
\proofsketch{
To show optimality: by \Cref{prop:lasso_equiv_primal}, the Lasso model \eqref{thm_eq:3lyrsign} is equivalent to the \nc{} training problem \eqref{equ:nn_train_signdetinput_rescale}. Note in \Cref{prop:lasso_equiv_primal}, the existence of a general map from the Lasso solution to a neural net achieving the same objective in the \nc{} primal \eqref{equ:nn_train_signdetinput_rescale}  was given through the definition and construction of the dictionary sets $A_L$. So we may assume strong duality a priori, and it suffices to show that the neural net defined in theorem \ref{thm:3lyr_sign_reconstruction} achieves the same loss in the primal \eqref{equ:nn_train_signdetinput_rescale}  as in the Lasso problem \eqref{thm_eq:3lyrsign}.}
}
\begin{proof}[Proof of \Cref{thm:3lyr_sign_reconstruction}]
By \Cref{thm:3lyr_sign} and \Cref{lemma:rescale}, it suffices to show the parameters without \unscal{}ing achieve the same objective in the rescaled problem \eqref{eq:noncvxtrain_genLlyrLLnorm_rescale} as \lasso{}. First, $|\mathcal{I}| \leq m_2$ and by \Cref{thm:3lyr_sign}, $m^{(i)} \leq m_1$ so the weight matrices are the correct size. 
%
Let $S^{(i)}_n$ be the number of times $\Amat_i$ switches until index $n$.
Since $x_1 > \cdots > x_N$, we get $\Xul{i}{2}_{n,j}=\sigma\left( \mathbf{X}  \Wul{i}{1} + \mathbf{1} \cdot \bul{i}{1} \right)_{n,j} = \sigma \left(x_n - x_{I^{(i)}_j-1} \right) = -\sigma \left( S^{(i)}_n-j \right) $. 
%
\iftoggle{long}{
So the $n^{th}$ row of $\sigma \left([\mathbf{X}, \mathbf{1}] \Wul{i}{1}\right)$ is $\left( -\mathbf{1}^T_{ S_i^{-1}(n)}, \mathbf{1}^T_{N-S_i^{-1}(n)} \right)$}
%
So 
$\displaystyle \Xul{i}{3}_n = \sigma\left( \Xul{i}{2} \Wul{i}{2} + \bul{i}{2} \mathbf{1} \right)_n = \sigma \left( \sum_{j=1}^{S_n^{(i)}} (-1)(-1)^{j+1} + \sum_{j=S_n^{(i)}+1}^{m^{(i)}} (1)(-1)^{j+1} - \mathbf{1} \left\{m^{(i)} \mbox{ odd} \right \}\right) = \sigma\left(-2 \cdot \mathbf{1} \left \{S^{(i)}_n \mbox { odd} \right \}\right) = (-1)^{S_n^{(i)}}  = \Amat{}_{n,i}$.
So,
$f_{3}(\mathbf{X}; \theta)  = \xi+ \sum_{i \in \mathcal{I}}  \alpha_i \Xul{i}{3}  =\mathbf{Az}$.
And, $||\boldsymbol{\alpha}||_1 = ||\mathbf{z}^*_I||_1 = ||\mathbf{z}^*||_1$. So the rescaled problem and \lasso{} achieve the same objective.
\iftoggle{long}{

	\begin{equation}
		\begin{aligned}
			\sigma \left([\mathbf{X}, \mathbf{1}] \Wul{i}{1}\right) & =
			\begin{pmatrix}
				x_1 & 1      \\
				x_2 & 1      \\
				    & \vdots \\
				x_N & 1      \\
			\end{pmatrix}
			\begin{pmatrix}
				1             & \cdots & 1                 \\
				-x_{S_i(1)-1} & \cdots & -x_{S_i(|S_i|)-1}
			\end{pmatrix}.
		\end{aligned}
	\end{equation}

	First suppose $|S_i|$ is even. We have

	\begin{equation}
		\begin{aligned}
			[\mathbf{X}, \mathbf{1}] \mathbf{\Tilde{W}^{(i,1)}} & =
			\begin{pmatrix}
				x_1 & 1      \\
				x_2 & 1      \\
				    & \vdots \\
				x_N & 1      \\
			\end{pmatrix}
			\begin{pmatrix}
				1             & \cdots & 1                 \\
				-x_{S_i(1)-1} & \cdots & -x_{S_i(|S_i|)-1}
			\end{pmatrix}.
		\end{aligned}
	\end{equation}

	Recall $x_1 > \cdots > x_N$, and the elements of $S_i$ are in ascending order.  So $ [\mathbf{X},\mathbf{1}] \mathbf{\Tilde{W}^{(i,1)}}$ is a matrix whose $j^{th}$ column is nonnegative at indices $k<S_i(j)$, and negative at all larger indices. So the $j^{th}$ row of $\sigma( [\mathbf{X},\mathbf{1}] \mathbf{\Tilde{W}^{(i,1)})}$ is $1$ at indices from 1 until $S_i(j)-1$, and $-1$ at all indices greater than $k$. In other words, the $j^{th}$ column of $\sigma( [\mathbf{X},\mathbf{1}] \mathbf{\Tilde{W}^{(i,1)}})$ is a row vector in $\{-1,1\}^N$ that switches at the index $S_i(j)$. Since the elements of $S_i$ are in increasing order, increasing columns of $\sigma( [\mathbf{X},\mathbf{1}] \mathbf{\Tilde{W}^{(i,1)}})$ have switches at increasing indices. Let $\mathbf{\Tilde{W}^{(i,1)}}_j$ be the  $j^{th}$ column of $\mathbf{\Tilde{W}^{(i,1)}}$.
	So, for odd $j$, $\mathbf{\Tilde{W}^{(i,1)}}_j - \mathbf{\Tilde{W}^{(i,1)}}_{j+1}$ is a vector that is $-2$ for indices $k < |S_i|$ such that $ S_i(j) \leq k \leq S_i(j+1)$, and $0$ at all other indices.

	Now suppose $|S_i|$ is odd. The above argument holds except for $j=|S_i|$, we have  $\mathbf{\Tilde{W}^{(i,1)}}_j - \mathbf{\Tilde{W}^{(i,1)}}_{j+1}$ is a vector that is $-2$ at indices $k \geq |S_i|$ and $0$ at all smaller indices.

	Therefore, in both cases of even and odd $|S_i|$, since $\mathbf{\Tilde{W}^{(i,2)}}$ is a vector alternating between $1$ and $-1$ starting at $1$, we have $\sigma( [\mathbf{X},\mathbf{1}] \mathbf{\Tilde{W}^{(i,1)})\Tilde{W}^{(i,2)}} = \sum_{j \mbox{ odd}} \mathbf{\Tilde{W}^{(i,1)}}_j - \mathbf{\Tilde{W}^{(i,1)}}_{j+1}$ is a vector that is $-2$ between (and including) indices $S_i(j)$ and $S_i(j+1)$ for odd $j$, and $0$ everywhere else.

	Since $\sigma(0)=1$ and $\sigma(-2)=-1$, we have $\sigma(\sigma( [X,1] \Tilde{W}^{(i,1)})\Tilde{W}^{(i,2)})$ is a vector in $\{-1,1\}^N$ whose switching indices are exactly those in $S_i$. Therefore $\sigma(\sigma( [X,1] \Tilde{W}^{(i,1)})\Tilde{W}^{(i,2)}) = a_i$.

	Therefore,

	\begin{equation}
		\begin{aligned}
			f_{3}(X) & = \sum_{i=1}^{m_2} \sigma_{s^{(i,2)}} \left ( \sigma_{s^{(i,1)}} \left([X,1] W^{(i,1)}  \right) W^{(i,2)}\right) \alpha_i + \xi \\
			         & = \sum_{i \in I} \sigma \left ( \sigma \left([x,1] W^{(i,1)}  \right) W^{(i,2)}\right) z_i                                      \\
			         & = \sum_{i \in I} \sigma \left ( \sigma \left([X,1] \Tilde{W}^{(i,1)}  \right) \Tilde{W}^{(i,2)}\right) z_i                      \\
			         & = \sum_{i \in I} a_i z_i                                                                                                        \\
			         & = Az_I                                                                                                                          \\
			         & = Az.
		\end{aligned}
	\end{equation}

}
\iftoggle{long}{

	So the neural net achieves the same function error $\mathcal{L}_{\mathbf{y}}(f_3(X)) = \mathcal{L}_{\mathbf{y}}(Az) $ in the primal \eqref{equ:nn_train_signdetinput_rescale} as the bidual solution $z$ does in Lasso \eqref{thm_eq:3lyrsign}. In addition, the regularization $||\boldsymbol{\alpha}||_1 = \beta||\mathbf{z}^*_I||_1 = ||\mathbf{z}^*||_1$. So the objective values in \eqref{equ:nn_train_signdetinput_rescale}  and \eqref{thm_eq:3lyrsign} are the same.
}
\end{proof}

\begin{remark}\label[remark]{rk:morelayersreconstruct}
    For a \deepsign{} network, a reconstruction similar to \Cref{thm:3lyr_sign_reconstruction} holds by setting additional layer weight matrices to the identity.
\end{remark}

\begin{proof}[Proof of \Cref{cor:3lyr_m1x2lyr} ]
%
%
By \Cref{rk:dictsgetbigger}, $p^*_{L=3,\beta} \leq p^*_{L=2,\beta}$. Let $\params^{(L)}$ and $\boldsymbol{\alpha}^{(L)}$ denote $\params$ and $\boldsymbol{\alpha}$ for a $L$-layer net. Since the training and rescaled problems have the same optimal value, to show $p^*_{L=2,\beta} \leq p^*_{L=3,m_1\beta}$, it suffices to show for any optimal $\params^{(3)}$, there is $\params^{(2)}$ with $f_{2}\left(\mathbf{X};\theta^{(2)}\right) = f_{3}\left(\mathbf{X};\theta^{(3)}\right)$ and $\left\|\boldsymbol{\alpha^{(2)}}\right \|_1 \leq m_1 \left\|\boldsymbol{\alpha^{(3)}}\right\|_1$. 
Let $\mathbf{z}^*$ be optimal in the $3$-layer \lasso{} problem \eqref{eq:genericlasso}. Let $m_3^* = ||\mathbf{z^*}||_0$ and let $\mathbf{z} \in \mathbb{R}^{m_3^*}$ be the subvector of nonzero elements of $\mathbf{z^*}$. Let  $m=m_1m^*_3$.
 By \Cref{thm:3lyr_sign_reconstruction} and its proof, there are $\Wul{i}{1} \in \mathbb{R}^{1 \times m_1}, \bul{i}{1} \in \mathbb{R}^{1 \times m_1}, \Wul{i}{2} \in \{1,-1,0\}^{m_1}, \bul{i}{2} \in \mathbb{R}$ such that $\Xul{i}{2}\Wul{i}{2} + \bul{i}{2} \mathbf{1} \in \{-2,0\}^N$ and 
 $\nn{3}{\mathbf{X}} =$
 \iftoggle{long}{ the first layer output is the matrix }
 \iftoggle{long}{, whose columns are $1$-switch vectors in $\{-1,1\}^N$, 
 and $\Xul{i}{1}\Wul{i}{2}\in \{-2,0\}^N$ and an output of the optimal neural network for the rescaled \eqref{equ:nn_train_signdetinput_rescale} evaluated on all the training data is }


	\begin{equation*}\label{eq:2layerto3layer_int}
 \scalemath{0.9}{
		\begin{aligned}
			& 
			\sum_{n=1}^{m_3^*} z_i \sigma \left(\Xul{i}{2} \Wul{i}{2} + \mathbf{1} \cdot \bul{i}{2}\right) 
			                   =   \sum_{i=1}^{m_2^*} z_i^* \left( \Xul{i}{2} \Wul{i}{2} + \mathbf{1} \cdot \bul{i}{2}+\mathbf{1}\right)  =                        \\
			                     &  \sum_{i=1}^{m_3^*} z_i^* \left(  \sigma \left( \mathbf{X} \Wul{i}{1} + \mathbf{1} \cdot \bul{i}{1}  \right )\Wul{i}{2} + \mathbf{1} \cdot \bul{i}{2} +\mathbf{1} \right)  =   \\
			           &  \sigma \left(\mathbf{X} 
              \underbrace{\left[ \Wul{1}{1}\cdots \Wul{m_3^*}{1}\right]}_{\left(\Wul{1}{1},\cdots,\Wul{m}{1} \right) \in \mathbb{R}^{1\times m}} + \mathbf{1} \cdot 
              \underbrace{\left[ \bul{1}{1}, \cdots, \bul{m_3^*}{1}\right]}_{\left(\bul{1}{1},\cdots,\bul{m}{1} \right) \in \mathbb{R}^{1 \times m}}
              \right)
              \underbrace{
			\begin{bmatrix}
				 & z_1  \Wul{1}{2}           \\
				 & \vdots                           \\
				 & z_{m_3^*}  \Wul{m_3^*}{2}
			\end{bmatrix}
   }_{\boldsymbol{\alpha^{(2)}}}
   +
  \mathbf{1} \underbrace{\sum_{i=1}^{m_3^*} z_i \left(1+ \bul{i}{2} \right)}_{\xi},
		\end{aligned}
  }
	\end{equation*}
which is $f_{2}\left(\mathbf{X};\params^{(2)}\right)$ with 
$m$ neurons.
\iftoggle{long}{

	which matches the form of a $2$-layer network defined by \eqref{eq:2layernn} with $m=m_1m^*_2$ neurons, $\xi =  \sum_{i=1}^{m_2^*} z_i(1+ \Wul{i}{2}_{m_1+1} )$, and $\boldsymbol{\alpha} = \begin{bmatrix}
			 & z_1  \Wul{1}{2}_{1:m_1}            \\
			 & \vdots                           \\
			 & z_{m_2^*} \Wul{m_2^*}{2}_{1:m_1}
		\end{bmatrix} \in \mathbb{R}^{m}$ , and $\mathbf{W} =\left[\Wul{1}{1} \cdots \Wul{m_2^*}{1}\right]  \in \mathbb{R}^{2 \times m}$.
Note $z_i \Wul{i}{2} \in \mathbb{R}^{m_1}$.
}
 And $\|\boldsymbol{\alpha}^{(2)}\|_1 \leq m_1 \| \mathbf{z}^* \|_1 = m_1 \| \boldsymbol{\alpha}^{(3)} \|_1 $.
%
\end{proof}

\iftoggle{long}{
\textbf{Proof of \Cref{thrm:lasso_deep_1nueron_ReLU}}

\proofsketch{
\iftoggle{long}{The proof follows the method \cite{pilanci2020neural} used in the conversion from the training problem \eqref{eq:background_ReLU} to its dual \eqref{eq:background_ReLU_dual}. But before taking a dual again to get to the bidual \eqref{eq:background_ReLU_bidual}, we leverage that for $1$-dimensional data the dual constraints can be significantly simplified. Therefore, the bidual in the $1D$ case yields  a LASSO problem similar to \eqref{eq:background_ReLU_bidual} in that it also consists of the original training loss function and involves a $l_1$ penalty; however the $1D$ bidual \eqref{gen_lasso_sol} is less complex in its variables, constraints and dictionary matrix parameters compared to the high dimensional bidual \eqref{eq:background_ReLU_bidual}.}}

\begin{proof}

	By \Cref{thrm:rescaling}, the training problem \eqref{eq:nonconvex_generic_loss_gen_L_intro} with $l_2$ penalty on all regularized parameters is equivalent to the rescaled problem \eqref{equ:nn_train_signdetinput_rescale} with $l_1$ penalty on the outer weights $\alpha$. By \Cref{lemma:dual_of_rescaled}, the dual of the rescaled problem  \eqref{equ:nn_train_signdetinput_rescale} is \eqref{gen_dual}.

    \iftoggle{long}{The constraint set in the dual \eqref{gen_dual} involves a single neuron, but constrains its output over all possible weights. This could potentially constitute a large number of constraints.}
	
	The constraint in the dual \eqref{gen_dual} is semi-infinite, but because the data is $1$-D it can be broken into a small number of explicit constraints that correspond to turning off a neuron for each training sample (\Cref{lemma:dual_finite}). 
 \iftoggle{long}{ Each constraint corresponds to a possible value for the weight $w$ and bias $b$ that set the argument $x_n w+b$ of a neuron $\sigma(x_n w+b)$ to the value of $0$ on a sample $x_n$. In other words, each constraint corresponds to a sample $x_n$ that the neuron memorizes.}
 By \Cref{lemma:lasso_without_simp}, the dual of the dual \eqref{gen_dual} is the bidual \eqref{gen_lasso_sol}.

The bidual \eqref{gen_lasso_sol} is a lower bound on its primal \eqref{equ:nn_train_signdetinput_rescale}. By \Cref{thrm:2layer_reconstruct}, an optimal optimal solution to \eqref{eq:nonconvex_generic_loss_gen_L_intro}

for every value the bidual \eqref{gen_lasso_sol}  can achieve with at most $||z||_0 \leq m$ nonzero bidual variables, there is a neural network with $m$ neurons feasible in the rescaled primal  \eqref{equ:nn_train_signdetinput_rescale} that achieves the same objective value in \eqref{equ:nn_train_signdetinput_rescale}. Therefore strong duality holds and the rescaled primal \eqref{equ:nn_train_signdetinput_rescale} and bidual \eqref{gen_lasso_sol} problems are equivalent. The rescaled primal \eqref{equ:nn_train_signdetinput_rescale} is equivalent to the original primal, so the original primal and the bidual \eqref{gen_lasso_sol} are equivalent.

Bounded activations that are piecewise linear have coefficients $c_1=c_2=0$ (\Cref{app_section:act_func}). In problem \eqref{gen_lasso_sol}, if  $c_1=c_2=0$ then $\xi=0$. So  simplification \ref{bounded_simp} for sign and threshold activations in theorem \ref{theo:1d_all} holds. Problem \eqref{gen_lasso_sol} is therefore the same as problem \ref{eqref:Lasso}.

	Lemmas \ref{lemma:sym_activation_simp} and \ref{lemma:sign_activation_simp} yield simplification \ref{symm_simp} in in theorem  \ref{theo:1d_all} for symmetric and sign activations, respectively. Note that setting $z_-=0$ in theorem  \ref{theo:1d_all} is equivalent to just having one dictionary matrix $A=A_+ \in \mathbb{R}^{N \times N}$ and variable $z=z_+ \in \mathbb{R}^N$.

\end{proof}

}


\subsection{Proofs of results for \nd{2} data}\label{app_section_2d_data}

\begin{proof}[Proof of \Cref{lemma:2D_upperhalfplane_signact}]
    \Cref{lemma:deeplasso_sign_L} holds for any dimension $d$, so the \lasso{} formulation in \Cref{thrm:lasso_deep_1nueron_ReLU} and \Cref{thm:3lyr_sign} similarly hold for $d>1$ but with a different dictionary $A_{L,par}$ and matrix $\Amat{}$. 
    
    Let $x'_n = \angle \mathbf{x}^{(n)}$ and order the data 
    so that ${x'_1} > {x'_2} > \dots {x'_N} $. Let $\mathbf{X'}=(x'_1 \cdots x'_N) \in \mathbb{R}^N$.
Let $\mathbf{w} \in \mathbb{R}^2$ with
$\angle \mathbf{w} \in \left[\frac{\pi}{2}, \frac{3\pi}{2} \right]$. Let $w' = \angle \mathbf{w} $.
Note $\mathbf{w} \mathbf{x}^{(n)} \geq 0$ if and only if $x_n' \in \left[w'-\frac{\pi}{2}, w'+\frac{\pi}{2} \right]$. Since $x_n' < \pi$ for all $n \in [N]$, this condition is equivalent to $x_n' \geq w'-\frac{\pi}{2}$, ie $n \leq \max \left\{n \in [N]: x_n' \geq w' - \frac{\pi}{2} \right\}$. So $ \left \{\sigma(\mathbf{X'w}): w' \in \left[\frac{\pi}{2}, \frac{3\pi}{2} \right] \right\} = \switchmatrix{1} \cup \{-1\}$. Similarly $ \left \{\sigma(\mathbf{X'w}): w' \in \left[-\frac{\pi}{2}, \frac{\pi}{2} \right] \right \}$ is the "negation" of this set. 
%
%
\iftoggle{long}{
Suppose the training data is on the upper half plane of $\mathbb{R}^{2}$: $x_i \in \mathbb{R}^2, {x_i}_2 \geq 0$. Suppose none of the training data have the same angle from the origin. 
}
\iftoggle{long}{
Then if we ignore the bias term, }
%
Therefore, the $L=2$ training problem is equivalent to the \lasso{} problem with dictionary $\switchmatrix{1}$. \Cref{prop:m_1_more_layers_same_size} holds analogously for this training data, so for $L>2$, the dictionary is $\switchmatrix{m_{L-1}}$.
%
%
\end{proof}

\iftoggle{unitcircle}{

\begin{theorem}\label{lemma:2D_unitcirlce_signact}
\iftoggle{long}{
	Let $X \in \mathbb{R}^{N \times d}$ consist of $d$-dimensional data. Define an $L\geq 2$-layer parallel NN with sign activation, i.e., 
 $\sigma(x)=s \cdot \mathrm{sign}(x)$ and }
 Suppose
 $m_1 = \cdots = m_{L-1} = m$. Suppose the data is on the unit circle:  $||\mathbf{x}_i||_2=||\mathbf{x}_j||_2$ for all $i,j \in [N]$: \\
 
 Then, the \nc{} training problem in \eqref{eq:nonconvex_generic_lossL_Lyr} is equivalent to
	\begin{align}\label{eqref:Lasso_L}
		\min_{\mathbf{z}} \frac{1}{2} \| \mathbf{Az} - \mathbf{y}\|_2^2 + \beta \|\mathbf{z}\|_1,
	\end{align}

 where 
 \begin{equation*}
     \Amat{}= \begin{cases}
				      \mathbf{H}_{2}    & \text{ if } L=2      \\
				      \mathbf{H}_{2m_1} & \text{ if } L \geq 3 \\
			      \end{cases}
 \end{equation*}

\end{theorem}

\begin{proof}

 By a similar observation as for data in the upper half plane, the indices $n \in [N]$ for which $\mathbf{ x_n w} \geq 0$ are consecutive, although it may exclude the first or last samples $\mathbf{x_1}, \mathbf{x_N}$. Therefore by a similar argument as for data in the upper half plane, $\mathcal{H}(\mathbf{X}')$ will consist of all vectors in $\{-1,1\}^N$ with at most $2$ switches, so $\Amat{} = \mathbf{H}_{2}$. By similar arguments as \Cref{prop:mswitches}, \Cref{prop:m_1_more_layers}, and \Cref{prop:m_1_more_layers_same_size}, for $L \geq 3$ layers, $\mathbf{A = }\mathbf{H}_{2m_1}$.

\end{proof}

\iftoggle{long}{
Theorem \ref{lemma:2D_upperhalfplane_signact} states that when the samples are anywhere in $\mathbb{R}$, i.e., $X \in \mathbb{R}^{N}$, and when the samples are on the upper half plane $\mathbb{R} \times \mathbb{R}^{\geq 0}$, their training problems \eqref{eq:nonconvex_generic_lossL_Lyr} turn out to be equivalent to the same LASSO problem \eqref{eqref:Lasso_L}.
}

Moreover, theorem \ref{lemma:2D_unitcirlce_signact} states when the training samples are $2$-dimensional and have the same Euclidean norm, in other words live on a circle of some radius $r=||x_1||_2 = \cdots ||x_N||_2$, then the LASSO problem \eqref{eqref:Lasso_L} corresponding to its training problem \eqref{eq:nonconvex_generic_lossL_Lyr} has a dictionary matrix with columns having up to twice as many switches as the one-dimensional case. Unit normalized data may occur from preprocessing steps 
or model layer batch normalization in the neural network. 
}

\begin{proof}[Proof of \Cref{lemma:2Doptnet}]
Observe  $n \leq i$ if and only if $\mathbf{x}^{(n)}\Wul{i}{1}\geq 0$ and so \\ $\sigma \left(\mathbf{X}\Wul{i}{1} \right) 
\left[\mathbf{1}^T_i, -\mathbf{1}^T_{N-i} \right]\\=
\Amat{}_i$. 
Thus $f_{L=2}(\mathbf{X};\theta) {=} \sum_{i} \alpha_i \sigma \left(\mathbf{X}\Wul{i}{1} \right) {=} \Amat\mathbf{z^*} $ so $\params$ achieves the same objective in the rescaled training problem \eqref{eq:gen_rescale}, as the optimal value of \lasso{} \eqref{eq:genericlasso}.  Unscaling (\Cref{parameterunscale}) optimal parameters of the rescaled problem makes them optimal in the training problem. 
\end{proof}



\iftoggle{todos}{
	\mert{Add a result for $d$ dimensional datasets which are rank-one, i.e., training data $X=x_1x_2^T$. Basically, the convex program reduces to the 1D case since the dual constraint simplifies:\\ $\max_{||u||_2=1,  u \in \mathbb{R}^d, b \in \mathbb{R}} \lambda^T(Xu+1b)_+ = \max_{||u||_2=1, u \in \mathbb{R}^d, b \in \mathbb{R}} \lambda^T((x_1x_2^T)u+1b)_+ =  \max_{|u^\prime|\le \|x_2\|_2, u^\prime,b \in \mathbb{R}} \lambda^T(x_1u^\prime+1b)_+ $
		The last step follows from $-\|x_2\|_2\le x_2^Tu \le \|x_2\|_2$.
	}
}

\iftoggle{long}{
The $1D$ structure of the data is only used to explicitely construct the dictionary matrix $\tilde{S}_{j:j\in [L]}(X)$ as a collection of vectors with a certain number of switches governed by the product of the number of neurons in each layer. If the data is \nd{2}, in certain situations this structure can also be recovered.
}

\iftoggle{todos}{\textcolor{red}{todo: recover neural net from Lasso}}

\section*{\secnameMinNorm}

\begin{remark}
    For each optimal $\mathbf{z}^*$ of the \lasso{} problem, minimizing the objective over $\xi$ gives the optimal bias term as 
    $\xi^* = \left(\mathbf{y}-\mathbf{1} \mathbf{1}^T \mathbf{y}\right) - \left( \Amat{} -\mathbf{1}\mathbf{1}^T \Amat{} \right) \mathbf{z}^*$.
    
\end{remark}

\begin{remark}\label[remark]{rk:strcvx}
    For a neural net with $L=2$ layers and sign activation, by \Cref{thrm:lasso_deep_1nueron_ReLU} 
    the \lasso{} problem has an objective function $f(\mathbf{z})=\frac{1}{2}\| \mathbf{Az} -\mathbf{y} \|^2_2 + \beta \|\mathbf{z}\|_1$ where $\Amat{} \in \mathbb{R}^{N \times N}$. By \Cref{lemma:A_full_rk}, $\Amat{}$ is full rank, which makes $f$ strongly convex. Therefore the \lasso{} problem has a unique solution $\mathbf{z}^*$ \cite{boyd:2004}. Moreover, for any \lasso{} problem, $\mathbf{z}^*$ satisfies the subgradient condition $\mathbf{0} \in \delta f(\mathbf{z})= \Amat{}^T (\mathbf{Az^*}-\mathbf{y})+\beta \partial \| \mathbf{z^*} \|_1$. Equivalently, 

\begin{equation*}
    \begin{aligned}
        \frac{1}{\beta}\Amat{}_n^T (\Amat{}\mathbf{z}^*-\mathbf{y}) \in  
        \begin{cases}
             \{ - \mathrm{sign}(\mathbf{z}^*_n)\} & \mbox{ if } \mathbf{z}^*_n \neq 0 \\
              [-1,1] & \mbox{ if } \mathbf{z}^*_n = 0 \\
        \end{cases},
        \quad n \in [N].
    \end{aligned}
\end{equation*}
\end{remark}

\subsection{Proofs of results in \Cref{sec:minnorm}}\label{app:minNorm}
Let $\mathbf{e}^{(n)} \in \mathbb{R}^N$ be the $n^{th}$ canonical basis vector, that is $\mathbf{e}^{(n)}_i = \mathbf{1}\{i=n\}$.

\begin{proof}[Proof of \Cref{prop:lasso_abs_solset}]
We analyze the solution set of $\mathbf{Az}+\xi\boldsymbol{1}=\mathbf{y}$. We note that
$
    \left(I-\mathbf{1}\mathbf{1}^T/N \right)\mathbf{Az}= \left(I-\mathbf{1}\mathbf{1}^T/N \right)\mathbf{y}.
$.
As $\mathbf{z}^*$ is optimal in 
\eqref{rm_gen_lasso_nobeta}
, this implies that $(I-\mathbf{1}\mathbf{1}^T/N)\Amat{}\mathbf{z}^*=(I-\mathbf{1}\mathbf{1}^T/N)\mathbf{y}$. This implies that
$
    (I-\mathbf{1}\mathbf{1}^T/N)\Amat{}(\mathbf{z}-\mathbf{z}^*)=0.
$
As $x_1>x_2>\dots>x_N$, we have $\Amat{}\left(\mathbf{e}^{(1)}+\mathbf{e}^{(N)}\right)\propto \mathbf{1}$. As $\Amat{}$ is invertible by \Cref{lemma:A_absval_full_rk} in \Cref{section:dic_matrices}, this implies that there exists $t\in\mathbb{R}$ such that
$
    \mathbf{z}-\mathbf{z}^* = t \left(\mathbf{e}^{(1)}+\mathbf{e}^{(N)}\right)
$.
It is impossible to have $z^*_1z^*_N>0$ from the optimality of $\mathbf{z}^*$. Otherwise, by taking $t=-\text{sign}(z^*_1)\min\{|z^*_1|,|z^*_n|\}$, we have $\|\mathbf{z}\|_1=\|\mathbf{z}^*\|_1-2\min\{|z^*_1|,|z^*_n|\}<\|\mathbf{z}^*\|_1$. Therefore, we have $z^*_1z^*_n\leq0$. We can reparameterize $\mathbf{z}=\mathbf{z}^*+t\text{sign}(z^*_1)(\mathbf{e}^{(1)}+\mathbf{e}^{(N)})$. It is easy to verify that for $t$ such that $-|z^*_1|\leq t\leq |z^*_n|$, we have $\|\mathbf{z}\|_1=\|\mathbf{z}_n\|_1$, while for other choice of $t$, we have $\|\mathbf{z}\|_1>\|\mathbf{z}_n\|_1$. Therefore, the solution set of 
\eqref{rm_gen_lasso_nobeta}
is given by \eqref{lasso:min_sol}. 
\end{proof}

\begin{proof}[Proof of \Cref{cor:sign_full_rk_unique}]
Follows from \Cref{rk:strcvx} describing the \lasso{} objective. 
\iftoggle{long}{
By \Cref{rk:strcvx}, \eqref{eq:genericlasso} has a unique solution. This is $\mathbf{z}^* = \Amat{}^{-1} \mathbf{y}$ when $\beta \to 0$.
	\Cref{thrm:lasso_deep_1nueron_ReLU} and its simplifications state that the \lasso{} problem \eqref{eq:genericlasso} is,

	\begin{equation}\label{eq:lasso_sol_L2norm}
		\begin{aligned}
			\min_{\mathbf{z} \in \mathbb{R}^N} & \frac{1}{2}||\Amat{}\mathbf{z}-\mathbf{y}||^2_2 + \beta ||\mathbf{z}||_1.
		\end{aligned}
	\end{equation}

By \Cref{rk:strcvx}, \eqref{eq:lasso_sol_L2norm} has a unique solution. When $\beta \to 0$, the solution to \eqref{eq:lasso_sol_L2norm} is $\mathbf{z}^* = \Amat{}^{-1} \mathbf{y}$.
}
\end{proof}









\iftoggle{long}{

\begin{proof}[\Proof of \Cref{lemma:thresh_informal_sol}]
    The result follows from \Cref{lemma:thresh_beta0_sol} and \Cref{lemma:Athresh_full_rk}.
\end{proof}
\begin{lemma}\label{lemma:thresh_beta0_sol}
    Consider a $2$-layer network with threshold activation. Suppose $y_1 \geq y_2 \geq \cdots \geq y_N \geq 0$. Then as $\beta \to 0$, $\mathbf{z}^*_{+}= \Amat{}^{-1}_{+} \mathbf{y}, \mathbf{z}^*_{-} = \mathbf{0}$ is the unique solution to \lasso{} \eqref{eq:genericlasso}.
\end{lemma}
}

\begin{proof}[Proof of \Cref{lemma:thresh_informal_sol}]
\iftoggle{long}{
	By simplification \ref{bounded_simp} in \Cref{thrm:lasso_deep_1nueron_ReLU}, the \lasso{} problem \eqref{eq:genericlasso} becomes,

	\begin{equation}\label{eq:Lasso_thresh}
		\min_{\mathbf{z} \in \mathbb{R}^{2N}} \frac{1}{2} \|\Amat{} \mathbf{z}  - \mathbf{y} \|^2_2 + \beta \|\mathbf{z}\|_1.
	\end{equation}
 
When $\beta \to 0$, \eqref{eq:Lasso_thresh} becomes the minimum $l_1$-norm problem, 

\begin{equation}\label{eq:thresh_beta0}
		\min_{\mathbf{z} \in \mathbb{R}^{2N}}
  \|\mathbf{z} \|_1, \mbox{ such that } \Amat{} \mathbf{z} = \mathbf{y}.
	\end{equation}
 
    \Cref{lemma:Athresh_full_rk} makes $\mathbf{z}^*$ well-defined and feasible in \eqref{eq:thresh_beta0}. 
    }
  %
    By \Cref{lemma:Athresh_full_rk}, for $n \in [N-1]$, ${z_+}^*_n =y_n - y_{n-1} \geq 0$ and ${z^*_+}_N = y_N \geq 0$. So $\mathbf{z}^*$ achieves an objective value of $\|\mathbf{z^*}\|_1 
    = y_1$ in \eqref{rm_gen_lasso_nobeta}. Now let $\mathbf{z}$ be any solution to \eqref{rm_gen_lasso_nobeta}. Then $\mathbf{Az} = \mathbf{y}$. Since the first row of $\Amat{}$ is $[\mathbf{1}^T, \mathbf{0}^T]$, we have $y_1 = (\Amat{} \mathbf{z})_1 = \mathbf{1}^T \mathbf{z}_+ \leq \|\mathbf{z}_+ \| \leq \|\mathbf{z}\|_1 \leq \|\mathbf{z}^*\| = y_1$. So $\|\mathbf{z}_+ \|_1 = \|\mathbf{z}\|_1 = y_1$, leaving $ \mathbf{z}_-  = \mathbf{0} = \mathbf{z}^*_-$. Therefore  $\mathbf{z}_+ = \Amat{}^{-1}\mathbf{y} = \mathbf{z}^*_+$. Applying \Cref{lemma:Athresh_full_rk} gives the result.
\end{proof}

\begin{proof}[Proof of \Cref{Cor:beta->0_overfit} ]
By \Cref{lemma:A_full_rk}, \Cref{lemma:A_absval_full_rk}, \Cref{lemma:Athresh_full_rk} and \Cref{lemma:ReLU_fullrk}, the dictionary matrix for the $2$-layer net is full rank for sign, absolute value, threshold and ReLU activations. 
The dictionary matrices for deeper nets with sign activation are also full rank by \Cref{rk:dictsgetbigger}. Let $\mathbf{u}=\Amat^T(\Amat \mathbf{z}^*-\mathbf{y})  $. By  \Cref{rk:strcvx}, as $\beta \to 0$, we have $\mathbf{u} \to \mathbf{0}$, so $\Amat \mathbf{z}-\mathbf{y}=(\Amat \Amat^T)^{-1}\Amat \mathbf{u} \to \mathbf{0}$. So as $\beta \to 0$, the optimal \lasso{} objective approaches $0$, and by \Cref{thm:3lyr_sign} and \Cref{thrm:lasso_deep_1nueron_ReLU}, so does the training problem. So $\nn{L}{\mathbf{X}} \xrightarrow{\beta\to 0} \mathbf{y}$.
%
%
\iftoggle{long}{So when $\beta \to 0$, $\mathbf{Az^*}=\mathbf{y}$
When $\beta \to 0$, the \lasso{} problem  for both the $L=2$ layer (\eqref{eq:background_threshold_bidual} with simplifications \ref{eqref:Lasso}, \eqref{symm_simp} ) and $L=3$ layer (\eqref{thm_eq:3lyrsign}) neural nets becomes the minimum $l_1$ norm solution to

	\begin{equation}\label{gen_lasso_nobeta}
		\min_\mathbf{z} \frac{1}{2} ||\mathbf{Az}-\mathbf{y}||^2_2
	\end{equation}
	where the dictionary matrix $\Amat{}$ depends on the number of layers $L$.
	 So an optimal solution $\mathbf{z}^*$ of \eqref{gen_lasso_nobeta} will satisfy $\Amat{}\mathbf{z}^* = \mathbf{y}$ and achieve an objective of $0$ in \ref{gen_lasso_nobeta}. By theorems \ref{theo:1d_all} optimal value of the Lasso problem \ref{gen_lasso_nobeta} is the same as that of the training problem. As $\beta \to 0$ the optimal value of the training problem becomes $\min_{\theta \in \Theta} \frac{1}{2}||f_L(\mathbf{X}) - \mathbf{y}||^2_2$, so its optimal value must also be $0$ and hence optimal parameters $\theta$ must satisfy $f_L(\mathbf{X}) = \mathbf{y}$.}
\end{proof}


%
%
\iftoggle{long}{
If the number of neurons $m$ in the neural network \eqref{eq:2layernn} is at least the number of nonzero elements in the optimal $\mathbf{z} \in \mathbb{R}^{2N}$ to the Lasso problem \eqref{eq:genericlasso}, then the optimal neural network \eqref{eq:2layernn} (ie optimal in the training problem \ref{eq:nonconvex_generic_loss_gen_L_intro})  can be reconstructed from the solution to \lasso{} \eqref{eq:genericlasso} as follows.
\begin{theorem}\label{thrm:2layer_reconstruct}
    Consider a $2$-layer network. Suppose $\mathbf{z}^*$ is optimal in \lasso{} \eqref{eq:genericlasso} and $m \geq \|\mathbf{z}^*\|_0$. Construct $\boldsymbol{\alpha}, \mathbf{W_0}, \mathbf{b}, s$ as given in \Cref{lemma:2layer_reconstruct2rescaled}. Then if the activation is ReLU, leaky ReLU or absolute value, apply the transformation described in \Cref{lemma:reconstruct_ReLU2original}. Otherwise if the activation is sign or threshold, apply the transformation in \Cref{lemma:reconstruct_bdd2original}. The resulting weights are globally optimal in the \nc{} training problem \eqref{eq:nonconvex_generic_loss_gen_L_intro}.
\end{theorem}

\begin{proof}[Proof of \Cref{thrm:lasso_deep_1nueron_ReLU} and \Cref{thrm:2layer_reconstruct}]
     Let $t^*, r^*, d^*, l^*$ be the optimal values of the training problem \eqref{eq:nonconvex_generic_loss_gen_L_intro}, rescaled problem \eqref{equ:nn_train_signdetinput_rescale},  problem \eqref{gen_dual}, and the \lasso{} problem \eqref{eq:genericlasso}, respectively. 
     By \Cref{thrm:rescaling}, the training problem \eqref{eq:nonconvex_generic_loss_gen_L_intro} is equivalent to its rescaling \eqref{equ:nn_train_signdetinput_rescale}.
     By \Cref{lemma:dual_of_rescaled}, the dual of the rescaled problem  \eqref{equ:nn_train_signdetinput_rescale} is \eqref{gen_dual}. By \Cref{lemma:lasso_without_simp}, the dual of the dual \eqref{gen_dual} is the \lasso{} \eqref{eq:genericlasso}. 

    So by weak duality, $l^* \leq d^* \leq r^* = t^*$. On the other hand by \Cref{lemma:2layer_reconstruct2rescaled}, \Cref{lemma:reconstruct_ReLU2original}, and \Cref{lemma:reconstruct_bdd2original}, the resulting weights in \Cref{thrm:2layer_reconstruct} achieve the same objective in the training problem \eqref{eq:nonconvex_generic_loss_gen_L_intro} as $\mathbf{z}^*$ in \lasso{} \eqref{eq:genericlasso}. So $l^* = t^*$ (strong duality holds) and the resulting weights are optimal.

    The simplifications in \Cref{thrm:lasso_deep_1nueron_ReLU} follow from applying 
	\Cref{lemma:sym_activation_simp}, \Cref{lemma:sign_activation_simp}, and the fact that $c_1 = c_2 = 0$ for sign and threshold activations (\Cref{app_section:act_func}), to \Cref{lemma:lasso_without_simp}. 
    
\end{proof}

\begin{remark}
The proof of \Cref{thrm:lasso_deep_1nueron_ReLU} starts off similarly to \cite{pilanci2020neural}. The key difference is that \nd{1} data simplifies the dual \eqref{gen_dual} constraints so that they correspond to turning off a neuron on each training sample (\Cref{lemma:dual_of_rescaled}). This can be interpreted as neurons ``memorizing" training data.
\end{remark}

\iftoggle{long}{

\begin{theorem}\label{thrm:3layer_bdd_reconstruct}
    Consider a $3$-layer network with sign activation. Suppose $\mathbf{z}^*$ is optimal in the $3$-layer \lasso{} \eqref{thm_eq:3lyrsign} and $m \geq \|\mathbf{z}^*\|_0$. Construct $\alpha_i, \Wul{i}{1}, \mathbf{W^{(i,2)}}$ and amplitude parameters as given in \Cref{thm:3lyr_sign_reconstruction}. Apply the transformation given in \Cref{lemma:reconstruct_bdd2original}. The resulting reconstructed weights are globally optimal in the \nc{} training problem \eqref{eq:nonconvex_generic_loss_gen_L_intro}.
\end{theorem}

\begin{proof}
    By \Cref{thm:3lyr_sign}, the \lasso{} \eqref{thm_eq:3lyrsign} and training problem \eqref{eq:nonconvex_generic_loss_gen_L_intro} have the same optimal value. So it suffices to show that the reconstructed neural net achieves the same loss in the training problem \eqref{eq:nonconvex_generic_loss_gen_L_intro}
 as in \lasso{} \eqref{thm_eq:3lyrsign}.
    And this is shown by \Cref{thm:3lyr_sign_reconstruction} and \Cref{lemma:reconstruct_bdd2original}.
\end{proof}

\begin{remark}\label{rk:all0sol}
    If the \lasso{} (\eqref{thm_eq:3lyrsign}, \eqref{eq:genericlasso}) solution is $\mathbf{z}^*=\mathbf{0}, \xi=0$ then a neural net with all weights being zero achieves the same objective in  \eqref{eq:nonconvex_generic_loss_gen_L_intro} as the optimal \lasso{} value and is therefore optimal.
\end{remark}

The following lemmas are used to prove \Cref{thrm:3layer_bdd_reconstruct} and \Cref{thrm:2layer_reconstruct}, and they are organized as follows. We provide a map to transform a \lasso{}  \eqref{eq:genericlasso} into an optimal solution for the rescaled training problem \eqref{eq:gen_rescale}.
Then we provide a map to transform the optimal solution from the rescaled problem \eqref{equ:nn_train_signdetinput_rescale} to the original, \unscal{}ed training problem \eqref{eq:nonconvex_generic_loss_gen_L_intro}.

\subsection{\lasso{}  \eqref{eq:genericlasso} to rescaled problem \eqref{eq:nonconvex_generic_loss_gen_L_intro}}

\begin{lemma}\label{lemma:2layer_reconstruct2rescaled}
Let $L=2$ layers. Suppose $\xi^*, \mathbf{z}^*$ is feasible in \lasso{} \eqref{eq:genericlasso} with $m \geq ||\mathbf{z}^*||_0$ . 

Let $\mathbf{W} = [\mathbf{1}^T, -\mathbf{1}^T] \in \mathbb{R}^{1 \times 2N}$, $\mathbf{b} = [-\mathbf{X}^T, \mathbf{X}^T] \in \mathbb{R}^{1 \times 2N}$, $I = \{i \in [N]: z^*_i \neq 0 \}$ be the set of indices where $z$ is nonzero. Then $\alpha = \mathbf{z}^*_I, \mathbf{W_0} = \mathbf{W}_I, \mathbf{b} = \mathbf{b}_I$ (and $s = 1$ for sign and threshold) achieves the same value in the rescaled problem \eqref{equ:nn_train_signdetinput_rescale} . 



\end{lemma}

\begin{proof}
    By definition of the dictionary matrix $\Amat{}$ as given in \Cref{thrm:lasso_deep_1nueron_ReLU}, $f_2(\mathbf{X};\theta)=\sigma(\mathbf{XW}) \boldsymbol{\alpha} + \xi \mathbf{1}= \sigma \left([\mathbf{X}, \mathbf{1}] \begin{bmatrix} \mathbf{W_0} \\ b\end{bmatrix} \right) \boldsymbol{\alpha} + \xi  = \sigma \left([\mathbf{X}, \mathbf{1}] \begin{bmatrix} \mathbf{1}^T & -\mathbf{1}^T \\ -\mathbf{X}^T & \mathbf{X}^T \end{bmatrix}_{1:2,I} \right)\mathbf{z}_I + \xi^* = \sigma \left([\mathbf{X}, \mathbf{1}] \begin{bmatrix} \mathbf{1}^T & -\mathbf{1}^T \\ 
    -\mathbf{X}^T & \mathbf{X}^T \end{bmatrix} \right) \mathbf{z} + \xi^* = \mathbf{Az} + \xi^* \mathbf{1}$, so the training loss is $\mathcal{L}_{\mathbf{y}}(f_2(\mathbf{X};\theta)) = \frac{1}{2}||\mathbf{Az}+\xi^*-\mathbf{y}||^2_2$. And the regularization penalty is $\beta||\boldsymbol{\alpha}||_1= \beta||\mathbf{z}_I||_1 = \beta||\mathbf{z}||_1$. So with these parameter values, the neural net achieves the same objective in the rescaled primal \eqref{equ:nn_train_signdetinput_rescale} as the \lasso{} problem  \eqref{eq:genericlasso}. And since $m \geq m^* = ||\mathbf{z}||_0 = |I|$, the resulting neural net $f$ has at most $m$ neurons. 
\end{proof}

\iftoggle{long}{
\textbf{\Cref{thm:3lyr_sign_reconstruction}} (Informal).
\textit{
	Consider a $3$ layer neural net with sign activation. Suppose $\mathbf{z}^*$ is a solution for \eqref{thm_eq:3lyrsign}, and $m_2 \geq ||\mathbf{z}^* ||_0$.
	Let $\xi=0, s^{(i,l)}=1, \mathbf{W^{(i,2)}} = (1, -1, 1, -1, \cdots) \in \mathbb{R}^{m_1}$. Let $I = \{i: z_i \neq 0\}$. For every $i \in I$, let $\alpha_i = z^*_i$ and the columns of $\Wul{i}{1}$ be $(1, - x_j)$ for indices $j$ at which the $i^{th}$ column of $\Amat{}$ switches. 
Then the $3$ layer neural net with these parameters achieves the same objective in 
the rescaled problem \eqref{equ:nn_train_signdetinput_rescale}.
}
}

}

\Cref{fig:3lyr_arch_sign_tikz} illustrates an optimal neural net following the reconstruction given in \Cref{thm:3lyr_sign_reconstruction}.

\subsection{rescaled problem \eqref{eq:nonconvex_generic_loss_gen_L_intro} to original problem \eqref{eq:nonconvex_generic_loss_gen_L_intro}}

\begin{lemma}\label{lemma:reconstruct_ReLU2original}
    Consider a $2$-layer net with ReLU, leaky ReLU, or absolute value activation. Suppose $\mathbf{W}, \mathbf{b}, \boldsymbol{\alpha} $ are feasible in the reconstructed problem \eqref{equ:nn_train_signdetinput_rescale}. Let $\gamma_i =  \sqrt{|\alpha_i|}$. Then setting $\alpha_i \to \frac{\alpha_i}{\gamma_i}, w_i \to \gamma_i w_i, b_i \to \gamma_i b_i$ achieves the same value in the original problem \eqref{eq:nonconvex_generic_loss_gen_L_intro}.
\end{lemma}

\begin{proof}
Recall the rescaled problem \eqref{equ:nn_train_signdetinput_rescale} has the constraint $|w_i| = 1$. Plugging in $\alpha_i \to \frac{\alpha_i}{\gamma_i}, w_i \to \gamma_i w_i, b_i \to \gamma_i b_i$ makes the objective of the $2$-layer original problem \eqref{eq:nonconvex_generic_loss_2lyr} evaluate to $\mathcal{L}_{\mathbf{y}} ( \sum_i \frac{\alpha_i}{\gamma_i}\sigma (\gamma_i w_i \mathbf{X} + \gamma_i b_i)  + \xi - \mathbf{y} )   + \frac{\beta}{2} (\sum_{i} (\gamma_i w_i)^2 + (\frac{\alpha_i}{\gamma_i})^2 ) = \mathcal{L}_{\mathbf{y}} ( \sum_i \alpha_i \sigma (w_i \mathbf{X} +  b_i)  + \xi - \mathbf{y} )   + \frac{\beta}{2} (\sum_{i}  (1)^2|\alpha_i| + |\alpha_i| ) = \mathcal{L}_{\mathbf{y}} ( \sum_i \alpha_i \sigma (w_i \mathbf{X} +  b_i)  + \xi - \mathbf{y} )   + \beta ||\boldsymbol{\alpha}||_1 $, which is the objective of the rescaled problem \eqref{equ:nn_train_signdetinput_rescale}.
\end{proof}

For sign and threshold activation, the magnitudes of the optimal inner weights and amplitude parameters of the original primal go to zero.

\begin{theorem}\label{thrm:2layer_reconstruct}
	Suppose $\xi^*, \mathbf{z}^*:=(\mathbf{z^*_+}, \mathbf{z^*_-}) $ is  optimal in \eqref{eq:genericlasso} and
	\iftoggle{long}{Let $m^*=||\mathbf{z}^*||_0$.  Suppose the number of neurons is }
	$m \geq ||\mathbf{z}^*||_0$.
	Let $I^{+} = \{i \in [N]: (z^*_+)_i \neq 0 \}, I^{-} = \{i \in [N]: (z^*_-)_i \neq 0 \}$.
	\iftoggle{long}{. Define the row vectors }
     Let $\gamma^+_i = \sqrt{|z^*_+|}_i$ and $ \gamma^-_i = \sqrt{|z^*_-|}_i$.
    
    For each $i \in I_+$, let ${\alpha^+}_i = \mbox{sign}({z^*_+}_i) \gamma^+_i , (\mathbf{W_0}^+)_i =  \gamma^+_i , (b^+)_i = - x_i \gamma^+_i $. Similarly for each $i \in I_-$, let ${\alpha^-}_i = \mbox{sign}({z^*_-}_i )\gamma^-_i, (\mathbf{W_0}^-)_i =  -\gamma^-_i, (b^-)_i =  x_i\gamma^-_i$. 

    Then $\xi = \xi^*$ and $\{(\alpha_j,\mathbf{W_0}_j, \gamma_j, s_j ): j \in [m_1]\} = \{(\alpha^+_i,  (\mathbf{W_0}^+)_i, \gamma^+_i, s^+_i): i \in I^+\} \cup \{(\alpha^-_i,  (\mathbf{W_0}^-)_i, \gamma^-_i,s^+_i ):i \in I^-\}$  is optimal in the \nc{} training problem \eqref{eq:2layernn}.
    

	\iftoggle{long}{$I$ is the set of indices where $z^*$ is nonzero, ie the support of $\mathbf{z}^*$. We have $m \geq |I|$. }





\end{theorem}

The proof of theorem \ref{thrm:2layer_reconstruct} is in appendix \ref{app_section:2lyr_nn}. 

}

\begin{proof}[Proof of \Cref{lemma:minNormCert2neurons}]
    It can be verified that as shown in \Cref{fig:ReLUfeatdiagram}, the \lasso{} features for a \symmeterized{} network all have slope magnitude $0,1$, or $2$. However, only monotone features contain a segment with slope magnitude $2$, and the training data $(x_n,y_n)$ in \Cref{fig:ReLU3lyrnet} is not monotone. There is a "left branch" consisting of $\left\{(x_5,y_5)=(-1,1), (x_4,y_4)=(0,0)\right\}$ and a "right branch" consisting of $\left\{(x_2,y_2)=(3,1),(x_1,y_1)=(4,2)\right\}$. Let $\mathbf{z}^*, \xi^*$ be a \lasso{} solution that fits the data exactly: $\mathbf{A}\mathbf{z}^*+\xi^*=\mathbf{y}$. Let $\mathcal{I}^-$ ($\mathcal{I}^+$) be the set of indices $i$ where the $i^{th}$ feature is monotone and has negative (positive) slope over the left (right) branch. Let $\mathcal{I}^0$ be the set of indices corresponding to features that are not monotone. Let $m^-$ and $m^+$ be the magnitude of the slopes of the left and right branch, respectively. Then $2\sum_{i \in \mathcal{I}^-} \left|z^*_i\right| + \sum_{i \in \mathcal{I}^0}  \left|z^*_i\right| \geq m^-$ and  $2\sum_{i \in \mathcal{I}^+} \left|z^*_i\right|  + \sum_{i \in \mathcal{I}^0}  \left|z^*_i\right| \geq m^+$. Note $\mathcal{I}^-,\mathcal{I}^+,\mathcal{I}^0$ are all pairwise disjoint. Therefore $\|\mathbf{z}^*\|_1 \geq \frac{m^++m^-}{2}=1$.
\end{proof}

\begin{proof}[Proof of \Cref{lemma:minNormCert}]
Since ReLU and absolute value activations have slopes $\pm 1$ or $0$, and the weights of \deepnarrow{} networks are $\pm 1$, the features $\xl{L}(x)$ have slopes $\pm 1$ or $0$. Observe $\mathbf{y} = \nn{L}{\mathbf{X}}  = \xi^*\mathbf{1}  + \Amat \mathbf{z}^* = \xi^*\mathbf{1} + \sum_{i} z^*_i \Amat_i $. For any $n \in [N-1]$, $\left| \slope_n \right| = \left| \frac{\nn{L}{\mathbf{X}}_{n+1}-\nn{L}{\mathbf{X}}_{n}}{x_{n+1}-x_n} \right| = \left| \frac{ \sum_{i} z^*_i \left( \Amat_{n+1,i} - \Amat_{n,i}\right) }{x_{n+1}-x_n}\right| = \left| \sum_{i} z^*_i \frac{   \xl{L}(x_{n+1}) - \xl{L}(x_{n}) }{x_{n+1}-x_n} \right|\\  \leq  \sum_{i} \left|z^*_i \right| \left| \frac{    \xl{L}(x_{n+1}) - \xl{L}(x_{n}) }{x_{n+1}-x_n}\right| \leq \sum_i |z^*_i| = \|\mathbf{z}^*\|$. 
\end{proof}


\begin{proof}[Proof of \Cref{lemma:minnormL2ReLUsol}]
%
Let $n \in [N-1]$. Let $\mathcal{S}^+_n = \{i \in [N]: i>n, {\zposscalar}_i \neq 0\}, \mathcal{S}^-_n = \{i \in [N]: i \leq n, {\znegscalar}_i \neq 0\}$. 
Observe $\mathcal{S}^+_{n+1}=\mathcal{S}^+_{n} - \{n+1\}$ if ${\zposscalar}_{n+1} \neq 0$ and $\mathcal{S}^+_{n+1}=\mathcal{S}^+_{n}$ otherwise. Similarly, $\mathcal{S}^-_{n+1}=\mathcal{S}^+_{n} \cup \{n+1\}$ if ${\znegscalar}_{n+1} \neq 0$ and $\mathcal{S}^-_{n+1}=\mathcal{S}^-_{n}$ otherwise.
Now, $\ReLU{+}{x_i}$ has slope $1$ after $x_i$ and $\ReLU{-}{x_i}$ has slope $1$ before $x_i$, so $\slope_n = \sum_{i \in \mathcal{S}^+_n} {\zposscalar}_i + \sum_{i \in \mathcal{S}^-_n}{\znegscalar}_i $. Therefore, 
$\left|\slope_{n}-\slope_{n+1}\right| = \left|-{\zposscalar}_{n+1} + {\znegscalar}_{n+1}\right| \leq  \left|{\zposscalar}_{n+1}\right| + \left|{\znegscalar}_{n+1}\right|$. So $\sum_{n=1}^{N-1}|\slope_{n}-\slope_{n+1}| \leq \|\mathbf{z}^* \|_1 - |{\zposscalar}_{1}| - |{\znegscalar}_1| \leq \|\mathbf{z}^*\|_1$.

Now, for any $n \in [N-1]$, $\left(\Amat \mathbf{z}{+}\xi \mathbf{1}\right)_n {-} \left(\Amat \mathbf{z}{+}\xi \mathbf{1}\right)_{n+1} {=} \sum_{i=1}^N \zposscalar_i \left( {\Apos}_{n,i} {-} {\Apos}_{n+1,i} \right) {=} \\ \sum_{i=1}^N \zposscalar_i \left( \left(x_n {-} x_i \right)_+  {-} \left(x_{n+1} {-} x_i \right)_+ \right) {=} \sum_{i=n+1}^N \left(\slope_{i-1} {-}\slope_i \right) \left( x_n - x_{n+1} \right)  {=} \left( x_n {-} x_{n+1} \right) \slope_{n} {=} y_n - y_{n+1}
$. And $(\Amat\mathbf{z}+\xi \mathbf{1})_N {=} \xi+\sum_{i=1}^N \zposscalar_i \left( x_N - x_i\right)_+ {=} y_N {+} \sum_{i=1}^N \zposscalar_i (0) {=} y_N$. So $\Amat\mathbf{z}{+}\xi \mathbf{1}{=}\mathbf{y}$. And $\|\mathbf{z}\|_1 = \sum_{n=1}^{N-1}\left| \slope_{n} {-}\slope_{n+1} \right|$, which exactly hits the lower bound on $\|\mathbf{z}^*\|_1 $. Therefore $\mathbf{z}, \xi$ is optimal.
\end{proof}

\begin{remark}\label[remark]{rk:diff_detector}
    By \Cref{lemma:A_absval_full_rk}, 
    the absolute value network ``de-biases" the \labelvec{}, normalizes is by the interval lengths $x_i - x_{i+1}$, and applies $\mathbf{E}$ (which contains the difference matrix $\Delta$) twice, acting as a second-order difference detector. By \Cref{lemma:A_full_rk}, the sign network's dictionary inverse contains $\Delta$ just once, acting as a first-order difference detector.
\end{remark}

\section*{Inverses of $2$-layer dictionary matrices}

\newcommand{\bigDelta}{\mbox{\normalfont\Large\bfseries $\Delta$}}

In this section, 
we consider the $2$-layer dictionary matrix $\Amat$ as defined in \Cref{Amat2lyrs}. 
Define the \textit{finite difference matrix} 

 \begin{equation}
\Delta =
\scalemath{0.75}{
    \begin{pmatrix}
        1 & -1 & 0 &   \cdots & 0 & 0 \\
        0 & 1 & -1 &    \cdots & 0 & 0\\
        0 & 0 & 1 &    \cdots & 0 & 0\\
        \vdots & \vdots  &\vdots  &\ddots & \vdots & \vdots\\
        0 & 0 & 0 &  \cdots & 1 & -1 \\
        0 & 0 & 0 &  \cdots & 0 & 1 
    \end{pmatrix}
    }
    \iftoggle{long}{
    \mbox{ i.e., }
    \begin{aligned}
         \Delta_{i,j} = 
         \begin{cases}
             1 & \mbox { if } i=j, i \in [N-1] \\
            -1 & \mbox{ if } j = i+1, i \in [N-2] \\
            0 & \mbox{ else }
         \end{cases}
     \end{aligned}
              }
              \in \mathbb{R}^{N-1 \times N-1}.
\end{equation}

Multiplying a matrix on its right by $\Delta$ subtracts its consecutive rows. Define the diagonal matrix $\mathbf{D} \in \mathbb{R}^{N \times N}$ by $\mathbf{D}_{i,i} = \frac{1}{x_{i}-x_{i+1}}$ for $i \in [N-1]$ and $\mathbf{D}_{N,N} = \frac{1}{x_1 - x_N}$. For $n \in \mathbb{N}$, let   

\begin{equation}\label{eq:AsnAtn}
		\setcounter{MaxMatrixCols}{20}
		\begin{aligned}
            \Asign{n} =
            \scalemath{0.75}{
			\begin{pmatrix}
				1      & 1      & 1            & \cdots       & 1      \\
				-1     & 1      & 1            & \cdots       & 1      \\
				-1     & -1     & 1            & \cdots       & 1      \\
				\vdots & \vdots & \vdots  & \ddots       & \vdots \\
				-1     & -1     & -1          & \cdots       & 1
			\end{pmatrix}} ,
		\setcounter{MaxMatrixCols}{20}
		\begin{aligned}
            \Athresh{n}  =
            \scalemath{0.75}{
			\begin{pmatrix}
				1      & 1      & 1           & \cdots       & 1      \\
				0     & 1      & 1           & \cdots       & 1      \\
				0     & 0     & 1          & \cdots       & 1      \\
				\vdots & \vdots & \vdots & & \ddots       & \vdots \\
				0     & 0     & 0         & \cdots       & 1
			\end{pmatrix} 
   }
		\end{aligned}
  \in \mathbb{R}^{n \times n}.
		\end{aligned}
	\end{equation}

\begin{remark}\label[remark]{remk:Asignform}
The dictionary matrices for sign and threshold activation satisfy $\Amat= \Asign{N}$ and $\Apos=\Athresh{N}$, respectively.

\iftoggle{long}{
    The dictionary matrix for sign activation 
    is $\Amat=\Amat{}^{(s)}_N$. 
 }
 
   \iftoggle{long}{    \begin{equation*}
		\setcounter{MaxMatrixCols}{20}
		\begin{aligned}
			\Amat{} & =
			\begin{pmatrix}
				1      & 1      & 1      & 1      & \cdots       & 1      \\
				-1     & 1      & 1      & 1      & \cdots       & 1      \\
				-1     & -1     & 1      & 1      & \cdots       & 1      \\
				-1     & -1     & -1     & 1      & \cdots       & 1      \\
				\vdots & \vdots & \vdots & \vdots & \ddots       & \vdots \\
				-1     & -1     & -1     & -1     & \cdots       & 1
			\end{pmatrix}  
  \in \{-1,1\}^{N \times N}.
		\end{aligned}
	\end{equation*}} 

\end{remark}

\iftoggle{long}{
\begin{remark}\label{remk:Athreshform}
    The partial dictionary matrix for threshold activation is $\Apos$
     $= \Amat{}^{(t)}_N $. 
    \iftoggle{long}{
    \begin{equation*}
		\setcounter{MaxMatrixCols}{20}
		\begin{aligned}
			\Amat{} & =
			\begin{pmatrix}
				1      & 1      & 1      & 1      & \cdots       & 1      \\
				0     & 1      & 1      & 1      & \cdots       & 1      \\
				0     & 0     & 1      & 1      & \cdots       & 1      \\
				0     & 0     & 0     & 1      & \cdots       & 1      \\
				\vdots & \vdots & \vdots & \vdots & \ddots       & \vdots \\
				0     & 0     & 0     & 0     & \cdots       & 1
			\end{pmatrix}  
  \in \{-1,1\}^{N \times N}.
		\end{aligned}
	\end{equation*}
 }
\end{remark}
}

\subsection{results about $2$-layer dictionary matrices}\label{section:dic_matrices}

\begin{lemma}\label[lemma]{lemma:A_full_rk}
	The dictionary matrix for $\sigma(x){=}\mathrm{sign}(x)$ 
 has inverse 
    $
    \Amat{}^{-1}{=}\frac{1}{2}
    \scalemath{0.7}{
    \begin{pNiceArray}{cccc|c}[margin] 
    \Block{4-4}<\Large>{\Delta} 
    & & & & 0 \\
    & & & & \vdots \\
    & & & & 0 \\
    & & & & -1 \\
    \hline
   1& 0 & \cdots& 0 & 1
   \end{pNiceArray}
   }
   $.

\iftoggle{long}{
    
 \begin{equation}
\Amat{}^{-1}=\frac{1}{2}
    \begin{pmatrix}
        1 & -1 & 0 & 0 & 0 & \cdots & 0 & 0 \\
        0 & 1 & -1 & 0 &  0 & \cdots & 0 & 0\\
        0 & 0 & 1 & -1 &  0 & \cdots & 0 & 0\\
        0 & 0 & 0 & 1 & -1 & \cdots & 0 & 0\\
        0 & 0 & 0 & 0 & 0 &\cdots & 0 & 0\\
        \vdots & \vdots &\vdots &\vdots &\vdots  &\ddots & \vdots & \vdots\\
        0 & 0 & 0 & 0  & 0 & \cdots & 1 & -1 \\
        1 & 0 & 0 & 0 & 0 & \cdots & 0 & 1 
    \end{pmatrix}, \mbox{ i.e., }
    \begin{aligned}
         \Amat{}^{-1}_{i,j} = \frac{1}{2}
         \begin{cases}
             1 & \mbox { if } i=j, i \in [N] \\
             1 & \mbox { if } (i,j)=(N,1) \\
            -1 & \mbox{ if } j = i+1, i \in [N-1] \\
            0 & \mbox{ else }
         \end{cases}.
     \end{aligned}
\end{equation}
}

\end{lemma}

\begin{proof}
Multiplying the two matrices (see \Cref{remk:Asignform}) gives the identity.
    \iftoggle{long}{
	Let $\mathbf{h} \in \mathbb{R}^N$. We show that we can solve the linear system $A\mathbf{w}=\mathbf{h}$ for $\mathbf{w}$.
	\iftoggle{long}{ The elements of $A$ are $1$ in the upper right corner up to and including the diagonal, and are $-1$ otherwise.}
	Let $\mathbf{w} \in \mathbb{R}^{N}$ be defined as follows.
	For $i \in [N-1]$, let $w_{i} = \frac{h_{i} - h_{{i+1}}}{2}$. And let $w_{N} = \frac{h_{1}+h_{N}}{2}$. Then for $i \in [n]$, $(Aw)_{i} = \sum_{j=1}^{i-1} (-1) w_j + \sum_{j=i}^{N-1} (1) w_j + w_N = -\frac{1}{2}\sum_{j=1}^{i-1} (h_{j} - h_{{j+1}})  + \frac{1}{2}\sum_{j=i}^{N-1} (h_{j} - h_{{j+1}}) + \frac{{h_1}+h_{N}}{2} = \frac{1}{2}(-(h_{1}-h_{i}) + (h_{i}-h_{N}) + (h_{n}+h_{1}))=h_{i}$.
	So $Aw=h$. 
 }
\end{proof}

\begin{lemma}\label[lemma]{lemma:A_absval_full_rk}
	The dictionary matrix $\Amat$ for absolute value activation 
 has inverse $\Amat{}^{-1}=\frac{1}{2} \mathbf{PEDE}$, where 
 \iftoggle{long}{ $\mathbf{P}$ is a permutation matrix, $\frac{1}{2}\mathbf{E}$ is the inverse matrix given in \Cref{lemma:A_full_rk} but with its last row negated, and $\mathbf{D}$ is a diagonal matrix. Specifically, $\mathbf{D}$ is defined by $\mathbf{D}_{i,i} = \frac{1}{x_{i}-x_{i+1}}$ for $i \in [N-1]$ and $\mathbf{D}_{N,N} = \frac{1}{x_1 - x_N}$. And the matrices $\mathbf{P}, \mathbf{E} \in \mathbb{R}^{N\times N}$ are given by,
}
%
    $
    \mathbf{P}=
    \scalemath{0.75}{
    \begin{pNiceArray}{cccc|c}[margin] 
    0& 0 & \cdots& 0 & 1 \\
    \hline
    \Block{4-4}<\Large>{I_{N-1}} 
    & & & & 0 \\
    & & & & \vdots \\
    & & & & 0 \\
    & & & & 0 \\
   \end{pNiceArray}},
    \mathbf{E}=
    \scalemath{0.75}{
    \begin{pNiceArray}{cccc|c}[margin] 
    \Block{4-4}<\Large>{\Delta} 
    & & & & 0 \\
    & & & & \vdots \\
    & & & & 0 \\
    & & & & -1 \\
    \hline
   -1& 0 & \cdots& 0 & -1
   \end{pNiceArray}}.
   $
%
\iftoggle{long}{and $\mathbf{D}$ is a diagonal matrix with $\mathbf{D}_{i,i} = \frac{1}{x_{i}-x_{i+1}}$ for $i \in [N-1]$ and $\mathbf{D}_{N,N} = \frac{1}{x_1 - x_N}$. }
\iftoggle{long}{
 \begin{equation}
\mathbf{P} = 
    \begin{pmatrix}
        0 & 0 & 0 & 0 & 0 & \cdots & 0 & 1 \\
        1 & 0 & 0 & 0 & 0 & \cdots & 0 & 0 \\
        0 & 1 & 0 & 0 & 0 & \cdots & 0 & 0 \\
        0 & 0 & 1 & 0 & 0 & \cdots & 0 & 0 \\
        0 & 0 & 0 & 1 & 0 &\cdots & 0  & 0 \\
        0 & 0 & 0 & 0 & 1 &\cdots & 0  & 0 \\
        \vdots & \vdots &\vdots &\vdots &\vdots  &\ddots & \vdots & \vdots\\
        0 & 0 & 0 & 0  & 0 & \cdots & 1 & 0  
    \end{pmatrix}, 
\mathbf{E} =
    \begin{pmatrix}
        1 & -1 & 0 & 0 & 0 & \cdots & 0 & 0 \\
        0 & 1 & -1 & 0 &  0 & \cdots & 0 & 0\\
        0 & 0 & 1 & -1 &  0 & \cdots & 0 & 0\\
        0 & 0 & 0 & 1 & -1 & \cdots & 0 & 0\\
        0 & 0 & 0 & 0 & 0 &\cdots & 0 & 0\\
        \vdots & \vdots &\vdots &\vdots &\vdots  &\ddots & \vdots & \vdots\\
        0 & 0 & 0 & 0  & 0 & \cdots & 1 & -1 \\
        -1 & 0 & 0 & 0 & 0 & \cdots & 0 & -1 
    \end{pmatrix}.  
\end{equation}
}
\end{lemma}

\begin{proof}
For $i \in [N-1], j \in [N]$, 
$
\begin{aligned}
      \Amat{}_{i,j} {-} \Amat{}_{i+1,j} {=} 
      \begin{cases}
          (x_j {-} x_i) {-} (x_j {-} x_{i+1}) {=} x_{i{+}1}{-}x_i &\mbox { if } i {>} j  \\
          (x_i {-} x_j) {-} (x_{i+1} {-} x_{j}) {=} x_{i}-x_{i+1} &\mbox { if }  i {\leq} j.
      \end{cases}
\end{aligned}    
$
\iftoggle{long}{So $\mathbf{a}^T_{i+1} - \mathbf{a}^T_i = (x_i - x_{i+1})(-\mathbf{1}^T_i, \mathbf{1}^T_{N-i}) $.}
And for all $j \in [N]$, $\Amat{}_{1,j}+\Amat{}_{N,j} = (x_1 - x_j) + (x_j - x_N) = x_1 + x_N$. 
\iftoggle{long}{So $\mathbf{a}^T_1 + \mathbf{a}^T_N = (x_1 + x_N) \mathbf{1}^T$. }
\iftoggle{long}{Multiplying $\mathbf{EA}$ by $\mathbf{D}$ normalizes elements to $1$. } Therefore,
 \begin{equation*} 
  \mathbf{DEA}=
  \scalemath{0.75}{
 \begin{pNiceArray}{c|ccc}[margin]
  -1       &       &                                        &    \\
 \vdots    &       &\mbox{\large{$\Asign{N-1}$} }    &    \\
  -1       &       &                                        &     \\
  \hline
  -1       &-1     &\cdots                                  &-1   \\
  \end{pNiceArray}}, \quad
      \frac{1}{2} \mathbf{EDEA}=
     \scalemath{0.75}{ 
 \begin{pNiceArray}{c|ccc}[margin]
  0       &       &                                        &    \\
 \vdots    &       &\mbox{\large{$I_{N-1}$} }    &    \\
  0       &       &                                        &     \\
  \hline
  1       &0     &\cdots                                  &0   \\
  \end{pNiceArray} }.
    \end{equation*}
    \iftoggle{long}{
      \begin{equation}
\mathbf{D E A }=
    \begin{pmatrix}
        -1 & 1 & 1 & 1 &  \cdots & 1 & 1 \\
        -1 & -1 & 1 & 1 &   \cdots & 1 & 1\\
        -1 & -1 & -1 & 1 &   \cdots & 1 & 1\\
        -1 & -1 & -1 & -1 &  \cdots & 1 & 1\\
        \vdots & \vdots &\vdots  &\vdots  &\ddots & \vdots & \vdots\\
        -1 & -1 & -1 & -1  & \cdots & -1 & 1 \\
        -1 & -1 & -1 & -1 &  \cdots & -1 & -1 
    \end{pmatrix}, 
    \frac{1}{2} \mathbf{EDEA} =
    \begin{pmatrix}
        0 & 1 & 0 & 0 &  \cdots & 0 & 0 \\
        0 & 0 & 1 & 0 & \cdots & 0 & 0 \\
        0 & 0 & 0 & 1 & \cdots & 0 & 0 \\
        0 & 0 & 0 & 0 & \cdots & 0 & 0 \\
        \vdots & \vdots  &\vdots &\vdots  &\ddots & \vdots & \vdots\\
        0 & 0 & 0 & 0  &  \cdots & 0 & 1 \\
        1 & 0 & 0 & 0 &  \cdots & 0 & 0 
    \end{pmatrix}.
    \end{equation}
    }
Applying the permutation $\mathbf{P}$ makes $\frac{1}{2}\mathbf{PEDE A} = \mathbf{I}$, so $\mathbf{A^{-1}} = \frac{1}{2}\mathbf{PEDE}$.
\end{proof}

\begin{lemma}\label[lemma]{lemma:Athresh_full_rk}
	The inverse of $\Apos$ for threshold activation 
is 
 %
%
  $
    \Amat{}_+^{-1}=
    \scalemath{0.75}{
    \begin{pNiceArray}{cccc|c}[margin] 
    \Block{4-4}<\Large>{\Delta} 
    & & & & 0 \\
    & & & & \vdots \\
    & & & & 0 \\
    & & & & -1 \\
    \hline
   0& 0 & \cdots& 0 & 1
   \end{pNiceArray}}.
   $
%
\iftoggle{long}{
 \begin{equation}
\Amat{}_+^{-1}=
    \begin{pmatrix}
        1 & -1 & 0 & 0 & 0 & \cdots & 0 & 0 \\
        0 & 1 & -1 & 0 &  0 & \cdots & 0 & 0\\
        0 & 0 & 1 & -1 &  0 & \cdots & 0 & 0\\
        0 & 0 & 0 & 1 & -1 & \cdots & 0 & 0\\
        0 & 0 & 0 & 0 & 0 &\cdots & 0 & 0\\
        \vdots & \vdots &\vdots &\vdots &\vdots  &\ddots & \vdots & \vdots\\
        0 & 0 & 0 & 0  & 0 & \cdots & 1 & -1 \\
        0 & 0 & 0 & 0 & 0 & \cdots & 0 & 1 
    \end{pmatrix}, \mbox{ i.e., }
    \begin{aligned}
         \Amat{}^{-1}_{i,j} = 
         \begin{cases}
             1 & \mbox { if } i=j, i \in [N] \\
            -1 & \mbox{ if } j = i+1, i \in [N-1] \\
            0 & \mbox{ else }
         \end{cases}.
     \end{aligned}
\end{equation}
}
\end{lemma}

\begin{proof}
Multiplying the two matrices (see \Cref{remk:Asignform}) gives the identity.
\end{proof}

\begin{lemma}\label[lemma]{lemma:ReLU_fullrk}
The submatrix $ \left[(\mathbf{A_+})_{1:N, 2:N}, (\mathbf{A_-})_{1:N,1} \right] \in \mathbb{R}^{N \times N}$ of the dictionary matrix for ReLU activation
has inverse $ \mathbf{ E_+ D E_-}$, where



\iftoggle{long}{$\Amat{}^{-1} = \mathbf{\tilde{E}\tilde{I} DE}$}

\iftoggle{long}{$\mathbf{D}_{i,i} = \frac{1}{x_{i}-x_{i+1}}$ for $i \in [N-1]$, and $\mathbf{D}_{N,N} = \frac{1}{x_1 - x_N}$ and,
}

  \begin{equation*}
    \mathbf{E}_+ =
    \scalemath{0.75}{
    \begin{pNiceArray}{cccc|c}[margin] 
    \Block{4-4}<\Large>{\Delta} 
    & & & & 0 \\
    & & & & \vdots \\
    & & & & 0 \\
    & & & & 1 \\
    \hline
   0& 0 & \cdots& 0 & 1
   \end{pNiceArray}} , \quad
    \mathbf{E}_- =
    \scalemath{0.75}{
    \begin{pNiceArray}{cccc|c}[margin] 
    \Block{4-4}<\Large>{\Delta} 
    & & & & 0 \\
    & & & & \vdots \\
    & & & & 0 \\
    & & & & -1 \\
    \hline
   0& 0 & \cdots& 0 & -1
   \end{pNiceArray}}.
    \end{equation*}

\iftoggle{long}{

  \begin{equation*}
   \mathbf{\tilde{E}}=
    \begin{pNiceArray}{cccc|c}[margin] 
    \Block{4-4}<\Large>{\Delta} 
    & & & & 0 \\
    & & & & \Vdots \\
    & & & & 0 \\
    & & & & 0 \\
    \hline
   0& 0 & \Cdots& 0 & 1
   \end{pNiceArray},
  \mathbf{\tilde{I}}=
    \begin{pNiceArray}{cccc|c}[margin] 
    \Block{4-4}<\Large>{I_{N-1}} 
    & & & & 1 \\
    & & & & \Vdots \\
    & & & & 1 \\
    & & & & 1 \\
    \hline
   0& 0 & \Cdots& 0 & 1
   \end{pNiceArray},
    \mathbf{E}=
    \begin{pNiceArray}{cccc|c}[margin] 
    \Block{4-4}<\Large>{\Delta} 
    & & & & 0 \\
    & & & & \Vdots \\
    & & & & 0 \\
    & & & & -1 \\
    \hline
   0& 0 & \Cdots& 0 & -1
   \end{pNiceArray}.
    \end{equation*}

 \begin{equation}
 \begin{aligned}
\mathbf{\tilde{I}} = 
    \begin{pmatrix}
        1 & 0 & 0 & 0 & 0 & \cdots & 0 & 1 \\
        0  & 1 & 0 & 0 & 0 & \cdots & 0 & 1 \\
        0 & 0 & 1 & 0 & 0 & \cdots & 0 & 1 \\
        0 & 0 & 0 & 1 & 0 & \cdots & 0 & 1 \\
        0 & 0 & 0 & 0 & 1 &\cdots & 0  & 1 \\
        0 & 0 & 0 & 0 & 0 &\cdots & 0  & 1 \\
        \vdots & \vdots &\vdots &\vdots &\vdots  &\ddots & \vdots & \vdots\\
        0 & 0 & 0 & 0  & 0 & \cdots & 0 & 1  
    \end{pmatrix}, 
\mathbf{E} =
    \begin{pmatrix}
        1 & -1 & 0 & 0 & 0 & \cdots & 0 & 0 \\
        0 & 1 & -1 & 0 &  0 & \cdots & 0 & 0\\
        0 & 0 & 1 & -1 &  0 & \cdots & 0 & 0\\
        0 & 0 & 0 & 1 & -1 & \cdots & 0 & 0\\
        0 & 0 & 0 & 0 & 0 &\cdots & 0 & 0\\
        \vdots & \vdots &\vdots &\vdots &\vdots  &\ddots & \vdots & \vdots\\
        0 & 0 & 0 & 0  & 0 & \cdots & 1 & -1 \\
        0 & 0 & 0 & 0 & 0 & \cdots & 0 & -1 
    \end{pmatrix}, \\
\mathbf{\tilde{E}} =
    \begin{pmatrix}
        1 & -1 & 0 & 0 & 0 & \cdots & 0 & 0 \\
        0 & 1 & -1 & 0 &  0 & \cdots & 0 & 0\\
        0 & 0 & 1 & -1 &  0 & \cdots & 0 & 0\\
        0 & 0 & 0 & 1 & -1 & \cdots & 0 & 0\\
        0 & 0 & 0 & 0 & 0 &\cdots & 0 & 0\\
        \vdots & \vdots &\vdots &\vdots &\vdots  &\ddots & \vdots & \vdots\\
        0 & 0 & 0 & 0  & 0 & \cdots & 1 & 0 \\
        0 & 0 & 0 & 0 & 0 & \cdots & 0 & 1 
    \end{pmatrix}       
 \end{aligned}
\end{equation}
}

\end{lemma}

\begin{proof}
For $i \in [N-1], j \in [N]$, \\
$
\begin{aligned}
      (\Apos{})_{i,j} - (\Apos{})_{i+1,j} = 
      \begin{cases}
          0 &\mbox { if } i \geq j \\
          (x_i - x_j) - (x_{i+1}- x_j) = x_i - x_{i+1} &\mbox { if }  i < j \\
      \end{cases}
\end{aligned}  
$ \\
and $(\Aneg{})_{i,1}-(\Aneg{})_{i+1,1}= (x_1 - x_i) - (x_1 - x_{i+1}) = x_{i+1}-x_i$.
%
\iftoggle{long}{  So the difference of rows $i,i+1$ of $\Amat{}$ is, $(x_i - x_{i+1})(0, \cdots, 0, 1, \cdots, 1 ) $. }
 Observe that
 $
\mathbf{D E_- A }=
\scalemath{0.75}{
    \begin{pNiceArray}{cccc|c}[margin] 
    \Block{4-4}<\Large>{ \Athresh{N-1} } 
    & & & & -1 \\
    & & & & {\vdots} \\
    & & & & -1 \\
    & & & & -1 \\
    \hline
   0& 0 & {\cdots}& 0 & 1
   \end{pNiceArray}
   }
   \iftoggle{long}{
    \begin{pmatrix}
        1 & 1 & 1 & 1 &  \cdots & 1 & -1 \\
        0 & 1 & 1 & 1 &   \cdots & 1 & -1\\
        0 & 0 & 1 & 1 &   \cdots & 1 & -1\\
        0 & 0 & 0 & 1 &  \cdots & 1 & -1\\
        \vdots & \vdots  &\vdots &\vdots  &\ddots & \vdots & \vdots\\
        0 & 0 & 0 & 0  &  \cdots & 1 & -1 \\
        0 & 0 & 0 & 0  &  \cdots & 0 & 1 
    \end{pmatrix}
    }
    , 
 $
    and applying $\mathbf{E_+}$ gives $\mathbf{I}$. 
\end{proof}

\iftoggle{long}{
\begin{remark}
    In \Cref{lemma:A_absval_full_rk}, $\mathbf{EA} + 2\mathbf{I} = \mathbf{A_{sign}}$, which was defined in \Cref{remk:Asignform}.
\end{remark}
}

\iftoggle{long}{
\begin{remark}
   Consider minimum regularization: $\beta \to 0$. \Cref{lemma:A_absval_full_rk} and \Cref{lemma:A_full_rk} imply that the sign activation network is a difference detector since its weights are derived from applying $\Delta$ to $\mathbf{y}$. The absolute value network is a second-order difference detector since $\Delta$ is applied  twice to $\mathbf{y}$. In this sense, the sign activation takes a discrete derivative of $\mathbf{y}$, while the sign activation takes a discrete second derivative. Interestingly, the sign activation is the derivative of the absolute value activation. This suggests the sign activation "compensates" by using one less derivative in its network compared to the sign activation.
\end{remark}
}


\iftoggle{long}{
\section{Properties of vectors in $\{-1,1\}^N$}\label{section:1D_dictionary}

In this section, 
we assume $\sigma$ is sign activation.
We will refer to the \switchsetname{} $\switchmatrix{n}$ defined in \Cref{sec:deepsign}. 

}
\iftoggle{long}{
\begin{definition}
Let $I_1 \cdots, I_m$ be integers such that $1 < I_1 < \cdots < I_m \leq N$. Let $h$ be a column of $\switchmatrix{m}$ that switches precisely at  $I_1 \cdots, I_m$ and $u= h+\mathbf{1}$.

$\mathcal{I} = \{I_1, \cdots, I_m\}$ be an ordered set of integers where $1 = I_0 < I_1 < I_2 < \dots < I_m \leq N$. Denote $I_0=1$ and $I_{m+1}=N+1$.
For index $n \in [N]$, let $g(n) = \max \{k \in \{0, \cdots, m\}: n \geq I_k\}$ be the last index in $\mathcal{I}$ before $n$. In other words, $n \in \left \{I_{g(n)}, \cdots I_{g(n)+1}-1 \right\}$.
Define $ h(\mathcal{I})  \in \{-1,1\}^N$ as a vector with $n^{th}$ element \\ 

$
	\begin{aligned}
		h(\mathcal{I})_n =
		\begin{cases}
			1 & \mbox { if } g(n) \mod 4 \mbox { is even }  \\
         -1 & \mbox { else } 
		\end{cases}
  =\begin{cases}
			1 \text{ if }  n \in \left \{I_{2i}, \cdots I_{2i+1}-1 \right\} \text{ for some }  i \geq 0 \\
			-1 \text{ else. }
		\end{cases}
	\end{aligned} \\
$

Let $ u(\mathcal{I}) \in \{-2,0\}^N$ as a vector with $n^{th}$ element

$
	\begin{aligned}
		u(\mathcal{I})_n =
  \begin{cases}
			0 & \mbox { if } g(n) \mod 4 \mbox { is even }  \\
         -2 & \mbox { else } 
		\end{cases}
  =
		\begin{cases}
			0 \text{ if }  n \in \{I_{2i}, \cdots I_{2i+1} \} \text{ for some }  i \geq 0 \\
			-2 \text{ else. }
		\end{cases}
	\end{aligned}
$

With abuse of notation, $h(I_1,\dots,I_m), u(I_1,\dots,I_m)$ will denote $h(\mathcal{I}), u(\mathcal{I})$, respectively.

\end{definition}
}

\iftoggle{long}{
The vector $h(\mathcal{I})$ starts at $h(\mathcal{I})_1 = 1$ and switches sign at indices in 
$\mathcal{I}$. Note $h(\mathcal{I})$ switches $|\mathcal{I}|$ times and is a column of $\switchmatrix{|I|}$. \Cref{fig:h_ex} illustrates an example of $h(\mathcal{I})$.
}

\iftoggle{long}{

\begin{equation*}
	\begin{aligned}
		h(I_1,\dots,I_m)_n =
		\begin{cases}
			1 \text{ if }  n \in [I_{2i}, I_{2i+1}) \text{ for some }  i \geq 0 \\
			-1 \text{ else }.
		\end{cases}
	\end{aligned}
\end{equation*}

In other words, the vector $h(I_1,\dots,I_m)$ starts at $h(I_1,\dots,I_m)_1 = 1$ and switches sign at indices $I_1, \cdots, I_m$. We also denote $h(\mathcal{I}) = h(I_1, \cdots, I_m)$ where $\mathcal{I }= \{I_1, \cdots, I_m\}$.  
Let $u(I_1, \cdots, I_m) \in \{-2,0\}^N$ be defined by, 

\begin{equation*}
	\begin{aligned}
		u(I_1,\dots,I_m)_n =
		\begin{cases}
			0 \text{ if }  n \in [I_{2i}, I_{2i+1}) \text{ for some }  i \geq 0 \\
			-2 \text{ else }
		\end{cases}.
	\end{aligned}
\end{equation*}
}


\iftoggle{long}{

\begin{remark}\label{remk:hu}
Let $I_1, \cdots, I_j$ and $I_k, \cdots, I_m$ be integers such that $1 < I_1 < \cdots < I_j < I_k < \cdots < I_m \leq N$. Let $u_1, u_2, u_3 \in \{0,-2\}^N$ starting with $0$ with switch points precisely at $I_1, \cdots, I_j$; $I_k, \cdots, I_m$; and $I_1, \cdots, I_j, I_k, \cdots, I_m$. Then $u_1+u_2=u_3$.
\iftoggle{long}{
Also, for any $m$, if $1 < I_1 <  \cdots < I_j < \cdots < I_k < \cdots I_m \leq N $ then $u(I_1,\cdots, I_j ) + u(I_k,\cdots, I_m) = u(\mathcal{I})$, and $u(I_1,\cdots, I_j ) + h(I_k,\cdots, I_m) = h(\mathcal{I})$. Additionally, $\sigma(h(\mathcal{I}))=h(\mathcal{I})$.
}
\end{remark}

\begin{figure}
    \centering
\begin{tikzpicture}
    \draw[color=black]    (0,1) node[anchor=south]  {$h(\mathcal{I})$}  -- (1,1) node[anchor = south] {$I_1$} -- (1,-1) -- (2,-1) -- (2,1) node[anchor = south] {$I_2$} -- (3,1) node[anchor = south] {$I_3$} -- (3,-1) -- (4,-1)  -- (4,1)  node[anchor = south] {$I_4$} -- (5,1) node[anchor = south] {$I_5$}  -- (5,-1) -- (6,-1) -- (6,1) node[anchor = south] {$I_6$} -- (7,1) node[anchor = south] {$I_7$} -- (7,-1) -- (8, -1)  -- (8,1) node[anchor = south] {$I_8$} -- (9,1) node[anchor = south] {$I_9$}  -- (9,-1)  -- (10,-1) -- (10,1) node[anchor = south] {$I_{10}$} -- (11,1)  ;
    \draw (11, 1) node[anchor=south]  {$1$};
    \draw  (11, -1) node[anchor=west] {$-1$};
\end{tikzpicture}
    \caption{An example of $h(\mathcal{I})$ where $\mathcal{I} = \{I_1, \cdots, I_{10}\}$.}
    \label{fig:h_ex}
\end{figure}

\iftoggle{long}{

\begin{lemma}
The vectors $h(i), h(j)$ satisfy the following properties:
\begin{enumerate}
		\item \label{h1} $h(i) = \sigma(h(i)-\mathbf{1})$ 
		\item \label{h2} $h(i,j) = \sigma(h(i)-h(j)), 1 < i < j \leq N$
        \item \label{h3}$\frac{1}{2}(h(i)-\mathbf{1}) = -\mathbf{1}[i,m] \leq 0$ 
        \item \label{h4} $ \frac{1}{2}(h(i)-h(j)) = -\mathbf{1}[i,j-1] \leq \mathbf{0}$. 
        \item \label{h5} $\sigma(a+b)=\sigma(a)+\sigma(b), a,b \in {\mathbb{R}^{N}_-}$
	\end{enumerate}

\end{lemma}

For any index $i \in \{2, \cdots, N\} $, the vector $h(i)-\mathbf{1} $ is $-2$ at indices $\geq i$ and $0$ otherwise. Therefore \iftoggle{long}{we have the "unit cliff vector"}
$h(i)=\sigma(h(i)-\mathbf{1})$.

For any indices $1 < i < j \leq m$, the vector $h(i,j)$ is $-1$ between indices $i$ and $j-1$, and is $1$ elsewhere. The vector $h(i)-h(j)$ is $-2$ between indices $i$ and $j-1$ and is $0$ elsewhere. 
So, we can decompose $h(i, j)  =
		\sigma(h(i)-h(j))$.
\iftoggle{long}{
So, the one-dip vector $h(i, j)$ can be decomposed into single-switch vectors $h(i)$ and $h(j)$ as the unit dip vector

\begin{equation}
	\begin{aligned}
		h(i, j) & =
		\sigma(h(i)-h(j))
	\end{aligned}
\end{equation}

}


We have the unit ``cliff" (a negative step) vector $\frac{1}{2}(h(i)-\mathbf{1}) = -\mathbf{1}[i,m] \leq 0$ and the unit ``dip" (a negative bump) vector $ \frac{1}{2}(h(i)-h(j)) = -\mathbf{1}[i,j-1] \leq \mathbf{0}$.
Note for any nonpositive vectors $a,b \in {\mathbb{R}^{N}}^{\leq0}, \sigma(a+b)=\sigma(a)+\sigma(b)$.
}

\iftoggle{long}{

\begin{lemma}\label{lemma:h+h-}
	Let $m \geq 1$ and suppose $h(\mathcal{I}) \in \{-1, 1\}^N$ is a vector with $2m$ switches at $\mathcal{I} = \{I_1, \cdots, I_{2m} \}$. There exist $h^+(\mathcal{I}), h^-(\mathcal{I}) \in \{-1,1\}^N$ each with $m$ switches such that $ h^+(\mathcal{I}) - h^-(\mathcal{I})=u(I)$ and therefore $h(\mathcal{I})=\sigma(h^+(\mathcal{I}) - h^-(\mathcal{I}))$.
\end{lemma}

\begin{proof}

	Let

	\begin{equation}\label{eq:m+m-}
		\begin{aligned}
			(m^+, m^-) & =
			\begin{cases}
				(2m-1, 2m  ) \text{ if } m \text{ is odd} \\
				(2m, 2m-1) \text{ if } m  \text{ is even} \\
			\end{cases}
		\end{aligned}
	\end{equation}

	\begin{equation}\label{eq:h+h-def}
		\begin{aligned}
			h^+(\mathcal{I}) & = h(I_1, I_4, I_5,I_8,I_9,I_{12},I_{13},\cdots,I_{4i},I_{4i+1},\cdots, I_{m^+}) \\
			h^-(\mathcal{I}) & = h(I_2, I_3,I_6,I_7,\cdots, I_{4i+2},I_{4i+3},\cdots, I_{m^-})                 \\
		\end{aligned}
	\end{equation}

	\Cref{fig:lemmah+h-} illustrates $h^+(\mathcal{I})$ and $ h^-(\mathcal{I})$  and it can be seen that
 \begin{equation*}
 ( h^+(\mathcal{I}) - h^-(\mathcal{I}))_n =
     \begin{cases}
      1-1=0,   & n \in \{2i, 2i+1\}, i \geq 0\\
      -1-1=-2, & n \in \{2i-1, 2i\}, i \geq 1 
     \end{cases} \quad = u(\mathcal{I})_n.
 \end{equation*}

And $h^+(\mathcal{I})$ and $h^-(\mathcal{I})$ each have $m$ switches. 

\begin{figure}
    \centering
\begin{tikzpicture}
    \draw[color=red]    (0,1) node[anchor=south]  {$h_+(\mathcal{I})$}  -- (1,1) node[anchor = south] {$I_1$} -- (1,-1)  -- (4,-1)  -- (4,1)  node[anchor = south] {$I_4$} -- (5,1) node[anchor = south] {$I_5$}  -- (5,-1)  -- (8, -1)  -- (8,1) node[anchor = south] {$I_8$} -- (9,1) node[anchor = south] {$I_9$}  -- (9,-1)  -- (11, -1) ;
     \draw[color=blue] (0,-1) node[anchor=south]  {$h_-(\mathcal{I})$} -- (2,-1) -- (2,1)  node[anchor = south] {$I_2$} -- (3,1) node[anchor = south] {$I_3$} -- (3,-1)  -- (6,-1)  -- (6,1) node[anchor = south] {$I_6$} -- (7,1) node[anchor = south] {$I_7$} -- (7, -1)  -- (10,-1)  -- (10,1) node[anchor = south] {$I_{10}$}  -- (11, 1)  ;
    \draw (11, 1) node[anchor=south]  {$1$};
    \draw  (11, -1) node[anchor=south] {$-1$};
\end{tikzpicture}
    \caption{An example of $h^{+}(\mathcal{I})=h(I_1, I_4, I_5, I_9), h^{-}(\mathcal{I})=h(I_2,I_3,I_6,I_7,I_{10})$ in \Cref{lemma:h+h-} when $h(\mathcal{I}) = h(I_1, \dots, I_{10} )$ has $2m=10$ switches. $h^+(\mathcal{I})$ and $h^-(\mathcal{I})$ interleave to form $h(\mathcal{I})$.}
    \label{fig:lemmah+h-}
\end{figure}

\iftoggle{long}{
	In other words, the ``hills" (the indices where the vectors are $1$) of two $m$-switch vectors $h^+$ and $h^-$ are interleaved to form a vector $ h(I_1,\cdots, I_{2k}) $ with $2m$ switches.
}
\end{proof}

\begin{remark}
    If $\mathcal{I}=\{I_1\}$ then set $h^-(\mathcal{I}) = -\mathbf{1}$.
\end{remark}

\begin{lemma}\label{lemma:hremainder}
	Let $m \geq 1$ and suppose $h(\mathcal{I}) \in \{-1,1\}^N$ is a vector with $r \in \{m+1, \cdots, 2m-1 \}$ switches at $\mathcal{I} = \{I_1, \cdots, I_{r} \}$. There exist $h^+(\mathcal{I}), h^-(\mathcal{I}) \in \{-1,1\}^N$ each with at most $m$ switches such that $h^+(\mathcal{I}) - h^-(\mathcal{I}) =  u(\mathcal{I})$ and therefore $ h(\mathcal{I})=\sigma\left(h^+(\mathcal{I}) - h^-(\mathcal{I})\right)$.
\end{lemma}

\begin{proof}
The proof is similar to that of \Cref{lemma:h+h-}.
If $r$ is even, let $m = r^+ = r^- = \frac{r}{2}$ and $m^+, m^-$ be defined by \eqref{eq:m+m-}. Otherwise if $r$ is odd, let $r^+ = \frac{r+1}{2}, r^- = \frac{r-1}{2}, m^+ = r$. Additionally if $r$ is odd, let $m^-=m^+-2$ if $r^+$ is odd, and otherwise let $m^-=m^+-1$.
Observe $r^+ \geq r^-, r = r^+ + r^-$ and $r^+ - r^- \in \{0,1\}$.
Using $m^-, m^+$ as defined above, let $h^+(\mathcal{I}), h^-(\mathcal{I})$ be defined by \eqref{eq:h+h-def}. A similar argument as the proof of \Cref{lemma:h+h-} holds.

\begin{figure}
    \centering
\begin{tikzpicture}
    \draw[color=red]    (0,1) node[anchor=south]  {$h_+(\mathcal{I})$}  -- (1,1) node[anchor = south] {$I_1$} -- (1,-1)  -- (4,-1)  -- (4,1)  node[anchor = south] {$I_4$} -- (5,1) node[anchor = south] {$I_5$}  -- (5,-1)  -- (8, -1)  -- (8,1) node[anchor = south] {$I_8$} -- (9,1) node[anchor = south] {$I_9$}  -- (9,-1)  -- (11, -1) ;
     \draw[color=blue] (0,-1) node[anchor=south]  {$h_-(\mathcal{I})$} -- (2,-1) -- (2,1)  node[anchor = south] {$I_2$} -- (3,1) node[anchor = south] {$I_3$} -- (3,-1)  -- (6,-1)  -- (6,1) node[anchor = south] {$I_6$} -- (7,1) node[anchor = south] {$I_7$} -- (7, -1)  -- (10,-1)  -- (10,1) node[anchor = south] {$I_{10}$}  -- (11, 1)  ;
    \draw (11, 1) node[anchor=south]  {$1$};
    \draw  (11, -1) node[anchor=south] {$-1$};
\end{tikzpicture}
    \caption{An example when $m=5$, $r=7$ in \Cref{lemma:hremainder}. Since $r$ is odd, $m^+ = r^+ = 4, r^- = 3$. Since $r^+$ is even, $m^- = 3$.}
    \label{fig:lemmarem}
\end{figure}
\end{proof}

}

\begin{lemma}\label{lemma:hdecomp}
	Let $m \geq 1, r \in \{m+1,\cdots,2m\}$. 
 Let $h \in \switchmatrix{r}$.
 Let $u=h-\mathbf{1}$. There exist $h^+, h^- \in \switchmatrix{m}$ such that $u=h^+ - h^-$.
 \iftoggle{long}{
 and suppose $h(\mathcal{I}) \in \{-1,1\}^N$ is a vector with $r \in \{m+1, \cdots, 2m \}$ switches at $\mathcal{I} = \{I_1, \cdots, I_{r} \}$. There exist sets $\mathcal{I}^+, \mathcal{I}^-$ of at most $m$ integers such that  $h(\mathcal{I}^+) - h(\mathcal{I}^-) =  u(\mathcal{I})$ and therefore $ h(\mathcal{I})=\sigma\left(h(\mathcal{I}^+) - h(\mathcal{I}^-)\right)$.
 }
\end{lemma}

\begin{proof}
Let $I_1, \cdots, I_r$ be the locations at which $h$ switches, where $I_1< \cdots< I_r$. Let $h^+, h^- \in \{-1,1\}^N$ starting with $1$, with switches precisely at $I_k$ for $k \in \{0,1\} \mod{4}$ and $k \in \{2,3\} \mod{4}$. Then $u = h^+ - h^-$.
The number of switches of $h^+, h^-$ differ by at most $1$, so they each have at most $\frac{r}{2} \leq m$ switches. So $h^+, h^- \in \switchmatrix{m}$.
\iftoggle{long}{

\iftoggle{long}{}
	Partition $\mathcal{I}$ into $\mathcal{I}^+ := \{i \in \mathcal{I}: i \mod 4 \in \{0,1\} \}=\{I_1, I_4, I_5,I_8,I_9,I_{12},I_{13},\cdots,I_{4i},I_{4i+1},\cdots\}$ and $ \mathcal{I}^- := \{i \in \mathcal{I}: i \mod 4 \in \{2,3\} \}=\{I_2, I_3,I_6,I_7,\cdots, I_{4i+2},I_{4i+3},\cdots\}$. 
 \iftoggle{long}{ At indices $i$ where $h(\mathcal{I}^+)_i=1$,  we have $-h(\mathcal{I}^-)_i=-1$. Similarly at indices $i$ where $-h(\mathcal{I}^-)_i=1$,  we have $h(\mathcal{I}^+)_i=-1$. 
 }
Observe $\Big||\mathcal{I}^+| -|\mathcal{I}^-| \Big| \leq 1$ and so  $|\mathcal{I}^+|, |\mathcal{I}^+| \leq \frac{|\mathcal{I}|}{2} \leq m$. 
For $n \in [N]$, \\

$ \left(h(\mathcal{I}^+)_n,-h(\mathcal{I}^-)_n\right) =
     \begin{cases}
         (1,-1) & \mbox { if } g(n) \mod 4 = 0 \\
         (-1,1) & \mbox { if } g(n) \mod 4 = 2 \\
         (-1,-1) & \mbox { if } g(n) \mod 4 \in \{1, 3\} . \\
     \end{cases}
$

Therefore, 
$\left(h(\mathcal{I}^+)-h(\mathcal{I}^-)\right)_n =
     \begin{cases}
         0 & \mbox { if } g(n) \mod 4 \in \{0,2\} \\
         -2 & \mbox { if } g(n) \mod 4 \in \{1, 3\} 
     \end{cases} 
     \quad
     = u(\mathcal{I})_n
$.
}
\begin{figure}
    \centering
\begin{tikzpicture}
    \draw[color=red]    (0,1) node[anchor=south]  {$h^+(\mathcal{I})$}  -- (1,1) node[anchor = south] {$I_1$} -- (1,-1)  -- (4,-1)  -- (4,1)  node[anchor = south] {$I_4$} -- (5,1) node[anchor = south] {$I_5$}  -- (5,-1)  -- (8, -1)  -- (8,1) node[anchor = south] {$I_8$} -- (9,1) node[anchor = south] {$I_9$}  -- (9,-1)  -- (11, -1) ;
     \draw[color=blue] (0,-1) node[anchor=south]  {$-h^-(\mathcal{I})$} -- (2,-1) -- (2,1)  node[anchor = south] {$I_2$} -- (3,1) node[anchor = south] {$I_3$} -- (3,-1)  -- (6,-1)  -- (6,1) node[anchor = south] {$I_6$} -- (7,1) node[anchor = south] {$I_7$} -- (7, -1)  -- (10,-1)  -- (10,1) node[anchor = south] {$I_{10}$}  -- (11, 1)  ;
    \draw (11, 1) node[anchor=south]  {$1$};
    \draw  (11, -1) node[anchor=south] {$-1$};
\end{tikzpicture}
    \caption{An example of $h(\mathcal{I}^-)=h(I_1, I_4, I_5, I_9), h(\mathcal{I}^+)=h(I_2,I_3,I_6,I_7,I_{10})$ in \Cref{lemma:hdecomp} when $h(\mathcal{I}) = h(I_1, \dots, I_{10} )$ has $2m=10$ switches. The vectors $h(\mathcal{I}^+)$ and $h(\mathcal{I}^-)$ indicated in red and blue interleave to form $h(\mathcal{I})$.}
    \label{fig:lemmahdecomp}
\end{figure}
\iftoggle{long}{
	In other words, the ``hills" (the indices where the vectors are $1$) of two $m$-switch vectors $h^+$ and $h^-$ are interleaved to form a vector $ h(I_1,\cdots, I_{2k}) $ with $2m$ switches.
}
\end{proof}
}



\section*{\secnameSignSol} \label{app_sec:squarewave-2lyr}

In this section 
we assume the neural net uses sign activation, and 
$d=1$. 
Recall $\mathbf{e}^{(n)}$ as the $n^{\text{th}}$ canonical basis vector (\Cref{app:minNorm}).
Note that in Figures \ref{fig:pulse_ex_overlap_2lyr} 
\iftoggle{arxiv}{, \ref{fig:feature_ex}, } 
and \ref{fig:z_bdry_cycle}, $\mathbf{y}=\hkt$, $N=40, T=10$, and vectors 
$\mathbf{v} = (v_1, \cdots, v_N)$ are depicted by plotting $(n,v_n)$ as a dot.

\begin{remark}\label{rmk:criticalbeta}
 A neural net with all weights being $0$ achieves the same objective in the training problem as the optimal \lasso{} value and is therefore optimal.
\end{remark}

In this section, we will find the critical value $\beta_c$ defined in \Cref{sec:squarewaveapp}. Then for $\beta < \beta_c$, we use the subgradient condition from \Cref{rk:strcvx} to solve the \lasso{} problem \eqref{eq:genericlasso}. Note when $L=2$, $\left(\Amat{}_n\right)^T = (
    \mathbf{1}_{1:n} ,
    -\mathbf{1}_{n+1:N} 
)$ switches at $n+1$.

\subsection{$L=2$} Assume the network has $2$ layers.

%


\begin{lemma}\label[lemma]{lemma:A^TA_2layer_sign}
 \iftoggle{long}{ the elements of the gram dictionary matrix  $\Amat{}^T \Amat{}$ are 
} The elements of $\Amat^T \Amat \in \mathbb{R}^{N \times N}$ are $(\Amat{}^T \Amat{})_{i,j}= N - 2|i-j|$.
\end{lemma}

\begin{proof}
 \iftoggle{long}{
 	Since the data is sorted in descending order 
 $x_1 > x_2 > \cdots > x_N$ (\Cref{section:notation}), the dictionary matrix $A \in \{-1,1\}^{N \times N}$ is of the form

	\begin{equation}\label{mat:A_2layer_sign}
		A =
		\begin{pmatrix}
			1      & 1      & 1      & 1      & \cdots & 1      & 1      \\
			-1     & 1      & 1      & 1      & \cdots & 1      & 1      \\
			-1     & -1     & 1      & 1      & \cdots & 1      & 1      \\
			-1     & -1     & -1     & 1      & \cdots & 1      & 1      \\
			-1     & -1     & -1     & -1     & \cdots & 1      & 1      \\
			\vdots & \vdots & \vdots & \vdots & \ddots & \vdots & \vdots \\
			-1     & -1     & -1     & -1     & \cdots & -1     & 1
		\end{pmatrix}.
	\end{equation}
 }
 \iftoggle{long}{
Recall \Cref{remk:Asignform}'s description of $\Amat{}$.
}
If $1 \leq i \leq j \leq N$ then
$
			(\Amat{}^T \Amat{} )_{i,j} 
			                {=} \sum_{k=1}^{i-1} \Amat{}_{i,k} \Amat{}_{j,k} {+} \sum_{k=i}^{j-1} \Amat{}_{i,k} \Amat{}_{j,k} {+} \sum_{k=j}^{N} \Amat{}_{i,k} \Amat{}_{j,k} 
			                \\{=} \sum_{k=1}^{i-1} (1) (1) {+} \sum_{k=i}^{j-1} (-1) (1) {+} \sum_{k=j}^{N} (-1) (-1)        
			                {=} (i-1) {-} (j{-}1{-}i{+}1) {+} (N{-}j{+}1)     {=}N {+}  2(i{-}j)\\                     {=} N{-}2|i{-}j|
$.
\iftoggle{long}{
	For example, if $i=1,j=N$ then $(A^TA)_{i,j}=N+2(1-N) = 2 - N$.
}
Since $\Amat{}^T \Amat{}$ is symmetric, if  $1 \leq j \leq i \leq N$ then 
$(\Amat{}^T \Amat{})_{i,j} = (\Amat{}^T \Amat{} )_{j,i} = N +  2(j-i)=N-2|i-j|$.
So for any $i, j \in [N]$, $(\Amat{}^T \Amat{} )_{i,j} =  N - 2|i-j|$.
\end{proof}

\iftoggle{arxiv}{
\begin{remark}
	By \Cref{lemma:A^TA_2layer_sign}, $\Amat{}^T\Amat{}$ is of the form
 \iftoggle{long}{ the gram matrix
 $A^TA$ for the $L=2$ layer sign activation can be written out as
 }

	\begin{equation}\label{eq:A^TA_writtenout}
		\setcounter{MaxMatrixCols}{20}
		\begin{aligned}
			\Amat{}^T \Amat{} & =  \begin{pmatrix}
				           N      & N-2    & N-4    & \cdots & 2      & 0      & -2     & \cdots & 6-N    & 4-N    & 2-N    \\
				           N-2    & N      & N-2    & \cdots & 4      & 2      & 0      & \cdots & 8-N    & 6-N    & 4-N    \\
				           N-4    & N-2    & N      & \cdots & 6      & 4      & 2      & \cdots & 10-N   & 8-N    & 6-N    \\
				           \vdots & \vdots & \vdots & \ddots & \vdots & \vdots & \vdots & \ddots & \vdots & \vdots & \vdots \\
				           6-N    & 8-N    & 10-N   & \cdots & 2      & 4      & 6      & \cdots & 10-N   & 8-N    & 6-N    \\
				           4-N    & 6-N    & 8-N    & \cdots & 0      & 2      & 4      & \cdots & 8-N    & 6-N    & 4-N    \\
				           2-N    & 4-N    & 6-N    & \cdots & -2     & 0      & 2      & \cdots & N-4    & N-2    & N
			           \end{pmatrix}.
		\end{aligned}
	\end{equation}

 \iftoggle{long}{
 
	\begin{equation}\label{eq:A^TA_writtenout}
		\setcounter{MaxMatrixCols}{20}
		\begin{aligned}
			A^T A & =
			\begin{pmatrix}
				1      & -1     & -1     & -1     & \cdots & -1     & -1     \\
				1      & 1      & -1     & -1     & \cdots & -1     & -1     \\
				1      & 1      & 1      & -1     & \cdots & -1     & -1     \\
				1      & 1      & 1      & 1      & \cdots & -1     & -1     \\
				1      & 1      & 1      & 1      & \cdots & 1      & -1     \\
				\vdots & \vdots & \vdots & \vdots & \ddots & \vdots & \vdots \\
				-1     & -1     & -1     & -1     & \cdots & 1      & 1
			\end{pmatrix}
			\begin{pmatrix}
				1      & 1      & 1      & 1      & \cdots & 1      & 1      \\
				-1     & 1      & 1      & 1      & \cdots & 1      & 1      \\
				-1     & -1     & 1      & 1      & \cdots & 1      & 1      \\
				-1     & -1     & -1     & 1      & \cdots & 1      & 1      \\
				-1     & -1     & -1     & -1     & \cdots & 1      & 1      \\
				\vdots & \vdots & \vdots & \vdots & \ddots & \vdots & \vdots \\
				-1     & -1     & -1     & -1     & \cdots & -1     & 1
			\end{pmatrix}                                                \\
			      & =  \begin{pmatrix}
				           N      & N-2    & N-4    & \cdots & 2      & 0      & -2     & \cdots & 6-N    & 4-N    & 2-N    \\
				           N-2    & N      & N-2    & \cdots & 4      & 2      & 0      & \cdots & 8-N    & 6-N    & 4-N    \\
				           N-4    & N-2    & N      & \cdots & 6      & 4      & 2      & \cdots & 10-N   & 8-N    & 6-N    \\
				           \vdots & \vdots & \vdots & \ddots & \vdots & \vdots & \vdots & \ddots & \vdots & \vdots & \vdots \\
				           6-N    & 8-N    & 10-N   & \cdots & 2      & 4      & 6      & \cdots & 10-N   & 8-N    & 6-N    \\
				           4-N    & 6-N    & 8-N    & \cdots & 0      & 2      & 4      & \cdots & 8-N    & 6-N    & 4-N    \\
				           2-N    & 4-N    & 6-N    & \cdots & -2     & 0      & 2      & \cdots & N-4    & N-2    & N
			           \end{pmatrix}.
		\end{aligned}
	\end{equation}

 }
\end{remark}

 An example of a column of $\Amat^T\Amat$ is plotted in the left plot of 	\Cref{fig:feature_ex}.
\begin{figure}[ht]
	\begin{tikzpicture}[scale=0.8]
		\centering
		\begin{axis}[axis x line=middle, axis y line=middle, xticklabels={,,}, xlabel = {$n$}]
			\addplot[ycomb,color=black,solid,mark=*]
			plot coordinates{
					(1,40-70)
					(2,40-68)
					(3,40-66)
					(4,40-64)
					(5,40-62)
					(6,40-60)
					(7,40-58)
					(8,40-56)
					(9,40-54)
					(10,40-52)

					(11,40-50)
					(12,40-48)
					(13,40-46)
					(14,40-44)
					(15,40-42)
					(16,40-40)
					(17,40-38)
					(18,40-36)
					(19,40-34)
					(20,40-32)

					(21,40-30)
					(22,40-28)
					(23,40-26)
					(24,40-24)
					(25,40-22)
					(26,40-20)
					(27,40-18)
					(28,40-16)
					(29,40-14)
					(30,40-12)

					(31,40-10)
					(32,40-8)
					(33,40-6)
					(34,40-4)
					(35,40-2)
					(36,40)
					(37,40-2)
					(38,40-4)
					(39,40-6)
					(40,40-8)
				};
		\end{axis}
	\end{tikzpicture}
    \begin{tikzpicture}[scale=0.8]
    \centering
    \begin{axis}[axis x line=middle, axis y line=middle, xticklabels={,,}, xlabel = {$n$}]
    \addplot[ycomb,color=black,solid,mark=*] 
    plot coordinates{
    (1,1) 
    (2,1) 
    (3,1) 
    (4,1)
    (5,1)
    (6,-1) 
    (7,-1) 
    (8,-1) 
    (9,-1)
    (10,-1)
    
    (11,1) 
    (12,1) 
    (13,1) 
    (14,1)
    (15,1)
    (16,-1) 
    (17,-1) 
    (18,-1) 
    (19,-1)
    (20,-1)
    
    (21,1) 
    (22,1) 
    (23,1) 
    (24,1)
    (25,1)
    (26,-1) 
    (27,-1) 
    (28,-1) 
    (29,-1)
    (30,-1)
    
    (31,1) 
    (32,1) 
    (33,1) 
    (34,1)
    (35,1)
    (36,-1) 
    (37,-1) 
    (38,-1) 
    (39,-1)
    (40,-1)
    };
    \end{axis}
    \draw [decorate,
    			decoration = {brace, mirror}] (3.4,5.9) --  (1.7,5.9);
    		\node[] at (2.6,6.2) {$T$};
    \end{tikzpicture}
	\caption{Left: Column $10$ of $\Amat{}^T \Amat{}$ for $N=40$. Right: Vector $\hkt$ for $T=10$.}
	\label{fig:feature_ex}
\end{figure}

}

\begin{remark}\label[remark]{rk:A^Ty}
 \iftoggle{long}{
 The dictionary matrix applied to a \labelvec{} is, 
 
Recall \Cref{remk:Asignform}.}
For $n \in [N]$,
$
			(\Amat{}^T \mathbf{y})_n = \sum_{i=1}^n \mathbf{y}_i - \sum_{i=n+1}^N \mathbf{y}_i
$.

\end{remark}

\iftoggle{arxiv}{
\begin{definition}\label[definition]{def:quotrem}
    For $a,b \in \mathbb{Z}$, let $\mathrm{Quot}(a,b) \in \mathbb{Z}$ and $ \mathrm{Rem}(a,b) \in \{0, \cdots, b-1\}$ be the quotient and remainder, respectively, when $a$ is divided by $b$. 
    The \emph{modified remainder} is
    \[
\mathrm{rem}(a,b) = 
\begin{aligned}
    \begin{cases}
   \mathrm{Rem}(a,b)  & \mbox { if }  \mathrm{Rem}(a,b) > 0 \\
   b & \mbox { if } \mathrm{Rem}(a,b) = 0 
\end{cases}
\quad \in [b].
\end{aligned}
\]
The \emph{modified quotient} is
\begin{equation*}
    \begin{aligned}
        \mathrm{quot}(a,b) = \frac{a-\mathrm{rem}(a,b)}{b} = 
        \begin{cases}
            \mathrm{Quot}(a,b)  & \mbox { if }  \mathrm{Rem}(a,b) > 0 \\
             \mathrm{Quot}(a,b)-1 & \mbox { if } \mathrm{Rem}(a,b) = 0 .
        \end{cases}
    \end{aligned}
\end{equation*}
%
\end{definition}
The quotient and remainder are modified to handle vector indices starting at $1$ instead of being zero-indexed. 

\begin{remark}
    The square wave has elements $\hkt_n =
		\begin{cases}
            -1 & \mbox { if } \quad \mathrm{rem}(T,n) \leq T/2 \\
			1  & \mbox { else. }
		\end{cases}$
\end{remark}
}

\iftoggle{long}{
\begin{definition}\label[definition]{def:h_{k,T}}

	Suppose \iftoggle{long}{ there are}
 $N=kT$
 \iftoggle{long}{ training samples}
 for integers $k,T$ where $T$ is even and at least $2$. 
 Define $\hkT \in \{-1,1\}^N$ by its $n^{\textrm{th}}$ element as
 \iftoggle{long}{
 the $i^{th}$ element of $h_{k,T}$ is
 }

	\begin{equation}
		\mathbf{h_{k,T}}(n) =
		\begin{cases}
            -1 & \mbox { if } \quad \textrm{rem}(T,n) \leq T/2 \\
			1  & \mbox { else }
		\end{cases},
  \quad \mbox{ for } n \in [N].
	\end{equation}

	In other words, $\mathbf{h_{k,T}}$ is a discretized square wave with amplitude $1$ and $k$ cycles of period $T$. The right plot of \Cref{fig:feature_ex} illustrates an example of $\mathbf{h_{k,T}}$ for $k=4,T=10$. 

\end{definition}
}

 \begin{remark}\label[remark]{rk:hkT_per_sum_0}
 Since $\hkt$ is periodic and zero mean, for $i,n \geq 0$,  
     $\sum_{j=iT+1}^{nT}  {\hkt}_j = 0.$ 
     \iftoggle{long}{
     $=(k-n) \sum_{j=1}^{T}  \mathbf{h_{k,T}}(j) = (k-n) \left( \sum_{j=1}^{\frac{T}{2}} (-1) + \sum_{j=\frac{T}{2}+1}^{T} (1) \right)=0$
     }
 \end{remark}

 \iftoggle{long}{
\begin{figure}[ht]
\begin{tikzpicture}
\centering
\begin{axis}[axis x line=middle, axis y line=middle, xticklabels={,,}, xlabel = {$n$}]
\addplot[ycomb,color=black,solid,mark=*] 
plot coordinates{
(1,1) 
(2,1) 
(3,1) 
(4,1)
(5,1)
(6,-1) 
(7,-1) 
(8,-1) 
(9,-1)
(10,-1)

(11,1) 
(12,1) 
(13,1) 
(14,1)
(15,1)
(16,-1) 
(17,-1) 
(18,-1) 
(19,-1)
(20,-1)

(21,1) 
(22,1) 
(23,1) 
(24,1)
(25,1)
(26,-1) 
(27,-1) 
(28,-1) 
(29,-1)
(30,-1)

(31,1) 
(32,1) 
(33,1) 
(34,1)
(35,1)
(36,-1) 
(37,-1) 
(38,-1) 
(39,-1)
(40,-1)
};
\end{axis}
\draw [decorate,
			decoration = {brace, mirror}] (3.4,5.9) --  (1.7,5.9);
		\node[] at (2.6,6.2) {$T$};
\end{tikzpicture}
\caption{\labelvec{} $\mathbf{y}=\hkT$ modeling non-linearly separable regions in $\mathbb{R}$ for $k=4,T=10$. 
}
\label{fig:pulse_ex}
\end{figure}
}

\iftoggle{long}{ 

\begin{figure}[ht]
	\begin{tikzpicture}
		\centering
		\begin{axis}[axis x line=middle, axis y line=middle, xticklabels={,,}, xlabel = {$x_1$}]
			\addplot[ycomb,color=black,solid,mark=*]
			plot coordinates{
					(1,1)
					(2,1)
					(3,1)
					(4,1)
					(5,1)
					(6,-1)
					(7,-1)
					(8,-1)
					(9,-1)
					(10,-1)

					(11,1)
					(12,1)
					(13,1)
					(14,1)
					(15,1)
					(16,-1)
					(17,-1)
					(18,-1)
					(19,-1)
					(20,-1)

					(21,1)
					(22,1)
					(23,1)
					(24,1)
					(25,1)
					(26,-1)
					(27,-1)
					(28,-1)
					(29,-1)
					(30,-1)

					(31,1)
					(32,1)
					(33,1)
					(34,1)
					(35,1)
					(36,-1)
					(37,-1)
					(38,-1)
					(39,-1)
					(40,-1)
				};
		\end{axis}
	\end{tikzpicture}
	\caption{Noiseless square wave $y=h_{k,T}$ with $k=4$ cycles and period $T=10$. The number of samples is $N=40$. Note the training samples are ordered in descending order, so to correspond to the \labelvec{} $\mathbf{y}$, the elements of $h_{k,T}$ are plotted from right to left.}
	\label{fig:pulse_ex_appendix}
\end{figure}
}

\begin{lemma}\label[lemma]{lemma:A^Th_kT}

The vector $\Amat^T \hkt$ is periodic with period $T$. For $n \in [T]$, 
\begin{equation*}
    \left(\Amat^T \hkt \right)_n = 2
    \begin{cases}
    n &\mbox { if } n \leq \frac{T}{2} \\
    T-n &\mbox { else}
\end{cases}
\quad \in [0, T].
\end{equation*}

\iftoggle{long}{

	For $n \in [N]$,

	\begin{equation}\label{eq:A^Th_k,Tn_cases}
		(\Amat{}^T \hkt)_n = 2 \cdot
		\begin{cases}
			-\mathrm{Rem}(n,T)   & \mbox{ if }  \mathrm{Rem}(n,T) \leq \frac{T}{2} \\
			\mathrm{Rem}(n,T) - T & \mbox{ else } 
		\end{cases}
  = 2
  	\begin{cases}
			-\mathrm{rem}(n,T)     & \mbox{ if }  \mathrm{rem}(n,T) \leq \frac{T}{2} \\
			 \mathrm{rem}(n,T) - T& \mbox{ else } 
		\end{cases}.
	\end{equation}

 In particular, 

 \begin{equation}\label{eq:A^Th_k,Tn_bdry}
		(\Amat^T \hkt)_n  
  = 2
  	\begin{cases}
			-n     & \mbox{ if }  \mathrm{rem}(n,T) \leq \frac{T}{2} \\
			n-N & \mbox{ if } n \geq N - \frac{T}{2} 
		\end{cases}.
	\end{equation}
}
	
\end{lemma}

\begin{proof}
By \Cref{rk:A^Ty}, \Cref{rk:hkT_per_sum_0}, and periodicity of $\hkt$, for $n \in [T], j \in [2k-1]$, 
\begin{equation*}
\begin{aligned}
    \left(\Amat^T \hkt\right)_{n{+}jT} 
                                        {=}  \sum_{i=jT+1}^{jT+n} {\hkt}_i  {-} \sum_{i{=}n{+}jT{+}1}^{(j{+}1)T} {\hkt}_i 
                                        {=}  \begin{cases}
                                             \displaystyle \sum_{i=1}^{n} 1 {-} \sum_{i=n+1}^{T/2}1 {+} \sum_{i=1+T/2}^{T}1  
                                             &\mbox{ if } n \leq \frac{T}{2} \\ 
                                              \displaystyle \sum_{i=1}^{T/2} 1 -\sum_{i=\frac{T}{2}+1}^{n} 1  + \displaystyle \sum_{i=n+1}^T 1 
                                              &\mbox{ else.} 
                                        \end{cases}   
\end{aligned}
\end{equation*}
Simplifying gives the result.
\iftoggle{long}{
	Denote the remainder when $n$ is divided by $T$  as $ \mathrm{rem}(n,T)$ (the period $T$ is implicit) and let $Q_n = \frac{n-\mathrm{rem}(n,T)}{T}$ be the quotient. So $n$ falls in the $q_n^{th}$ cycle of $h_{n,T}$. 
 }
\iftoggle{long}{
 Let $Q_n = \textrm{Quot}(n,T) \in \{0, \cdots, k\}$. So $n = Q_n T + \mathrm{Rem}(n,T)$. Note $Q_n = k$ only if $n=N$.
	For $n \in [N]$, by \Cref{rk:A^Ty} (with $\mathbf{y} = \hkt$) and \Cref{rk:hkT_per_sum_0}, 
	\begin{equation}\label{eq:A^Thw}
		\begin{aligned}
			\left( \Amat{}^T \hkt \right)_n & =
			\sum_{i=1}^n {\hkt}_i   - \sum_{i=n+1}^N {\hkt}_i                                         \\
			                & =  \left( \sum_{j=1}^{Q_n T}  {\hkt}_i
			+ \sum_{i=Q_nT+1}^{n}  {\hkt}_i \right)
			- \left( \sum_{i=n+1}^{ (Q_n+1) T} {\hkt}_i +   \sum_{i=(Q_n+1) T + 1}^{ k T} {\hkt}_i  \right) \mathbf{1} \{n < N\}\\
			                & = \sum_{i=Q_n T+1}^{n}  {\hkt}_i
			-   \left( \sum_{i=n+1}^{ (Q_n+1) T} {\hkt}_i  \right) \mathbf{1} \{n < N\} .                       %
		\end{aligned}
	\end{equation}
If $\mathrm{Rem}(n,T) >0$, then  $\mathrm{Rem}(n,T) = \mathrm{rem}(n,T) \in [T-1]$, and $n < N$.
Specifically, if $1 \leq \mathrm{Rem}(n,T) \leq T/2$, we have
	\begin{equation}\label{eq:Rem<T/2}
		\begin{aligned}\
			(\Amat{}^T \mathbf{h_{k,T}})_n
			 & = \sum_{j=Q_nT+1}^{n}  h_{k,T}(j) - \left( \sum_{j=n+1}^{ Q_n T+T/2}  h_{k,T}(j)
			+  \sum_{j= Q_n T+T/2 +1}^{ (Q_n + 1) T} h_{k,T} (j) \right)                                   \\
			 & = \sum_{j= Q_n T+1}^{n} (-1) - \left( \sum_{j=n+1}^{ Q_n T+T/2} (-1 )
			+  \sum_{j= Q_n T+T/2 +1}^{ (Q_n+1) T} (1 )\right)                                           \\
			 & = -\mathrm{Rem}(n,T) - \left( -(\frac{T}{2}- \mathrm{Rem}(n,T)) + \frac{T}{2} \right)                  \\
			 & = -2 \mathrm{Rem}(n,T).
		\end{aligned}
	\end{equation}
	Similarly if $T/2 + 1 \leq \mathrm{Rem}(n,T) \leq T-1$, then
	\begin{equation}\label{eq:Rem>T/2}
		\begin{aligned}
			(\Amat{}^T \mathbf{h_{k,T}})_n
			 & = \sum_{j=Q_n T+1}^{Q_n T + T/2}  h_{k,T}(j) + \sum_{j=Q_n T + T/2 +1}^{n}  h_{k,T}(j) - \sum_{j=n+1}^{(Q_n + 1) T} h_{k,T} \\
			 & = \sum_{j=Q_n T+1}^{Q_n T + T/2}  (-1) + \sum_{j=Q_n T + T/2 +1}^{n}  (1) - \sum_{j=n+1}^{(Q_n + 1) T} (1)                  \\
			 & = -\frac{T}{2} + \left(\mathrm{Rem}(n,T) - \frac{T}{2} \right) - (T- \mathrm{Rem}(n,T))                                                                   \\
			 & = -2 T + 2\mathrm{Rem}(n,T).
		\end{aligned}
	\end{equation}
When Rem$(n,T)=0$, we have rem$(n,T)=T$, ie $n = Q_n T$, so \eqref{eq:A^Thw} implies that $\left( \Amat{}^T \mathbf{h_{k,T}} \right)_n = 0 = -2\mathrm{Rem}(n,T) = 2(\mathrm{rem}(n,T)-T)$.
Combining this with \eqref{eq:Rem<T/2} and \eqref{eq:Rem>T/2} proves \eqref{eq:A^Th_k,Tn_cases}. 
Equation \eqref{eq:A^Th_k,Tn_bdry} follows from the fact that if $n \leq T$ then $\textrm{rem}(n,T) = n$. And if $n > N - T$, then $\textrm{rem}(n,T)=n-(N-T)$.
To show \eqref{eq:A^Th_k,Tn_simp}, observe if $ \mathrm{rem}(n,T) \leq \frac{T}{2}$, then $q_n$ is even and rem$(n,T) = \mathrm{rem}\left(n,\frac{T}{2}\right)$.
If $\frac{T}{2} < \mathrm{rem}(n,T) < T$, then $q_n$ is odd and $\mathrm{rem}(n,T)= \frac{T}{2} + \mathrm{rem}(n,\frac{T}{2})$. So if $\mathrm{rem}(n,T) < T$ then \eqref{eq:A^Th_k,Tn_simp} is equivalent to \eqref{eq:A^Th_k,Tn_cases}. If $\mathrm{rem}(n,T)=T$, then $q_n$ is odd and $\mathrm{rem}\left(n,\frac{T}{2}\right) = \frac{T}{2}$ and so the expression in \eqref{eq:A^Th_k,Tn_simp} evaluates to $0$, which is consistent with \eqref{eq:A^Th_k,Tn_cases}.
}
\end{proof}

\iftoggle{arxiv}{
\begin{lemma}\label[lemma]{lemma:A^Th^kTwithrem}

\iftoggle{long}{	Since $(A^T h_{k,T})_n $ is the same whether $\mathrm{rem}(n,T)$  is $0 $ or $T$ in equation \eqref{eq:A^Th_k,Tn_cases}, the remainder can be defined to be in the range from $0$ to the divisor minus $1$, or from $1$ to the divisor.
	Let the quotient $q_n = \frac{n-\textrm{rem}(n,\frac{T}{2})}{\frac{T}{2}}$  be the half-cycle containing index $n$, where we define the remainder to satisfy $\textrm{rem}(n,\frac{T}{2}) \in [\frac{T}{2}]$. \\
 }
Let $q_n = \mathrm{quot}\left(n,\frac{T}{2}\right) \in \{0, \cdots, 2k-1\}$. Then 

	\begin{equation}\label{eq:A^Th_k,Tn_simp}
		\begin{aligned}
			\left( \Amat{}^T \hkt \right)_n= 2  (-1)^{q_n+1} \mathrm{rem}\left(n,\frac{T}{2}\right)
			-\mathbf{1}\{q_n \textrm{ } \mathrm{ odd}\} T.
		\end{aligned}
	\end{equation}
\end{lemma}

\begin{proof}
    Follows from \Cref{lemma:A^Th_kT}.
\end{proof}
} 

\begin{corollary}\label[corollary]{cor:bdry_gram}
  Suppose $\mathbf{z} = \mathbf{e}^{({\frac{T}{2}})} + \mathbf{e}^{\left({N-\frac{T}{2}}\right)}$. Then for $ n \leq \frac{T}{2} $ and $ n \geq N-\frac{T}{2}$, $\frac{1}{2}(\Amat{}^T \Amat{} \mathbf{z})_n = \left( \Amat{}^T \hkt \right)_n $. 
\iftoggle{long}{
  
  \[
  \begin{aligned}
(\Amat{}^T \Amat{} \mathbf{z})_n = 2\left( \Amat{}^T \mathbf{h}_{k,T} \right)_n = 
4 \begin{cases}
				 n     & \mbox{ if } n \leq \frac{T}{2}                      \\
				 N-n & \mbox{ if } n \geq N- \frac{T}{2}
			\end{cases}.   
\end{aligned}
\]
}
And if $\frac{T}{2} \leq n \leq  N - \frac{T}{2} $, then 
$ 
(\Amat{}^T \Amat{} \mathbf{z})_n =  2T
$.    
\end{corollary}

\begin{proof}
    By \Cref{lemma:A^TA_2layer_sign}, for $n \in [N]$,
$(\Amat{}^T \Amat{} \mathbf{z})_n 
			                    =2 \left( N- \left|n-\frac{T}{2}\right| -\left|n-N+\frac{T}{2}\right| \right). \\
$
\iftoggle{long}{
$
            = 2 \cdot
			\begin{cases}
				N + (n-\frac{T}{2}) + \left(n - N + \frac{T}{2}\right) = 2n     & \mbox{ if } n \leq \frac{T}{2}                      \\
				N - \left(n-\frac{T}{2}\right) + (n - N + \frac{T}{2}) =  T     & \mbox{ if } \frac{T}{2} \leq n \leq N - \frac{T}{2} \\
				N - (n-\frac{T}{2}) - (n - N + \frac{T}{2}) = 2(N-n) & \mbox{ if } n \geq N- \frac{T}{2}.
			\end{cases}$
   }
 %
Simplifying and applying \Cref{lemma:A^Th_kT} gives the result. 
\end{proof}
\iftoggle{long}{
\begin{definition}
	Let $i \in [N], s \in \mathbb{R}$. Define the impulse noise vector $\delta_{s,i} \in \mathbb{R}^N$ by its $n^{\textrm{th}}$ element as, 

	\begin{equation}\label{def:delta_si}
		\delta_{s,i}(n) :=
		\begin{cases}
			s & \mbox { if } \quad n = i \\
			0 & \mbox { else }.
		\end{cases},
 \quad n \in [N].
	\end{equation}

\end{definition}
}
%

%
\iftoggle{long}{

\begin{observation}

	If $s=0$ then for any $i \in [N]$, adding the vector $\delta_{s,i}$ to the \labelvec{} $y$ does not change $y$, and $\delta_{s,i}$ represents zero noise. If $y=h_{k,T}$, $i \in [N]$ and $s=-2y_i$ then adding $\delta_{s,i}$ to $y$ flips the sign of $y_i$.

\end{observation}
}
\iftoggle{long}{

\begin{figure}[ht]
	\begin{tikzpicture}
		\centering
		\begin{axis}[axis x line=middle, axis y line=middle, xticklabels={,,}, xlabel = {$x_1$}]
			\addplot[ycomb,color=black,solid,mark=*]
			plot coordinates{
					(1,1)
					(2,1)
					(3,1)
					(4,1)
					(5,1)
					(6,-1)
					(7,-1)
					(8,-1)
					(9,-1)
					(10,-1)

					(11,1)
					(12,1)
					(13,1)
					(14,1)
					(15,1)
					(16,-1)
					(17,-1)
					(18,-1)
					(19,-1)
					(20,-1)

					(21,1)
					(22,1)
					(23,1)
					(24,0.5)
					(25,1)
					(26,-1)
					(27,-1)
					(28,-1)
					(29,-1)
					(30,-1)

					(31,1)
					(32,1)
					(33,1)
					(34,1)
					(35,1)
					(36,-1)
					(37,-1)
					(38,-1)
					(39,-1)
					(40,-1)
				};

			\addplot[ycomb, color=red,solid,mark=*]
			plot coordinates{
					(24,0.5)
				};

		\end{axis}
	\end{tikzpicture}
	\caption{Noisy square wave $y=h_{k,T}$ with $k=4$ cycles and period $T=10$, and additive noise $\delta_{s,i}$ at index $i=17$ and amplitude $s=-0.5$. The number of samples is $N=40$. Note the training samples are ordered in descending order, so to correspond to the \labelvec{} $\mathbf{y}$, the elements of $h_{k,T}$ are plotted from right to left.}
	\label{fig:pulse_ex_noise}
\end{figure}
}
%
\iftoggle{long}{
\begin{lemma}\label{lemma:A^Tdelta_si}
	For $n \in [N], s\in \mathbb{R}$, the $L=2$ layer sign activation dictionary matrix applied to the impulse noise $\delta_{s,i}$ is

	\begin{equation}
		\begin{aligned}
			(A^T\delta_{s,i})_n & =
			\begin{cases}
				-s & \textrm{ if } n < i \\
				s  & \textrm{ else }     \\
			\end{cases}
		\end{aligned}
	\end{equation}
\end{lemma}
\begin{proof}
	Apply \Cref{rk:A^Ty} to $\delta_{s,i}$ as defined in equation \eqref{def:delta_si}.
\end{proof}
}
\iftoggle{long}{

\begin{lemma}\label{}
	Suppose $y=h_{k,T}+\delta_{s,i}$. In other words, the \labelvec{} $y$ consists of a square wave $h_{k,T}$ possibly corrupted by noise at a single index $i \in [N]$. Then the $L=2$ layer sign activation dictionary matrix applied to the \labelvec{} is given by, for $n \in [N]$,

	\begin{equation}
		\begin{aligned}
			(A^T y)_n & =
			\begin{cases}
				(A^T h_{k,T})_n -s \textrm{ if } n < i \\
				(A^T h_{k,T})_n +s \textrm{ if } n \geq i
			\end{cases} \\
			          & = s \sigma(n-i) + 2 \cdot
			\begin{cases}
				-\mathrm{rem}(n,T)    & \mathrm{ if }  \quad 0 \leq \mathrm{rem}(n,T) \leq T/2 \\
				\mathrm{rem}(n,T) - T & \mathrm{ if } \quad  T/2 \leq \mathrm{rem}(n,T) \leq T \\
			\end{cases}
		\end{aligned}
	\end{equation}

	where $\sigma(x)$ is $1$ if $x\geq 0$ and $-1$ otherwise.
\end{lemma}

\begin{proof}
	Observe
	\begin{equation}
		A^T y
		= A^T h_{k,T} + A^T\delta_{s,i}
	\end{equation}
	Apply lemma \ref{lemma:A^Tdelta_si} and lemma \ref{lemma:A^Th_kT}.
\end{proof}
}

\iftoggle{long}{

\begin{figure}[ht]
	\begin{tikzpicture}
		\centering
		\begin{axis}[axis x line=middle, axis y line=middle, xticklabels={,,}, xlabel = {$x_1$}]
			\addplot[ycomb,color=black,solid,mark=*]
			plot coordinates{
					(1,+0.5)
					(2,-2+0.5)
					(3,-4+0.5)
					(4,-6+0.5)
					(5,-8+0.5)
					(6,-10+0.5)
					(7,-8+0.5)
					(8,-6+0.5)
					(9,-4+0.5)
					(10,-2+0.5)
					(11,+0.5)

					(12,-2+0.5)
					(13,-4+0.5)
					(14,-6+0.5)
					(15,-8+0.5)
					(16,-10+0.5)
					(17,-8+0.5)
					(18,-6+0.5)
					(19,-4+0.5)
					(20,-2+0.5)

					(21,+0.5)
					(22,-2+0.5)
					(23,-4+0.5)
					(24,-6+0.5)
					(25,-8-0.5)
					(26,-10-0.5)
					(27,-8-0.5)
					(28,-6-0.5)
					(29,-4-0.5)
					(30,-2-0.5)

					(31,-0.5)
					(32,-2.5)
					(33,-4.5)
					(34,-6.5)
					(35,-8.5)
					(36,-10-0.5)
					(37,-8-0.5)
					(38,-6-0.5)
					(39,-4-0.5)
					(40,-2-0.5)
				};

			\addplot[ycomb, color=red,solid,mark=*]
			plot coordinates{
					(24,-6+0.5)
				};

		\end{axis}
	\end{tikzpicture}
	\caption{$A^Ty$ when $y=h_{k,T}+\delta_{s,i}$ with $k=4$ cycles and period $T=10$, and additive noise $\delta_{s,i}$ at index $i=17$ and amplitude $s=-0.5$. The number of samples is $N=40$. Note the training samples are ordered in descending order, so to correspond to the \labelvec{} $\mathbf{y}$, the elements of $h_{k,T}$ are plotted from right to left.}
	\label{fig:A^T(h+delta)}
\end{figure}
}

\iftoggle{long}{
\begin{definition}
	For $i \in [N]$, let $e_i \in \mathbb{R}^N$ denote the $i^{th}$ canonical basis vector. For $j \in [N]$,

	\begin{equation}
		\begin{aligned}
			e_i(j) & = \begin{cases}
				           1 & \textrm{ if } i=j \\
				           0 & \textrm{ else }
			           \end{cases}
		\end{aligned}
	\end{equation}
\end{definition}
}

\iftoggle{long}{
\begin{figure}[ht]
	\begin{tikzpicture}
		\centering
		\begin{axis}[axis x line=middle, axis y line=middle, xticklabels={,,}, xlabel = {$x_1$}]
			\addplot[ycomb,color=black,solid,mark=*]
			plot coordinates{
					(1,0.8)
					(2,0.8)
					(3,0.8)
					(4,0.8)
					(5,0.8)
					(6,-0.6)
					(7,-0.6)
					(8,-0.6)
					(9,-0.6)
					(10,-0.6)

					(11,0.6)
					(12,0.6)
					(13,0.6)
					(14,0.6)
					(15,0.6)
					(16,-0.6)
					(17,-0.6)
					(18,-0.6)
					(19,-0.6)
					(20,-0.6)

					(21,0.6-0.1)
					(22,0.6-0.1)
					(23,0.6-0.1)
					(24,0.6-0.1)
					(25,0.6-0.1)

					(26,-0.6)
					(27,-0.6)
					(28,-0.6)
					(29,-0.6)
					(30,-0.6)

					(31,0.6)
					(32,0.6)
					(33,0.6)
					(34,0.6)
					(35,0.6)
					(36,-0.8)
					(37,-0.8)
					(38,-0.8)
					(39,-0.8)
					(40,-0.8)
				};

			\addplot[ycomb, color=red,solid,mark=*]
			plot coordinates{
					(21,0.6-0.1)
					(22,0.6-0.1)
					(23,0.6-0.1)
					(24,0.6-0.1)
					(25,0.6-0.1)
				};

		\end{axis}
	\end{tikzpicture}
	\caption{LASSO solution with $\beta=2$, and the \labelvec{} being the noisy square wave $y=h_{k,T}$ with $k=4$ cycles and period $T=10$, and additive noise $\delta_{s,i}$ at index $i=17$ and amplitude $s=-0.5$. The number of samples is $N=40$. Note the training samples are ordered in descending order, so to correspond to the \labelvec{} $y$, the elements of $h_{k,T}$ are plotted from right to left. The magnitude of the first and last half-cycles are $-2w_{bdry}-w_{cycle} = 1-\frac{\beta}{T}=1-\frac{2}{10}=0.8$. The magnitude of the middle cycles (excluding the first, last, and noise-contaminated cycle) are $w_{cycle}=1-2\frac{\beta}{T}=1-2\frac{2}{T}=0.6$.} Note the half-cycle containing noise is $q_i=3$. The magnitude of the noise cycle is $w_{cycle} + 2 w_{noise}= 1-2\frac{\beta}{T}+(-1)^{q_i+1}2\frac{s}{T}=1-2\frac{\beta}{T}+(-1)^{3+1}2\frac{(-0.5)}{10}=1-\frac{4}{10}-\frac{1}{10}=0.5$.
	\label{fig:pulse_ex_noise_sol}
\end{figure}
}
\iftoggle{long}{

\begin{definition}\label{def:q_i}
	Given the \labelvec{} $y = h_{k,T}+\delta{s,i}$, let $q_i = \frac{i -\textrm{rem}(\frac{T}{2}, i) }{T/2}$ be the half-cycle of $y$ that contains the noise at index $i$.
\end{definition}
}

\iftoggle{long}{
\begin{lemma}\label{lemma:switch_ay}
	Let $m \in \mathbb{N}$.
 \iftoggle{long}{Let $\mathbf{y} \in \mathbb{R}^N$.}
  For $\mathbf{v} \in \mathbb{R}^N$, let $\mathcal{I}(\mathbf{v}) = \{i \in \{2,\cdots,N\}: 0 \neq \mbox{sign}(\mathbf{v}_{i-1}) \neq \mbox{sign}(\mathbf{v}_i) \neq 0\}$ be the set of indices at which $\mathbf{v}$ switches sign. Let $\mathcal{A} = \{\mathbf{a} \in \{-1,1\}^N: \mbox{sign}(\mathbf{a}_1)= \mbox{sign}(\mathbf{y}_1), |\mathcal{I}(\mathbf{a})| \leq m \}$ be a set of vectors with at most $m$ switches and having the same sign as $\mathbf{y}$ in the first element. Suppose $\mathbf{a} \in \mbox{argmax}_{\mathbf{a} \in \mathcal{A}}\mathbf{y}^T \mathbf{a}$. Then for every index $i \in \mathcal{I}(\mathbf{a})$, if $\mathbf{y}_i \neq 0 \neq \mathbf{y}_{i-1}$ then $i \in \mathcal{I}(\mathbf{y})$.
\end{lemma}

 \Cref{lemma:switch_ay} states that a vector $\mathbf{a}$ that maximizes $\mathbf{y}^T\mathbf{a}$ must switch sign only at indices that $\mathbf{y}$ also switches sign.

\begin{proof}
	Suppose to the contrary there was an index $i \in \mathcal{I}(\mathbf{a}) \cap \mathcal{I}(\mathbf{y})^c$ such that $\mathbf{y}_i \neq 0 \neq \mathbf{y}_{i-1}$. In other words, $\mathbf{a}$ switches sign at index $i$ but $\mathbf{y}$ does not. First, observe that for $j \in \{2, \cdots, N-1\}$, if $\mbox{sign}(\mathbf{a}_j)\neq\mbox{sign}(\mathbf{a}_{j-1})$ or $\mbox{sign}(\mathbf{a}_j)\neq\mbox{sign}(\mathbf{a}_{j+1})$ then flipping the sign of $\mathbf{a}_j$ does not increase the number of switches of $\mathbf{a}$; it merely shifts an index at which $\mathbf{a}$ switches sign. Next, for $k \in [N]$, let $\mathbf{a^{(k)}}$ be the same vector as $\mathbf{a}$ but with its $k^{\textrm{th}}$ element negated: $\mathbf{a^{(k)}}_i=\mathbf{a}_i$ for $i \neq k$ and  $\mathbf{a^{(k)}}_k=-\mathbf{a}_k$. Consider two cases: 

	\textbf{Case 1: $\mbox{sign}(\mathbf{a}_i) \neq \mbox{sign}(\mathbf{y}_i)$}

 Since $\mbox{sign}(\mathbf{a}_i) \neq \mbox{sign}(\mathbf{a}_{i-1})$, $|\mathcal{I}(\mathbf{a^{(i)}})| \leq |\mathcal{I}(\mathbf{a})| \leq m $. Observe $\mathbf{y}_i \mathbf{a}_i < 0$. Therefore $\mathbf{y}^T \mathbf{a^{(i)}} =  \left(\sum_{j \neq i} \mathbf{y}_j \mathbf{a}_j \right) - \mathbf{y}_i \mathbf{a}_i = \mathbf{y}^T \mathbf{a} - 2\mathbf{y}_i \mathbf{a}_i > \mathbf{y}^T \mathbf{a}$, contradicting $\mathbf{a} \in \mbox{argmax}_{\mathbf{a} \in \mathcal{A}}\mathbf{y}^T \mathbf{a}$.

	\textbf{Case 2: $\mbox{sign}(\mathbf{a}_i) = \mbox{sign}(\mathbf{y}_i)$}

	Since $i \notin \mathcal{I}(\mathbf{y})$ but $i \in \mathcal{I}(\mathbf{a})$ and $\mathbf{y}_i \neq 0 \neq \mathbf{y}_{i-1}$, we have $\mbox{sign}(\mathbf{y}_{i-1})=\mbox{sign}(\mathbf{y}_{i})=\mbox{sign}(\mathbf{a}_i)=-\mbox{sign}(\mathbf{a}_{i-1})$.  
 Since 
 \iftoggle{long}{
 $\mbox{sign}(\mathbf{a}_{1+(i-1)}) = $ 
 }
 $\mbox{sign} \left(\mathbf{a}_{i} \right)  \neq \mbox{sign} \left(\mathbf{a}_{(i-1)} \right)$, $|\mathcal{I}(\mathbf{a^{(i-1)}})| \leq \mathcal{I}(\mathbf{a})\leq m $. And $\mathbf{y}_{i-1} \mathbf{a}_{i-1} < 0$. Therefore  $y^T \mathbf{a^{(i-1)}} = \sum_{j \neq i-1} \mathbf{y}_j \mathbf{a}_j - \mathbf{y}_{i-1} \mathbf{a}_{i-1} = \mathbf{y}^T \mathbf{a} - 2 \mathbf{y}_{i-1}\mathbf{a}_{i-1} > \mathbf{y}^T \mathbf{a}$, a contradiction.
 \iftoggle{long}{
 , contradicting $a \in \mbox{argmax}_{a \in \mathcal{A}}y^T a$.}

\end{proof}

\begin{remark}
	By a similar argument, \Cref{lemma:switch_ay} also holds for minimizing $\mathbf{a}^T \mathbf{y}$.
\end{remark}
}

\begin{lemma}\label[lemma]{lemma:2lyr_critbeta>=}
	If $\mathbf{y}=\hkt$, then the critical $\beta$ (defined in \Cref{section_main:square_wave}) is $\beta_c = T$ .
 \iftoggle{long}{
 If $\beta < T$ then the optimal solution to the \lasso{} problem \eqref{eq:genericlasso} is $\mathbf{z}^* =\mathbf{0}$.
 }
\end{lemma}

\begin{proof}
\iftoggle{long}{
	By \Cref{rk:A^Ty} and \Cref{rk:hkT_per_sum_0}, $|\Amat{}_N^T \mathbf{y}| = |\mathbf{1}^T \mathbf{y}| = |\mathbf{1}^T \mathbf{h_{k,T}}|=0$. By \Cref{lemma:A^Th_kT}, $\mathrm{argmax}_{n \in [N]} |\Amat{}_n^T \mathbf{h_{k,T}}|$ is an integer multiple of $\frac{T}{2}$. Let $n \in [N-1]$. 
  Let $q_n = \mathrm{quot}\left(n,\frac{T}{2}\right) \in \{0, \cdots, 2k-1\}$.
  \iftoggle{long}{
$q_n $ is the quotient when $j$ is divided by $\frac{T}{2}$
 } 
When $n$ is a multiple of $\frac{T}{2}$, $\mathrm{rem}\left(n,\frac{T}{2}\right)=\frac{T}{2}$, so by 
\Cref{lemma:A^Th_kT},

$( \Amat{}^T \mathbf{y})_n= 
\begin{aligned}
    \begin{cases}
0  & \mbox{ if } q_n \mbox { odd }\\
-T
			 & \mbox{ if } q_n \mbox { even }\\
    \end{cases}. \quad
\end{aligned}
$
}
	By \Cref{rmk:criticalbeta}, $\displaystyle \beta_c = \max_{n \in [N]} |\Amat{}_n^T \mathbf{y}| = \max_{n \in [N]} \left|\left(\Amat{}^T \hkt\right)_n\right| = T$. 
\end{proof}

\iftoggle{long}{

\begin{lemma}\label{lemma:2lyr_critbeta>=noisy}
	Let $T>2$. Suppose $s > -1$ if $q_i$ is odd and $s<1$ if $q_i$ is even. Suppose the label $y=h_{k,T}+\delta_{s,i}$ is a noisy square wave with $k \geq 3 $ cycles. Let $q_i \in \{0, \cdots, 2k-1\}$ be the quotient when noise index $i$ is divided by $\frac{T}{2}$. Suppose $3 \geq q_i \leq 2k-2$. In other words, the noise is not in the first three half cycles or the last half-cycle.  Let $\beta_c = T - |s|$. If $\beta < \beta_c$ then the optimal bidual variable is $z=\mathbf{0}$, and the neural net is the constant zero function.
\end{lemma}

\begin{proof}
	Let $\Amat{}_n$ be the $n^{\textrm{th}}$ column of the dictionary matrix $\Amat{}$. By \Cref{rk:hkT_per_sum_0}, $\mathbf{1}^T \mathbf{h_{k,T}}=0$. Therefore by \Cref{rk:A^Ty}, $|\Amat{}_N^T \mathbf{y}| = |\mathbf{1}^T \mathbf{y}| = |\mathbf{1}^T(h_{k,T} + \delta_{s,i})|= |s|$. Consider columns $\Amat{}_j$ of $\Amat{}$ for $j \in [N-1]$. By  \Cref{lemma:switch_ay},  $|\Amat{}_j^T \mathbf{y}|$ is maximized if $\Amat{}_j$ switches at a switching index of $\mathbf{y}$. Since  $s > -1$ if $q_i$ is odd and $s<1$ if $q_i$ is even, $\mbox{sign}(\mathbf{y})=\mbox{sign}(\mathbf{h_{k,T}})$. So the switchpoints of $\mathbf{y}$ are the same as the switchpoints of $\mathbf{ h_{k,T}}$, which are at multiples of $\frac{T}{2}$. Let $q_j  \in \{0, \cdots, 2k-1\}$ be the quotient when $j$ is divided by $\frac{T}{2}$. For $j$ being a multiple of $\frac{T}{2}$,

	\begin{align*}
		y^T a_j & = (h_{k,T} + \delta_{s,i})^T a_j                                                                                                                                            \\
		        & = h_{k,T}^T a_j + \delta_{s,i}^T a_j                                                                                                                                        \\
		        & = \begin{cases}
			            \frac{T}{2}(1) + \frac{T}{2}(1) + \frac{N-T}{2}(\frac{T}{2}(1) +  \frac{T}{2}(-1) ) + s(-1)^{\mathbf{1}_{q_j\geq q_i}} & \mbox{ if } q_j \mbox { odd } \\
			            \frac{N}{2}(\frac{T}{2}(1) +  \frac{T}{2}(-1) )  + s(-1)^{\mathbf{1}_{q_j\geq q_i}}                                    & \mbox{ if } q_j \mbox {even } \\
		            \end{cases} \\
		        & = \begin{cases}
			            T + s(-1)^{\mathbf{1}_{q_j\geq q_i}} & \mbox{ if } q_j \mbox { odd }   \\
			            s(-1)^{\mathbf{1}_{q_j\geq q_i}}     & \mbox{ if } q_j \mbox { even }.
		            \end{cases}
	\end{align*}

	If $s>0$, since the half cycle $q_i\geq 3$, there exists an odd half cycle $q_j<q_i$, so that $y^Ta_j = T+s=T+|s|$, the largest it can be.

	If $s<0$, since the half cycle $q_i \leq 2k-2$, there exists an odd half cycle $q_j \geq q_i$, so that $y^Ta_j = T-s=T+|s|$, the largest it can be.

	The smallest $y^Ta_j$ can be is $-s$, and $|-s|=|s|<T+|s|$. So the largest $|y^Ta_j|$ can be is $T+|s|$.



	So the critical $\beta$ is $\max_{j \in [N]} |a_j^T y| = T+|s|$.  \cite{efron2004least}
\end{proof}

\begin{remark}
	If there is no noise, i.e. $s=0$, then lemma \ref{lemma:2lyr_critbeta>=} holds for $k=1$ cycle as well, as the proof is the same.
\end{remark}

}

\begin{lemma}\label[lemma]{lemma:2lyr_sign_square_wave_big_beta}
	Let $\mathbf{y}=\hkt$. 
 \iftoggle{long}{
 is a noiseless square wave of $k$ cycles of length $T$. Let $\beta_c = T$. }
 If $\betafrac \geq \frac{1}{2}$ then the solution to the \lasso{} problem \eqref{eq:genericlasso} is
$
		\mathbf{z}^* = \frac{1}{2}\left(1-\betafrac\right)_+\left(\mathbf{e}^{\left({\frac{T}{2}}\right)} + \mathbf{e}^{\left({N-\frac{T}{2}}\right)}\right).
$


\end{lemma}

\begin{proof}
	By \Cref{lemma:2lyr_critbeta>=}, $\beta_c = T$. By \Cref{lemma:2lyr_critbeta>=}, if $\betafrac \geq 1$ then $\mathbf{z}^* = \mathbf{0}$ as desired.
    Now suppose $\frac{1}{2} \leq \betafrac \leq 1$. 
   Let $\delta = 1-\betafrac, \mathbf{g} = \Amat{}_n^T \left(\Amat{}\mathbf{z}^*-\mathbf{y}\right)$. By \Cref{cor:bdry_gram} and \Cref{lemma:A^Th_kT},
\begin{equation*}
\mathbf{g}=
\begin{cases}
  \begin{aligned}
  (\delta-1) \left(\Amat^T \hkt \right)_n &= -2 \betafrac n                  &\in [-\beta,0]    && \mbox { if }  n \leq \frac{T}{2}\\
  \left( \delta T - \left(\Amat{}^T \mathbf{h_{k,T}} \right)_n \right) &= \left( \beta_c - \beta - (\Amat{}^T \hkt)_n \right) 
  & \in [-\beta, \beta] && \mbox { if } \frac{T}{2} \leq n \leq N - \frac{T}{2} \\
    (\delta-1)\left(\Amat^T \hkt \right)_n &= -2\betafrac(N-n)                &\in  [- \beta, 0]     && \mbox { if } N -  \frac{T}{2}  \leq n 
  \end{aligned}
\end{cases},
\end{equation*}
where the second set inclusion follows from $\left( \Amat^T \hkt \right)_n \in \left[0, \beta_c \right]$ by \Cref{lemma:A^Th_kT} so that $-\beta \leq \mathbf{g} \leq \beta_c - \beta \leq \beta $.
%
Therefore, $|\Amat{}_n^T \left(\Amat{}\mathbf{z}^*-\mathbf{y}\right)| \leq \beta$, and at $n \in \{n: \mathbf{z}^*_n \neq 0\} = \left\{\frac{T}{2},N - \frac{T}{2}\right\}$, we have $\Amat{}_n^T \left(\Amat{}\mathbf{z}^*-\mathbf{y}\right)  = -\betafrac\frac{T}{2} = -\beta = -\beta \mbox{sign}(\mathbf{z}^*_n)$. By \Cref{rk:strcvx}, $\mathbf{z}^*$ is optimal.
\iftoggle{long}{
    
	First suppose $\mathbf{z}^*_n \neq 0$. So $n \in \left\{\frac{T}{2}, N-\frac{T}{2} \right\}$. This makes $q_n$ even in the proof of \Cref{lemma:2lyr_critbeta>=}, so $|\Amat{}_j^T \mathbf{y}| =T= \beta_c$.  Also, $\mathbf{z}^*_n>0$. Since  $-\Amat{}_{\frac{T}{2}}, \Amat{}_{N-\frac{T}{2}} \in \{-1,1\}^N$ differ only at the first and last $\frac{T}{2}$ indices, $\Amat{}_{\frac{T}{2}}^T \Amat{}_{N-\frac{T}{2}}=2T-N$.
     Therefore, $\Amat{}_n^T (\Amat{}\mathbf{z}^*-\mathbf{y}) = \frac{(\beta_c - \beta)_+}{2T} \Amat{}_n^T (\Amat{}_{\frac{T}{2}} + \Amat{}_{N-\frac{T}{2}}) - \Amat{}_n^T \mathbf{y} = \frac{\beta_c - \beta}{2T}\left(N+  \Amat{}_{\frac{T}{2}}^T \Amat{}_{N-\frac{T}{2}}\right)   -\beta_c = -\beta = -\beta \mbox{sign}(\mathbf{z}^*_n)$, as desired in \Cref{rk:strcvx}.

     Nex, suppose $\mathbf{z}^*_n = 0$, so $n \in [N]- \left\{\frac{T}{2}, N-\frac{T}{2}\right \}$. Note $\Amat{}_{\frac{T}{2}} + \Amat{}_{N-\frac{T}{2}} = 2 
     \begin{pmatrix}
         \mathbf{1}_{\frac{T}{2}}\\
         \mathbf{0}_{N-T}\\
         -\mathbf{1}_{\frac{T}{2}}\\
     \end{pmatrix}
$ 
\iftoggle{long}{
     
 Let $f(\mathbf{z}) = \frac{1}{2}||\mathbf{Az}-\mathbf{y}||^2_2 + ||\mathbf{z}||_1$. By \Cref{rk:strcvx}, it suffices to show that $\mathbf{0} \in \partial f(\mathbf{z}^*)$ .
 Therefore, 
 $\partial f(\mathbf{z}) = \Amat{}^T (\mathbf{Az}-\mathbf{y})+\beta \partial \| \mathbf{z} \|_1$.

	\begin{align*}
		(A^T(Az-y) + \beta \partial g(z))_j & = (A^T(Az-y))_j +\beta \partial g(z)_j                                                                           \\
		                                    & = a_j^T (Az-y) + \beta\partial g(z)_j                                                                            \\
		                                    & = a_j^T (\frac{1}{2T}(\beta_c - \beta)_+(a_{\frac{T}{2}} + a_{N-\frac{T}{2}})-y) + \beta \partial g(z)_j         \\
		                                    & = a_j^T (\frac{1}{2T}(\beta_c - \beta)_+(a_{\frac{T}{2}} + a_{N-\frac{T}{2}})) - a_j^T y + \beta \partial g(z)_j \\
		                                    & = (\frac{1}{2T}(\beta_c - \beta)_+(N+  a_{\frac{T}{2}}^T a_{N-\frac{T}{2}}))   -\beta_c + \beta                  \\
		                                    & = (\frac{1}{2T}(\beta_c - \beta)_+(N+ 2T-N) - \beta_c  + \beta                                                   \\
		                                    & = \frac{1}{2T}(\beta_c - \beta)_+(2T) - \beta_c  + \beta                                                         \\
		                                    & = 0 .
	\end{align*}

	}
\iftoggle{long}{
	By equation \eqref{2lyr_z_betac}, $Az = \frac{(\beta_c - \beta)_+}{2T}(a_{\frac{T}{2}} + a_{N-\frac{T}{2}})$. }
	The vector $\Amat{}_{\frac{T}{2}} + \Amat{}_{N-\frac{T}{2}}$ is $2$ for the first $\frac{T}{2}$ indices, $0$ for the next $N-T$ indices, and then $-2$ for the last $\frac{T}{2}$ indices. So

	\begin{equation}\label{remark:2lyr_pred_betac}
		\begin{aligned}
			(\Amat{}\mathbf{z}^*)_n = \frac{\beta_c - \beta}{2T}  (\Amat{}_{\frac{T}{2}} + \Amat{}_{N-\frac{T}{2}})_n =\begin{cases}
				        \frac{\beta_c - \beta}{T}  & \mbox{if } n \leq \frac{T}{2}                  \\
				        0                              & \mbox{if } n \in \left[\frac{T}{2}+1,N-\frac{T}{2}\right] \\
				        -\frac{\beta_c - \beta}{T} & \mbox{if } n > N-\frac{T}{2}                   \\
			        \end{cases},
           \quad n \in [N]
		\end{aligned}
	\end{equation}

	Let $q_n = \mathrm{quot}(n,\frac{T}{2}) \in \{0, \cdots, 2k-1\}$. 
 \iftoggle{long}{
 be the quotient when $j$ is divided by $\frac{T}{2}$. So index $j$ is in the $q_j^{th}$ half-cycle (where $h_{k,T}$ is negative), indexing from $0$. }
 Note $a_1$, the first column of (negated) $A$ is $a_1 = (-1, 1, 1, \cdots, 1)^T$, which switches at index $2$. So $a_{\frac{T}{2}}$ switches at index $\frac{T}{2}+1$ and its first $\frac{T}{2}$ elements are $-1$, and the rest are $1$. Note $a_N = -\mathbf{1}$.

	\begin{equation}\label{eq:(Az-y)_j_critzone_beta_2lyr}
		\begin{aligned}
			(Az-y)_j =\begin{cases}
				          \frac{(\beta_c - \beta)_+}{T} -1\leq 0 & \mbox{if} j \leq \frac{T}{2}                  \\
				          (-1)^{q_j+1}                           & \mbox{if} j \in [\frac{T}{2}+1,N-\frac{T}{2}] \\
				          1-\frac{(\beta_c - \beta)_+}{T} \geq 0 & \mbox{if} j \geq N-\frac{T}{2} + 1            \\
			          \end{cases}
		\end{aligned}
	\end{equation}

	Note that  $(Az-y)$ contains zeros if and only if $\beta_c = \beta$, if and only if the set of its zero indices is $\mathcal{Z} = \{j:(Az-y)_j=0\} = \{1, \cdots, \frac{T}{2}\}\cup\{1+N-\frac{T}{2}, \cdots, N\}$. If $\beta_c = \beta$, for any $j \in \{1, \cdots, \frac{T}{2}\}$, $a_j^T y = a_{\frac{T}{2}}^T y$ and for any $j \in \{1+N-\frac{T}{2}, \cdots, N\}$, $a_j^T y = a_{1+N-\frac{T}{2}}^T y$, so for the purposes of maximizing $a_j^T (Az-y)$, if $j \in \mathcal{Z}$ it suffices to consider just $j \in \{ a_{\frac{T}{2}}, a_{1+N-\frac{T}{2}} \} $, which are still switch points of $Az-y$ in \eqref{eq:(Az-y)_j_critzone_beta_2lyr}. Note by symmetry $a_N^T (Az-y)=0$. Every other $a_j$ for $j \in [N-1]$ contains a single switch. Combining the $\beta_c = \beta$ case with lemma \ref{lemma:switch_ay}, $a_k^T (Az-y)$ is maximized if $a_j$ switches at a switch point of $Az-y$. By equation \eqref{eq:(Az-y)_j_critzone_beta_2lyr}, the switch points of $Az-y$ are at multiples of $\frac{T}{2}$. There are two cases, depending on if the switch point occurs at the beginning of an odd or even half cycle. For column $a_j$ of $A$, where $j$ is a multiple of $ \frac{T}{2}$,

	\begin{equation}
		\begin{aligned}
			|a_j^T(Az-y)| & =
			\begin{cases}
				 & \frac{T}{2}(\frac{(\beta_c - \beta)_+}{T}-1)(-1) + \frac{N-T}{2}(\frac{T}{2}(1)+\frac{T}{2}(-1)) + \frac{T}{2}(1-\frac{(\beta_c - \beta)_+}{T})(1)                                           \\ &\mbox{ if } q_j \mbox{ is odd} \\
				 & \frac{T}{2}(\frac{(\beta_c - \beta)_+}{T}-1)(1) + \frac{T}{2}(-1)(-1)+ \frac{T}{2}(1)(1) + \frac{N-2T}{2}(\frac{T}{2}(1)+\frac{T}{2}(-1)) + \frac{T}{2}(1-\frac{(\beta_c - \beta)_+}{T})(-1) \\ &\mbox{ if } q_j \mbox{ is even} \\
			\end{cases} \\
			              & =
			\begin{cases}
				T(1-\frac{(\beta_c - \beta)_+}{T})     & \mbox{ if } q_j \mbox{ is odd}  \\
				T - T(1-\frac{(\beta_c - \beta)_+}{T}) & \mbox{ if } q_j \mbox{ is even} \\
			\end{cases}                                                                                                                                                                     \\
			              & =
			\begin{cases}
				T-(\beta_c - \beta)_+ & \mbox{ if } q_j \mbox{ is odd}  \\
				(\beta_c - \beta)_+   & \mbox{ if } q_j \mbox{ is even} \\
			\end{cases}                                                                                                                                                                                      \\
			              & =
			\begin{cases}
				\beta    & \mbox{ if } q_j \mbox{ is odd}  \\
				T- \beta & \mbox{ if } q_j \mbox{ is even} \\
			\end{cases}
		\end{aligned}
	\end{equation}

	assuming $\beta \leq \beta_c$ and using that $\beta_c = T$ for $2$ layer sign networks.

	Therefore

	\begin{equation}
		\begin{aligned}
			\max_{j \in [N]} |a_j^T(Az-y)| & = \max \{\beta, T -\beta\}.
		\end{aligned}
	\end{equation}

	So $\max_{j \in [N]} |a_j^T(Az-y)| \leq \beta$ if
	$T - \beta \leq \beta \leq \beta_c$, ie $\frac{T}{2} \leq \beta \leq \beta_c $. So the critical region is $\beta \in [\frac{T}{2},T]$.

	Therefore if $\beta \in [\frac{T}{2},T]$ then $|| A^T(Az-y)||_{\infty} \leq \beta$ and hence $0 \in A^T(Az-y) + \beta [-1,1] = \partial L(z)_j$ for $j \in [N]-\{\frac{T}{2}, N-\frac{T}{2}\}$.
}
\end{proof}

\iftoggle{long}{
\begin{remark}
	$z$ as defined in lemma \ref{eq:(Az-y)_j_critzone_beta_2lyr} is the minimizer

	\begin{align}\label{eq:subprob}
		\min \frac{1}{2}||\Tilde{A}\Tilde{z}-y||^2_2 + \beta \mathbf{1}^T \Tilde{z} ,
	\end{align}

	where  $\Tilde{A}=A\Tilde{I}$ is the submatrix of $A$, taking the columns with indices odd multiples of $\frac{T}{2}$, $\Tilde{z} = \Tilde{I}^Tz$ denotes a subvector $z$ with only the elements at every odd multiple of $\frac{T}{2}$. Indeed, \eqref{eq:subprob} is

	solved by setting the gradient to $0$:

	\begin{align*}
		\Tilde{A}^T(\Tilde{A}\tilde{z}-y) + \beta\mathbf{1} & =0                                                                 \\
		\tilde{z}                                           & = (\Tilde{A}^T\Tilde{A})^{-1}(\Tilde{A}^Ty-\beta\mathbf{1})        \\
		                                                    & =  (\Tilde{A}^T\Tilde{A})^{-1}(\beta_c \mathbf{1}-\beta\mathbf{1}) \\
		                                                    & (\beta_c - \beta)_+ (\Tilde{A}^T\Tilde{A})^{-1}\mathbf{1}.         \\
	\end{align*}

	To check that $\Tilde{z} \geq 0$ it suffices to show that $(\Tilde{A}^T\Tilde{A})^{-1}\mathbf{1} \geq 0$. Note $\Tilde{I}^T A^T A \Tilde{I}$ subsamples the matrix $A^TA$ at intervals of $T^{th}$ in the columns and then the rows starting at $\frac{T}{2}$.

	From equation \eqref{eq:A^TA_writtenout},

	\begin{equation}
		\setcounter{MaxMatrixCols}{20}
		\begin{aligned}
			\Tilde{I}^T A^T A \Tilde{I}
			 & =  \Tilde{I}^T\begin{pmatrix}
				                 N      & N-2    & N-4    & \cdots & 2      & 0      & -2     & \cdots & 6-N    & 4-N    & 2-N    \\
				                 N-2    & N      & N-2    & \cdots & 4      & 2      & 0      & \cdots & 8-N    & 6-N    & 4-N    \\
				                 N-4    & N-2    & N      & \cdots & 6      & 4      & 2      & \cdots & 10-N   & 8-N    & 6-N    \\
				                 \vdots & \vdots & \vdots & \ddots & \vdots & \vdots & \vdots & \ddots & \vdots & \vdots & \vdots \\
				                 6-N    & 8-N    & 10-N   & \cdots & 2      & 4      & 6      & \cdots & 8-N    & 6-N    & 4-N    \\
				                 4-N    & 6-N    & 8-N    & \cdots & 0      & 2      & 4      & \cdots & 6-N    & 4-N    & 2-N    \\
				                 2-N    & 4-N    & 6-N    & \cdots & -2     & 0      & 2      & \cdots & 4-N    & 2-N    & N
			                 \end{pmatrix}\Tilde{I} \\
			 & =  \begin{pmatrix}
				      N      & N-2T   & N-4T   & \cdots & 2T     & 0      & -2T    & \cdots & 6T-N   & 4T-N   & 2T-N   \\
				      N-2T   & N      & N-2T   & \cdots & 4T     & 2T     & 0      & \cdots & 8T-N   & 6T-N   & 4T-N   \\
				      N-4T   & N-2T   & N      & \cdots & 6T     & 4T     & 2T     & \cdots & 10T-N  & 8T-N   & 6T-N   \\
				      \vdots & \vdots & \vdots & \ddots & \vdots & \vdots & \vdots & \ddots & \vdots & \vdots & \vdots \\
				      6T-N   & 8T-N   & 10T-N  & \cdots & 2T     & 4T     & 6T     & \cdots & 8T-N   & 6T-N   & 4T-N   \\
				      4T-N   & 6T-N   & 8T-N   & \cdots & 0      & 2T     & 4T     & \cdots & 6T-N   & 4T-N   & 2T-N   \\
				      2T-N   & 4T-N   & 6T-N   & \cdots & -2T    & 0      & 2T     & \cdots & N-4T   & N-2T   & N
			      \end{pmatrix}            \\
		\end{aligned}
	\end{equation}

	which has inverse,

	\begin{equation}
		\setcounter{MaxMatrixCols}{20}
		\begin{aligned}
			(\Tilde{I}^T A^T A \Tilde{I})^{-1} & = \frac{1}{4T}
			\begin{pmatrix}
				2      & -1     & 0      & 0      & 0      & \cdots & 0      & 0      & 1      \\
				-1     & 2      & -1     & 0      & 0      & \cdots & 0      & 0      & 0      \\
				0      & -1     & 2      & -1     & 0      & \cdots & 0      & 0      & 0      \\
				0      & 0      & -1     & 2      & -1     & \cdots & 0      & 0      & 0      \\
				\vdots & \vdots & \vdots & \vdots & \vdots & \ddots & \vdots & \vdots & \vdots \\
				0      & 0      & 0      & 0      & 0      & \cdots & -1     & 2      & -1     \\
				1      & 0      & 0      & 0      & 0      & \cdots & 0      & -1     & 2      \\
			\end{pmatrix}. \\
		\end{aligned}
	\end{equation}


	So,

	\begin{equation}
		\begin{aligned}
			\Tilde{z} & = (\Tilde{I}^T A^T A \Tilde{I})^{-1}\mathbf{1} \\
			          & = \frac{1}{2T}
			\begin{pmatrix}
				1      \\
				0      \\
				\vdots \\
				0      \\
				1
			\end{pmatrix}
			          & \geq \mathbf{0}
		\end{aligned}
	\end{equation}

	as consistent with lemma \ref{eq:(Az-y)_j_critzone_beta_2lyr}.

\end{remark}

}


\iftoggle{arxiv}{
\begin{lemma}\label[lemma]{lemma:sum_formula}
      Let $a,b,c,d \in \mathbf{Z}_+, d \in \mathbb{R}, r = 1 - \mathrm{rem}(b-a,2)$. Then
      \begin{equation}\label{eq:sum_formula}
          \begin{aligned}
         \sum_{j{=}a}^b ({-}1)^j \left(c {-} jd\right) {=} ({-}1)^{a} \left( (c{-}ad)r {+} ({-}1)^{r} \frac{(b{-}(a{+}r){+}1)d}{2} \right) 
    {=} ({-}1)^a
        \begin{cases}
           \frac{(b{-}a{+}1)d}{2} & \mbox { if } r{=}0 \\
            c{-}ad {-} \frac{(b{-}a)d}{2}& \mbox { else. }
        \end{cases}
      \end{aligned}
      \end{equation}

\end{lemma}

\begin{proof}
We have

 \[
 \begin{aligned}
 \sum_{j=a}^b (-1)^j \left(c - jd\right) &= (-1)^{a}(c-ad)r + \sum_{j=a + r}^b (-1)^j \left(c - j d\right) \\
 %
 %
  &= (-1)^{a}(c-ad)r + (-1)^{a+r} \sum_{\substack{{a + r \leq j \leq b-1} \\ { j-(a+r) \textrm{ is even }}}}  \left(c - j d\right) - \left(c - (j+1) d\right) \\
  %
 \end{aligned}
 \]
Simplifying gives \eqref{eq:sum_formula}.
\end{proof}
}

\begin{lemma}\label[lemma]{lemma:lasso_sol_square_wave_noise}
	Let $L{=}2$, $\mathbf{y} {=} \hkt$, $0 {<} \beta {<} \frac{\beta_c}{2}$. 
Let $w_{\mathrm{bdry}}   {=} 1{-} \frac{3}{2}\betafrac, w_{\mathrm{cycle}} {=} 2\betafrac {-} 1$.
Let $\mathbf{z_{\mathrm{bdry}}} {=} \\ w_{\mathrm{bdry}}\left( \mathbf{e}^{\left( {\frac{T}{2}}\right)} {+} \mathbf{e}^{\left( {N{-}\frac{T}{2}} \right)}\right), \mathbf{z}_{\mathrm{cycle}}  {=} w_{\mathrm{cycle}} \sum_{i{=}2}^{2k{-}2} (-1)^{i} \mathbf{e}^{\left({\frac{T}{2}i} \right)}$.
Then $\mathbf{z}^* {=} \mathbf{z}_{bdry}{+}\mathbf{z}_{cycle}$ solves the \lasso{} problem \eqref{eq:genericlasso}.
\end{lemma}

\iftoggle{long}{

\begin{remark}\label{remark:pred_2layer} 
	The optimal neural net evaluated on the training data, $\mathbf{Az^*}$, has the similar square wave shape as $\mathbf{y}$, with the same number of cycles and same cycle length. However, the amplitudes of each half-cycle for $\mathbf{Az^*}$ are less than $\mathbf{y}$.
	The amplitude of the first and last half cycles are $-2w_{bdry}-w_{cycle}=1-\frac{\beta}{T}$, and the amplitude of the inner half cycles is $w_{cycle}=1-2\frac{\beta}{T}$.
\end{remark}
}

\iftoggle{long}{

\begin{remark}
	Each $z_j$ corresponds to a feature $a_j$ which is a column of $A^TA$ as shown in the left plot of \Cref{fig:feature_ex}, and $a_j$ is shaped like a peak. $A^T y$ is a zig-zag pattern as seen in figure \ref{fig:A^T(h+delta)}. The signs of $z_j$ alternate with $j$, corresponding to pointing the peak of $a_j$ up or down, which switches the direction of the zig-zag edge from rising or falling. The cycles in $y$ are all identical and can therefore be expressed by one number, $w_{cycle}$ which is the same coefficeient for all the terms in $z_{cycle}$. The boundary vector $z_{bdry}$ adjusts the beginning and endpoints of $A^T Az$ to match $A^T y$. The $z_i$ for the half-cycle $i$ containing noise is modified to account for the noise by adding $z_{noise}$.
\end{remark}

}

\begin{proof}
 We show $\mathbf{z}^*$ is optimal using the subgradient condition in \Cref{rk:strcvx}.
 \iftoggle{long}{
 with $T/2$ quotient and remainder }
By \Cref{cor:bdry_gram},

	\begin{equation}\label{eq:A^TAz_bdry}
		\begin{aligned}
			\frac{1}{2w_{\mathrm{bdry}}}(\Amat{}^T \Amat{} \mathbf{z}_{\mathrm{bdry}})_n   & =
			\begin{cases}
				\left(\Amat{}^T \hkt \right)_n & \mbox{ if } n \leq \frac{T}{2} \mbox{ or } n \geq N- \frac{T}{2}                      \\
				T                & \mbox{ if } \frac{T}{2} \leq n \leq N - \frac{T}{2} 
			\end{cases}, 
   \quad n \in [N].
		\end{aligned}
	\end{equation}
%
Next, 
\begin{equation}\label{eq:atazycycle}
  \frac{1}{w_{\mathrm{cycle}}}(\Amat{}^T \Amat{}\mathbf{z}_{\mathrm{cycle}})_n  =  \sum_{j=2}^{2k-2} (-1)^{j}\left(N-2\left|n-j\frac{T}{2}\right|\right).  
\end{equation}
%
%
\iftoggle{long}{
	\begin{figure}[ht]
		\begin{tikzpicture}
			\centering
    \scalebox{0.85}{
			\begin{axis}[axis x line=middle, axis y line=middle, xticklabels={,,}, xlabel = {$n$}]
				\addplot[ycomb,color=black,solid,mark=*]
				plot coordinates{
						(1,0)
						(2,-2)
						(3,-4)
						(4,-6)
						(5,-8)

						(6,-10)
						(7,-10)
						(8,-10)
						(9,-10)
						(10,-10)

						(11,-10)
						(12,-10)
						(13,-10)
						(14,-10)
						(15,-10)
						(16,-10)
						(17,-10)
						(18,-10)
						(19,-10)
						(20,-10)

						(21,-10)
						(22,-10)
						(23,-10)
						(24,-10)
						(25,-10)
						(26,-10)
						(27,-10)
						(28,-10)
						(29,-10)
						(30,-10)

						(31,-10)
						(32,-10)
						(33,-10)
						(34,-10)
						(35,-10)

						(36,-10)
						(37,-8)
						(38,-6)
						(39,-4)
						(40,-2)
					};

			\end{axis}
		\end{tikzpicture}
  }
		\caption{Example of $-\frac{1}{2w_{bdry}}\Amat{}^T\mathbf{Az_{bdry}}$\iftoggle{long}{ 
  with $T=10$,
  } 
  illustrating \eqref{eq:A^TAz_bdry}.}
		\label{fig:z_bdry}
	\end{figure}
}
\begin{center}
\begin{figure}[t]
\begin{tikzpicture}[scale=0.6]
    \centering
			\begin{axis}[axis x line=middle, axis y line=middle, xticklabels={,,}, xlabel = {$n$}]
				\addplot[ycomb,color=black,solid,mark=*]
				plot coordinates{
						(1,0)
						(2,2)
						(3,4)
						(4,6)
						(5,8)

						(6,10)
						(7,10)
						(8,10)
						(9,10)
						(10,10)

						(11,10)
						(12,10)
						(13,10)
						(14,10)
						(15,10)
						(16,10)
						(17,10)
						(18,10)
						(19,10)
						(20,10)

						(21,10)
						(22,10)
						(23,10)
						(24,10)
						(25,10)
						(26,10)
						(27,10)
						(28,10)
						(29,10)
						(30,10)

						(31,10)
						(32,10)
						(33,10)
						(34,10)
						(35,10)

						(36,10)
						(37,8)
						(38,6)
						(39,4)
						(40,2)
					};

			\end{axis}
\end{tikzpicture}
\begin{tikzpicture}[scale=0.6]
			\begin{axis}[axis x line=middle, axis y line=middle, xticklabels={,,}, xlabel = {$n$}]
				\addplot[ycomb,color=black,solid,mark=*]
				plot coordinates{
						(1,0)
						(2,2)
						(3,4)
						(4,6)
						(5,8)

						(6,10)
						(7,12)
						(8,14)
						(9,16)
						(10,18)

						(11,20)
						(12,18)
						(13,16)
						(14,14)
						(15,12)

						(16,10)
						(17,12)
						(18,14)
						(19,16)
						(20,18)

						(21,20)
						(22,18)
						(23,16)
						(24,14)
						(25,12)

						(26,10)
						(27,12)
						(28,14)
						(29,16)
						(30,18)

						(31,20)
						(32,18)
						(33,16)
						(34,14)
						(35,12)

						(36,10)
						(37,8)
						(38,6)
						(39,4)
						(40,2)
					};

			\end{axis}
\end{tikzpicture}
	\caption{Examples of $\frac{1}{2w_{bdry}}\Amat{}^T\mathbf{Az_{bdry}}$ \iftoggle{long}{ 
  with $T=10$,
  } 
 (left) and $\frac{1}{w_{cycle}}\Amat{}^T\mathbf{Az_{cycle}}$ \iftoggle{long}{with $T=10$,}
 (right).}
  \label[figure]{fig:z_bdry_cycle}
\end{figure}
\end{center}
%
%
See \Cref{fig:z_bdry_cycle}. 
\iftoggle{arxiv}{If $n{\leq}\frac{T}{2}$ or $n {\geq} N{-}\frac{T}{2}$ then there is $s \in \{-1,1\}$ such that for all $2{\leq}j{\leq}2k{-}2$, $n{-}j\frac{T}{2}=s(n{-}j\frac{T}{2})$. Applying \Cref{lemma:sum_formula} to \eqref{eq:atazycycle} and simplifying gives $(\Amat{}^T \mathbf{Az_{cycle}})_n {=}  \\ w_{\mathrm{cycle}}\left(N{-} skT {+}2sn\right)$. Comparing with \Cref{lemma:A^Th_kT} gives
%
%
\begin{equation}\label{eq:A^TAZcycle_edge}
    \frac{1}{w_{cycle}}(\Amat{}^T \mathbf{Az}_{\mathrm{cycle}})_n = 
2 
\begin{cases}
    n & \mbox{ if }  n \leq \frac{T}{2} \\
    N-n & \mbox{ if }  n \geq N - \frac{T}{2} 
\end{cases}
\quad = (\Amat{}^T{\hkt})_n.
\end{equation}

Next suppose $\frac{T}{2} \leq n \leq N - \frac{T}{2}$. Let $q_n = \textrm{quot} \left(n , \frac{T}{2} \right), r_n = \textrm{rem} \left(n , \frac{T}{2} \right)$. Then
\begin{equation}\label{eq:atazcyclemiddle}
  \frac{1}{w_{cycle}}(\Amat{}^T \mathbf{Az_{cycle}})_n =   \sum_{j=2}^{q_n} (-1)^{j} \left( N-2\left(n-j\frac{T}{2}\right) \right) + \sum_{j=q_n +1}^{2k-2} (-1)^{j}\left(N+ 2\left(n-j\frac{T}{2}\right) \right).
\end{equation}
%
%
%
Applying \Cref{lemma:sum_formula} to \eqref{eq:atazcyclemiddle} and simplifying 
gives 
\begin{equation*}
    \begin{aligned}  
     \frac{1}{w_{cycle}} \left(\Amat{}^T \mathbf{Az}_{\mathrm{cycle}} \right)_n  
        &{=} \begin{cases}
            N{-}2n{+}2T{+}\frac{q_n {-} 2 }{2}T {-} \frac{2k{-}2{-}q_n}{2}T {=} 2 \left(T {-} \textrm{rem}\left(n, \frac{T}{2} \right) \right) &\mbox { if } 
            q_n \mbox { is even } \\
            \frac{1 {-} q_n }{2}T  {+} N{+}2n {-}(1{+}q_n)T {-} \frac{2k{-}3 {-}q_n}{2}T {=} T {+} 2\textrm{rem}\left(n, \frac{T}{2}\right) &\mbox { if } 
            q_n \mbox { is odd } 
        \end{cases}\\
        &{=}(2-\mathbf{1}\{q_n \textrm{ odd}\} )T  {+} 2  (-1)^{q_n{+}1} \textrm{rem}\left(n,\frac{T}{2}\right) 
        {=} 2T {-} \left(\Amat{}^T \mathbf{h_{k,T}}\right)_n,
    \end{aligned}
\end{equation*}
where the last equality follows from \Cref{lemma:A^Th^kTwithrem}.
%
%
%
\iftoggle{long}{

	\begin{figure}[ht]
		\begin{tikzpicture}
			\centering
			\begin{axis}[axis x line=middle, axis y line=middle, xticklabels={,,}, xlabel = {$n$}]
				\addplot[ycomb,color=black,solid,mark=*]
				plot coordinates{
						(1,0)
						(2,2)
						(3,4)
						(4,6)
						(5,8)

						(6,10)
						(7,12)
						(8,14)
						(9,16)
						(10,18)

						(11,20)
						(12,18)
						(13,16)
						(14,14)
						(15,12)

						(16,10)
						(17,12)
						(18,14)
						(19,16)
						(20,18)

						(21,20)
						(22,18)
						(23,16)
						(24,14)
						(25,12)

						(26,10)
						(27,12)
						(28,14)
						(29,16)
						(30,18)

						(31,20)
						(32,18)
						(33,16)
						(34,14)
						(35,12)

						(36,10)
						(37,8)
						(38,6)
						(39,4)
						(40,2)
					};

			\end{axis}
		\end{tikzpicture}
		\caption{Example of $\frac{1}{w_{cycle}}\Amat{}^T\mathbf{Az_{cycle}}$ \iftoggle{long}{with $T=10$,}
  illustrating \eqref{eq:A^TAZcycle_edge}.}
		\label{fig:z_cycle}
	\end{figure}
}
Combining with \eqref{eq:A^TAZcycle_edge} gives

	\begin{equation}\label{eq:A^TAzcycle}
		\begin{aligned}
			(\Amat{}^T\mathbf{Az_{cycle}})_n 
			                  & = w_{cycle} \cdot
			\begin{cases}
				\left(\Amat{}^T \hkt\right)_n     & \mbox{ if } n \leq \frac{T}{2}     \mbox{ or } n \geq N- \frac{T}{2}                  \\
				2T - \left(\Amat{}^T \hkt\right)_n & \mbox{ if } \frac{T}{2} \leq n \leq N - \frac{N}{2}. 
			\end{cases}
		\end{aligned}
	\end{equation}

Add \eqref{eq:A^TAz_bdry} and \eqref{eq:A^TAzcycle} and plug in $\mathbf{y}=\hkt$ to get

} 

	\begin{equation}\label{eq:A^TAzn}
		\begin{aligned}
			(\Amat{}^T\mathbf{Az})_n 
			& {=}  
			\begin{cases}
				(w_{cycle}{+}2w_{bdry})(\Amat{}^T \mathbf{y})_n  {=} \left(1{-}\betafrac\right)(\Amat{}^T \mathbf{y})_n    & \mbox{ if } n {\leq} \frac{T}{2} \mbox{ or }   n {\geq} N{-} \frac{T}{2}                    \\
				(w_{bdry} {+} w_{cycle})2T {-} w_{cycle} (\Amat{}^T \mathbf{y})_n {=} \beta {+} \left(1 {-} 2\betafrac \right) (\Amat{}^T \mathbf{y})_n   & \mbox{ if } \frac{T}{2} {\leq} n \leq N {-} \frac{N}{2}. 
			\end{cases} 
		\end{aligned}
	\end{equation}

 Therefore, 

 \begin{equation}\label{eq:A^TAz-y}
		\begin{aligned}
			-\frac{1}{\beta}\Amat{}^T(\mathbf{Az}-\mathbf{y})_n 
			& =  
			\begin{cases}
				\frac{1}{\beta_c}(\Amat{}^T \mathbf{y})_n      & \mbox{ if } n \leq \frac{T}{2} \mbox{ or }   n \geq N- \frac{T}{2}                    \\
				 1  + \frac{2}{\beta_c} (\Amat{}^T \mathbf{y})_n   & \mbox{ if } \frac{T}{2} \leq n \leq N - \frac{N}{2} .
			\end{cases} 
		\end{aligned}
	\end{equation}

By \Cref{lemma:2lyr_critbeta>=}, $\beta_c = T$. 
By \Cref{lemma:A^Th_kT}, $0 \leq \frac{1}{\beta_c}(\Amat{}^T \hkt)_n = \frac{1}{\beta_c}(\Amat{}^T \mathbf{y})_n \leq 1$, and $\frac{1}{\beta_c}(\Amat{}^T \mathbf{y})_n=1$ when $n$ is an odd multiple of $\frac{T}{2}$. Since $0 < \beta < \frac{\beta_c}{2}$, we have $w_{bdry}>0$ and $w_{cycle}<0$. Therefore $\frac{1}{\beta}\Amat{}^T(\mathbf{Az}-\mathbf{y})_n = -\mbox{sign}(\mathbf{z}^*_n)$ when $\mathbf{z}^*_n \neq 0$, ie $n$ is an integer multiple of $\frac{T}{2}$. And for all $n \in [N]$, $\left|\frac{1}{\beta}\Amat{}^T(\mathbf{Az}-\mathbf{y})_n \right| \leq 1$. By \Cref{rk:strcvx}, $\mathbf{z}^*$ is optimal.
\end{proof}


\iftoggle{arxiv}{

The nonzero indices of $\mathbf{z_{bdry}}$ and  $\mathbf{z_{cycle}}$ partition those that are  multiples of $\frac{T}{2}$. 

}



\iftoggle{long}{
\begin{lemma}(Informal lemma \ref{lemma:3layer_criticalbeta}) \\
	Suppose the \labelvec{} of a $3$-layer network is a square wave with one noisy point. If the noise is sufficiently small that it does not cause a sign change, then at a critical value of the regularization coefficient $\beta$, only one final-layer neuron will be active, and the predicted neural net will be the unnoisy square wave with its entire amplitude scaled according to the noise and $\beta$.

\end{lemma}
}

\begin{proof}[Proof of \Cref{thm:no_noise_square_wave_2lyr}]
    By \Cref{lemma:2lyr_sign_square_wave_big_beta} and \Cref{lemma:lasso_sol_square_wave_noise}, \\
    $
		\begin{aligned}
			\mathbf{z}^* & =
			\begin{cases}
				\frac{1}{2}\left(1-\betafrac\right)_+
                \left(\mathbf{e}^{\left({\frac{T}{2}}\right)}+ \mathbf{e}^{\left({N-\frac{T}{2}}\right)}\right) & \mbox{ if } \betafrac \geq  \frac{1}{2} \\    \mathbf{z_{bdry}}+\mathbf{z_{cycle}} &\mbox { if } 0 < \betafrac \leq \frac{1}{2}.
			\end{cases}
		\end{aligned}
	$
%
%
\end{proof}

\begin{proof}[Proof of \Cref{cor:solfor2lyrwave}]
    Note that \unscal{}ing (defined in \Cref{section:nn_arch}) does not change the neural network as a function. The reconstructed neural net (\Cref{def:reconstructfunc}) before \unscal{}ing is $\nn{2}{x}  {=} \sum_{i=1}^N z^*_i \sigma(x{-}x_i)$. For $\frac{1}{2} {\leq} \betafrac{} {\leq} 1 $, $ \nn{2}{x} {=}\frac{1}{2}\left(1{-}\betafrac \right)_+ \left(\sigma \left(x{-}x_{\frac{T}{2}}\right) {+} \sigma\left(x{-}x_{N{-}\frac{T}{2}} \right)\right)$. We can compute $\nn{2}{x} $ similarly for $\betafrac{} \leq \frac{1}{2} $.
\end{proof}

\subsection{$L=3$ layer nets}\label{appsec:squarewaveL3}

\begin{proof}[\Cref{thm:3layer_no_noise}]
 Since $\mathbf{y}{=}\hkt$ switches $2k{-}1$ times, by \Cref{thm:3lyr_sign}, $\Amat{}_i {=}\hkt$ for some $i$. 
Since $\mathbf{y}{=}{-}\Amat{}_i$, and for all $n \in [N], \|\Amat{}_n \|_2 {=} N$, we have $i {\in} \mbox{argmax}_{n \in N} |\Amat{}_n^T \mathbf{y}|$. By \Cref{rmk:criticalbeta}, $\beta_c {=} \max_{n \in N} |\Amat{}_n^T \mathbf{y}| {=} \mathbf{y}^T \Amat{}_i {=} N$. So if $\beta {>} \beta_c$ then $\mathbf{z}^*{=}0$, consistent with $z_i {=} {-}\left( 1 {-} \betafrac\right)_+$.

Next, $\mathbf{z}^*$ satisfies the subgradient condition in \Cref{rk:strcvx}, since
	for $n \in [N]$, $\displaystyle 
 \left|\Amat{}_n^T(\Amat{}\mathbf{z}^*-\mathbf{y}) \right| \\ =\left|\Amat{}_n^T(z_i \Amat{}_i - \Amat{}_i)\right|  = (z_i-1) \left |\Amat{}_n^T\Amat{}_i \right| = \left|\frac{\beta}{\beta_c}  \Amat{}_n^T\mathbf{y} \right| \leq \frac{\beta}{\beta_c} \mbox{argmax}_{n \in N} \left|\Amat{}_n^T \mathbf{y} \right| = \leq \beta$. Since $\mathbf{z}^*_i< 0$, when $i=n$, $\Amat{}_i^T(\Amat{}\mathbf{z}^*-\mathbf{y})=\beta = -\beta \mbox{sign}(\mathbf{z}_i^*)$. 
 By \Cref{rk:strcvx}, 
$\mathbf{z}^*$ is optimal.
\end{proof}

\begin{proof}[\Cref{cor:solfor3lyrwave}]
Follows from the reconstruction in \Cref{thm:3lyr_sign_reconstruction}. 
\end{proof}

\iftoggle{long}{

\begin{figure}[ht]
	\centering
	\begin{tikzpicture}
		\centering
		\begin{axis}[axis x line=middle, axis y line=middle, xticklabels={,,}, xlabel = {$n$}]
			\addplot[ycomb,color=black,solid,mark=*]
			plot coordinates{
					(1,1)
					(2,1)
					(3,1)
					(4,1)
					(5,1)
					(6,-1)
					(7,-1)
					(8,-1)
					(9,-1)
					(10,-1)

					(11,1)
					(12,1)
					(13,1)
					(14,1)
					(15,1)
					(16,-1)
					(17,-1)
					(18,-1)
					(19,-1)
					(20,-1)

					(21,1)
					(22,1)
					(23,1)
					(24,1)
					(25,1)
					(26,-1)
					(27,-1)
					(28,-1)
					(29,-1)
					(30,-1)

					(31,1)
					(32,1)
					(33,1)
					(34,1)
					(35,1)
					(36,-1)
					(37,-1)
					(38,-1)
					(39,-1)
					(40,-1)
				};

			\addplot[ycomb,color=magenta,solid,mark=*]
			plot coordinates{
					(1,1/2)
					(2,1/2)
					(3,1/2)
					(4,1/2)
					(5,1/2)
					(6,-1/2)
					(7,-1/2)
					(8,-1/2)
					(9,-1/2)
					(10,-1/2)

					(11,1/2)
					(12,1/2)
					(13,1/2)
					(14,1/2)
					(15,1/2)
					(16,-1/2)
					(17,-1/2)
					(18,-1/2)
					(19,-1/2)
					(20,-1/2)

					(21,1/2)
					(22,1/2)
					(23,1/2)
					(24,1/2)
					(25,1/2)
					(26,-1/2)
					(27,-1/2)
					(28,-1/2)
					(29,-1/2)
					(30,-1/2)

					(31,1/2)
					(32,1/2)
					(33,1/2)
					(34,1/2)
					(35,1/2)
					(36,-1/2)
					(37,-1/2)
					(38,-1/2)
					(39,-1/2)
					(40,-1/2)
				};
		\end{axis}

		\draw [decorate,
			decoration = {brace, mirror}] (0.9,4.3) --  (0.9,5.7);
		\node[] at (1.2,5) { $\frac{\beta}{N}$};

	\end{tikzpicture}
	\caption{$L=3$ layer neural net prediction plots illustrating \Cref{thm:3layer_no_noise}. The  \labelvec{} $\mathbf{y}=\mathbf{h_{k,T}}$ is plotted in black. The 2 layer neural net prediction on the training data at each index is plotted in magenta, where  $\beta= 20 = \frac{N}{2} \leq N$. The magnitude of the predicted value is $1-\frac{\beta}{N} = 1- \frac{1}{2} = \frac{1}{2}$.
	}
	\label{fig:pulse_overlap_3lyr}
\end{figure}

}






\section*{\secnameSolSet}

\subsection{Proofs for results in \Cref{sec:solsetslassovsnn}}\label{section:solsetav}

\input{appendices/solution_set}

\begin{lemma}\label[lemma]{prop:stat_nn}
Suppose $L=2$, and the activation is ReLU, leaky ReLU or absolute value. 
Suppose $m^* \leq m \leq 2N$.  Since $m \leq 2N$, we can 
let $\paramspace^{\mathrm{\lasso}, \mathrm{stat}} = \{ \params: \exists j \in [N] \text{ s.t. } b_i = -x_j w_i  \} \subset \paramspace$. 
Let $\params^* \in \paramspace^{\mathrm{\lasso}, \mathrm{stat}} \cap C(\beta)$. Then $\params^*$ is a minima of the \lasso{} problem.
\end{lemma}
\begin{proof}
We can ignore $\xi$ since its derivative and reconstruction are straightforward, and it does not interact with any other parameters. We denote vector-vector operations as being performed elementwise.

Since $m^* \leq m$, a neural net reconstructed from \lasso{} is optimal in the training problem. Let $F(\params)$ and $F^{\mathrm{\lasso}}\left(\mathbf{z}\right)$ be the objectives of the \nc{} training problem \eqref{eq:nonconvex_generic_loss_gen_L_intro}
 and \lasso{}  \eqref{eq:genericlasso}, respectively. 
 The parameters $\params$ are stationary if $\params{} \in C(\beta)$, i.e., $0 \in \partial F(\params)$.

  Let  $\paramspace^{\mathrm{\lasso}} {=} \left \{\params: \exists j {\in} [N] \text{ s.t. } |w_i| {=}  |\alpha_i|, b_i {=} {-} x_j w_i   \right \} \subset \paramspace^{\mathrm{\lasso}, \mathrm{stat}}$.  
  By a similar argument as the proof of Theorem 3 in \cite{wang2021hidden}, 
  since $0 {\in} \partial F(\params^*)$, we have $|w_i| {=} |\alpha_i|$ for all neurons $i$. Therefore $\params^* {\in} \paramspace^{\mathrm{\lasso{}}} $. Thus for $\alpha^* {\in} \params^*$, observe that $\params{}^* {=} R^{\alpha,w,b\to \params}(\boldsymbol{\alpha}^*)$.
 Let $\tilde{F}(\boldsymbol{\alpha}){=}F \left(R^{\alpha,w,b \to \params}(\boldsymbol{\alpha}) \right) $. 
 
Since $0 {\in} \partial F(\params^*)$ 
at $(\boldsymbol{\alpha}^*) {=} \left({R^{\alpha,w,b \to \params}}\right)^{-1}(\params^*)$, we have $\tilde{F}(\boldsymbol{\alpha}^*){=}F(\theta^*)$. Perform the following operations elementwise. The chain rule gives $\partial \tilde{F}(\boldsymbol{\alpha}^*) {=} \partial F(\theta^*) \partial R^{\alpha,w,b \to \params}(\boldsymbol{\alpha}^*) {\ni} \mathbf{0}$.
Let
$ R^{\alpha {to} z}(\boldsymbol{\alpha}) 
=\mbox{sign}(\boldsymbol{\alpha})\boldsymbol{\alpha}^2$.
At $\mathbf{z}^* {=} R^{\alpha \to z}\left(\boldsymbol{\alpha}^*\right) $, we have $F^{\mathrm{\lasso}}(\mathbf{z}^*){=} \tilde{F}(\boldsymbol{\alpha}^*) $. The chain rule gives $\partial F^{\mathrm{\lasso}}(\mathbf{z}^*)  {=}  \partial \tilde{F}(\boldsymbol{\alpha}^*) \partial  R(\mathbf{z}^*) {\ni} \mathbf{0}$. Since the \lasso{} problem \eqref{eq:genericlasso} is convex, the result holds.
\end{proof}

 \iftoggle{long}{

\begin{proposition}\label{prop:stat_nn}
  Consider training a $2$-layer neural net $\nn{2}{\mathbf{X}}$ and $m^* \leq m \leq |\dicindexset{}|$ neurons. Then $R^{-1}$ 
  applied to any stationary point $\params^*$ of the \nc{} training problem \eqref{eq:nonconvex_generic_loss_gen_L_intro} that satisfies $b_i = -x_j w_i$ for all neurons $i \in \dicindexset{}$ yields a minima of the \lasso{} problem \eqref{eq:genericlasso}.
  
\end{proposition}
   
\begin{proof}[Proof of \Cref{prop:stat_nn}]
%

By \Cref{rk:Rinv},
 $\params^* \in \paramspace^{\lasso}$. For any $\params \in \paramspace^{\lasso}$,  $\params$ is determined by $\boldsymbol{\alpha}$ and $\xi$ as $(\boldsymbol{\alpha}, \xi, \mathbf{w}, \mathbf{b}) = g(\boldsymbol{\alpha},\xi):=(\boldsymbol{\alpha}, \xi, \mathbf{s}|\boldsymbol{\alpha}|, -\mathbf{s}\mathbf{X}|\boldsymbol{\alpha}|)$, where operations are taken elementwise. 
 Let $\tilde{F}(\boldsymbol{\alpha},\xi)=F(g(\boldsymbol{\alpha},\xi))$. 
 
Since $0 \in \partial F(\params^*)$ 
at $(\boldsymbol{\alpha}^*,\xi^*) = g^{-1}(\params^*)$, we have $\tilde{F}(\boldsymbol{\alpha}^*,\xi^*)=F(\theta^*)$. The chain rule gives $\partial_{(\boldsymbol{\alpha},\xi)} \tilde{F}(\boldsymbol{\alpha}^*,\xi^*) = \partial_{\theta} F(\theta^*) \partial_{(\boldsymbol{\alpha},\xi)} g(\boldsymbol{\alpha}^*,\xi^*) \ni \mathbf{0}$.
Next, at $\mathbf{z}^* = (R^{-1}\left(\boldsymbol{\alpha}^*\right))$, we have $F^{\lasso}(\mathbf{z}^*)= \tilde{F}(\boldsymbol{\alpha}^*) $. The chain rule gives $\partial F^{\lasso}(\mathbf{z}^*)  =\partial \tilde{F}(\boldsymbol{\alpha}^*) \partial R(\mathbf{z}^*) \ni \mathbf{0}$. Since the  \lasso{} problem \eqref{eq:genericlasso} is convex, the result holds.
\end{proof}

}

\begin{proof}[Proof of \Cref{thm:statpts_equiv_set}]
Observe that
$
   R(\lassoSolSet{}(\beta)) \subseteq  \tilde{\mathcal{C}}(\beta) \cap \paramspace^{\mathrm{\lasso{}}, \mathrm{stat}}  \subseteq  \mathcal{C}(\beta) \cap \paramspace^{\mathrm{\lasso{}}, \mathrm{stat}}  \subseteq    R(\lassoSolSet{}(\beta)),
$
where the first and last subset inequality follow from \Cref{thrm:lasso_deep_1nueron_ReLU} and \Cref{prop:stat_nn}, respectively. Therefore all subsets in the above expression are equal. Observe that \\ $P \left (\paramspace^{\mathrm{\lasso{}}, \mathrm{stat}} \right) {=} \paramspace^P$ and so $P \left(C(\beta) \cap \paramspace^{\mathrm{\lasso{}}, \mathrm{stat}} \right) {=} C(\beta) \cap P \left(\paramspace^{\mathrm{\lasso{}}, \mathrm{stat}}\right) {=}C(\beta) \cap \paramspace^P $ and similarly $P \left(\tilde{C}(\beta) \cap \paramspace^{\mathrm{\lasso{}}, \mathrm{stat}} \right){=}\tilde{C}(\beta) \cap \paramspace^P  $. Now apply $P$ to all subsets above. 
\end{proof}


\section*{Numerical results}

\iftoggle{long}{
By \Cref{lemma:minNormCert}, the $L=3$ \lasso{} solution found numerically is optimal in both the $L=3$ and $L=4$ \lasso{} problems, while the $L=4$ \lasso{} solution found numerically is not.
In \Cref{fig:3vs4sim}, $\xi=0$ and $z_i = \frac{3}{4}$ corresponding to a single \deepnarrow{} feature $\ReLUdicfun{L}=\ramp{+}{2}{6}$ in the $L=3$ and $L=4$ dictionaries achieves $\xi \mathbf{1} + \Amat \mathbf{z} = \mathbf{y}$ as can be seen from the figure, and also achieves $\|\mathbf{z}\|_1 = z_i =  \frac{3-0}{6-2}= \frac{y_2-y_3}{x_2-x_3}$. 
By \Cref{lemma:minNormCert}, the optimal value of the min norm problem \eqref{rm_gen_lasso_nobeta} for $L=3,4$ is $\|\mathbf{z}^*\|=\frac{3}{4}$. However, in \Cref{fig:3vs4sim}, 
the $L=4$ solution that is found satisfies $3 =\|\mathbf{z}^*\|_1 > \frac{3}{4}$. The $L=4$ solution that is found is a near-optimal solution with optimality gap less than $10^{-7}$ in the (non-min norm) training problem. However, this solution is suboptimal. In \Cref{fig:3vs4sim}, the $L=3$ solution is optimal for $L=4$.

In \Cref{fig:zigzag}, $\xi=0$ and $z_{i_1} = \frac{13}{4}$ corresponding to the feature $\ReLU{+}{2}$ and $z_{i_2} = \frac{3}{4}$ corresponding to the feature $\ReLU{+}{6}$ satisfies $\xi \mathbf{1} + \Amat \mathbf{z} = \mathbf{y}$ and $\|\mathbf{z}\|_1 = z_{i_1}+z_{i_2} =  4 = \frac{6-3}{7-6} = \frac{y_1-y_2}{x_1-x_2}$. These features are in the dictionaries for $L=2,3,4$. By \Cref{lemma:minNormCert}, the optimal value of the min norm problem \eqref{rm_gen_lasso_nobeta} for $L=2,3,4$ is $\|\mathbf{z}^*\|=4$.
In \Cref{fig:zigzag}, for $L=2,3$, the \lasso{} solution that is found satisfies $\|\mathbf{z}^*\|_1 = 3.25+0.75=4$ but for $L=4$, $\|\mathbf{z}^*\|_1 = 2.5+6.5=9>4$. 
The $L=4$ solution that is found is a near-optimal solution with optimality gap less than $10^{-7}$ in the (non-min norm) training problem. However, this solution is suboptimal. The dictionary for $L=4$ contains the dictionary for $L=3$, which contains the dictionary for $L=2$. 
}

\subsection{Autoregression figures}\label{sec:App_autoreg}
In all figures except for the regularization path, the horizontal axis is the training epoch.

\iftoggle{long}{ 
\begin{figure}[t]
\centering
\begin{minipage}[t]{0.9\textwidth}
\centering
    \includegraphics[width=\linewidth]{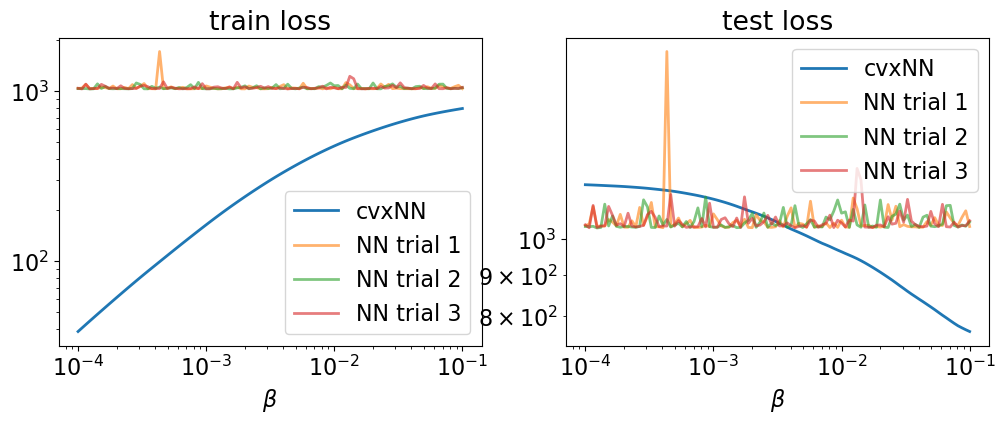}
\caption*{Without skip connection.}
\end{minipage}
\begin{minipage}[t]{0.9\textwidth}
\centering
    \includegraphics[width=\linewidth]{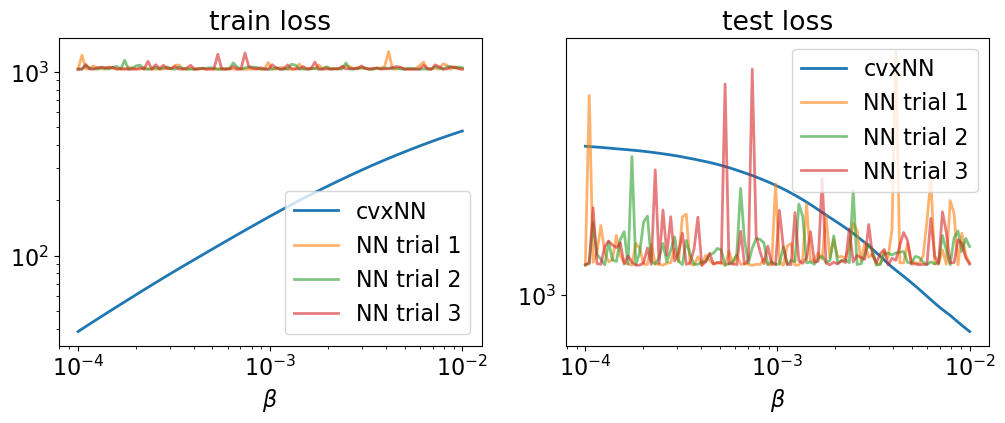}
\caption*{With skip connection.}
\end{minipage}
\caption{BTC-hourly. Regularization path.}
\end{figure}
}

\begin{figure}[t]
\centering
\begin{minipage}[t]{0.9\textwidth}
\centering
\includegraphics[width=\linewidth]{figures/BTC_2017min_T_2000_MOSEK_lbd1_m1000.png}
\caption*{BTC-2017min. }
\end{minipage}
\begin{minipage}[t]{0.9\textwidth}
\centering
\includegraphics[width=\linewidth]{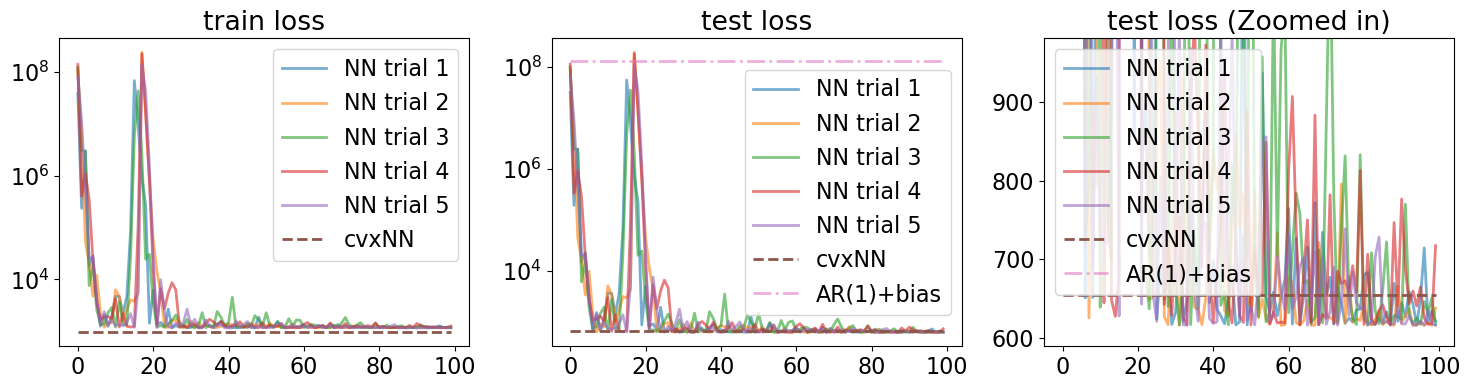}
\caption*{BTC-hourly.}
\end{minipage}
\caption{Regression with L2 loss.}\label{fig:l2}
\end{figure}

\begin{figure}[t]
\centering

\iftoggle{long}{ 
\begin{minipage}[t]{0.9\textwidth}
\centering
\includegraphics[width=\linewidth]{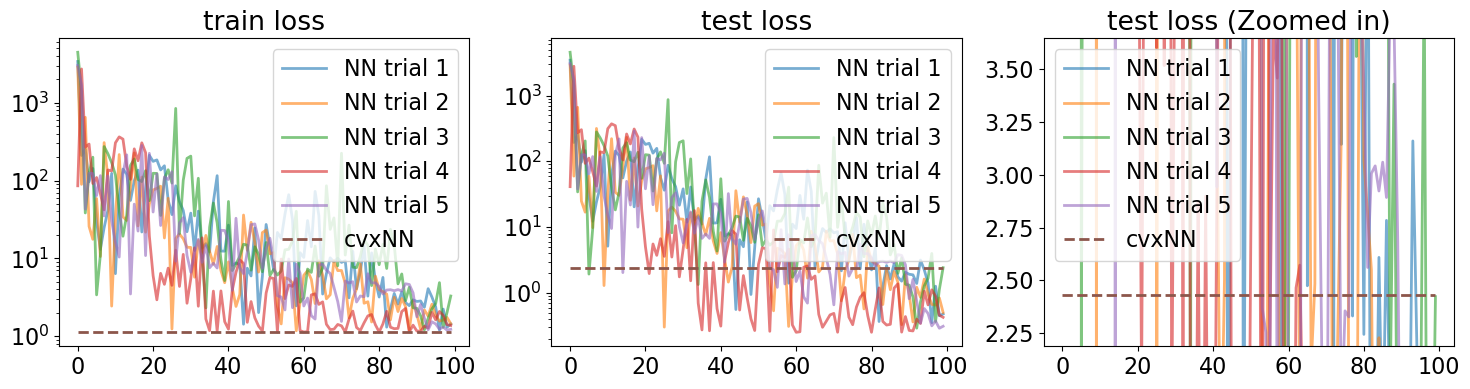}
\caption*{BTC-2017min. }
\end{minipage}
}

\begin{minipage}[t]{0.9\textwidth}
\centering
\includegraphics[width=\linewidth]{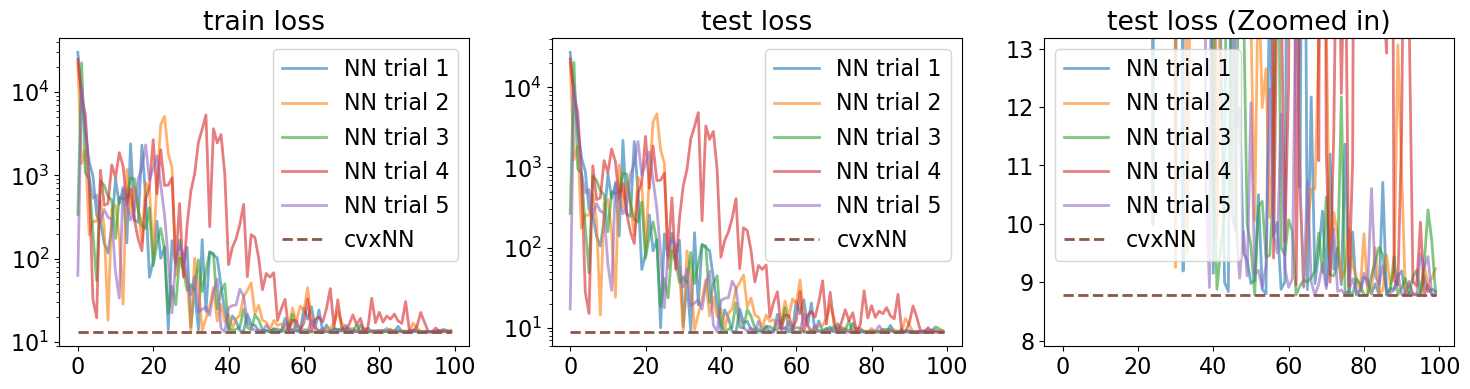}
\caption*{BTC-hourly.}
\end{minipage}
\caption{Regression with quantile loss. $\tau=0.3$}\label[figure]{fig:quantile_0.3}
\end{figure}

\begin{figure}[!htb]
\centering
\begin{minipage}[t]{0.9\textwidth}
\centering
\includegraphics[width=\linewidth]{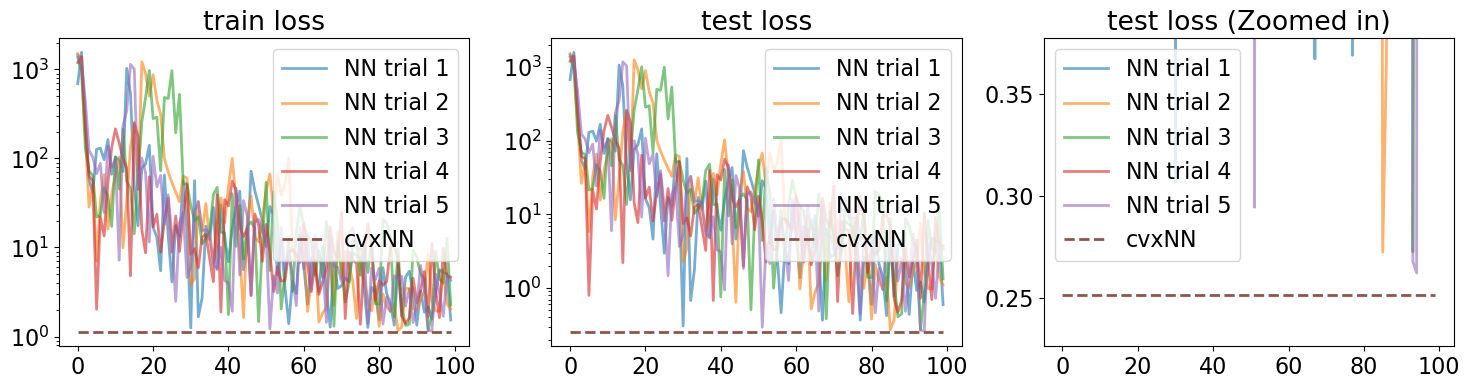}
\caption*{BTC-2017min. }
\end{minipage}

\iftoggle{long}{ 
\begin{minipage}[t]{0.9\textwidth}
\centering
\includegraphics[width=\linewidth]{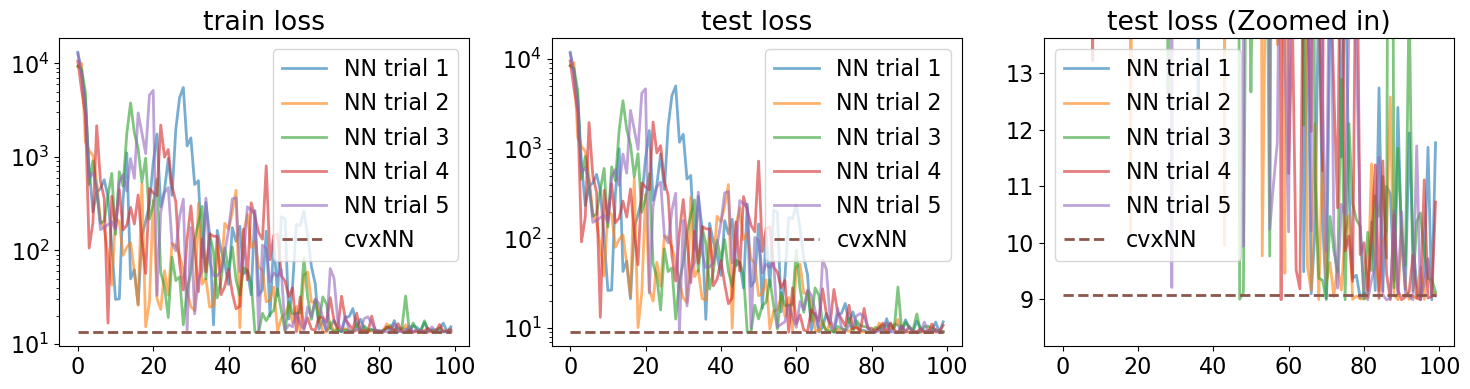}
\caption*{BTC-hourly.}
\end{minipage}
}
\caption{Regression with quantile loss. $\tau=0.7$}\label[figure]{fig:quantile_0.7}
\end{figure}

\iftoggle{long}{
\begin{figure}[!htb]
\centering
\begin{minipage}[t]{0.9\textwidth}
\centering
\includegraphics[width=\linewidth]{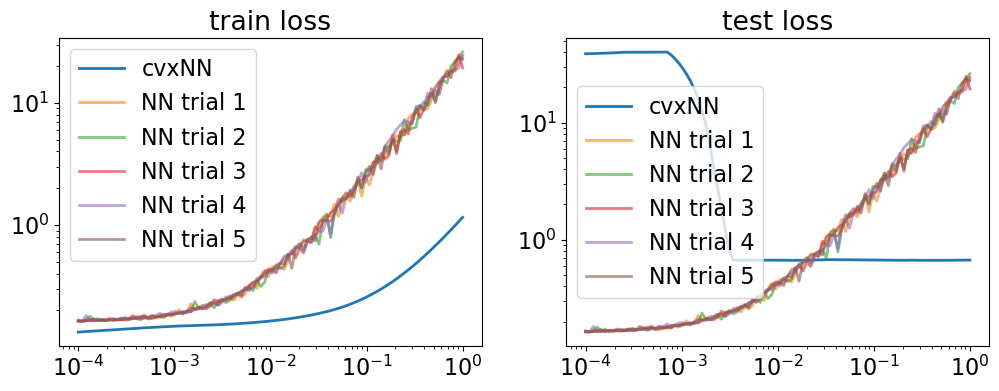}
\end{minipage}
\caption{The regularization path. BTC2017-min.} 
\end{figure}
}

\iftoggle{todos}{\textcolor{red}{todo:use restatable and link proof sections to paper}}

\iftoggle{todos}{\emi{todo: alternate sign definition also used in appendix?, for fitting the neural net to the square wave.}}



\iftoggle{long}{
A parallel neural network can be also represented as a standard network. A parallel net is $
 \nn{3,par}{\mathbf{X}}
 := \sum_{i=1}^{m_2} \sigma \left ( \left [ \sigma \left([\mathbf{X},1] \Wul{i}{1}   \right), \mathbf{1} \right ] \mathbf{W^{(i,2)}} \right) \alpha_i + \xi
$. Let $\tilde{\Wul{i}{l}} = \Wul{i}{l} $ with $\left(0, \cdots, 0, 1 \right)^T$ appended to the rightmost column. Let $\mathbf{W}^{(1)}_s = \left[ \Tilde{\Wul{1}{1}} \cdots \Tilde{\Wul{m_2}{1}} \right], \mathbf{W}^{(2)}_s = \mathrm{blockdiag}\left( \Wul{1}{2} \cdots \Wul{m_2}{2} \right)$ with $\mathbf{0}^T$ appended to the bottom row. Then $\left[[\mathbf{X},1] \mathbf{W^{(1)}_s} \right] =  \left[[\mathbf{X},1]\Tilde{\Wul{1}{1}} \cdots  [\mathbf{X},1]\Tilde{\Wul{m_2}{1}}  \right] = \left[ \left[[\mathbf{X},1]\Wul{1}{1}\right], \mathbf{1}  \cdots \left[ [\mathbf{X},1]\Wul{m_2}{1}\right], \mathbf{1}  \right] $. Therefore there is  an equivalent standard neural net with larger weight matrices,

$\nn{3,std}{\mathbf{X}} =   \sigma \left ( \left[ \sigma \left([\mathbf{X},1] \mathbf{W^{(1)}_s} \right), \mathbf{1}  \right ]   \mathbf{W^{(2)}_s} \right) \boldsymbol{\alpha} + \xi = \nn{3,par}{\mathbf{X}} $. Here we assume $\sigma(1)=1$, which holds for all activations considered except leaky ReLU (see \Cref{app_section:act_func}), in which case we can append $\left(0, \cdots, 0, \frac{1}{c_2}\right)^T$ instead of $\left(0, \cdots, 0, 1\right)^T$ to $\Wul{i}{l}$ to create $\tilde{\Wul{i}{l}}$.
 }
 

\clearpage


\iftoggle{long}{

\section{3 layer networks}
Consider a \nd{1} $L=3$-layer parallel network \eqref{eq:3lyr_nn_par}, where each parallel unit has a single neuron in each hidden layer (i.e., two neurons in series):

\begin{equation}\label{3lyr_seriesnn}
    \nn{3}{x} = \sum_{j=1}^{m_2} \sigma \left(\sigma \left( x w^{(j,1)} + b^{(j,1)} \right) w^{(j,2)} + b^{(j,2)}  \right) \alpha_i + \xi.
\end{equation}

Consider the \nc{} training problem

\begin{equation}\label{eq:noncvxtrain_3lyrL3norm}
    \min_{\params \in \Theta}  \mathcal{L}_{\mathbf{y}} (\nn{L}{\mathbf{X}})
     + \frac{\beta}{3}  \|\regparams\|_3^3,
\end{equation}

where the regularized parameters are $\regparams = \left \{w^{(j,l)}, \alpha_j:j \in [m_2], l \in [2] \right \} $, and they are penalized by their $l_3$ norm as $ \|\regparams\|_3^3 = \sum_{j=1}^{m_2} \left( |w^{(j,1)}|^3 + | w^{(j,2)}|^3 + \alpha_j^3 \right)$. 

\begin{theorem}\label{thrm:3lyrReLUlong}
The training problem \eqref{eq:noncvxtrain_3lyrL3norm} for the $3$-layer neural network \eqref{3lyr_seriesnn} is equivalent to the \lasso{} problem
     \begin{equation}\label{L3lasso}
     \min_{\mathbf{z}, \xi \in \mathbb{R}} & \mathcal{L}_{\mathbf{y}}(\mathbf{ A z} + \xi \mathbf{1} )+ \beta \|\mathbf{z}\|_1, 
 \end{equation}

 where for $s_1, s_2 \in \{1, -1\}$ and $ j_1, j_2 \in [N]$, $\Amat{}_{i, (s_1,s_2),(j_1,j_2)} = \sigma \left( s_2 \left( \sigma \left( s_1 \left( x_i - x_{j_1} \right)  \right)   -\sigma \left(s_1 \left( x_{j_2}   - x_{j_1} \right)  \right)  \right) \right)$.
\end{theorem}

The dictionary matrix $\Amat{}$ has $4N^3$ elements and is indexed by a multiset. However, depending on the activation, $\Amat{}$ may contain redundant copies of columns. 

\begin{proof}[Proof of \Cref{thrm:3lyrReLUlong}]

By the AM-GM inequality, $|w^{(j,1)}|^3 + | w^{(j,2)}|^3 + \alpha_j^3 \geq 3 |w^{(j,1)}w^{(j,2)}\alpha_j|$, with equality achieved when $|w^{(j,1)}|=|w^{(j,2)}|=|\alpha_j|$. For a homogenous activation, e.g. ReLU, leaky ReLU, or absolute value, a similar argument as \Cref{prop:2layerhomrescale} can be used to rescale \eqref{eq:noncvxtrain_3lyrL3norm} to the equivalent problem 

\begin{equation}\label{eq:noncvxtrain_3lyrL3norm_rescale}
    \min_{\params: |w^{(i,l)}| = 1, l \in [2] }  \mathcal{L}_{\mathbf{y}} (\nn{L}{\mathbf{X}})
     + \beta \|\regparams\|_1.
\end{equation}

By a similar argument as \Cref{lemma:dual_of_rescaled}, taking the dual of \eqref{eq:noncvxtrain_3lyrL3norm_rescale} over the outer weights gives a lower bound

\begin{equation}\label{gen_dual_gen_arch_outer_L3}
		\begin{aligned}
			\min_{  |w^{(1)}| = 1, b^{(1)} }  \max_{{\lambda} \in \mathbb{R}^N} & -\mathcal{L}_{\mathbf{y}}^*( {\lambda} )                                            \\
			\text{ s.t. }                   & \max_{|w^{(2)}|=1, b^{(2)}}\left | \lambda^T \sigma \left(\sigma \left( \mathbf{X} w^{(1)} + b^{(1)} \right) w^{(2)} + b^{(2)} \mathbf{1}  \right) \right| \leq \beta
		\end{aligned} .
	\end{equation}

 A lower bound on \eqref{gen_dual_gen_arch_outer_L3} is 

\begin{equation}\label{gen_dual_gen_arch_outer_lb_L3}
		\begin{aligned}
			\max_{{\lambda} \in \mathbb{R}^N} & -\mathcal{L}_{\mathbf{y}}^*( {\lambda} )                                            \\
			\text{ s.t. }                   & \max_{|w^{(l)}|=1, b^{(l)} : l \in [2] }\left | \lambda^T \sigma \left(\sigma \left( \mathbf{X} w^{(1)} + b^{(1)} \right) w^{(2)} + b^{(2)} \mathbf{1} \right) \right| \leq \beta
		\end{aligned}.
	\end{equation}

 By a similar argument as the proof of \Cref{lemma:dual_finite}, the optimal $b^{(2)}$ in the constraint of \eqref{gen_dual_gen_arch_outer_lb_L3} will be $-\sigma \left( x_{n_2} w^{(1)} + b^{(1)} \right)w^{(2)}$ for some $n_2 \in [N]$. Then, the optimal $b^{(1)}$ will be $ - x_{n_1} w^{(1)} $ for some $n_1 \in [N]$. Plugging these $b^{(1)}, b^{(2)}$ into  \eqref{gen_dual_gen_arch_outer_lb_L3} gives the equivalent problem 
 
 \begin{equation}\label{gen_dual_gen_arch_outer_lb_L3-equiv}
		\begin{aligned}
			\max_{{\lambda} \in \mathbb{R}^N} & -\mathcal{L}_{\mathbf{y}}^*( {\lambda} )                                            \\
			\text{ s.t. }                   & 
   \left | \lambda^T \sigma \left( w^{(2)} \left( \sigma \left( w^{(1)}\left( \mathbf{X} - x_{n_1} \right)  \right)   -\sigma \left(w^{(1)}\left( x_{n_2}   - x_{n_1} \right)  \right) \mathbf{1} \right) \right) \right| \leq \beta: w^{(1)},w^{(2)} \in \{-1,1\}, n_1, n_2 \in [N] \\
   & \lambda^T \mathbf{1} = 0
		\end{aligned}.
	\end{equation}

 By similar arguments as \Cref{lemma:lasso_without_simp}, the dual of \eqref{gen_dual_gen_arch_outer_lb_L3-equiv} is \eqref{L3lasso}.

 Now suppose $(\mathbf{z}^*, \xi^*)$ is a solution to the \lasso{} problem \eqref{L3lasso} such that $\|\mathbf{z^*}\|_0 \leq m_2$. For the nonzero indices $i$ of $\mathbf{z^*}$, setting $\alpha_j = \mathbf{z^*}_j, w^{(j,1)} = s_1, b^{(j,1)} = -s_1 x_{j_1}, w^{(j,2)} = s_2, b^{(j,2)} =-s_2 \sigma \left(s_1 \left( x_{j_2}   - x_{j_1} \right)  \right), \xi = \xi^*$ and then rescaling parameters so that $|w^{(j,1)}|=|w^{(j,2)}|=|\alpha_j|$ gives a neural network $\nn{3}{\mathbf{X}}$ that achieves the same value in the \nc{} problem \eqref{3lyr_seriesnn} as its lower bound \lasso{} problem \eqref{L3lasso}. Therefore, the \lasso{} problem \eqref{L3lasso} is equivalent to the training problem \eqref{3lyr_seriesnn}.

\end{proof}
}

\iftoggle{long}{

\begin{lemma}\label{lemma:reconstruct_bdd2original}
    Consider a $L$-layer parallel neural net with sign or threshold activation, with $L \geq 2$ layers. Suppose $\alpha_i, s^{(i, L-1)}$ are the final layer weight and amplitude parameters of the $i^{th}$ parallel unit feasible in the rescaled problem \eqref{equ:nn_train_signdetinput_rescale}. Let $\gamma_i = \sqrt{|\alpha_i|}$. Then with $\alpha_i \to \frac{\alpha_i}{\gamma_i} $ and $ s^{i,L-1} \to \gamma_i s^{i,L-1}$, the limit of the objective of the original problem \eqref{eq:nonconvex_generic_loss_gen_L_intro} as  $||\mathbf{W}^{(i,l)}||_F \to 0 $ for $l \in [L-1]$, $s^{(i,l)} \to 0 $ for $l \in [L-2]$ is the objective of the rescaled problem \eqref{equ:nn_train_signdetinput_rescale}.
\end{lemma}

\begin{proof}
Recall the amplitude parameters $s^{(i,l)}$ of the rescaled problem \eqref{equ:nn_train_signdetinput_rescale} are set to $1$.
Plugging in the transformed variables into the objective of the original problem \eqref{eq:nonconvex_generic_loss_gen_L_intro} for $L$-layer parallel networks \eqref{eq:generic_NN} gives $\mathcal{L}_{\mathbf{y}} (\xi + \sum_{i=1}^{m_L} s^{(i,L-1)} \gamma_i \sigma \left ( \mathbf{X}^{(i,L-2)} \mathbf{W}^{(i,L-1)} \right) \frac{\alpha_i}{\gamma_i}  + \frac{\beta}{2}  (\sum_i (\gamma_i s^{i,L-1})^2 + \sum_{l=1}^{L-2} ||\mathbf{W}^{(i,l)}||_F^2 + (s^{(i,L-1)})^2 ) = 
\mathcal{L}_{\mathbf{y}} (\xi + \sum_{i=1}^{m_L}  \sigma \left ( \mathbf{X}^{(i,L-2)} \mathbf{W}^{(i,L-1)} \right) \alpha_i  + \frac{\beta}{2}  ( \sum_i |\alpha_i| + |\alpha_i| + \sum_{l=1}^{L-2} ||\mathbf{W}^{(i,l)}||_F^2 + (s^{(i,L-1)})^2 ) \xrightarrow[]{\mathbf{||W^{(i,l)}}||_F, s^{(i,l)} \to 0 } \mathcal{L}_{\mathbf{y}} (\xi + \sum_{i=1}^{m_L}  \sigma \left ( \mathbf{X}^{(i,L-2)} \mathbf{W}^{(i,L-1)} \right) \alpha_i  +  \beta ||\boldsymbol{\alpha}||_1 $, which is the objective of the rescaled problem \eqref{equ:nn_train_signdetinput_rescale}.
 \end{proof}

}

\iftoggle{long}{   
\section{$L$ layers, one neuron in each layer of each parallel unit}\label{app:Llyr1neuronReLU}

\begin{lemma}
The training problem for \eqref{eq:noncvxtrain_LlyrLLnorm} is equivalent to   \begin{equation}\label{eq:noncvxtrain_3lyrLLnorm_rescale}
    \min_{\params: |w^{(j,l)}| = 1, l \in [L-1], j \in [m_{L-1}] }  \mathcal{L}_{\mathbf{y}} (\nn{L}{\mathbf{X}})
     + \beta \|\boldsymbol{\alpha}\|_1.
\end{equation}
\end{lemma}

\begin{proof}
Since the activation is homogeneous, there is only one neuron per layer, and bias parameters are unregularized, the weight magnitudes $\left|w^{(j,l)}\right|$ can be factored out of the neural network so that  \eqref{eq:noncvxtrain_LlyrLLnorm} is equivalent to 
\begin{equation}\label{eq:Llayer_pullout_wts}
    \min_{\params \in \Theta, s^{(j,l)} = \pm 1}   \mathcal{L}_{\mathbf{y}} \left(\sum_{j=1}^{m_{L-1}} \sigma \left( \cdots \sigma \left(\sigma \left( x s^{(j,1)} + b^{(j,1)} \right)  \cdots \right) s^{(j,L-1)} + b^{(j,L-1)}\right )\alpha_j\prod_{l=1}^{L-1}\left|w^{(j,l)}\right| + \xi \right)
     + \frac{\beta}{L}  \|\regparams\|_L^L.
\end{equation}

    By the AM-GM inequality, 
    \begin{equation}
        \frac{1}{L} \|\regparams\|^L_L =  \sum_{j=1}^{m_{L-1}} \frac{\left|\alpha_j \right|^L+\sum_{l=1}^{L-1} \left|w^{(j,l)}\right|^L }{L} \\
    \geq \sum_{j=1}^{m_{L-1}} \left(\left|\alpha_j \right|^L\prod_{l=1}^{L-1} \left|w^{(j,l)}\right|^L \right)^{1/L}
    = \sum_{j=1}^{m_{L-1}}\prod_{q \in \regparams^{(j)}} |q|. 
    \end{equation}

    So a lower bound on \eqref{eq:Llayer_pullout_wts} is 
    \begin{equation}\label{eq:Llayer_prodform}
    \min_{\params \in \Theta, s^{(j,l)} = \pm 1}   \mathcal{L}_{\mathbf{y}} \left(\sum_{j=1}^{m_{L-1}} \sigma \left( \cdots \sigma \left(\sigma \left( x s^{(j,1)} + b^{(j,1)} \right)  \cdots \right) s^{(j,L-1)} + b^{(j,L-1)}\right )\prod_{q \in \regparams^{(j)}} |q|+ \xi \right)
     +  \beta \sum_{j=1}^{m_{L-1}}\prod_{q \in \regparams^{(j)}} |q|.
\end{equation}
    
    Moreover, given a solution for \eqref{eq:Llayer_pullout_wts}, a solution for \eqref{eq:Llayer_prodform} that achieves the same objective can be found as follows: for each $j \in [m_{L-1}]$,
   let $\gamma^{(j)} = \left( \prod_{q \in \regparams^{(j)}} |q|  \right)^{1/L}$ and apply the transformation $\alpha^{(j,l)} \to \mbox{sign} \left(\alpha^{(j,l)}\right) \gamma, |w^{(j,l)}| \to \gamma$. So, \eqref{eq:Llayer_pullout_wts} and \eqref{eq:Llayer_prodform} are equivalent. Finally, for each $j \in [m_{L-1}]$, $\prod_{q \in \regparams^{(j)}} |q|$ can be combined into one variable $\alpha_j$ so that \eqref{eq:Llayer_prodform} becomes \eqref{eq:noncvxtrain_3lyrLLnorm_rescale}.
\end{proof}

\begin{lemma}\label{lemma:Llayer_dual_lb}
 Let $\mathbf{X}^{(0)} = \mathbf{X}$, and for  $l \in [L-1]$, let $\mathbf{X}^{(l)}= \sigma \left( \mathbf{X}^{(l-1)}w^{(l)}+b^{(l)} \mathbf{1}\right) \in \mathbb{R}^N$ where $w^{(l)} \in \{-1,1\}, b^{(l)} \in \mathbb{R}$.
 A lower bound on \eqref{eq:noncvxtrain_3lyrLLnorm_rescale} is 
     \begin{equation}\label{gen_dual_gen_arch_outer_lb_LL}
		\begin{aligned}
			\max_{{\lambda} \in \mathbb{R}^N} & -\mathcal{L}_{\mathbf{y}}^*( {\lambda} )                                            \\
			\text{ s.t. }                   & \max_{|w^{(l)}|=1, b^{(l)} : l \in [L-1] }\left | \lambda^T \mathbf{X}^{(L-1)} \right| \leq \beta
		\end{aligned}.
	\end{equation}
\end{lemma}

\begin{proof}
    By a similar argument as \Cref{lemma:dual_of_rescaled}, taking the dual of \eqref{eq:noncvxtrain_3lyrLLnorm_rescale} over the outer weights gives a lower bound

\begin{equation}\label{gen_dual_gen_arch_outer_LL}
		\begin{aligned}
			\min_{ |w^{(l)}| = 1,  b^{(l)}: l \in [L-2]}  \max_{{\lambda} \in \mathbb{R}^N} & -\mathcal{L}_{\mathbf{y}}^*( {\lambda} )                                            \\
			\text{ s.t. }                   & \max_{|w^{(L-1)}|=1, b^{(L-1)}}\left | \lambda^T \mathbf{X}^{(L-1)} \right| \leq \beta
		\end{aligned} .
	\end{equation}
 A lower bound on \eqref{gen_dual_gen_arch_outer_LL} is \eqref{gen_dual_gen_arch_outer_lb_LL}.
\end{proof}

}

\end{document}

%% file: sections/solution_set.tex
We have shown that training neural networks on \nd{1} data
is equivalent to fitting a \lasso{} model. 
Now we develop analytical expressions for all minima of the \lasso{} problem and its relationship to the set of all minima of the training problem. 
These results, which build on the existing literature for convex reformulations
\citep{mishkin2023solution} as well as characterizations of the
\lasso{} \citep{efron2004least}, illustrate that
the \lasso{} model provides insight into \nc{} networks.
We focus on $2$-layer models with ReLU, leaky ReLU and absolute value activations, although our results can be extended to other architectures by considering the corresponding neural net reconstruction. 
Proofs are deferred to \cref{section:solsetav}. 

We start by characterizing the set of global optima to the  \lasso{} problem \eqref{eq:genericlasso}.
Suppose
\( (\mathbf{z}^*, \xi^*) \) is a solution to
the convex training problem.
In this notation, the optimal model fit $\hat{\mathbf{y}}$ and equicorrelation set $\mathcal{E}_\beta $ are given by
\[
	\hat{\mathbf{y}} = \mathbf{A} \mathbf{z}^* + \xi^* \mathbf{1},
	\quad \quad
	\mathcal{E}_\beta =
	\left\{
	i 
 : |\mathbf{A}_{i}^\top (\hat{\mathbf{y}} - \mathbf{y})| = \beta
	\right\},
\]
where \( \hat{\mathbf{y}} \) is unique over the optimal set
\citep{vaiter2012dof, tibshirani2013unique}.
The equicorrelation set contains the features
maximally correlated with the residual $\hat{\mathbf{y}} - \mathbf{y}$
and plays a critical role in the solution set.
\begin{proposition}\label[proposition]{prop:lasso-solution-set}
	Suppose \( \beta {>} 0 \).
	Then the set of global optima of the \lasso{} problem \eqref{eq:genericlasso} is 
	\begin{equation}\label{eq:lasso-optimal}
		\lassoSolSet{}^*(\beta) =
		\left\{
		(\mathbf{z}, \xi) :
		z_i \! \neq \! 0 \Rightarrow \mathrm{sign}(z_i) \! = \! \mathrm{sign}\left(\mathbf{A}_{i}^\top (\hat{\mathbf{y}} - \mathbf{y})\right), \,
		z_i = 0 \,  \forall i \not \in \mathcal{E}_\beta, \,
		\mathbf{A} \mathbf{z} + \xi \mathbf{1} = \hat{\mathbf{y}}.
		\right\}
	\end{equation}
\end{proposition}
The solution set \( \lassoSolSet{}^*(\beta) \) is polyhedral
and its vertices correspond exactly to minimal models,
i.e. models with the fewest non-zero elements of \( \mathbf{z} \)
\citep{mishkin2023solution}.
Let $R$ be the reconstruction function described in \cref{rk:2lyrRecon}. All networks generated from applying $R$ to a \lasso{} solution are globally optimal in the training problem. The next result gives a full description of such networks. The $2$-layer parameter notation defined in \Cref{sec:deepnarrownets} is used.
\begin{proposition}
    \label[proposition]{thm:nc-solution-set}
	Suppose \( \beta > 0 \) and the activation is ReLU, leaky ReLU or absolute value. 
 The set of all $2$-layer \lasso{}-reconstructed networks is   
	\begin{equation}\label{eq:nc-optimal}
		\begin{aligned}
			R(\lassoSolSet{}(\beta)) =
			\bigg\{
			( & \mathbf{w}, \mathbf{b}, \mathbf{\alpha}, \xi) :
			\mathbf{\alpha}_i \! \neq \! 0 \Rightarrow
			\mathrm{sign}(\mathbf{\alpha}_i) \! = \!
			\mathrm{sign}\left(\mathbf{A}_{i}^\top (\mathbf{y} - \hat{\mathbf{y}})\right),
			b_i = - x_i \frac{\salpha_i}{\sqrt{|\alpha_i|}}, 
			\\
			  & w_i = \frac{\salpha_i}{\sqrt{|\alpha_i|}}; \,
			\mathbf{\alpha}_i = 0 \,  \forall i \notin \mathcal{E}_\beta, \,
			\nn{2}{\mathbf{X}} = \hat{\mathbf{y}}
			\bigg\}.
		\end{aligned}
	\end{equation}
\end{proposition}
\cref{thm:nc-solution-set} shows that all neural nets trained using our \lasso{} and reconstruction share the same
model fit whose set of active neurons is at most the equicorrelation
set.
By finding just \emph{one} optimal neural net that solves \lasso{}, we can form
\( R(\lassoSolSet{}^{\beta}) \) and compute all others. 


The min-norm solution path is continuous for the \lasso{} 
problem \citep{tibshirani2013unique}.
Since the solution mapping in \cref{def:reconstructfunc}, \cref{app:deepReLU} is continuous,
the corresponding reconstructed neural net path is also continuous
as long as the network is sufficiently wide.
Moreover, we can compute this path efficiently using the LARS algorithm
\citep{efron2004least}.
This is in contrast to the under-parameterized setting, where the regularization path is discontinuous \citep{mishkin2023solution}.

	


What subset of optimal, or more generally, stationary, points of the \nc{} training problem \eqref{eq:nonconvex_generic_loss_gen_L_intro} consist of \lasso{}-generated networks $R(\lassoSolSet{}(\beta))$? First, $R(\lassoSolSet{}(\beta))$ can generate additional optimal networks through neuron splitting, described as follows.
 Consider a single neuron  $\alpha \sigma_s(w x + b) $ (where 
$\alpha, w, b \in \mathbb{R}$), and let $\{\gamma_i\}_{i=1}^{n} \subset [0,1]^n$ be such that $\sum_{i=1}^n \gamma_i = 1$. The neuron can be \textit{split} into $n$ neurons $ \left \{\sqrt{\gamma_i} \alpha  \sigma \left(\sqrt{\gamma_i} w x + \sqrt{\gamma_i} b \right) \right \}_{i=1}^n  $
 \iftoggle{long}{
 when $\sigma$ is ReLU, leaky ReLU or absolute value
 ; and $ \left \{\sqrt{\gamma_i} \alpha  \sigma_{\sqrt{\gamma_i} s} \left(w x +  b \right) \right \}_{i=1}^n  $  when $\sigma$ is sign or threshold }
 \cite{wang2021hidden}. 
 For any collection $\paramspace$ of parameter sets $\params$, let $P(\paramspace)$ be the collection of parameter sets generated by all possible neuron splits and permutations of each $\params \in \paramspace$.
 Next, let $\mathcal{C}(\beta)$ and $\tilde{C}(\beta)$ be the sets of Clarke stationary points and solutions to the \nc{} training problem \eqref{eq:nonconvex_generic_loss_gen_L_intro}, respectively. 

\begin{proposition}\label[proposition]{thm:statpts_equiv_set}
	Suppose $L=2, \beta > 0$, the activation is ReLU, leaky ReLU or absolute value and 
  $m^* \leq m \leq 2N$.
 Let $\paramspace^P = \{ \params: \forall i \in [m], \exists j \in [N] \ \mathrm{ s.t.}\  b_i = -x_j w_i  \} $. %
	Then
 \begin{equation}
P(R(\lassoSolSet{}(\beta))) = \tilde{\mathcal{C}}(\beta) \cap \paramspace^{P} = \mathcal{C}(\beta) \cap \paramspace^{P} .
\end{equation}
	
\end{proposition}

\cref{thm:statpts_equiv_set} states that up to neuron splitting and permutation, our \lasso{} method gives all stationary points in the training problem satisfying $b_i = -x_i w_i$. Moreover, all such points are optimal in the training problem, similar to \cite{Feizi:2017}. 

Since optimal solutions are stationary, a neural net reconstructed from the \lasso{} model is in $\tilde{C}(\beta) \subset \mathcal{C}(\beta)$. However, $C(\beta) \not \subset \paramspace^P$. 
 This is because there may be other neural nets with the same output on $\mathbf{X}$ as the reconstructed net so that they are all in $\mathcal{C}(\beta)$, but that differ in the the unregularized parameters $\mathbf{b}$ and $\xi$, so that they are not in $\paramspace^{P}$. For example, if $\beta$ is large enough, the \lasso{} solution is $\mathbf{z}=\mathbf{0}$ \citep{efron2004least}, so the reconstructed net will have $\boldsymbol{\alpha}=\mathbf{0}$, which makes the neural net invariant to $\mathbf{b}$.
%
In this section, we analyzed the general structure of the \lasso{} solution set when $\beta > 0$. Next, we analyze the \lasso{} solution set for specific activations and training data when $\beta \to 0$.

%% file: appendices/solution_set.tex

\begin{proof}[\cref{prop:lasso-solution-set}]
The result is almost a sub-case of that given by Mishkin and Pilanci
\citep{mishkin2023solution} with the exception that the bias parameter \( \xi \),
is not regularized.
Therefore optimality conditions do not impose a sign
constraint and it is sufficient that
$
	\mathbf{1}^\top (\mathbf{A} \mathbf{z} + \xi \mathbf{1} - \mathbf{y}) = 0
$
for \( \xi \) to be optimal.
This stationarity condition is guaranteed by
$
	\mathbf{A} \mathbf{z} + \xi \mathbf{1} = \hat{\mathbf{y}}.
$
Now let us look at the parameters \( z_i \).
If \( i \not \in \mathcal{E}(\beta) \), then
\( z_i = 0 \) is necessary and sufficient from standard results
on the \lasso{} \citep{tibshirani2013unique}.
If \( i \in \mathcal{E}(\beta) \)
and \( {z_i} \neq 0 \),
then
$
	\mathbf{A}_{i}^\top (\hat{\mathbf{y}} - \mathbf{y}) = \beta \text{sign}({z}_i),
$
which shows that \( \mathbf{z}_i \) satisfies first-order conditions.
If \( {z_i} = 0 \), then first-order optimality is immediate since
$
	|\mathbf{A}_{i}^\top (\hat{\mathbf{y}} - \mathbf{y})| \leq \beta,
$
holds.
Putting these cases together completes the proof.
\end{proof}


\begin{proof}[\cref{thm:nc-solution-set}]
This result follows from applying the reconstruction in
\cref{def:reconstructfunc} to each optimal point in \( \lassoSolSet{} \).
The reconstuction sets
\( {\alpha}_{i} = \text{sign}({z}_{i})\sqrt{|{z}_{i}|} \).
From this we deduce
$
	\text{sign}({\alpha}_{i}) \! = \!
	\text{sign}\left(\mathbf{A}_{i}^\top (\mathbf{y} - \hat{\mathbf{y}})\right).
$
\iftoggle{long}{
If \( {z}_{i} = 0 \), then from the reconstruction we see that 
\( w_i = b_i = 0 \);
these neurons are usually omitted by reducing the width of the non-convex model,
but we retain them for simplicity.
Otherwise, 
}
The solution mapping determines the values of
\( w_i \) and \( b_i \) and in terms of \( \mathbf{\alpha}_{i} \).
Finally, the constraint
$
	\nn{2}{\mathbf{X}} = \hat{\mathbf{y}}
$
follows immediately by equality of the convex and non-convex prediction
functions on the training set.
\end{proof}